\documentclass[12pt,oneside,letterpaper]{report}

\usepackage{float}
\usepackage{algorithm}
\usepackage[noend]{algpseudocode}

\usepackage{amssymb}
\usepackage{amsthm}
\usepackage{xurl}
\usepackage{contour}
\usepackage{amsmath,amsfonts,amssymb}
\DeclareMathAlphabet{\mathbbold}{U}{bbold}{m}{n}
\usepackage{multirow}
\usepackage[many]{tcolorbox}    
\usepackage{ragged2e}

\newcommand{\iid}{i.i.d.}
\newcommand{\ood}{o.o.d.}

\newcommand{\thesistitle}{AI for the Open-World: \\the learning principles}
\newcommand{\thesisauthor}{Jianyu Zhang}
\newcommand{\thesisadvisor}{L{\'e}on Bottou, Yann Lecun}
\newcommand{\thesisdept}{Center for Data Science}
\newcommand{\gradmonth}{July}
\newcommand{\gradyear}{2025}

\theoremstyle{plain}
\newtheorem{theorem}{Theorem}[section]
\newtheorem{proposition}[theorem]{Proposition}

\theoremstyle{definition}
\newtheorem{definition}[theorem]{Definition}

\theoremstyle{remark}

\RequirePackage[margin=1in, includefoot, letterpaper]{geometry}

\definecolor{LightCyan}{rgb}{0.7,1,1}
\definecolor{SchoolColor}{rgb}{0.3412, 0.0235, 0.5490} %
\definecolor{chaptergrey}{rgb}{0.2600, 0.0200, 0.4600} %
\definecolor{midgrey}{rgb}{0.4, 0.4, 0.4}

\usepackage{hyperref}

\hypersetup{colorlinks,
  linkcolor=darkblue,
  citecolor=darkblue,
  urlcolor=darkblue,
  filecolor=darkblue}
  
\RequirePackage[labelfont={bf,sf,small,singlespacing},
                textfont={sf,small,singlespacing},
                margin=0pt,
                figurewithin=chapter,
                tablewithin=chapter]{caption}

\usepackage{fix-cm}
\RequirePackage[raggedright,sc]{titlesec}
\definecolor{gray75}{gray}{0.75}
\newcommand{\hsp}{\hspace{20pt}}

\titleformat{\chapter}[hang]
{\Huge\sc}
{\textcolor{SchoolColor}{\thechapter}\hsp\textcolor{gray75}{|}\hsp}
{0pt}{\Huge\sc\raggedright}
\usepackage{setspace}
\usepackage{lipsum}

\input{defs}

\begin{document}

\pagenumbering{roman}
\thispagestyle{empty}

\vspace*{25pt}
\begin{center}

  {
    \begin{doublespace}
    \Huge
      {\textcolor{SchoolColor}{AI for the Open-World: the Learning Principles}}
    \end{doublespace}
  }
    
  \vspace{.2in}
    by \\
    \vspace{.4in}
  \thesisauthor \\
  \vspace{2.5in}

  \begin{doublespace}
    \textsc{
    PhD Thesis \\
    \thesisdept~New York University\\
    \gradmonth, \gradyear}
  \end{doublespace}
    Advisors: \thesisadvisor 
    \vspace{.4in}
\end{center}
\vfill

 \begin{center}
    \textit{This is not a compilation of published papers, but a new one.}
    \end{center}

\newpage

\thispagestyle{empty}
\vspace*{25pt}
\begin{center}
  \scshape \noindent \small \copyright \  \small  \thesisauthor \\
  all rights reserved, \gradyear
\end{center}
\vspace*{0in}
\newpage

\chapter*{Abstract}
\addcontentsline{toc}{chapter}{Abstract}

During the past decades, numerous successes of Artificial Intelligence (AI) has been made on ``specific capabilities'', named \emph{closed-world}, such as artificial environments or specific real-world tasks. This well-defined narrow capability brings two nice benefits, a clear criterion of success\footnote{For example, scores in game environments, precision or recall metrics for real-world tasks.} and the opportunity to collect a lot of examples.    
The criteria not only reveal whether a machine has achieved a goal, but also reveal how the machine falls short of the goal. As a result, human designers can fix the problems one after the other until the machine is deemed good enough for the task.\footnote{For example, Montezuma’s Revenge game is hard for reinforcement learning algorithms because of the sparse reward signals. However, by creating intermediate rewards, RL algorithms accomplish this game easily.} Furthermore, the large set of collected examples reduces the difficulty of this problem-fixing process (by the central limit theorem).

Do the success in \emph{closed-world} translate into broad \emph{open-world}, where a machine is required to perform any task that a human could possibly undertake with fewer examples and less priori knowledge from human designers (i.e. \textsc{Turing Test})? No. Because competence in a specific task provides little insight in handling other tasks, the valuable criteria for specific tasks become helpless when handling broader unseen tasks.
Furthermore, due to the shortage of examples in unseen tasks, central limit theorem does not stand on our side. At the end, human designers lose the oscilloscope to ``hack'' an AI system for the open-world.\footnote{The number of unseen tasks is too large to consider thoroughly before encountering them in real scenarios.}

Achieving AI for the open-world requires unique \emph{learning principles} and innovated techniques, which are different from the ones in building AI for the closed-world. This thesis explores necessary learning principles required to construct AI for the open-world, including \emph{rich features} (analogy a large tool box), \emph{disentangled representation} (analogy an organized tool box), and \emph{inference-time learning} (analogy a tool-savvy hand). Driven by the learning principles, this thesis further proposes innovated techniques to utilize the learning principles, conducts enormous large-scale experiments to verify the learning principles.

\begin{boxH}
\end{boxH}

\newpage

\renewcommand{\appendixtocname}{List of appendices}
\renewcommand{\appendixpagename}{Appendices}

\makeatletter
\let\oldappendix\appendices

\renewcommand{\appendices}{%
  \clearpage
  \let\tf@toc\tf@app
  \addtocontents{app}{\protect\setcounter{tocdepth}{1}}
  \immediate\write\@auxout{%
    \string\let\string\tf@toc\string\tf@app^^J
  }
  \oldappendix
}%

\newcommand{\listofappendices}{%
  \begingroup
  \renewcommand{\contentsname}{\appendixtocname}
  \let\@oldstarttoc\@starttoc
  \def\@starttoc##1{\@oldstarttoc{app}}
  \tableofcontents%
  \endgroup
}

\tableofcontents
\listofappendices

\pagenumbering{arabic} %

\chapter{Introduction}
\label{chp:introduction}

\begin{boxH}
\textit{We propose that a 2 month, 10 man study of artificial intelligence be carried out during the summer of 1956 at Dartmouth College in Hanover, New Hampshire. The study is to proceed on the basis of the conjecture that every aspect of learning or any other feature of intelligence can in principle be so precisely described that a machine can be made to simulate it. An attempt will be made to find how to make machines use language, form abstractions and concepts, solve kinds of problems now reserved for humans, and improve themselves. We think that a significant advance can be made in one or more of these problems if a carefully selected group of scientists work on it together for a summer.} 
\begin{flushright}
--- McCarthy, Minsky, Rochester, and Shannon, 1955
\end{flushright}
\end{boxH}

It is well known that the ambition of creating artificial intelligence (AI) for the \emph{open-world} -- a machine capable of performing any task that a human could possibly undertake -- did not go well. Over the years, many successes in AI research are instead associated with the \emph{closed-world}, includes artificial environments (e.g. games \cite{samuel1959some,berliner1980backgammon,tesauro1995temporal,silver2016mastering}) and specific real-world tasks (e.g. handwriting recognition \cite{matan1991multi}, face recognition \cite{taigman2014deepface}, chat\&language \cite{eliza,SHRDLU,winograd1971procedures}). 
These successes were mainly due to clear criteria for success and a lot of examples. The criteria in closed-world (e.g. precision or recall) do not only reveal whether a machine has achieved a goal, but also reveals how the machine falls short of the goal. The large set of examples further reduces the difficulty of constructing AI systems (thank central limit theorem).

With a bit of AI research experience for specialized tasks in the setup of \emph{closed-world}, one would realize that the criteria can be ``hacked'' by understanding how machines fall short of their goals and then incorporating priori knowledge of designers, such as feature engineering or reward shaping. One (human designers) can fix the problems one after the other until the machine is deemed good enough for the task. In the end, instead of proving that our machine is intelligent, we often find satisfaction in proving instead that we are intelligent \cite{ai-in-open-world}. Clearly, this is not the right AI we want to create. 

What we want is AI for the \emph{open-world}:
\begin{boxA}
    AI for the open-world requires a machine to learn on a \textbf{wide range} of new tasks/domains quickly using \textbf{fewer examples} and \textbf{less task-specific priori knowledge} from human designers. 
\end{boxA}

In contrast to AI for the closed-world which handles one \iid~distribution\footnote{ Shuffling the examples from the mixture of multiple distributions results in a \iid~distribution.}, AI for the \emph{open-world} aims at achieving versatility -- being able to carry out any task a human could possibly undertake, not just specialized capacities.
This aligns with the measurement of \textsc{Turing Test} proposed by the great Alan Turing in 1950 \cite{turing-test}, where the competence on a specific task provides little insight into passing \textsc{Turing Test}. %
This requirement contrasts sharply with that of closed-world AI. Thus, open-world AI requires unique \emph{learning principles} and {techniques} that differ from those used in {\iid} training for closed-world scenarios.

{This thesis explores three key learning principles and the corresponding techniques required to build AI for the open-world:}
\vspace{-0.6em}
\begin{itemize}
     \setlength\itemsep{-0.3em}
    \item[\textbf{1)}] \emph{rich features} (analogy a large tool box) , a richer set of features beyond the need of \iid~generalization, helps the learning of a broader range of unseen tasks; 
    \item[\textbf{2)}] \emph{disentangled representation} (analogy an organized tool box), driven by a cheap yet reliable pressure \emph{predictive disentanglement}, reduces the number of examples required on unseen tasks; 
    \item [\textbf{3)}] \emph{inference-time learning} paradigm (analogy a tool-savvy hand) reduces the relying of designers’ priori knowledge. It employs memory-based methods at inference-time, leveraging rich features and disentangled representation constructed during pretraining. 
\end{itemize}
\vspace{-0.6em}

These principles are explored in detail in the following chapters. Chapters \ref{chap:rich_representation_ood_scope}, \ref{chap:disentanglement_and_sample_complexity}, \ref{chap:inference-time-learning} study {rich features}, {disentangled representation}, {inference-time learning} principles, respectively, as well as corresponding techniques. After that, Chapter \ref{chap:future_directions} discusses potential problems, remaining difficulties, and future directions.\footnote{Source code at: \url{https://github.com/TjuJianyu/AI_for_the_open_world}}

This thesis allocates a couple pages to introduce the learning principles, and hundred pages to techniques and experiments. These techniques and experiments are used to support and verify the learning principles. The research of \emph{AI for the open-world} is in its early stages and requires numerous experimental discoveries to nurture the research direction. Thus, this is not a thesis about language models or vision models or any applications of deep learning, but exploring learning principles to build AI for the open-world.

\chapter{Rich Features}
\label{chap:rich_representation_ood_scope}

It has been widely confirmed that the knowledge encoded in the feature extraction process is crucial for learning. This includes the practice of transferring features across related tasks \citep{tr-bottou-2011,collobert-2011,oquab-2014} and hand-crafted feature engineering \citep{automatic_feature_engineering}. Meanwhile, these feature extraction processes are inherently task-specific, meaning that each task requires its own set of features.

On the other hand, the challenge in the context of AI for \emph{open-world} is to learn a wide range of tasks quickly with fewer examples and less task-specific priori knowledge (from designers). Therefore, it is essential for an open-world AI system to prepare \emph{rich features} before encountering any new task. This chapter studies such \emph{rich features} aimed at building AI for the open-world. Specifically, this chapter addresses the following questions:
\begin{itemize}
     \setlength\itemsep{-0.3em}
    \item[\textbf{1)}] Do \emph{rich features} truly help AI for the open-world? 
    \item[\textbf{2)}] Does the traditional \iid~training paradigm for close-world AI successfully discover {rich features}? If not, what are the reasons? 
    \item[\textbf{3)}] What approaches can be used to discover {rich features}?
\end{itemize}

This chapter is organized as follows: Section \ref{sec:features} introduces a feature learning framework that studies \emph{rich features} and theoretically highlights the limitations of \iid~training in discovering \emph{rich features}. After that, Section \ref{sec:randomness} presents experiments in various domains and scenarios, illustrating these limitations, showing the benefits of rich features on learning a broad range of unseen tasks. Section \ref{sec:bonsai} and \ref{sec:very-large-dropout} provide two case studies of \emph{rich features} in invariant-learning and \ood~fine-tuning domains, respectively. 
Finally, Section \ref{sec:rich_feature_conclusion} provides conclusions.

\section{Features and representations framework}
\label{sec:features}

This section introduces a conceptual framework \cite{zhang2023learning} for talking about richness and diversity of representations. Although it seems natural to compare representations using information theory concepts such as mutual information, this approach is fraught with problems. For instance, the simplest way to maximize the mutual information $M(\Phi(x),y)$ between the representation $\Phi(x)$ and the desired output $y$ consists of making $\Phi$ equal to the identity. The information theoretic approach overlooks the main role of a feature extraction function, which is not filtering the information present in the inputs $x$, but formatting it in a manner exploitable by a simple learning system such as a linear classifier or a linear regression.\footnote{We choose linear classifiers as the ``simple learning system'' in our framework for the ease of theoretical analysis. This does not imply non-linear classifiers would behave differently. In fact, we empirically investigate another simple learning system, a cosine classifier, in the appendix Table \ref{tab:few_shot_synt_cat}.} The following framework relies on the \textbf{linear probing error} instead.

\subsection{Framework}

 This framework calls \emph{feature} a function $x\,{\mapsto}\,\varphi(x)\,{\in}\,\R$, and calls \emph{representation} a set $\Phi$ of features. Uses the notation $\w^\top\Phi(x)$ to denote the dot product $\sum_{\varphi\in\Phi} w_\varphi\,\varphi(x)$ where the coefficients $w_\varphi$ of vector $\w$ are indexed by the corresponding feature $\varphi$ and are assumed zero if $\varphi\notin\Phi$.

For simplicity, this framework assumes that the representations are exploited with a linear classifier trained with a convex loss~$\ell$. The expected loss of classifier $f$ is
\[ C_P(f) = \E_{(x,y)\sim P}\:\big[\,\ell(f(x),y)\,\big]\]
and the optimal cost achievable with representation $\Phi$ 
\begin{equation}
    \label{eq:ecost}
    C^*_P(\Phi) = \min_\w ~ C_P(f) ~~\text{with}~~ f: x\mapsto \w^\top \Phi(x)~.
\end{equation}
This construction ensures:
\begin{proposition}
\label{prop:min}
$C^*_P(\Phi_1 \cup \Phi_2) \leq C^*_P(\Phi_2)$ for all $\Phi_1$, $\Phi_2$.
\end{proposition}
Intuitively, if the combined representation $\Phi_1 \cup \Phi_2$ performs better than $\Phi_2$, then $\Phi_1$ must contain something useful that $\Phi_2$ does not. We formalize this using the word \emph{information} to actually mean \emph{linearly exploitable information about $y$}.

\begin{definition}
\label{def:more}
$\Phi_1$ contains information not present in $\Phi_2$
iff~
$C^*_P(\Phi_1\cup\Phi_2) < C^*_P(\Phi_2)$.
\end{definition}
Thanks to proposition~\ref{prop:min}, the opposite property becomes\,:
\begin{definition}
\label{def:notmore}
$\Phi_2$ contains all the information present in $\Phi_1$
iff~  $C^*_P(\Phi_1\cup\Phi_2) = C^*_P(\Phi_2)$.
\end{definition}
Finally we say that $\Phi_1$ and $\Phi_2$ carry equivalent information when $\Phi_2$ contains all the information present in $\Phi_1$, and $\Phi_1$ contains all the information present in $\Phi_2$\,:
\begin{definition}
\label{def:equiv}
$\Phi_1$ and $\Phi_2$ carry equivalent information
iff~ 
$C^*_P(\Phi_1) = C^*_P(\Phi_1 \cup \Phi_2) = C^*_P(\Phi_2)$.
\end{definition}
This definition is stronger\footnote{This is also weaker than using the quantity of information~$H$\,: writing $H(\Phi_1){=}H(\Phi_1{\cup}\Phi_2){=}H(\Phi_2)$ would imply that $\Phi_1$ and $\Phi_2$ are equal up to a bijection. Theorems~\ref{prop:ensemble} and~\ref{prop:opt} are important because this is not the case here.} than merely requiring equality $C^*_P(\Phi_1)\,{=}\,C^*_P(\Phi_2)$. In particular, we cannot improve the expected cost by constructing an ensemble\,:
\begin{theorem}
\label{prop:ensemble}
Let representations $\Phi_1$ and $\Phi_2$ carry equivalent information. Let $f_i(x){=}\w_i^{*\top}\Phi_i(x)$, for $i{\in}\{1,2\}$, be corresponding optimal classifiers. Then, for all $0{\leq}\lambda{\leq}1$\,,
\[  C^*_P(\lambda f_1 + (1-\lambda) f_2) = C^*_P(f_1) = C^*_P(f_2). \]
\end{theorem}
\begin{proof}
Let $\Phi=\Phi_1\cup\Phi_2$. Because the loss $\ell$ is assumed convex, the solutions of optimization problem \eqref{eq:ecost} form a convex set $S$. Since $C^*_P(\Phi_1){=}C^*_P(\Phi_1 \cup \Phi_2){=}C^*_P(\Phi_2)$, set $S$ contains $w^*_1$ and $w^*_2$, as well as any mixture thereof.
\end{proof}

 We now turn our attention to representations constructed by optimizing 
 both the representation $\Phi$ and the weights $\w$: 
 \begin{equation}
     \label{eq:optrepr}
     \min_\Phi C^*_P(\Phi)  = \min_{\Phi} \min_\w \E_{(x,y)\sim P} [ \ell( \w^\top \Phi(x), y) ]\,.
 \end{equation}
This idealized formulation optimizes the expected error without constraints on the nature and number of features. All its solutions problem carry equivalent information\,:
 \begin{theorem}
 \label{prop:opt}
 Let $\Phi_1$ and $\Phi_2$ be two solutions of problem~\eqref{eq:optrepr}.
 Then $\Phi_1$ and $\Phi_2$ carry equivalent information.
 \end{theorem}
\begin{proof}
Proposition~\ref{prop:min} implies $C^*_P(\Phi_1\cup\Phi_2) \leq C^*_P(\Phi_1)$.
Since $\Phi_1$ and $\Phi_2$ are both solutions of problem~\ref{eq:optrepr}, $C^*_P(\Phi_1) = C^*_P(\Phi_2) \leq C^*_P(\Phi_1\cup\Phi_2) \leq C^*_P(\Phi_1)$. 
\end{proof}

\subsection{In-distribution viewpoint}

Consider a deep network that is sufficiently overparameterized to accommodate any useful representation in its penultimate layer. Assume that we are able to optimize its expected cost on the training distribution, that is, optimize its in-distribution generalization error.  Although repeated optimization episodes need not return exactly the same representations, Theorem~\ref{prop:opt} tells us that these representations \emph{carry equivalent information}; Definition~\ref{def:equiv} tells us that we cannot either improve the in-distribution test error by linear probing, that is, by training a linear layer on top of the concatenated representations; and Theorem~\ref{prop:ensemble} tells us that we cannot improve the test error with an ensemble of such networks. The performance of ensembles depends on the diversity of their components~\cite{dietterich2000ensemble,ganaie2021ensemble}, and nothing has been done here to obtain diverse networks.

In practice, we cannot truly optimize the expected error of an overparameterized network. The representations obtained with separate training episodes tend to carry equivalent information but will not do so exactly.\footnote{Experience shows however that repeated trainings on large tasks, such as \textsc{ImageNet}, yields networks with remarkably consistent training and testing performances.} Although an ensemble of such identically trained networks can still improve both the training and testing errors, using such similarly trained networks remains a poor way to construct ensembles when one can instead vary the training data, the hyper-parameters, or vary the model structure \cite{ganaie2021ensemble}. Engineering diversity escapes the setup of Theorem~\ref{prop:opt} because each component of the ensemble then solves a different problem. This is obviously better than relying on how the real world deviates from the asymptotic setup.

\subsection{Out-of-distribution viewpoint}

Assume now that we train our network on a first data distribution $P(x,y)$, but plan to use these networks, or their representations, or their inner layers, with data that follow a different distribution $Q(x,y)$. Doing so also escapes the assumptions of our framework because the definition of representation carrying similar information (Definition~\ref{def:equiv}) critically depends on the data distribution. Representations that carry equivalent information for the training distribution $P$ need not carry equivalent information for a new distribution $Q$ at all.\footnote{Information theoretical concepts are also tied to the assumed data distribution. For instance, whether two features have mutual information critically depends on the assumed data distribution.}

Consider again representations obtained by performing multiple training episodes of the same network that only differ by their random seed.\footnote{The random seed here may determine the initial weights, the composition of the mini-batches, or the data augmentations. It does not affect the data distribution, the model structure, or even the training algorithm hyper-parameters.}  These representations roughly carry equivalent information with respect to the training distribution, but, at the same time, may be very far from carrying equivalent information with respect to a new distribution. 

If this is indeed the case, \emph{constructing an ensemble of such similarly trained networks can have a far greater effect on out-of-distribution data than in-distribution.} 
Experimental results reported in the following sections will demonstrate this effect. In fact, since we cannot know which of these representations or features might prove more informative on the new distribution, it seems wise to keep them all. \emph{Premature feature selection is not a smart way to prepare for distribution changes.}

\subsection{Optimization dynamics}

There is growing evidence that implicit regularization in deep learning networks (including $L_2$ weight decay) is related to various flavors of sparsity (\eg. \citealp{andriushchenko2022sgd,blanc2020implicit}). In an oversimplified account of this complex literature, the learning process explores the feature space more or less randomly; features that carry incrementally useful information stick more than those who do not. Consider, for instance, a network with representation~$\Phi_t$ at iteration $t$ and a feature $\varphi\in\Phi_t$ whose information is already present in $\Phi_t\!{\smallsetminus}\{\varphi\}$ in the sense of Definition~\ref{def:notmore}. This feature does not incrementally improve the performance of the training distribution and therefore may not stick. Yet this feature might contain useful information when compared to a different representation, or when compared to $\Phi_t\!{\smallsetminus}\{\varphi\}$ under a different distribution.

Explicit regularization in deep networks, such as the ubiquitous slight weight decay, also tends to destroy features that appear redundant. \citet{papyan-2020} describes how representations collapse when one trains a network for a very long time. \citet{shwartz2017opening} describe competing processes that create representations and prune representations in all layers at once. 

\begin{table}[t]
    \par\vspace*{-2ex}
    \caption{Impact of L2 weight decay on supervised transfer learning between \textsc{Cifar10} and \textsc{Cifar100}.}
    \par\vspace*{-1ex}
    \label{tab:supervise_transfer_cifar}
    \bigskip\centering
    \resizebox{0.5\textwidth}{!}{
    \begin{tabular}{c|cc}
    \toprule
     L2 weight decay    & $0$  & $5e-4$  \\
         \midrule
         \textsc{Cifar10} &
         91.41$\pm$0.81 & \textbf{94.89$\pm$0.23} \\
         \textsc{Cifar10$\rightarrow$Cifar100} &  
          \textbf{49.68$\pm$0.72} & 29.17$\pm$0.50 \\
        \midrule
         \textsc{Cifar100} &
         70.37$\pm$1.49 &\textbf{76.78$\pm$0.36} \\         
          \textsc{Cifar100$\rightarrow$Cifar10} &  
           \textbf{78.87$\pm$0.98} & 75.92$\pm$0.54 \\
    \bottomrule
    \end{tabular}
    }
    \par\vspace*{-2ex}
\end{table}

Table~\ref{tab:supervise_transfer_cifar} reports on a simple experiment to illustrate how capacity control with regularization can help in-distribution performance but hurt when the distribution changes. We pre-train a \textsc{resnet18} on the \textsc{Cifar10} task and transfer its learned representation to a \textsc{Cifar100} task by linear probing{ (see setups in appendix \ref{apx:cifar10_100_sl})}. %
Although the best in-distribution performance, 94.9\%, is achieved using a slight weight decay, the  representation learned \emph{without weight decay} transfers far better (49.7\% versus 29.2\%). The same observation holds when one reverses the role of the \textsc{Cifar10} and \textsc{Cifar100} datasets.

\section{Learning rich features}
\label{sec:randomness}

Does the dominant representation learning approach (\iid~training, foundational model), learning representation as a side effect of optimizing an expected cost for a single training distribution with a single episode, remain a good feature learning approach when we deal with multiple distributions in the \emph{open-world}? Section \ref{sec:features} theoretically proves that this approach lacks the motivation to construct a representation with rich features, regardless of model size and data size. This section uses various experiments \cite{zhang2023learning} to argue that \emph{the open-world scenario (multiple distributions) are better served by rich features, a representation that is richer than those obtained with a single optimization episode.}

The experiments in this section uses an apparently na\"{\i}ve ensembling technique: concatenating the representations obtained from multiple training episodes using the same data, model, algorithm, and hyper-parameters, but different random seeds. These independently trained networks perform similarly. Yet, in a number of scenarios involving new distributions, the concatenated representation performs substantially better than an equivalently sized network trained with a single training run. 
This proves that the representations constructed by multiple training episodes are, in fact, different. Although their concatenation carries little additional information about the training task under the training distribution, it becomes substantially more informative when tasks or distributions change. Meanwhile, a single training episode is unlikely to yield such a redundant representation because the optimization process has no reason to accumulate features that do not incrementally improve training performance.

Subsections \ref{sec:supervisedtransfer}, \ref{sec:ssltransfer}, \ref{sec:metalearning}, and \ref{sec:oodlearning} carry out experiments on supervised transfer-learning, self-supervised transfer-learning, meta\&few-shot learning, and \ood~robustness areas, respectively.

\subsection{Supervised transfer learning}
\label{sec:supervisedtransfer}

This section focuses on supervised transfer learning scenarios in which the representation learned using an auxiliary supervised task, such as the \textsc{ImageNet} object recognition task \citep{imagenet_cvpr09}, is then used for the target tasks, such as, for our purposes, the \textsc{Cifar10}, \textsc{Cifar100}, and \textsc{Inaturalist18} (\textsc{Inat18}) object recognition tasks \citep{krizhevsky2009learning,van2018inaturalist}. We distinguish the \emph{linear probing} scenario where the penultimate layer features of the pre-trained network are used as inputs for linear classifiers trained on the target tasks, and the \emph{fine tuning} scenario which uses back-propagation to further update the transferred features using the target task training data.\footnote{Code is available at \url{https://github.com/TjuJianyu/RRL}}

\begin{table}[ht!]
    \caption{Supervised transfer learning from \textsc{ImageNet} to \textsc{Inat18}, \textsc{Cifar100}, and \textsc{Cifar10} using linear probing. The \textsc{erm} {(empirical risk minimization)} rows provide baseline results. The \synthcat$n$ rows use the concatenated representations of $n$ separately trained networks.}
    \label{tab:imagenet_sl_lineareval}
    \bigskip
    \centering
    
    \resizebox{0.75\textwidth}{!}{
    \begin{tabular}{c cc  |c| ccc }
    \toprule
                 &              &        &    {\small ID}             &      \multicolumn{3}{c}{\small Linear Probing (OOD)} \\
         method  & architecture     & params & \textsc{\small imagenet} &  \textsc{\small inat18}  & \textsc{\small cifar100} & \textsc{\small cifar10} \\
         \midrule
             \textsc{erm} & \textsc{resnet50}     & 23.5\textsc{m} & 75.58 & 37.91 & 73.23 & 90.57   \\
             \textsc{erm} & \textsc{resnet50w2}   & 93.9\textsc{m} & 77.58	& 37.34 & 72.65 & 90.86		\\
             \textsc{erm} & \textsc{resnet50w4}   & 375\textsc{m}  & 78.46	& 38.71 & 74.81	& 92.13	\\
            \midrule
             \textsc{erm} & 2$\times$\textsc{resnet50}    & 47\textsc{m}   & 75.03	& 39.34 & 74.36	& 90.94	\\
             \textsc{erm} & 4$\times$\textsc{resnet50}    & 94\textsc{m}   & 75.62	& 41.89 & 74.06	& 90.61	\\
             \midrule
             {\synthcat}2& 2$\times$\textsc{resnet50}     & 47\textsc{m}   & 77.57 & 43.26	& 76.10	& 91.86	\\
             {\synthcat}4& 4$\times$\textsc{resnet50}     & 94\textsc{m}   & 78.15 & 46.55	& 78.19	& 93.09	\\
            {\synthcat}10& 10$\times$\textsc{resnet50}    & 235\textsc{m}  & 78.36 & 49.65 	& 79.61	& 93.75	\\
        \bottomrule
    \end{tabular}
    }
\end{table}

\begin{figure*}[t]
    \centering
    \includegraphics[width=\textwidth]{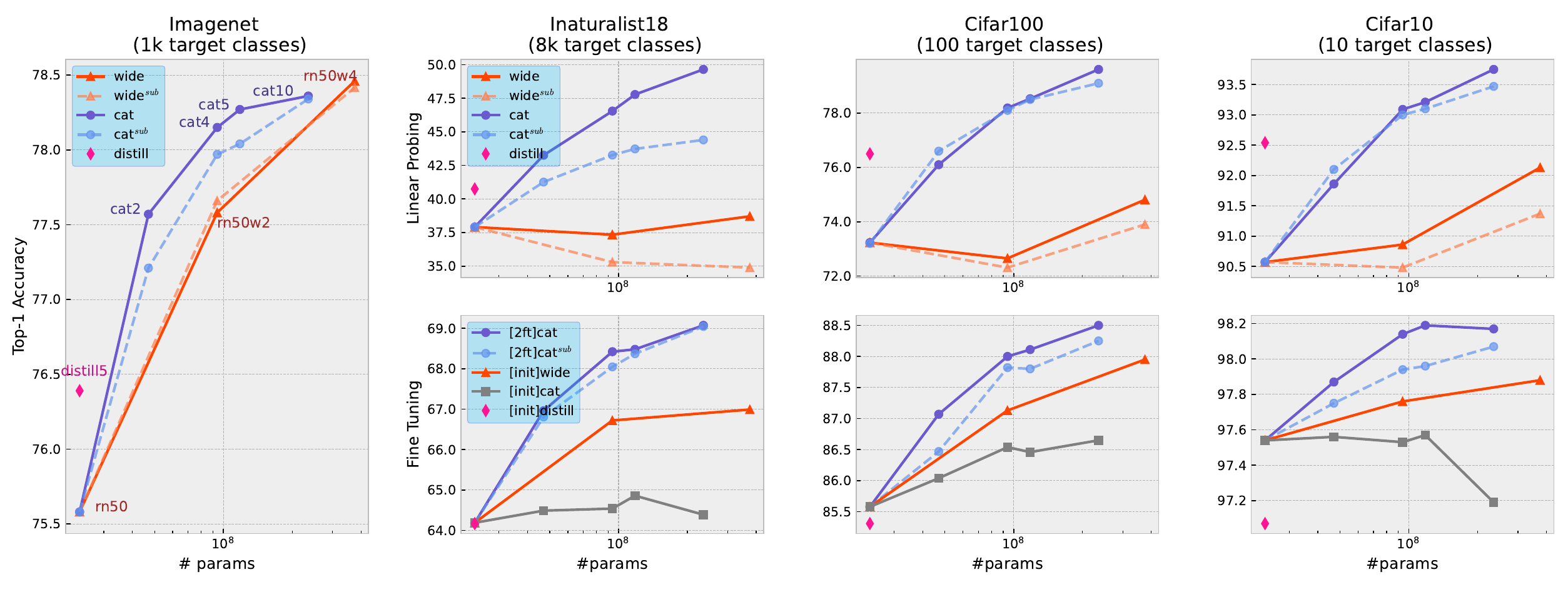}
    \par\vspace*{-2ex}
    \caption{Supervised transfer learning from \textsc{ImageNet} to \textsc{Inat18}, \textsc{Cifar100}, and \textsc{Cifar10}.  The top row shows the superior linear probing performance of the \synthcat$n$ networks (blue, \plotlabel{cat}). The bottom row shows the performance of fine-tuned \synthcat$n$, which is poor with normal fine-tuning (gray, \plotlabel{[init]cat}) and excellent for two-stage fine tuning (blue, \plotlabel{[2ft]cat}). \synthdistill$n$ (pink, \plotlabel{distill}) representation is obtained by distilling \synthcat$n$ into one \textsc{resnet50} (we omit $\synthdistill$ in this section due to the space limit. see details in the appendix \ref{apx:imagenet_sl}). 
    }
    \label{fig:imagenet_sl_ft_2ft}
    \par\vspace*{-1ex}
\end{figure*}
\bfparagraph{Linear probing}

The first three rows of Table~\ref{tab:imagenet_sl_lineareval}, labeled \textsc{erm}, provide baselines for the linear probing scenario, using respectively a \textsc{resnet50} network \citep{he-2016}, as well as larger variants \textsc{resnet50w}$n$ with $n$ times wider internal representations and roughly $n^2$ times more parameters.  The following two rows provide additional baseline results using networks \textsc{$n\times$resnet50} composed of respectively $n$ separate \textsc{resnet50} networks joined by concatenating their penultimate layers. Although these networks perform relatively poorly on the pre-training task \textsc{ImageNet}, their linear probing performance is substantially better than that of the ordinary \textsc{resnet}s.

The final three rows of Table~\ref{tab:imagenet_sl_lineareval}, labeled \synthcat$n$, are obtained by training $n$ separate \textsc{resnet50} networks on \textsc{ImageNet} with different random seeds, and using their concatenated representations as inputs for a linear classifier trained on the target tasks. This approach yields linear probing performances that substantially exceed that of comparably sized baseline networks. Remarkably, \synthcat$n$, with separately trained components, outperforms the architecturally similar \textsc{$n\times$resnet50} trained as a single network. See appendix \ref{apx:imagenet_sl} for experimental details.

These results are succinctly\footnote{In order to save space, all further results in the main text of this contribution are presented with such plots, with result tables provided in the appendix.} represented in the top row of Figure~\ref{fig:imagenet_sl_ft_2ft}. For each target task \textsc{Inat18}, \textsc{Cifar100}, and \textsc{Cifar10}, the solid curves show the linear probing performance of the baseline \textsc{resnet50w}$n$ (red, labeled \plotlabel{wide}) and of the \synthcat$n$ networks (blue, \plotlabel{cat}) as a function of the number of parameters of their inference architecture. 

The left plot (double height) of Figure~\ref{fig:imagenet_sl_ft_2ft} provides the same information in-distribution, that is, using the pre-training task as target task. In-distribution, the advantage of \synthcat$n$ vanishes when the networks become larger, possibly large enough to approach the conditions of Theorem~\ref{prop:opt}. The out-of-distribution curves (top row) are qualitatively different because they show improved performance all along.

An ensemble of $n$ \textsc{resnet50} networks is architecturally similar to the \synthcat$n$ models. Instead of training a linear classifier on the concatenated features, the ensemble averages $n$ classifiers
independently trained on top of each network. Whether this is beneficial depends on the nature of the target task and its training data (dashed blue, labeled \plotlabel{cat$^{\sf sub}$}). For completeness, we also present an ensemble baseline (dashed red plot, labeled \plotlabel{wide$^{\sf sub}$}) averaging $n$ linear classifiers trained on top of a random partition of the corresponding wide network features.

\bfparagraph{Fine-tuning} 

Having established, in the linear probing case, that transferring concatenated representations \synthcat$n$ outperforms transferring the representation of an equivalently sized network, we turn our attention to fine-tuning.

Fine-tuning is usually achieved by setting up a linear classifier on top of the transferred feature and training it on the target task data while allowing back-propagation to update the transferred features as well. The bottom row of Figure~\ref{fig:imagenet_sl_ft_2ft} shows the performance of this approach using the baseline network representations (red curve, 
 labeled \plotlabel{[init]wide}) and the concatenated representations (gray curve, labeled \plotlabel{[init]cat}), The latter perform very poorly.\footnote{The poor performance of plain fine-tuning had already been pointed out by \citet{kumar2022finetuning} and \citet{kirichenko2022last}.}

 \begin{figure}[t]
\centering
\includegraphics[height=0.24\linewidth]{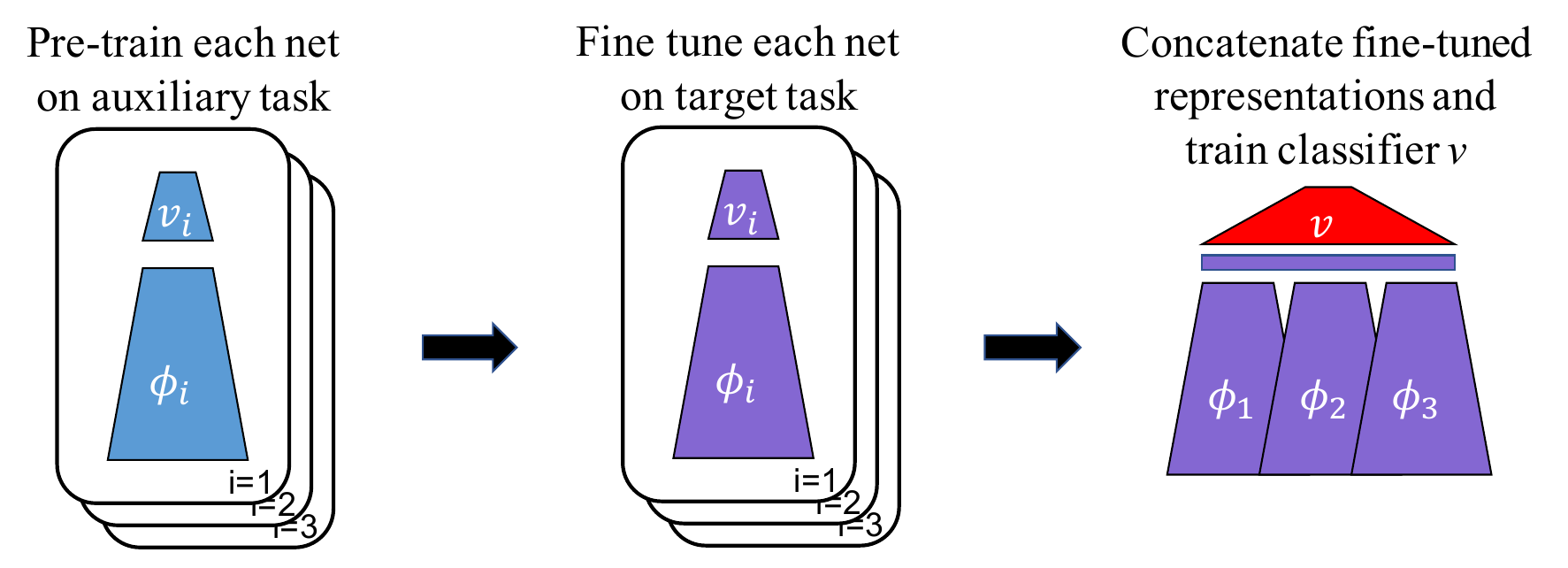}
\caption{\emph{Two-stage fine-tuning} consists of fine-tuning each network separately, then concatenating their feature extractors, now frozen, and training a final classifier. }
\label{fig:twostageft}
\end{figure}

\begin{figure*}[ht]
    \centering
    \includegraphics[height=0.395\textwidth]{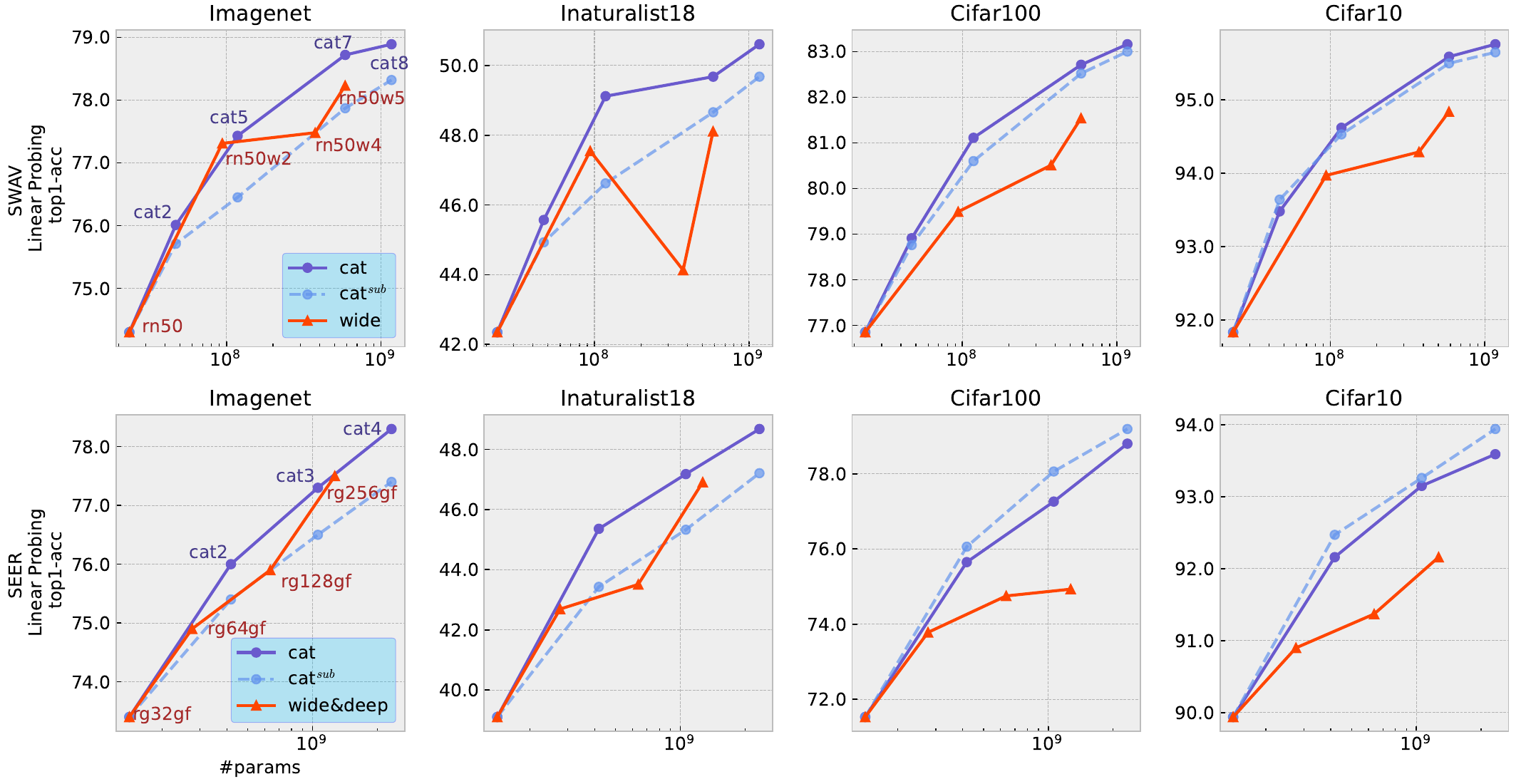}~~~~~
    \includegraphics[height=0.395\textwidth]{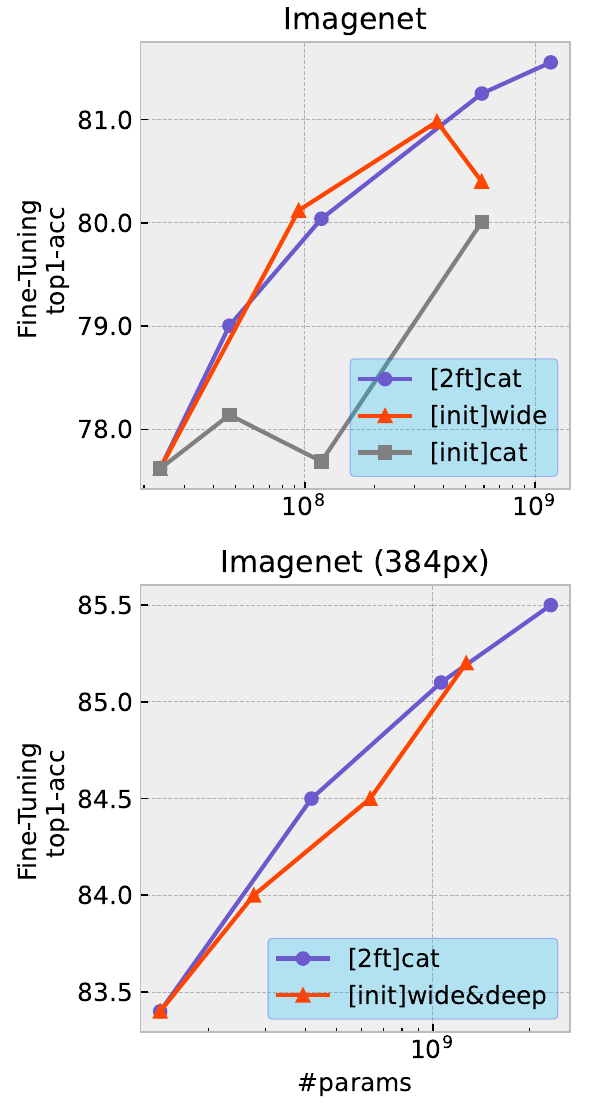}
    \caption{Self-supervised transfer learning with \textsc{swav} trained on unlabeled \textsc{ImageNet(1k)} (\emph{top row}) and with \textsc{seer} on \textsc{Instagram1B} (\emph{bottom row}). The constructed rich representation, \synthcat$n$, yields the best linear probing performance (\plotlabel{cat} and \plotlabel{cat$^{\sf sub}$}) for supervised \textsc{ImageNet}, \textsc{Inat18}, \textsc{Cifar100}, and \textsc{Cifar10} target tasks. The two-stage fine-tuning (\plotlabel{[2ft]cat}) matches equivalently sized baseline models (\plotlabel{[init]wide} and \plotlabel{[init]wide\&deep}), but with much easier training. The sub-networks of \synthcat5 (and \synthcat2) in \textsc{swav} hold the same architecture. Due to the space limitation, we put other fine-tuning curves in appendix \ref{apx:swav_additional_exp}.} 
    \label{fig:ssl_tf}
\end{figure*}

\begin{figure}
    \centering
    \includegraphics[width=0.6\textwidth]{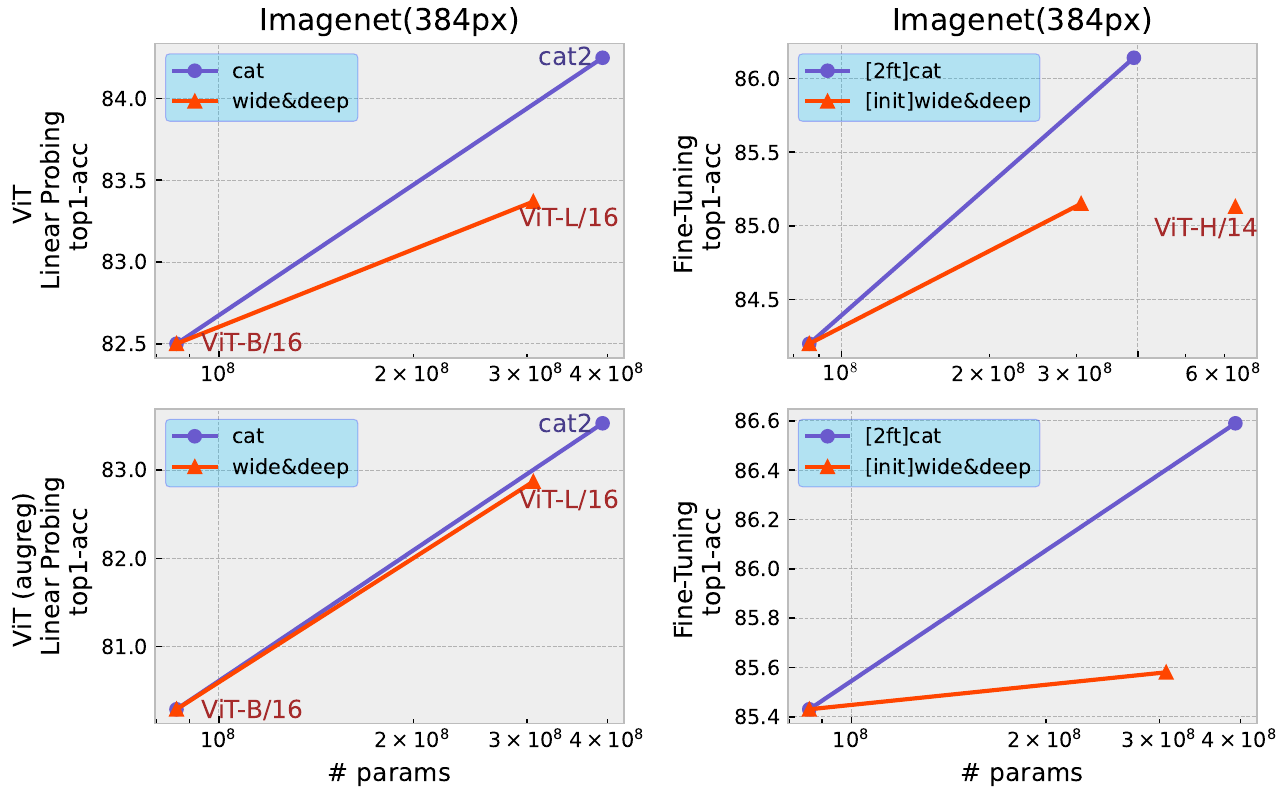}
    \caption{{Supervised transfer learning from \textsc{ImageNet21k} to \textsc{ImageNet} on vision transformers.}}
    \label{fig:imagenet_vit_sl_tf}
\end{figure}

We posit that fine-tuning with a single training episode impoverishes the initially rich representation. Instead, we propose \emph{two-stage fine tuning} which consists
of separately training $n$ networks on the pre-training task, separately fine-tuning them on the target task, and finally training a linear classifier on top of the concatenation of the $n$ separately fine-tuned representations (Figure~\ref{fig:twostageft}). The superior performance of two-stage fine-tuning is clear in the bottom row of Figure~\ref{fig:imagenet_sl_ft_2ft} (blue solid curve, labeled \plotlabel{{[2ft]}cat}). Ensembles of fine-tuned networks perform almost as well (blue dashed curve, labeled \plotlabel{[2ft]cat$^{\sf sub}$}).

\bfparagraph{Vision transformers}

Figure~\ref{fig:imagenet_vit_sl_tf} shows that transformer networks behave similarly. We carried out supervised transfer experiments using the original vision transformer, \textsc{ViT}, \citep{dosovitskiy2020image}, and using a more advanced version using carefully crafted data augmentations and regularization, \textsc{ViT(augreg)},  \citep{steiner2021train}. 
We use two transformers of two different sizes, {ViT-B/16} and {ViT-L/16}, pre-trained on \textsc{ImageNet21k}.\footnote{Checkpoints provided at \url{https://github.com/google-research/vision_transformer}.} Supervised transfer baselines (red, \plotlabel{wide\&deep} or \plotlabel{[init]wide\&deep}) are obtained by linear-probing and by fine-tuning on \textsc{ImageNet(1k)}. These baselines are outperformed by respectively linear-probing and \emph{two-stage fine tuning} on top of the concatenation of their final representations (\synthcat2). 

An even larger transformer architecture, {ViT-H/14}, yields about the same \textsc{ImageNet1k} fine-tuning performance as {ViT-L/16}, but lags 1\% behind \synthcat2, despite having twice as many parameters \cite{dosovitskiy2020image}. 
Experiments with two-stage fine-tuned \synthcat2 in \textsc{ViT(augreg)} show even better results, possibly because changing the random seed does not just changes the initial weights and the mini-batch composition, but also affects the data augmentations of the \textsc{ViT(augreg)} networks.

\subsection{Self-supervised transfer learning}
\label{sec:ssltransfer}

In self-supervised transfer learning (SSL), transferable representations are no longer constructed using a supervised auxiliary task, but using a training criterion that does not involve tedious manual labeling. We focus on schemes that rely on the knowledge of a set of acceptable pattern transformations. The training architecture then resembles a siamese network whose branches process different transformations of the same pattern. The SSL training objective must then balance two terms: on the one hand, the representations computed by each branch must be close or, at least, related; on the other hand, they should be prevented from collapsing partially \citep{jing2021understanding} or catastrophically \citep{chen2020simsiam}.  Although this second term tends to fill the representation with useful features, what is necessary to balance the SSL training objective might still exclude potentially useful features for the target tasks.

This section presents results obtained using \textsc{swav} pre-training using 1.2 million \textsc{ImageNet} images \citep{caron2020unsupervised} and using \textsc{seer} pre-training using 1 billion \textsc{Instagram1B} images \citep{goyal2022vision}. These experiments leverage the pre-trained models made available by the authors: five \textsc{resnet50} (four from our reproduction), one \textsc{resnet50w2}, one \textsc{resnet50w4} and one \textsc{resnet50w5} for the \textsc{swav} experiments;\footnote{\url{https://github.com/facebookresearch/swav}} one \textsc{regnet32gf}, one \textsc{regnet64gf}, one \textsc{regnet128gf}, and one \textsc{regnet256gf} (1.3B parameters) for the \textsc{seer} experiments.\footnote{\url{https://github.com/facebookresearch/vissl/tree/main/projects/SEER}}

The first four columns of Figure~\ref{fig:ssl_tf} present linear probing results for four target object recognition tasks: supervised \textsc{ImageNet}, \textsc{Inaturalist18}, \textsc{Cifar100}, and \textsc{Cifar10}. The baseline curves (red, labeled \plotlabel{wide} or \plotlabel{wide\&deep}) plot the performance of linear classifiers trained on top of the pre-trained SSL representations. The solid \synthcat$n$ curves were obtained by training a linear classifier on top of the concatenated representations of the $n$ smallest SSL pre-trained representations (solid blue, \plotlabel{cat}). The dash \synthcat$n$ curves train an ensemble of $n$ small classifiers on subsets of the concatenated representation (dash blue, \plotlabel{cat$^{\sf sub}$}).\footnote{Likewise the supervised transfer learning experiments, each small classifier learns on the representation of a sub-network (e.g. \textsc{regnet32gf}, \textsc{regnet64gf}). Now the representation subset cannot be treated as random subsets of the concatenated representation anymore, because the model architectures are not always the same. So we omit the ensemble classifiers for red curves.} Overall, the \synthcat$n$ approach offers the best performance.

The last column of Figure~\ref{fig:ssl_tf} presents results with fine-tuning for the supervised \textsc{ImageNet} task. Our \emph{two-stage fine-tuning} approach (as Figure~\ref{fig:twostageft}) matches the performance of equivalently sized baseline networks. In particular, the largest \synthcat4 model using \textsc{seer} pre-training, with 2.3B parameters, achieves 85.5\% correct classification rate, approaching the 85.8\% rate of the largest baseline network in \textsc{seer} \citep{goyal2022vision}, \textsc{regnet10B} with 10B parameters. Of course, separately training and fine-tuning the components of the $\synthcat$4 network is far easier than training a single \textsc{regnet10B} network.

Additional results using \textsc{SimSiam} \citep{chen2020simple} and with distillation are provided in appendix \ref{apx:simsiam_cifar}. Other experiment details are provided in appendix \ref{apx:ssl}.

\subsection{Meta-learning \& few-shots  learning}
\label{sec:metalearning}

\begin{figure}[ht]
    \centering
    \includegraphics[width=0.55\textwidth]{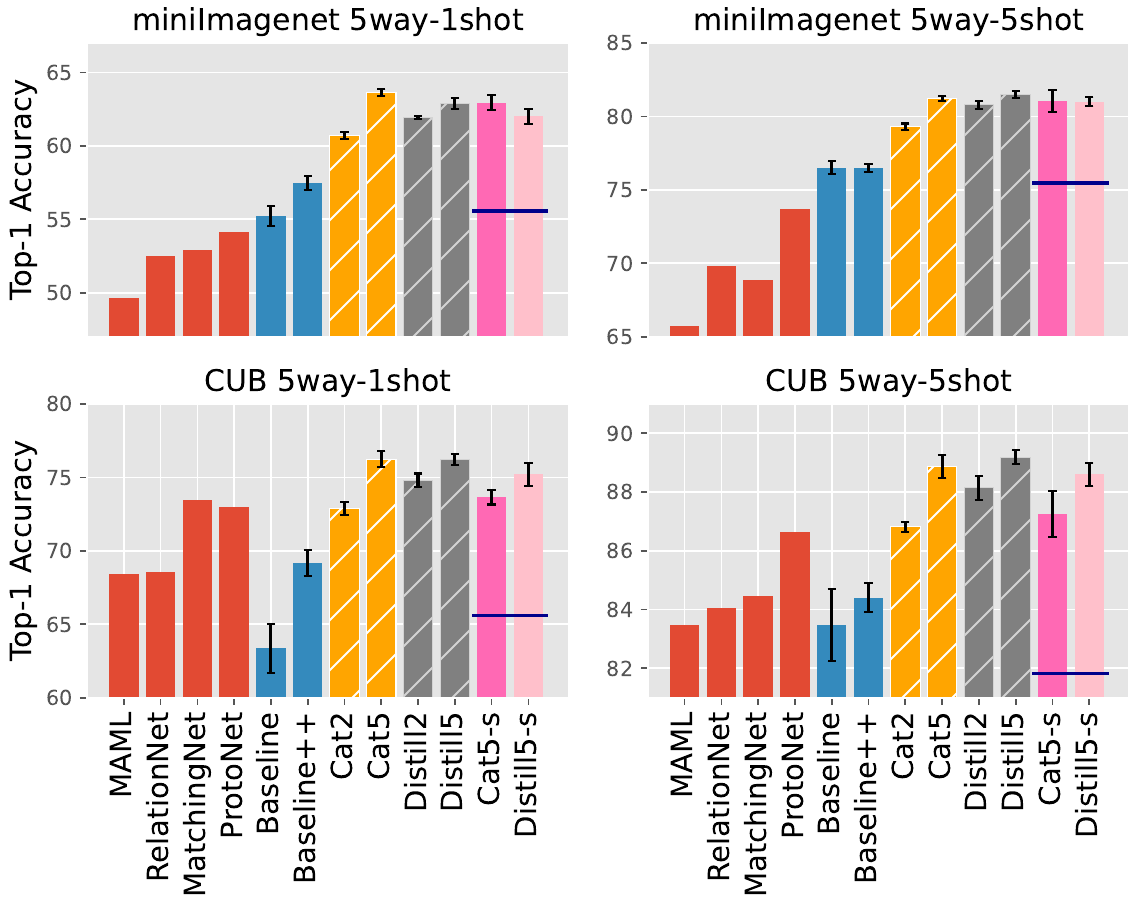}
    \caption{Few-shot learning performance on \textsc{MiniImageNet} and \textsc{Cub}. Four common few-shot learning algorithms are shown in red (results from \citet{closelookatfewshot}). Two supervised transfer methods, with either a linear classifier (\textsc{Baseline}) or cosine-based classifier (\textsc{Baseline++}) are shown in blue. The $\synthdistill$ and $\synthcat$ results, with a cosine-base classifier, are respectively shown in orange and gray. The \synthcat5\textsc{-s} and \synthdistill5\textsc{-s} results were obtained using five snapshots taken during a single training episode with a relatively high step size. The dark blue line shows the best individual snapshot. Standard deviations over five repeats are reported.}
    \label{fig:few_shots}
\end{figure}
Each target task in the few-shots learning scenario comes with only a few training examples. One must then consider a large collection of target tasks to obtain statistically meaningful results.

We follow the setup of \citet{closelookatfewshot}\footnote{We are aware of various existing few-shot benchmarks, such as MetaDataset \citep{triantafillou2019meta}, that contain more datasets than \citet{chen2020simple}. We choose \citet{chen2020simple}, because it is enough to validate our ideas in section \ref{sec:features}.
} in which the base task is an image classification task with a substantial number of classes and examples per class, and the target tasks are five-way classification problems involving novel classes that are distinct from the base classes and come with only a few examples. Such a problem is often cast as a \emph{meta learning} problem in which the base data is used to learn how to solve a classification problem with only a few examples. \citet{closelookatfewshot} find that excellent performance can be achieved using simple baseline algorithms such as supervised transfer learning with linear probing (\textsc{Baseline}) or with a cosine-based final classifier (\textsc{Baseline++}). These baselines match and sometimes exceed the performance of common few-shots algorithms such as \textsc{maml} \citep{finn2017model}, \textsc{RelationNet} \citep{sung2018learning}, \textsc{MatchingNet} \citep{matchingnet}, and \textsc{ProtoNet} \citep{protonet}.

Figure~\ref{fig:few_shots} reports results obtained with a \textsc{resnet18} architecture on both the \textsc{MiniImageNet} \citep{matchingnet} and \textsc{Cub} \citep{wah2011caltech} five ways classification tasks with either one or five examples per class as set up by \citet{closelookatfewshot}. The \textsc{maml}, \textsc{RelationNet}, \textsc{MatchingNet}, and \textsc{ProtoNet} results (red bars) are copied verbatim from \citep[table A5]{closelookatfewshot}. The \textsc{Baseline} and \textsc{Baseline++} results were further improved by a systematic L2 weight decay search procedure (see appendix \ref{apx:meta_baseline_pretrain}). All these results show substantial variations across runs, about $4\%$ for \textsc{Cub} and $2\%$ for \textsc{MiniImageNet}.

\begin{figure}
\centering
    \includegraphics[width=0.5\textwidth]{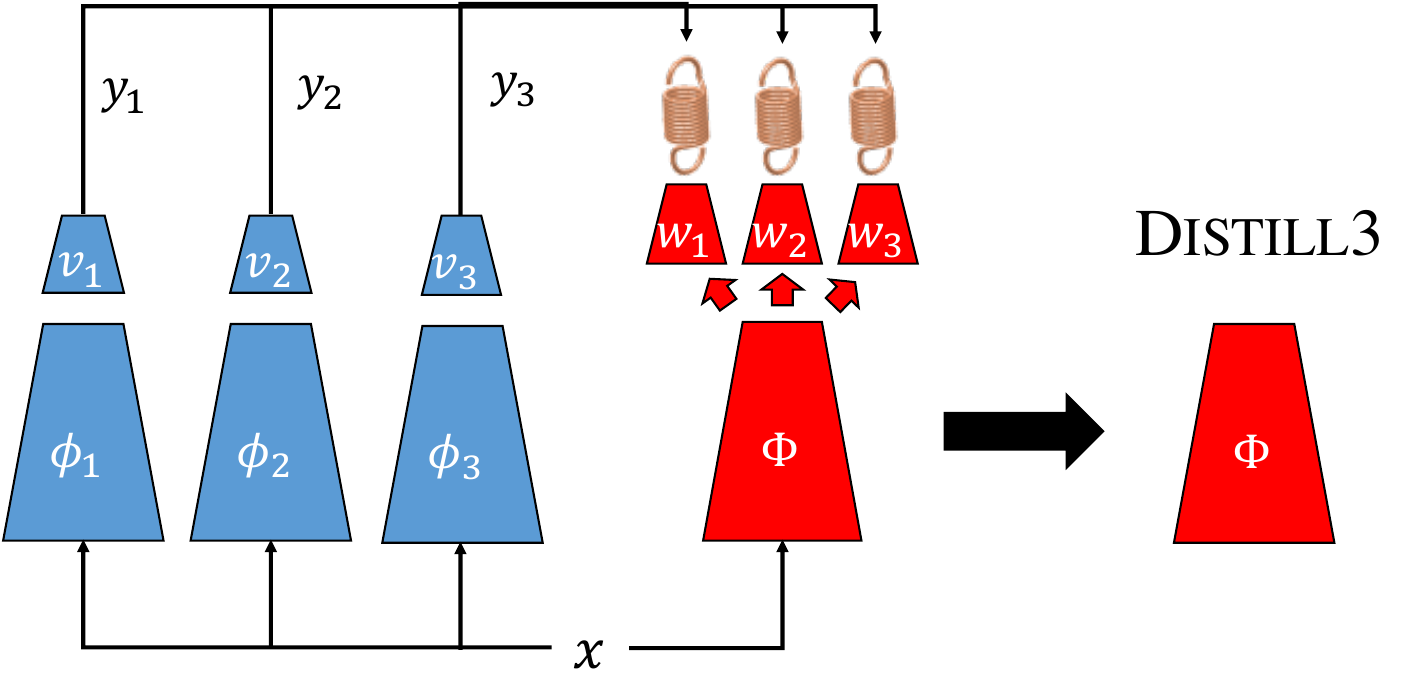}
    \caption{(\synthdistill$n$) A multiple head network (red) trained to predict the outputs of the pre-trained networks  $\Phi_1,\Phi_2,\cdots$ (blue) must develop a representation $\Phi$ that subsumes those of all the blue networks. The same distillation process is used by the \textsc{Bonsai} algorithm \citep{zhang2022rich} but after training the networks with adversarially re-weighted data. }
    \label{fig:distill} 
\end{figure}

The \synthcat$n$ and \synthdistill$n$ results were then obtained by first training $n$ \textsc{resnet18} on the base data with different seeds, constructing a combined (rich) representation by either concatenation or  distillation (as Figure~\ref{fig:distill}), then, for each task, training a cosine distance classifier using the representation as input. Despite the high replication variance of the competing results, both $\synthdistill$ and $\synthcat$ show very strong performance. Note that naively increasing model architecture, e.g. from \textsc{resnet18} to \textsc{resnet34}, can only gain limited improvements ($\leq 1\%$, \citet{chen2020simple}, table A5) and is still lagging behind $\synthcat$ and $\synthdistill$. 

The pink bars (\synthcat5\textsc{-s} and \synthdistill5\textsc{-s}) in Figure~\ref{fig:few_shots}, concatenate or distill five snapshots taken at regular intervals during a single training episode with a relatively high step size (0.8), achieve a similar few-shots learning performance as \synthcat5 and \synthdistill5, perform substantially better than the best individual snapshot (dark blue line). \emph{It implies that diverse features are discovered and then abandoned but not accumulated during the optimization process.} More results and details, as well as a comparison with conditional meta-learning algorithms \cite{wang2020structured, denevi2022conditional, rusu2018meta}, are shown in appendix \ref{apx:meta-learning}.

\subsection{Out-of-distribution generalization}
\label{sec:oodlearning}

In the out-of-distribution generalization scenario, we seek a model that performs well on a family of data distributions, also called environments, on the basis of a finite number of training sets distributed according to some of these distributions. \citet{irm} propose an invariance principle to solve such problems and propose the \textsc{IRMv1} algorithm which searches for a good predictor whose final linear layer is simultaneously optimal for all training distributions.
Since then, a number of algorithms exploiting similar ideas have been proposed, such as \textsc{vREx} \citep{vrex}, \textsc{Fishr} \citep{rame2021fishr}, or \textsc{CLOvE} \citep{wald2021calibration}. Theoretical connections have been made with multi-calibration \citep{hebertjohnson18a,wah2011caltech}. Alas, the performance of these algorithms remains wanting \citep{gulrajani2021in}.
\citet{zhang2022rich} attribute this poor performance to the numerical difficulty of optimizing the complicated objective associated with these algorithms. They propose to work around these optimization problems by providing initial weights that already extract a rich palette of potentially interesting features constructed using the \textsc{Bonsai} \citep{zhang2022rich}  algorithm.

Following \citet{zhang2022rich}, we use the \textsc{Camelyon17} tumor classification dataset \citep{bandi2018detection} which contains medical images collected from five hospitals with potentially different devices and procedures. As suggested in \citet{koh2021wilds}, we use the first three hospitals as training environments and the fifth hospital for testing. \textsc{ood}-tuned results are obtained by using the fourth hospital to tune the various hyper-parameters. \textsc{iid}-tuned results only use the training distributions (see details in appendix \ref{apx:ood}). The purpose of our experiments is to investigate whether initializing with the {\synthdistill} or {\synthcat} algorithm provides a computationally attractive alternative to \textsc{Bonsai}.

\begin{table}[t]
    \caption{Test accuracy on the \textsc{Camelyon17} dataset with \textsc{DenseNet121}. We compare various initialization (\textsc{ERM}, \synthcat$n$, \synthdistill$n$, and \textsc{Bonsai}) for two algorithms \textsc{vREx} and \textsc{ERM} using either the \textsc{iid} or \textsc{ood} hyperparameter tuning method. The standard deviations over 5 runs are reported.    }
    \label{tab:camelyon17_synt_cat}
    \centering
    \bigskip
    \resizebox{0.585\textwidth}{!}{
    \begin{tabular}{c | cc |cc}
    \toprule
    &      \multicolumn{2}{c|}{\small IID-Tune}      & \multicolumn{2}{c}{\small OOD-Tune} \\
     & {\small \textsc{vREx}} & {\small \textsc{ERM}} & {\small \textsc{vREx}} & {\small \textsc{ERM}} \\
                   \midrule
{\small \textsc{ERM}} & 69.6$\pm$10.5   &   66.6$\pm$9.8   &   70.6$\pm$10.0  &   70.2$\pm$8.7 \\
\midrule
{\synthcat}2 & 74.3$\pm$8.0   &   74.3$\pm$8.0   &   73.7$\pm$8.1   &   74.2$\pm$8.1 \\
{\synthcat}5 & 75.2$\pm$2.9   &   75.0$\pm$2.7   &   74.9$\pm$3.3   &   75.1$\pm$2.8 \\
{\synthcat}20 & 76.4$\pm$0.5   &   76.5$\pm$0.5   &   76.8$\pm$0.9   &   76.4$\pm$0.9 \\
\midrule
{\synthdistill}2 & 67.1$\pm$4.7   &   66.9$\pm$4.8   &   67.4$\pm$4.3   &   66.7$\pm$4.2 \\
{\synthdistill}5 & 69.9$\pm$7.4   &   69.9$\pm$6.9   &   71.8$\pm$5.0   &   69.9$\pm$6.3 \\

{\synthdistill}20 & 73.3$\pm$2.5   &   73.2$\pm$2.3   &   74.8$\pm$3.2   &   73.1$\pm$2.7 \\

   \midrule
    \textsc{Bonsai}2\tablefootnote{We apply \textsc{Bonsai} algorithm with 2 discovery episodes.} &  77.9$\pm$2.7 &  78.2$\pm$2.6 & 79.5$\pm$2.7 & 78.6$\pm$2.6 \\ 
    \bottomrule
    \end{tabular}
    }
\end{table}

Table~\ref{tab:camelyon17_synt_cat} compares the test performance achieved by two algorithms, \textsc{vREx} and \textsc{ERM}, after initializing with \textsc{ERM}, \synthcat$n$, \synthdistill$n$, and \textsc{Bonsai}, in both the \textsc{iid}-tune and \textsc{ood}-tune scenarios.  The {\synthcat} and {\synthdistill} initialization perform better than \textsc{ERM} but not as well as \textsc{Bonsai}. \emph{This result clearly shows the need to research better ways to train networks in a manner that yields diverse representations.} Although this contribution shows that simply changing the seed (as in {\synthcat} and {\synthdistill}) can achieve good results, the experience of deep ensembles \citep{gontijo-lopes2022no} suggests that more refined diversification methods might yield substantially better representations.

\subsection*{Conclusion}

Using a simple theoretical framework and a broad range of experiments, we show that deep learning scenarios that involve changing tasks or distributions are \emph{better served by representations that are richer than those obtained with a single optimization episode (\iid~training).} In a time where many organizations
deploy considerable resources training huge foundational
models, this conclusion should be sobering. 

The simple multiple-training-episode approach $\synthcat$ constructs such richer representation with excellent performances in various scenarios. The \emph{two-stage fine tuning} method works around the poor performance of normal fine-tuning in various transfer scenarios.

More importantly, this section provides a lot of room for new representation learning algorithms that move away from relying solely on a single optimization episode.

\section{Case study 1: Rich feature in invariant-learning}
\label{sec:bonsai}

The theoretical framework and experimental results presented in Section \ref{sec:features} reveal the limitation of \iid~training in discovering rich features, highlights the benefits of rich features in transfer-learning across different distributions. This section provides a case study of rich features in the invariant-learning domain, where multiple training environments are provided to illustrate the range of potential distribution changes. In this case study, rich features are created by the \textsc{Bonsai}\footnote{Bonsai, also known as \emph{penjing} (tray planting), refers to the art of growing small trees in trays using obsessive trimming techniques to impede their growth and produce miniature versions of real-life trees. Likewise, the Bonsai algorithm impedes the learning process in order to obtain diverse representations.} algorithm \cite{zhang2022rich} through an adversarial process. The experimental results not only demonstrate the significant benefits of rich features to invariant learning, but also reveal that \emph{it is {rich feature} rather than the commonly believed invariant-learning penalties that matters in the invariant-learning domain} (Table \ref{tab:camelyon_full_results}). That is, rich feature changes the field of invariant-learning.

This section is organized as follows. Section \ref{sec:spurious_tour} introduces the background of invariant learning, and an awkward optimization-generalization dilemma posed by many invariant learning methods \cite{zhang2022rich}. Following that, Section \ref{sec:bonsai} introduces the \textsc{Bonsai} method, which constructs rich features through adversarial discovery. Finally, Section \ref{sec:experiments} shows interesting results in both synthetic and realistic tasks, showing a significant improvement in rich features for all invariant-learning algorithms.

\subsection{Invariant-learning}
\label{sec:spurious_tour}

In order to achieve good performance on testing (inference) data that follows a different distribution from the training data, many invariant-learning methods assume access to multiple training datasets, or environments, whose different distributions illustrate a range of potential distribution changes. 

To this goal, on disrection is to learn representation such that the optimal classifier built on top is the same for all training environments: IRMv1/IRM \cite{irm}, MAML-IRM \cite{mamlirm}, CLOvE \cite{wald2021calibration}. Another line of work introduces gradient alignment constraints across training environments using dot-product (Fish \cite{gm}), squared distance of gradients (IGA \cite{iga}), or squared distance of gradients variance (Fishr \cite{rame2021fishr}).
Methods such as vREx \cite{vrex} and GroupDRO \cite{groupDRO} aim at finding a solution that performs equally well across training environments. 

In practice, these learning constraints are encoded as additional penalty terms apart from the empirical risk minimization term (ERM):
\begin{align}
    \frac{1}{n_e} \sum_{e} C_e(\theta) + \kappa \sum_e\Omega_e(\theta)\,,
\end{align}
where $\Omega_e(\theta)$ refers to the additional penalty term on environment (or training dataset) $e$, $C_e$ refers to the ERM term on $e$, $\theta$ indicates the parameters to be optimized. However, it is tricky to schedule the penalty weight $\kappa$ to satisfy both the \ood~generalization goal and an easy optimization process at the same time. This awkward stage is referred as \textit{optimization-generalization dilemma}.

The follow paragraphs present experiments that illustrate the optimization-generalization dilemma that plagues invariant-learning methods. All these experiments are carried out on the \textsc{ColorMNIST} task \citep{irm}. In this task, the relation between the robust feature (the digit class) and output label is invariant in all training and testing  environments. In contrast, although the spurious feature (the digit color) is more predictive on the training environments, its relation with the output labels is not invariant across environments.
We report results on a variety of published invariant-learning algorithms: IRMv1 \cite{irm}, Fish \cite{gm}, IGA \cite{iga}, vREx \cite{vrex}, Spectral Decoupling (SD) \cite{sd}, Fishr \cite{rame2021fishr}, RSC \cite{rsc}, LfF \cite{lff}, and CLOvE \cite{wald2021calibration}. We do not report results on MAML-IRM \cite{mamlirm} because it is equivalent to Fish+vREx, and we do not report results on GroupDRO \citep{groupDRO} because it performs like vREx (see Appendix \ref{apdix:mamlirm=vrex+gm} and \ref{apdix:groupdro_vrex_inter_extra} for details).

\bfparagraph{Invariant penalties make the optimization challenging}
\label{sec:importance_of_initialization_ood}

\begin{figure}[th!]
    \centering
    \includegraphics[width=0.55\textwidth]{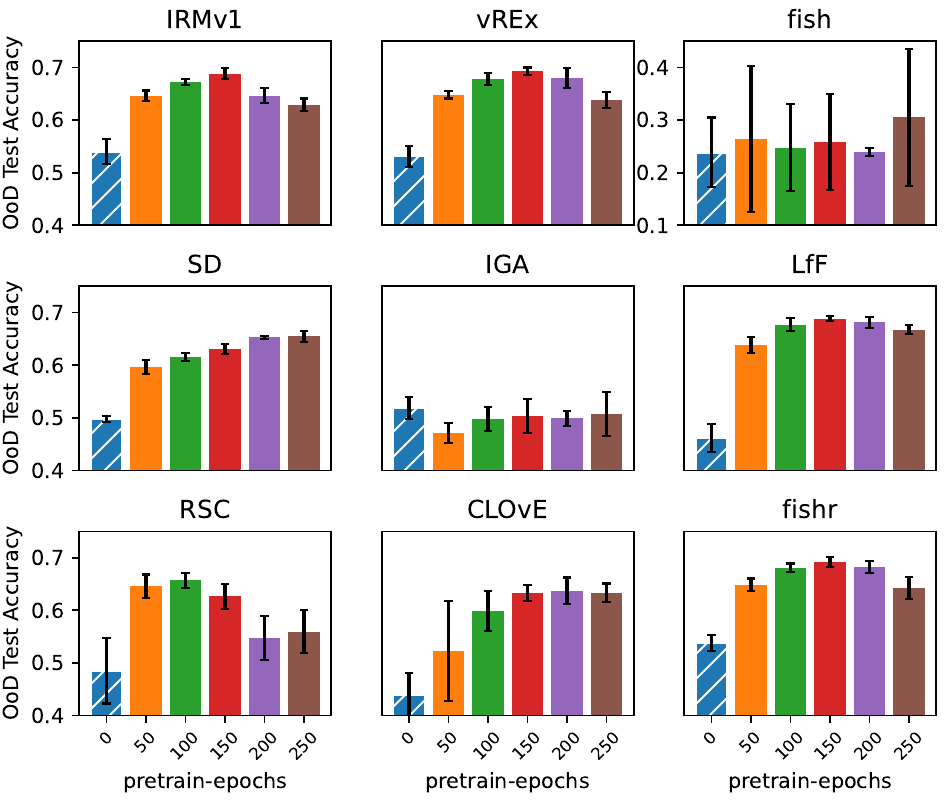}
    \caption{Test performance of nine penalized invariant-learning methods as a function of the number of epochs used to pre-train the neural network with ERM. The final \ood~testing performance is very dependent on choosing the right number of pretraining epochs, illustrating the challenges of these optimization problems.}
    \label{fig:importance_of_rep}
\end{figure}

Because their optimization is difficult, most authors recommend to pre-train the network with ERM before applying their invariant penalty. 
Figure~\ref{fig:importance_of_rep} shows the final \ood~test performance of models trained with each method as a function of the number $n_p\in\{0, 50, 100, 150, 200, 250\}$ of ERM pretraining epochs. During the execution of the invariant-learning algorithm, we choose one of five penalty weights and select the best early-stopping epoch by directly peeking at the \ood~test performance. All other hyper-parameters are copied from the \textsc{ColorMNIST} task \cite{irm}.  Appendix~\ref{apdix:train_details_colorminst} discusses these experiments with further details.

Figure~\ref{fig:importance_of_rep} shows that optimizing from
a random initialization (blue bars, 0 pretraining epochs) fails for all nine algorithms and all five penalty weights. Although pretraining with ERM helps, the final performance of the competitive algorithms depends on the number of pretraining epochs in rather inconsistent. Too much pretraining can cause performance drops in excess of 20\%. Even when one guesses the right amount of pretraining, the final performance comes short of the oracle performance ($0.721\pm0.002$) achieved by a network that is trained only on the robust feature.

We also showcase the optimization difficulty of several invariant-learning methods from a loss landscape's view on a low-dimensional case. See Appendix \ref{apdix:lowdim_losslandscape} for details. 

\begin{figure}[th!]
    \centering
    \includegraphics[width=0.55\textwidth]{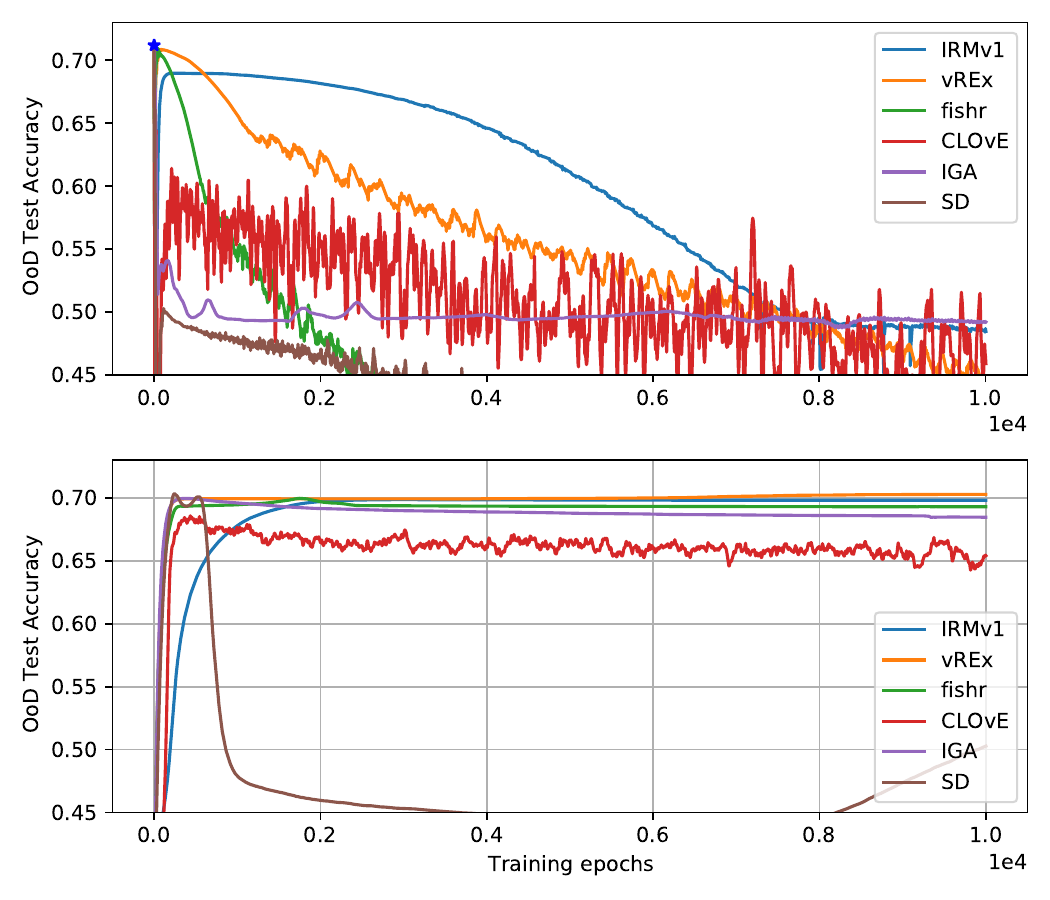}
    \caption{\ood~performance of invariant-learning methods as a function of training epochs. Top: Six invariant-learning methods are trained from a `perfect' initialization where only the robust feature is well learned. The blue star indicates the initial test accuracy. Bottom: Six invariant-learning methods are trained from a rich representation, constructed by (frozen) \textsc{Bonsai} representation.}
    \label{fig:generalization_difficulty}
\end{figure}

\bfparagraph{Invariant penalties do not enforce the constraints}
\label{sec:generalization_difficulty}

The previous section shows that the penalties introduced by these invariant-learning methods are too strong to allow reliable optimization. We now show that they are also too weak to enforce the constraints they are meant to enforce.

To substantiate this assertion, we initialize a network with the correct solution, that is, the solution obtained by training the network on a variant of the \textsc{ColorMNIST} dataset in which the
spurious color feature was removed. In order to keep the network from deviating from the target constraint, we use the largest penalty weight in the search space in each invariant-learning method. We do not report results on the Fish method because it failed to learn the task. We do not report on RSC and LfF because their test accuracy drops too fast.

The top plot in figure \ref{fig:generalization_difficulty} shows how the \ood~testing performance of six algorithms deviates from the performance of our perfect initialization. This might happen because
the chosen constraints have spurious solutions \citep{Kamath2021} or because the penalty terms are too weak to enforce the target constraints. Instead, the training process pulls the perfect initialization in the direction of the spurious feature (the color) which happens to be more predictive on the training data.

\begin{boxA}
    \textbf{Generalization-optimization dilemma} in penalty-based invariant learning algorithms: On one hand, the invariant penalties are too strong to optimize smoothly. On the other hand, the penalties are too weak to enforce the invariant constraints. 
\end{boxA}

The bottom plot in Figure~\ref{fig:generalization_difficulty} shows that rich features (constructed by \textsc{Bonsai}) help most invariant learning methods escape the generalization-optimization dilemma.\footnote{The penalizes in SD algorithm, a $L_2$ norm of logits, drives away the network from the oracle weights.}

\subsection{Bonsai method}
\label{sec:rfc}

This section presents tools for constructing rich representations. First, describes a mathematically sound approach to the problem of constructing a rich set of diverse features and we introduce the notions of \emph{discovery} and \emph{synthesis} episodes. Then shows how to use Distributionally Robust Optimization (DRO) \cite{rahimian2022frameworks} to cut on the synthesis episodes. Finally, presents the practical \textsc{Bonsai} algorithm that we use in Section~\ref{sec:experiments}.

\bfparagraph{Feature discovery}

Intuitively, constructing additional features is desirable when using these features increases the system performance on a pertinent subset of examples. Best would of course achieve a large performance increase on large subsets of examples.

For the purposes of this section, let $\Phi_k(x)$ be a large vector containing all previously constructed $k$ features for pattern $x$. The first step consists of defining an ensemble $P=\{D^1\dots{D^i}\dots\}$ of pertinent subsets $D^i$ of the training set. An effective way to choose a good ensemble of subsets is discussed at the end of this section. Having defined such subsets, we can define costs $C_i(\Phi,w)$ that measure
the quality of a feature set $\Phi$ measured on subset $D^i$:
\[
   C_i(\Phi,w) = \frac{1}{|D^i|} \sum_{(x,y)\in D^i} \ell(y, w^\top \Phi(x))
\]
where $w$ represent the weights of a linear layer and $\ell(y,\hat{y})$ is a convex loss. In the context of deep learning, considering a linear layer operating a large feature vector is not an unreasonable way to investigate the effectiveness of a representation \citep{ntk}. We can reweigh the training data in a manner that emphasizes the weaknesses of our current set of features, that is,
\begin{equation}
    \label{eq:r-rw}
    R_{rw} = \max_\lambda \min_w \sum_i \lambda_i C_i(\Phi_k,w)
\end{equation}
where the $\lambda_i$ coefficients are positive and sum to 1.  Let $\lambda^*_i$ be the pessimal mixture coefficients resulting from
optimization problem~\eqref{eq:r-rw}.  We can then learn a new set of features that help performance on this pessimal mixture,
\begin{equation}
    \label{eq:rprime-rw}
    R'_{rw} = \min_{w,\Phi} \sum_i \lambda^*_i C_i(\Phi,w)~.
\end{equation}
The main difference is that we are now training the features, yielding a new feature vectors $\Phi^*(x)$. If $R'_{rw}$  \eqref{eq:rprime-rw} is smaller than $R_{rw}$ \eqref{eq:r-rw}, then we know that $\Phi^*$ contains \emph{new useful features that were not present} in $\Phi_k$. This is the \emph{discovery phase}.

The next step consists in forming new feature vectors $\Phi_{k+1}(x)$ that contain the features present in both $\Phi_k$ and $\Phi^*$, a \emph{synthesis phase}. We can then iterate and obtain additional useful and diverse features at each iteration. The synthesis phase can be as simple as a vector concatenation. In the context of deep learning, however, one often has to use distillation, as discussed later in section~\ref{sec:rfc}.

The selection of a pertinent ensemble of subsets certainly affects which new features will be constructed at each iteration. In particular, it is desirable to make $R_{rw}$ as high as possible using a minimal number of subsets. This goal can be easily achieved by forming subsets containing examples that were either correctly classified or misclassified by the learning systems constructed by problem \eqref{eq:rprime-rw}.

\bfparagraph{Using DRO}

We now show how a DRO reformulation of this process can cut the intermediate synthesis phase. Because the $C_i$ are convex in $w$,
we can first apply von Neumann's minimax theorem \citep[theorem~3]{simons1995} to problem \eqref{eq:r-rw} and obtain a DRO problem \citep{Ben-TalGN09}:
\begin{align}
\label{eq:r-dro}
    R_{rw} &= \max_\lambda \min_w \sum_i \lambda_i C_i(\Phi_k,w)\notag  \\ 
        &= \min_w \max_\lambda \sum_i \lambda_i C_i(\Phi_k,w) \notag \\ 
        &= \min_w \max_i C_i(\Phi_k,w) ~=~  R_{dro}~. 
\end{align}

The next step is to run this same DRO optimization while also learning the features
\begin{equation}
    \label{eq:rprime-dro}
    R'_{dro} = \min_{w,\Phi} \max_i C_i(\Phi,w)~.
\end{equation}
To understand how quantity $R'_{dro}$ relates to $R'_{rw}$, we can use the max-min inequality as follows:
\begin{align}
  R'_{dro} &= \min_{w,\Phi} \max_\lambda \sum_i\lambda_i C_i(\Phi,w) \notag \\
    &\ge \max_\lambda \min_{w,\Phi} \sum_i \lambda_i C_i(\Phi,w) \notag \\
    &\ge \min_{w,\Phi} \sum_i \lambda^*_i C_i(\Phi,w) = R'_{rw}~. 
\end{align}
In other words, if $R'_{dro}$ is smaller than $R_{dro}$, then $R'_{rw}$ is smaller than $R_{rw}=R_{dro}$, and the new feature vector $\Phi$ contains new and useful features. The advantage of this approach is that
problem \eqref{eq:rprime-dro} does not involve mixture coefficients $\lambda^*$. Therefore there is no need to solve \eqref{eq:r-dro} or \eqref{eq:r-rw}, and no need for a synthesis phase at each iteration. The synthesis phase is only needed to construct the final rich representation after the last iteration.

\subsubsection{The practical Bonsai algorithm}

We now describe a practical algorithm that implements the ideas discussed in the previous subsection in a manner that is usable with ordinary deep networks. 
The workhorse of this algorithm is the Robust Empirical Risk Minimisation (RERM) algorithm (Algorithm~\ref{alg:rerm}) which takes an ensemble of datasets $D^k$ representing multiple distributions and seeks neural network weights that simultaneously yields small errors for all these distributions. RERM is in fact a minimal form of DRO with overfitting control by cross-validation.

\begin{algorithm}[ht]
\caption{Robust Empirical Risk Minimization (RERM)}
\label{alg:rerm}
\begin{algorithmic}[1]
\State  \textbf{Required:} datasets $D^k = \{(x^k_i, y^k_i)\}_{i=1}^{n^k}$, for $k = 1, \ldots, N$; model $f$; learning rate $\alpha$
\State  Randomly initialize $f$
\While{no overfit}  \texttt{  \quad\quad\quad\quad\quad\quad\quad\quad\quad\quad // By validation }
\State  Train on  datasets $D^1,\ldots,D^N$ by DRO: \\ \quad\quad\quad\quad\quad\quad $f \leftarrow f - \alpha \cdot \nabla_f \left[\max_k \left(\frac{1}{|D^k|} \sum_{(x^k_i, y^k_i) \in D^k} \ell(f(x^k_i), y^k_i)\right)_{k=1}^N\right]$
\EndWhile
\State \textbf{return} $f$
\end{algorithmic}
\end{algorithm}

\begin{algorithm}[ht]
\caption{Bonsai algorithm}
\label{alg:rich_feature}
\begin{algorithmic}[1]
\State \textbf{Input:} dataset $D$; the number of discovering rounds $K$
\State \texttt{// Discovery episodes}
\State $f_1 \leftarrow \text{RERM}(\{D\})$ 
\State Split $D$ into groups $A_1,B_1$ according to $f_1$. ($A_1=$ examples correctly classified by $f_1$, $B_1 = D \backslash A_1$)   
\State Available groups $P = \{A_1,B_1\}$
\For{$k \in [2,\ldots, K]$}
\State $f_k \leftarrow \text{RERM}(P)$
\State Split $D$ into groups $A_k,B_k$ according to $f_k$
\State $P \leftarrow P \cup \{A_k, B_k\}$
\EndFor
\State \texttt{// Synthesis episode}
\State Pick a feature extractor function $\Phi$, and $K$ linear classifiers $\omega_1,...\omega_k$ at random
\State Create $K$ groups of pseudo-labels ${y}^k$ by applying each $f_k$ on $D$ 
\State $A = A_i \cap ... \cap A_K$
\State Update $\Phi, \omega$ such that each pseudo-label ${y}^k$ is well learned by the corresponding classifier $\omega_k$ and $\Phi$: 
$\sum_{k=1}^{K} \frac{1}{|{A}|}\sum_{(x_i,{y}_i^k)\in {A}}\ell(\omega_k  \circ \Phi(x_i), {y}^k_i) + \frac{1}{|D\backslash {A}|}\sum_{(x_i,{y}_i^k)\notin {A}} \ell(\omega_k  \circ \Phi(x_i), {y}^k_i)$
\State \textbf{return} $\Phi, \{\omega_k\}_{k=1}^K$
\end{algorithmic}
\end{algorithm}

The \textsc{Bonsai} algorithm (Algorithm~\ref{alg:rich_feature}) first performs a predefined number of \emph{discovery episodes}, using RERM to repeatedly solve an analogue of problem \eqref{eq:rprime-dro} that constructs a model $f_k$ at each iteration, using an ensemble of subsets formed by selecting which examples were correctly or incorrectly recognized by the models $f_0\dots f_{k-1}$ constructed during the previous iterations.

The \textsc{Bonsai} algorithm performs a distillation-based \emph{synthesis episode}. The goal is to learn a representation network $\Phi(x)$ such that we can emulate the functions $f_k$ using a simple network with weights $w_k$ on top of $\Phi(x)$. To that effect, we use the $f_k$ models to compute pseudo-labels $y^k(x)$ for each example $x$. We then train a composite model with parameters $\Phi$, $w_1$, \dots, $w_K$ whose $k$ outputs are trained to replicate the pseudo-labels. 

\paragraph{Why use linear classifiers in synthesis episode (line 11)?}

The goal is to perform the synthesis step by distillation into a network whose architecture is as close as possible as the architecture of the “source” networks. However the distillation network needs one head for each source network. The least intrusive way to implement multiple heads is to duplicate the very last layer, hence linear. The opposite approach would be to claim the whole network is a classifier and the feature extractor is the identity. In this case, we can get a perfect synthesis loss (Alg \ref{alg:rich_feature}, line 14) with an identity feature extractor which is obviously useless. We leave the non-linear classifier in \emph{synthesis phase} as a future work.

\paragraph{What if the first \textsc{rerm} round achieves zero errors (line~3)?}
The training set of the first RERM round is the union of the data associated with all \ood~training environments. Since RERM avoids overfitting using a validation set (Alg \ref{alg:rerm}, line 3), a perfect accuracy on both the merged training and validation sets means that the features discovered in the first round are already invariant in all training environments and perfectly predictive (100\% accuracy). Therefore no further rounds are necessary since we already have a solution. This is in fact a degeneracy of the invariant training concept.

\subsection{Experiments on synthesis data}
\label{sec:experiments}

This section presents experimental results that illustrate how the rich representations constructed with \textsc{Bonsai} can help the \ood~performance and reduce the variance of invariant-learning methods, using a synthesis ColoredMNIST dataset.

\subsubsection{Bonsai representation helps all methods}
\label{sec:rfc_ood}

All experiments reported in this section use the \textsc{ColorMNIST} task \cite{irm} which consists of predicting labels that indicate whether the class of a colored digit image is less than 5 or not. The target label is noisy and only matches the digit class with probability 0.75 (correlation coefficient 0.5). Two training sets are provided where a spurious feature, the color of the digit, correlates with the target label with respective probabilities 0.8 and 0.9 (correlation coefficients 0.6 and 0.8). However, in the \ood~testing set, the digit color is negatively correlated with the label (correlation coefficient $-$0.8). This testing protocol hits algorithms that rely on the spurious color feature because it happens to be more predictive than the robust feature in both training environments.
 
We compare six invariant-learning methods (IRMv1, vREx, SD, IGA, Fishr, CLOvE) and ERM after four types of initialization/representation:\footnote{``Representation'' indicates the output of a function. ``Initialization'' not only indicates the output of a function, but also indicates the parameterization of the function. This section perfers ``initialization'', because the generalization-optimization dilemma in invariant-learning methods is connected to the parameterization of functions.  }
\begin{itemize}
\setlength\itemsep{-0.3em}
    \item \textbf{Rand}: a random initialization with the popular Xavier method \citep{pmlr-v9-glorot10a},
    \item \textbf{ERM}: a random initialization followed by several epochs of ERM, 
    \item \textbf{Bonsai}: a initialization with \textsc{Bonsai} representations,
    \item \textbf{Bonsai-cf}: a frozen \textsc{Bonsai} representations: the training algorithm is not allowed to update them.
\end{itemize} 
The ERM initialization essentially consists of turning off the penalty terms defined by the various invariant-learning method. This is comparable to the delicate penalty annealing procedures that are used by most authors \cite{irm, vrex, sd, rame2021fishr}. The \textsc{Bonsai} initialization was computed by two discovery phase iterations, containing richer features.

Table~\ref{tab:rich_rep_col025} reports the \ood~testing accuracies obtained by six invariant learning methods and four initialization/representation. Experimental details can be found in Appendix~\ref{apdix:train_details_colorminst}. Bonsai initialization (rich feature) helps the \ood~performance of most invariant-learning algorithms. Interestingly, the best results are achieved by freezing the Bonsai representation, which is consistent with the results of Section~\ref{sec:generalization_difficulty} showing that \emph{the penalties for the invariant learning algorithm are in fact insufficient to maintain the desired invariance constraints, even when initialized with the oracle weights}.

\begin{table}[ht]
    \caption{OoD testing accuracy achieved on the \textsc{ColorMNIST}. The first six rows of the table show the results achieved by six invariant-learning methods using respectively random initialization (Rand), ERM initialization (ERM), Bonsai initialization (Bonsai). The last column, (Bonsai-cf), reports the performance achieved by running the \ood~algorithm on top of the frozen Bonsai representations. The seventh row reports the results achieved using ERM under the same conditions. The last row reminds us of the oracle performance achieved by a network using data from which the spurious feature (color) has been removed.}
    \label{tab:rich_rep_col025}
    \centering
    \begin{tabular}{l|c|c|c|c}
\toprule
 & Rand & ERM & 
 Bonsai& Bonsai-cf \\
\midrule
IRMv1 & 54.0$\pm$2.4  & 68.9$\pm$1.1  &  66.5$\pm$1.5         &\textbf{69.9$\pm$0.6}  \\
vREx  & 53.1$\pm$2.0  & 69.3$\pm$0.7  & \textbf{70.3$\pm$0.4} & 69.9$\pm$0.4  \\
SD    & 49.8$\pm$0.6  & 65.5$\pm$1.1  & 69.8$\pm$0.6          &\textbf{70.4$\pm$0.4}  \\
IGA   & 51.8$\pm$2.1  & 50.7$\pm$4.2  & 69.4$\pm$0.7          &\textbf{70.0$\pm$0.8}  \\
fishr & 53.9$\pm$1.5  & 69.2$\pm$0.9  & \textbf{70.2$\pm$0.4} & 69.4$\pm$0.8  \\
CLOvE & 43.9$\pm$4.2  & 63.7$\pm$2.5  & 67.1$\pm$3.8          & \textbf{68.4$\pm$0.8} \\
\midrule
ERM &  27.3$\pm$0.4 & 27.3$\pm$0.4 & 43.4$\pm$2.8 & 35.6$\pm$1.2 \\
\midrule
oracle & \multicolumn{4}{c}{72.1 $\pm$ 0.2 } \\
\bottomrule
    \end{tabular}
\end{table}
\subsubsection{Aiming for the second easiest-to-find feature}
\label{sec:2nd_easy_feature}

Recent work \cite{liu2021just,lff, Bao2021, ahmed2020systematic, creager2021environment} claims to achieve \ood~generalization by discovering and using only the second easiest-to-find features. Although this strategy often works on datasets that were constructed to showcase \ood~problems, the assumption that the second easiest features are the robust ones is unreasonable.

To illustrate this claim we construct a variant of the \textsc{ColoredMNIST} dataset by changing the noise levels to make the robust feature (the digit shapes) more predictive than the spurious features (the digit color). 

Table \ref{tab:pi_rfc} compares the six invariant-learning methods using the frozen \textsc{Bonsai} representation on both \textsc{ColoredMNIST} and \textsc{InverseColoredMNIST}. All six methods achieve very comparable \ood~testing accuracies. The ERM method fails on \textsc{ColoredMNIST} but performs quite well on \textsc{InverseColoredMNIST} because relying on the most predictive features is a good strategy for this task. In contrast, the algorithm PI \citep{Bao2021}, which aims for the second easiest features, performs well on \textsc{ColoredMNIST} but far worse on the easier \textsc{InverseColoredMNIST} task.

\begin{table}[t]
\caption{OoD test accuracy of PI and OoD/ERM methods on \textsc{ColoredMNIST} and \textsc{InverseColoredMNIST}. The OoD/ERM methods use a frozen Bonsai representation (Bonsai-cf).}
    \label{tab:pi_rfc}
    \centering
    \begin{tabular}{c|c|c}
    \toprule
    Methods &\makecell{\textsc{ColoredMNIST}} & \makecell{\textsc{Inverse ColoredMNIST}} \\
    \midrule
        IRMv1 & 69.9$\pm$0.6 &  80.3$\pm$2.2 \\
        vREx  & 69.9$\pm$0.4 &  84.0$\pm$1.2\\
        SD    & 70.4$\pm$0.4 &  81.9$\pm$1.3\\
        IGA   & 70.0$\pm$0.8 &  78.5$\pm$3.6 \\
        fishr & 69.4$\pm$0.8 &  82.6$\pm$1.4 \\ 
        CLOvE & 68.4$\pm$0.8 &  71.9$\pm$0.7\\
        ERM   & 35.6$\pm$1.2 &  71.7$\pm$0.7 \\
        \midrule
         PI      &  70.9$\pm$0.3  &51.0$\pm$4.7 \\
    \bottomrule
    \end{tabular}
\end{table}

\subsection{Experiments on a real-world task}
\label{sec:camelyon}

\bfparagraph{Experimental setups} The \textsc{Camelyon17} dataset \citep{bandi2018detection} contains histopathological images accompanied by a label indicating whether the central region of the image contains a tumor. The images were collected from five different hospitals with potentially different imaging hardware and different procedures. The WILDS benchmark \citep{wilds2021} contains a task that uses this dataset with a very clear specification of which three hospitals are to be used as training data (302,436 images), which hospital is to be used for \ood~generalization testing (85,054 images). The task also specifies multiple runs with different seeds in order to observe the result variability. Finally the task defines two ways to perform hyper-parameter selection: ``IID Tune'' selects hyper-parameters based on model performance on 33,560 images held out from the training data, ``OoD Tune'' selects hyper-parameters on the model performance observed on the fifth hospital (34,904 images).

We compare four different learning methods (ERM, IRMv1, vREx, and CLOvE) on two initializations (ERM, 2-Bonsai). We also compare the four learning methods on corresponding frozen representations (ERM-cf, 2-Bonsai-cf, 3-Bonsai-cf).

We strictly follow these procedures as well as the experimental settings suggested in the WILDS task. The network is a DenseNet121 model \citep{huang2017densely} trained by optimizing a cross-entropy loss with $L_2$ weight decay=$0.01$ using SGD with learning rate=$0.001$, momentum=$0.9$ and batch size=$32$. The penalty weights are selected from $\{0.5, 1, 5, 10, 50, 100, 500, 1000\}$ for IRMv1 and vREx, $\{0.0005, 0.001,0.005,0.01,0.05,0.1,0.5,1\}$ for CLOvE. The number of ERM pre-training iterations is selected in set from $\{0, 500, 1000, 5000, 10000\}$. Further details are provided in Appendix~\ref{apdix:camelyon17}. More examples of real-work tasks can be found in related work \citet{chen2023towards}. 

\bfparagraph{Main results} Table \ref{tab:camelyon_full_results} reports the \ood~testing accuracies obtained by using both the IID and OoD hyper-parameter tuning approach. Accuracies were averaged over five repetitions with different random seeds. The first block of rows reports accuracies obtained by all four methods using ERM initialization. These accuracies come with large error bars because they considerably vary across repetitions. As a consequence, the accuracies differences observed in this block are not significant. The second block of rows shows that freezing the representations does not significantly improve this situation. In contrast, using a Bonsai representation with two discovery rounds (2-Bonsai) consistently improves the accuracies obtained by all four methods using either the IID or OoD tuning approaches (third block of rows). Freezing the Bonsai representation provides an additional boost (fourth block of rows).

\begin{table}[ht!]
 \caption{Test Accuracy on the \textsc{camelyon17} dataset. The hyper-parameter tuning process is performed on either the iid validation or the \ood~validation set (``IID/OoD Tune''). We test ERM pretrained initialization, 2-rounds, and 3-rounds Bonsai representation. As to the learning methods, we test ERM, IRMv1, vREx, and CLOvE. When freezing the representation and training the top-layer classifier only, we get ``-cf'' methods. The standard deviation is calculated on 5 random seeds [0-4]. \textbf{It worth noting that given Bonsai rich feature ERM approach performs as good as other invariant-learning methods, showing that it is the rich feature rather than invariant-learning penalty matters in invariant-learning. }}
    \label{tab:camelyon_full_results}
    \centering
    \resizebox{0.5\textwidth}{!}{
    \begin{tabular}{l|l|c|c}
        \toprule
        Network        & Methods & \multicolumn{2}{c}{Test Acc} \\
        Initialization &         & IID Tune & \ood~Tune \\
        \midrule
        $\times$ & ERM & 66.6$\pm$9.8 & 70.2$\pm$8.7 \\
        ERM & IRMv1 & 68.6$\pm$6.8 & 68.5$\pm$6.2\\
        ERM & vREx & 69.1$\pm$8.1 & 69.1$\pm$13.2\\
        ERM & CLOvE & 71.7$\pm$10.2 & 69.0$\pm$12.1 \\
        \midrule
        ERM-cf &ERM & $\times$ & $\times$ \\
        ERM-cf &IRMv1 & 69.6$\pm$10.5 & 70.7$\pm$10.0  \\
        ERM-cf &vREx & 69.6$\pm$10.5 & 70.6$\pm$10.0   \\
        ERM-cf &CLOvE & 69.6$\pm$10.5 & 69.2$\pm$9.5   \\
         \midrule
        2-Bonsai & ERM   & 72.8$\pm$3.2   & 74.7$\pm$4.3 \\
        2-Bonsai & IRMv1 & 71.6$\pm$4.2 & 75.3$\pm$4.8  \\
        2-Bonsai & vREx  &  73.4$\pm$3.3 & 76.4$\pm$5.3  \\
        2-Bonsai & CLOvE & 74.0$\pm$4.6 &76.6$\pm$5.3    \\
        \midrule
        2-Bonsai-cf & ERM   & 78.2$\pm$2.6 & 78.6$\pm$2.6\\
        2-Bonsai-cf & IRMv1 & 78.0$\pm$2.1 & 79.1$\pm$2.1 \\
        2-Bonsai-cf & vREx  & 77.9$\pm$2.7 & 79.5$\pm$2.7 \\
        2-Bonsai-cf & CLOvE & 77.8$\pm$2.2 & 78.6$\pm$2.6 \\
        \midrule
        3-Bonsai-cf & ERM   & 72.9$\pm$5.3 & 73.3$\pm$5.3 \\
        3-Bonsai-cf & IRMv1 & 72.7$\pm$5.5 & 75.5$\pm$3.8 \\
        3-Bonsai-cf & vREx  & 72.7$\pm$5.4 & 75.1$\pm$5.3 \\
        3-Bonsai-cf & CLOvE  & 72.8$\pm$5.4 & 73.2$\pm$7.1 \\
         \bottomrule
    \end{tabular}
    }
\end{table}

\begin{figure*}[t]
    \centering
    \includegraphics[width=\textwidth]{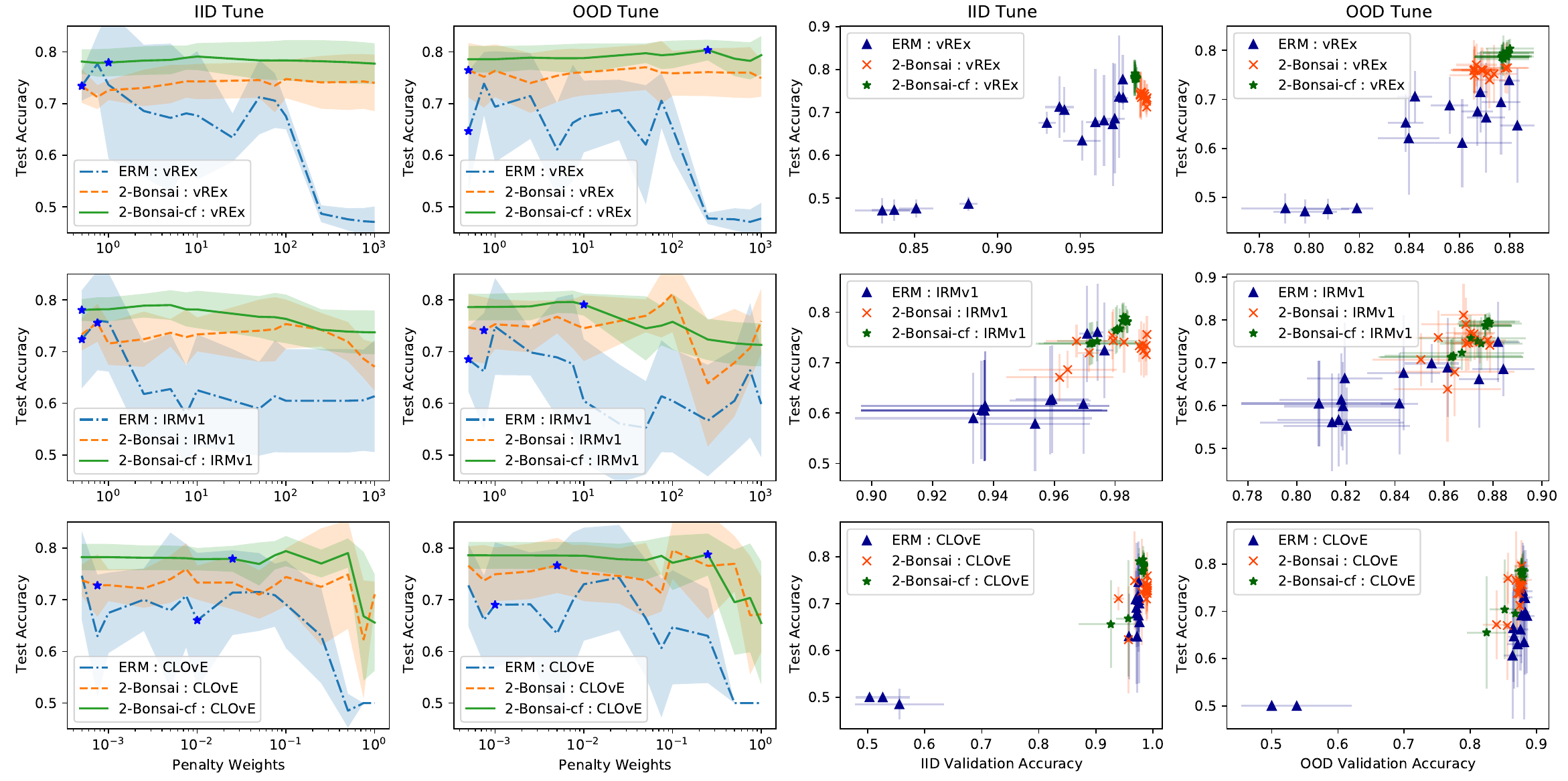}
    \caption{\textbf{Left half}: \ood~testing accuracy as a function of the penalty weight. The six plots correspond to the IRMv1, vREx, and CLOvE algorithms with all other hyper-parameters selected using either the IID or OoD tuning method. Bonsai initialization makes these curves far more predictable than ERM initialization. Starts indicate the final penalty weight choice.~
    \textbf{Right half}: \ood~testing accuracy as a function of the validation accuracy. Bonsai initialization reduces the variance of both the \iid~and \ood~validation performances, making them far more reliable indicators of the actual \ood~testing performance.}
    \label{fig:lambda_valid_test}
\end{figure*}

\citet{rosenfeld2022domain} claimed ERM may already discover enough features in the representation for \ood~generalization. The second block in Table \ref{tab:camelyon_full_results} shows the ERM learned representation is not rich enough in the \textsc{camelyon17} case, supporting the theoretical framework in Section \ref{sec:features}. 

Using a Bonsai representation with three discovery rounds (3-Bonsai-cf) does not work as well. In fact, the features extracted during the third discovery phase round are not as predictive as the first two rounds (Table~\ref{tab:camelyon17_rfc_performance}). More discovery rounds also increase the difficulty of the synthesis phase, as we want to distillate more features (including poor ones) into the same fixed-size representation.

Much to our surprise, Bonsai initialization consistently boosts the accuracies of both the ERM and invariant-learning methods, using either the IID or \ood~tuning method. The frozen Bonsai representations can even help ERM outperform earlier comparable results reported on the WILDS leaderboard\footnote{\url{https://wilds.stanford.edu/leaderboard}} by about 5\%.

\bfparagraph{Hyper-parameter tuning}
\label{sec:hyper-paraemeter-model-selection-difficulty}

Figure \ref{fig:importance_of_rep} and \ref{fig:generalization_difficulty} illustrate how the \ood~generalization performance of many invariant-learning methods depends strongly on hyper-parameters such as the number of pretraining epochs, the penalty weights, the learning epochs. This is in fact a consequence of the optimization-generalization dilemma itself. It is simply difficult to simultaneously ensure good \ood~generalization performance and run a stable and efficient optimization process.

The left half of Figure~\ref{fig:lambda_valid_test} shows \ood~testing accuracies for the \textsc{Camelyon17} task as a function of the penalty weights, with all other hyper-parameters chosen using either the IID or OoD tuning method. With ERM pretraining, the \ood~testing performance of all three invariant-learning methods (IRMv1, vREx, CLOvE) depends very chaotically on the penalty weight. In contrast, with a frozen Bonsai representation, the \ood~testing performance of invariant-learning methods, as a function of the penalty weight, follows a much smoother curve.

The right half of Figure~\ref{fig:lambda_valid_test} shows the relation between the \iid/\ood~validation accuracies and the \ood~testing accuracies for three invariant-learning methods using both ERM and Bonsai initialization. Bonsai initialization reduces the variance of both the \iid~and \ood~validation performances, making them far more reliable indicators of the actual \ood~testing performance.

\begin{table}[ht]
 \setlength{\tabcolsep}{6.5mm}
\caption{OoD test accuracies for the models constructed by the first three discovery rounds for the \textsc{Camelyon17} task. The first round amounts to performing ERM. The second round extracts a useful set of features. The third round extracts comparatively weaker features. All these accuracies remain substantially worse than those achieved by training a system on top of the combined representation computed during the synthesis phase (see Table~\ref{tab:camelyon_full_results}).}
    \label{tab:camelyon17_rfc_performance}
    \centering
    \resizebox{0.5\textwidth}{!}{
    \begin{tabular}{c|c|c}
    \toprule
    Round 1 & Round 2 & Round 3 \\
     \midrule
66.6$\pm$9.8 & 73.2$\pm$5.7 & 61.8$\pm$10.2 \\
\bottomrule
    \end{tabular}
    }
\end{table}

\bfparagraph{The value of the synthesis episode}
\label{sec:discovery_phase_performance_camelyon17}

The \textsc{ColoredMNIST} and \textsc{InverseColoredMNIST} experiments (Table~\ref{tab:pi_rfc}) show that the robust feature can be discovered during different rounds of the discovery phase. We can therefore wonder whether the discovery phase already produces the correct invariant representation during one of its successive rounds.

This is not the case in general. Table~\ref{tab:camelyon17_rfc_performance} reports the \ood~testing accuracies of the classifiers constructed during the first three rounds of the discovery phase. All three accuracies are substantially worse than the accuracies achieved by any algorithm using a frozen 2-Bonsai-cf representation (Table~\ref{tab:camelyon_full_results}). This indicates that these higher accuracies are obtained by simultaneously exploiting features discovered by different rounds of the discovery phase. Making them all simultaneously available is indeed the role of the synthesis phase.

\subsection*{Discussion}

This section makes several contributions:
\begin{itemize}
\setlength\itemsep{-0.3em} 
\item points out the severity of the optimization-generalization dilemma in invariant-learning domain, showing that the various penalties introduced by invariant-learning methods are either too strong to optimize or too weak to achieve their goals.
\item proposes to work around the problem by seeding the networks with a rich representation that contains a diversity of features readily exploitable by the algorithm. Formalizes this objective and describes an algorithm, \textsc{Bonsai}, that constructs such rich representations.
\item shows that \textsc{Bonsai} helps a variety of invariant-learning methods achieve a better \iid~performance. 

\item More importantly, these results (Table \ref{tab:camelyon_full_results}) reveal that \emph{it is rich feature rather than the commonly believed invariant-learning penalties that matters in invariant-learning.}
\end{itemize}

\section{Case study 2: Rich features in \ood~fine-tuning}
\label{sec:very-large-dropout}
In the study of Section \ref{sec:features}, \ref{sec:randomness}, and \ref{sec:bonsai}, we have shown that the hope of constructing rich features by merely optimizing the expectation of a suitable loss function for a single training distribution (e.g., using stochastic gradient techniques) is contradicted by the implicit sparsity bias of stochastic gradient algorithms \citep{andriushchenko2023sgd,blanc2020implicit} and thus fails. In a nutshell, a feature only survives when it brings an incremental training error advantage relative to what can be achieved using all the other features already present in the network. {We slightly abuse the terminology and call them ``strongly relevant''. However, features that are not strongly relevant might nevertheless 
\begin{itemize}
\setlength\itemsep{-0.3em}
    \item[\textbf{(a)}] be incrementally useful when the data follows a different distributions of interest, or
    \item[\textbf{(b)}] be useful under the training distribution when added to certain subsets of the other existing features instead of all of them (``weakly relevant'').
\end{itemize}
It is therefore tempting to ``enrich" the representation with features of type (b), which can be found using the training data, and hope that some of these will turn out to also be features of type (a) whose inclusion helps when the data distribution changes. To ``enrich'' representation, section \ref{sec:bonsai} introduces adversarial discovery, section \ref{sec:randomness} utilizes ensemble\footnote{Averaging weights \citep{rame2023model, rame2022diverse,wortsman2022robust} can be viewed as an approximation of ensemble in the optimization-friendly fine-tuning scenario.} in the scratch-training scenario. This section explores a simple approach, \emph{very large dropout} \cite{zhang2024fine}, in the \ood~fine-tuning scenario (optimization-friendly).

The dropout technique \citep{srivastava2014dropout} seems well suited to find {weakly relevant features} because randomly masking units of a representation layer during training amounts to forming random subsets of all other available features. However, in order to form small subsets, one would have to use very high levels of dropout. Unfortunately, training a sizable deep network from scratch with such a large dropout is practically impossible. {Instead, computationally demanding methods, such as adversarial sampling \citep{zhang2022rich,chen2023towards} and representation ensembles \citep{zhang2023learning}, have been proposed to find {weakly relevant features} while training a network from scratch.}

There is however a practically meaningful scenario {in which we can use an extremely aggressive dropout: fine-tuning a pre-trained network using a comparatively small dataset.  This is possible because such a fine-tuning operation makes only modest changes to the network weights.} For example, several authors \citep{rame2022diverse, wortsman2022model} argue that fine-tuned networks remain ``{linearly connected}'', that is averaging the parameters of multiple fine-tuned networks approximate the ensemble of these networks.  \citet{evci2022head2toe} even show that a linear classifier on top of the union of internal-layer features of pre-trained residual networks can match or exceed the performance of fine-tuning.

The \emph{very large dropout} approach introduced in this section simply \emph{fine-tune using very large dropout levels}, randomly masking {above} 90\% of the units in the penultimate representation layer. This simple approach \emph{exceeds the performance of both ensemble and weight-averaging methods}. This result is not only \emph{practically meaningful}, but also clarifies the idea of \emph{rich features}.

\subsection{Example Analysis}

\begin{boxA}
    \textbf{The {\textit{two-distributions}} setup} is commonly used for transfer learning. In this setup, features $\Psi$ are obtained by pre-training a network on a large training set associated with a first distribution $\mathcal{T}_{\text{p}}$. These features are then used to construct or initialize a new model $\omega_{\text{d}} \circ \Psi$, which is then trained using a smaller training set associated with a second distribution $\mathcal{T}_{\text{d}}$. The question is to determine which pre-training approach is most likely to make the features $\Psi$ useful for the transfer task $\mathcal{T}_{\text{d}}$.
\end{boxA}

\begin{boxA}
    \textbf{The {\textit{three-distributions}} setup} \citep{rame2022diverse} views the pre-trained model as a base model that is assumed very rich but whose training process is beyond our control (e.g., a fundational model). The features $\Psi$ of the pre-trained model are then incorporated into a new model $\omega_{\text{d}} \circ \Psi$ that is fine-tuned using a second distribution $\mathcal{T}_{\text{d}}$ and eventually tested on a third distribution $\Tilde{\mathcal{T}}_{\text{d}}$ illustrating the same general task as the second distribution (e.g., using the same classification labels.) The question is then to determine which fine-tuning approach is most likely to produce a model that will perform robustly under the eventual testing distribution $\Tilde{\mathcal{T}}_{\text{d}}$. This section focuses on this three-distribution setup.

\end{boxA}

Considering a logistic regression with parameter $\omega\in\mathbb{R}^n$ operating on a vector $\Psi$ of $n$ features and predicting a binary target $Y$ representing our ${\mathcal{T}}_{\text{d}}$ distribution. Assume further that each individual feature $\Psi_i, i \in [1,\dots, n]$ perfectly predicts $Y$, that is, zero classification error can be achieved with a regression $\omega$ whose only nonzero parameter marks the $i$-th feature. During gradient-based optimization, achieving zero loss by using only one feature prevents the system from using the other features, because of the ``gradient starving'' phenomenon \citep{pezeshki2021gradient}.   
We now evaluate this trained system on a target distribution $\Tilde{\mathcal{T}}_{\text{d}}$ that only differs from ${\mathcal{T}}_{\text{d}}$ because some features were missing and have been replaced by zeroes. If our trained system (on ${\mathcal{T}}_{\text{d}}$) depends only on one feature, we better hope that this is not one of the missing ones in target distribution $\Tilde{\mathcal{T}}_{\text{d}}$.

\bfparagraph{In this linear case,} the following three strategies are equivalent in terms of encouraging the optimization process to learn more features (Check \citet{srivastava2014dropout} for the proof): 
\begin{itemize}
    \setlength\itemsep{-0.3em}
    \item[1)] feature-bagging (ensemble) \cite{bryll2003attribute},
    \item[2)] Dropout \cite{srivastava2014dropout},
    \item[3)] $L_2$ regularization on $\omega$. 
\end{itemize}
Furthermore, the feature-bagging approach solves the above problem by construction. Thus, in the linear case, all three strategies solve the above problem.

\bfparagraph{In the case of a multilayer network,} however, this equivalence is broken. In particular, $L_2$ regularization on the inner layer parameters plays the different role of encouraging sparse representations \cite{blanc2020implicit, andriushchenko2023sgd}. Dropout and deep ensembles may achieve comparable error rates in distribution but differ sharply when it comes to estimating prediction uncertainty \citep{ashukha2020pitfalls}. These differences become very important when training models with the pursue of \ood~generalization ability, making deep ensembles and weight averaging ensembles more attractive than dropout for \ood~generalization \cite{rame2022diverse, rame2023model, wortsman2022model, swad, wa}. However, in the case of 

In contrast, this section shows that using a very large dropout rate during fine-tuning (rather than during scratch-training) substantially improves on the \ood performance of both ensemble and weight-averaging. This simple approach was not considered before, possibly because such large dropout rates are not usable during pretraining, resulting in poor performance overall.

\subsection{Very-large dropout Method}
\label{sec:dropout_method}
The key results described later in this paper have been obtained with a very simple method.  The base model is a deep learning network with residual connections trained on data $\mathcal{T}_{\text{p}}$ that is related to but substantially larger than the datasets illustrating the task of interest. Some of these datasets ($\mathcal{T}_{d}$) are used to fine-tune the base model. Performance is reported on both held-out data from the fine-tuning datasets (\iid. performance on $\mathcal{T}_{d}$) and data from the remaining datasets (\ood. performance on $\Tilde{\mathcal{T}}_{d}$). 

{We focus on residual networks because fine-tuning has been found to hardly change the inner layers of non-residual networks (\citealp{raghu2019rapid}, fig 2). 
In contrast, skip connections in residual networks expose the inner block features in such a manner that the fine-tuning process can utilize these features in a near-linear way \citep{evci2022head2toe}. }

Fine-tuning is carried out with a standard stochastic learning procedure (e.g. \textsc{sgd} or \textsc{adam}) after applying a very large dropout to the penultimate layer representation $\Phi$. {Unlike \citep{srivastava2014dropout}, we only apply dropout on the penultimate layer representation $\Phi$, because skip connections in residual networks expose many inner-layer features to the last linear layer, as illustrated by the decomposition of residual networks proposed by \citet{veit2016residual},}
\begin{align}
\label{eq:residual_feature}
    \Phi(x) &= \underbrace{\vphantom{f_1(x)}x}_{\mathclap{\phi_0(x)}} + \underbrace{f_1(x)}_{\phi_1(x)} + \underbrace{f_2(x+f_1(x))}_{\phi_2(x)} + \dots = \sum_{i\in [0,\dots,l]} \phi_i(x)\,,
\end{align}
where $f_i$ represents the function implemented by the $i$-th residual block, and
\begin{align}
\label{eq:dropout}
  \Phi_{\mathrm{dropout}}(x) = \frac{m(\lambda)}{1-\lambda} \odot \Phi(x) \, ,
\end{align}
where $\odot$ represents the component-wise product and $m(\lambda)$ is a vector of random Bernoulli variables equal to $0$ with probability $\lambda$ and $1$ with probability $1-\lambda$. The additive decomposition of $\Phi(x)$ in equation \eqref{eq:residual_feature} makes clear that applying dropout to $\Phi(x)$ simultaneously blocks the contributions~$\phi_i(x)$ of all residual blocks.

In the scratch training scenario, many papers \cite{li2019understanding,kim2023use} show the favor of mild dropout rates. In contrast, this section uses counterintuitively large dropout rates (more than 90\%) in the \ood~fine-tuning scenario. Hence, the approach is named \emph{very-large dropout}.

\subsection{Experiments in \ood~Fine-tuning}
\label{sec:experiements}

\bfparagraph{Dataset} 

We perform most experiments using \textsc{pacs} \citep{PACS}, \textsc{vlcs} \citep{vlcs}, \textsc{office\,home} \citep{officehome}, and \textsc{terra\, incognita} \citep{terraincognita} datasets. 
These datasets spam in diverse domains, from wild images with different environment conditions to artificial sketching and painting, from natural animals to home furniture. With $9,991$ to $24,788$ examples, these datasets are substantially smaller than the pre-training dataset \textsc{ImageNet} with 1.2\textsc{m} examples. 

Each of these datasets is divided into four sub-datasets that share the same target label categories but follow a different distribution. For example, one sub-dataset of \textsc{pacs} contains simple sketch images of `dog' and `elephant', while another sub-dataset contains real photos of `dog' and `elephant'. This makes it possible to conveniently evaluate \ood~performance by fine-tuning on three sub-datasets and testing on the fourth one. 
\bfparagraph{Models}
We carry out experiments using two wisely used residual architectures: \textbf{convolutional network} and \textbf{visual transformer}. For the {convolutional network} experiments, we use a \textsc{ResNet50} architecture \citep{he2016deep} with 25\textsc{m} parameters.\footnote{\url{https://pytorch.org/blog/how-to-train-state-of-the-art-models-using-torchvision-latest-primitives/}} For the {visual transformer} experiments, we use the large vision transformer \textsc{ViT-L-16} \citep{dosovitskiy2020image} with 304\textsc{m} parameters.\footnote{\url{https://github.com/pytorch/vision/tree/main/references/classification\#vit_l_16}} 

\bfparagraph{Pre-training} 

Unless otherwise stated, all experiments are carried out using networks
pre-trained using refined data augmentations initially introduced in the context of residual networks: \textsc{trivialaugment} \citep{muller2021trivialaugment}, \textsc{cutmix} \citep{yun2019cutmix}, and \textsc{random\,erasings} \citep{zhong2020random}. We use these augmentations to mimic the properties of large foundational models trained using very large and diverse pre-training data.

\bfparagraph{Baselines} 

Using these same datasets, \citet{gulrajani2020search} argue that plain Empirical Risk Minimization ({\textsc{erm}}) equals and often betters the \ood~performance of purposefully designed methods, such as {\textsc{coral}} \citep{sun2016deep}, {\textsc{dro}} \cite{Sagawa2019DistributionallyRN},  {\textsc{mldg}} \cite{li2018learning},  {\textsc{dann}} \cite{Ganin2015DomainAdversarialTO},  {\textsc{c-dann}} \cite{Li2018DomainGV},  {\textsc{mmd}} \cite{lidomainadversarial},  {\textsc{vrex}} \citep{krueger2021out}, and  {\textsc{irm}} \citep{arjovsky2019invariant}.  More recently, \citet{wa}, \citet{swad}, \citet{rame2022diverse}, and \citet{rame2023model} find that {ensemble} and {weight averaging} methods consistently outperform the \ood~performance of {\textsc{erm}}. 

Therefore, it is sufficient to compare our results with those of the \textbf{ensemble}, \textbf{weight averaging}, and \textbf{\textsc{erm}} methods which are the strongest available baselines.\footnote{\citet{gulrajani2020search, wa, swad, rame2022diverse, rame2023model} provide the details about how ensemble and weighting averaging outperform other baseline methods. }

\begin{table*}[th]
    \caption{\ood~performance comparison between very large dropout, ensembles, and weight averaging methods after hyperparameter selection. The hyperparameter is selected according to the best \iid. performance.}
    \label{tab:diverse_solutions_dominate_2d}
    \medskip
    \centering
    \setlength{\tabcolsep}{2mm} 
    \resizebox{\textwidth}{!}{
    \begin{tabular}{c| c|ccc|c|cc}
    \toprule
       & dataset & \textsc{erm} & \makecell{weight average\\ (single run)} & \makecell{ensemble\\ (single run)} & \makecell{very-large \\ dropout} & \makecell{weight average\\(multi run)} & \makecell{ensemble\\(multi run)} \\
    \midrule
       &  \textsc{vlcs} & 78.3 & 79.4 & 79.6 & \textbf{80.1} & 78.8 & 79.1 \\
      &  \textsc{office home} & 71.4 & 72.2 & 72.3 & \textbf{73.6} & 71.3 & 71.3 \\
      {\textsc{ResNet}} &  \textsc{pacs} & 87.3 & 86.9 & 87.3 & \textbf{88.5} & 87.0 & 87.1 \\
      &  \textsc{terra incognita} & 51.0 & 53.1 & 52.3 & \textbf{53.9} & 52.0 & 52.5 \\
        \cmidrule{2-8}
     & \textbf{Average} & 72.0 & 72.9 & 72.9& \textbf{74.0} & 72.3 &72.5 \\
     \midrule
     \midrule
    &   \textsc{vlcs} & 78.1 & 78.1 & 77.9 & \textbf{79.0} & 78.4 & 78.4 \\
    &    \textsc{office home} & 74.6 & \textbf{74.8} & \textbf{74.8} & 74.6 & 74.5 & 74.6\\
    \textsc{ViT-L-16} &    \textsc{pacs} & 85.0 & 84.2 & 84.3 & \textbf{86.0} & 84.7 & 84.8 \\
     &   \textsc{terra incognita} & 44.4 & 45.1 & 44.8 & \textbf{45.8} & 44.1 & 44.0\\
        \cmidrule{2-8}
     &   \textbf{Average} & 70.5 & 70.6 & 70.5 & \textbf{71.4} & 70.4 & 70.5\\
    \bottomrule
    \end{tabular}
    }
\end{table*}

\subsubsection{Very large dropout yields better \ood~performance}
\label{sec:dropout_exp}

Table \ref{tab:diverse_solutions_dominate_2d} shows our \textbf{main results} that comparing our very-large dropout approach and baseline methods on four \ood~datasets and two pretrained backbones.\footnote{Code: \url{https://github.com/TjuJianyu/verylarge_dropout}}

\begin{itemize}

 \item \textbf{{\textsc{erm}}} results are obtained by fine-tuning \textsc{ResNet50} or \textsc{ViT-L-16} using \textsc{sgd} with 0.9 momentum for $10,000$ iterations.\footnote{We use a batch size $32$ for all \textsc{ResNet} fine-tunings, and reduce the batch size to $16$ for all \textsc{ViT-L-16} fine-tunings due to the \textsc{vram} constraint.} A 10\% learning rate decay is applied at $5000^{th}$ iterations. For each choice of three training sub-datasets, we repeat three experiments for each combination of learning rate in $\{10^{-3}, 5.10^{-4}\}$ and L2 weight decay in $\{10^{-4}, 5.10^{-5}, 10^{-5}\}$. Following \citet{gulrajani2020search}, we prevent overfitting by early-stopping on 20\% hold-out \iid. validation examples, select hyperparameter (for each choice of training sub-datasets) according to the best \iid. performance. Finally, we evaluate the selected models on the fourth sub-dataset and average the four choices of training sub-datasets.  

    \item \textbf{Ensemble (single run)} results are obtained by an ensemble of checkpoints collected (every 300 iterations) along each fine-tuning trajectory. 

 \item \textbf{Weight average (single run)} results approximate the corresponding ensemble (single run) results by averaging the model weights instead of averaging the model outputs. 

 \item \textbf{Ensemble (multi run)} results are obtained by an ensemble of final checkpoints collected along all fine-tuning trajectories with different hyper-parameters ($2\times3=6$ in total). 

 \item \textbf{Weight average (multi run)} results approximate the corresponding ensemble (multi run) results by averaging the model weights. 

 \item \textbf{Very-large dropout} results are obtained using the same protocol but using a 90\% dropout rate on the penultimate layer representation.

\end{itemize}

As expected, both ensemble methods \citep{ensemble,dietterich2000ensemble}
and their weight averaging approximation \citep{rame2022diverse,wortsman2022model}
improve \textsc{erm} {on the} \ood~ performance. However, fine-tuning with a very large dropout outperforms the \ood~performance of both ensemble and weight averaging methods.

Because \textsc{ResNet50} produces a better performance than \textsc{ViT-L-16} on these \ood~fine-tuning tasks, our experiments in the following sections will be conducted on \textsc{ResNet50}.

\subsubsection{Very-large dropout $+$ other fine-tuning techniques}
\label{sec:exp_popular_finetuning_techniques}

\begin{table*}[h]
    \centering
    \caption{Very-large dropout $+$ a $10\times$ larger learning rate in the last layer. The first two columns show that this $10\times$ last-layer learning rate is helpful to \textsc{erm}. Then the middle two columns show that using a large dropout rate vastly improves the \ood~performance of merely using the increased learning rate ($\sim$$1.3\%$). The last two columns reveals that using this $10\times$ larger last-layer training rate yields small or zero incremental improvements over only using a large dropout rate ($\sim$$0.2\%$).}
    \medskip
    \setlength{\tabcolsep}{2mm} 
    \resizebox{0.7\textwidth}{!}{
    \begin{tabular}{c|ccccc}
        \toprule
        dataset & \textsc{erm} & \makecell{10$\times$ last-layer lr} &  \makecell{very-large dropout} & \makecell{very-large dropout\\+ 10$\times$ last-layer lr}\\
        \midrule
        \textsc{vlcs} & 78.3 & 79.9 {\scriptsize(+1.6)} & 80.1 {\scriptsize(+1.8)} & \textbf{80.5 {\scriptsize(+2.2)}} \\
        \textsc{office home} & 71.4 & 71.8 {\scriptsize(+0.4)}  & \textbf{73.6 {\scriptsize(+2.2)}}  & 73.3 {\scriptsize(+1.9)} \\
        \textsc{pacs} & 87.3 & 87.0 {\scriptsize(-0.3)} & \textbf{88.5 {\scriptsize(+1.2)}} &   88.3 {\scriptsize(+1.0)} \\
        \textsc{terra incognita} & 51.0 & 52.2 {\scriptsize(+1.2)} & 53.9 {\scriptsize(+2.9)} & \textbf{54.9 {\scriptsize(+3.9)}} \\
        \midrule
        Average & 72.00 & 72.73 & 74.03 & 74.25 \\
        \bottomrule
    \end{tabular}
    }    
    \label{tab:diverse_solutions_dominate_10}
\end{table*}

Various fine-tuning techniques have been proposed to improve the fine-tuning ability to leverage the representations learned by a pre-trained model, such as using a larger learning rate on the last layer \citep{caron2020unsupervised, bardes2021vicreg, kumar2022fine} or, as discussed above, using weight averaging and ensemble methods \citep{rame2022diverse,rame2023model,wa}. In this section, we show that incorporating these techniques \emph{in additional to very-large dropout} can further enhance \ood~performance, i.e. very-large dropout approach is compatible to these existing fine-tuning techniques. 

More importantly, very-large dropout approach dominates the \ood~performance improvements. i.e., all these finetuning techniques do not yield much \ood~performance improvements over using large dropout rates alone. %

\begin{table*}[th!]
    \caption{Very-large dropout $+$ ensembles or weight averagings. The \textsc{erm} and very-large dropout results are the same as those reported in Table~\ref{tab:diverse_solutions_dominate_2d}. In contrast, the ensemble and weight averaging results are now obtained by averaging the output or the weights of models fine-tuned \emph{with large dropouts}. Ensemble and weight averaging techniques provide a marginal \ood~performance improvement on \textsc{vlcs} or \textsc{office\,home} and a negligible \ood~performance improvement on \textsc{pacs} or \textsc{terra\,incognita}.}
    \label{tab:incremental_benefits_of_ensemble}
    \medskip
    \centering
    \setlength{\tabcolsep}{2mm} 
    \resizebox{\textwidth}{!}{
    \begin{tabular}{c|cccccc}
    \toprule
       dataset & \textsc{erm} & \makecell{very-large\\dropout} & \makecell{very-large dropout\\+ weight average\\ (single run)} &  \makecell{very-large dropout\\+ ensemble\\ (single run)} & \makecell{very-large dropout\\+ weight average\\ (multi run)} &  \makecell{very-large dropout\\+ ensemble\\ (multi run)} \\
    \midrule
    \textsc{vlcs}& 78.3 & 80.1 & 80.6 & 80.5 & 80.4 & 80.3 \\
    \textsc{office home}&71.4 & 73.6 & 74.2 & 74.3 & 74.4 & 74.2 \\
    \textsc{pacs}& 87.3 & 88.5 & 88.6 & 88.8 & 89.0 & 89.0 \\
    \textsc{terra incognita}& 51.0 & 53.9 & 54.0 & 54.7 & 52.3 & 54.7 \\
    \midrule
    \textbf{Average} &72.0 & 74.0 & 74.4 & 74.6 & 74.0 & 74.6 \\
    \bottomrule
    \end{tabular}
    }
\end{table*}

\bfparagraph{Very-large dropout $+$ large learning rates for the last layer}

Several authors routinely use a larger training rate on the last layer on the intuition that fine-tuning a pre-trained deep network on a different target task entails training a new last layer from scratch \citep{caron2020unsupervised, bardes2021vicreg, kumar2022fine}.

Table~\ref{tab:diverse_solutions_dominate_10} follows a similar fine-tuning process as in Table~\ref{tab:diverse_solutions_dominate_2d} but uses a $10\times$ larger training rate for the last layer classifier. Comparing the last two columns in Table~\ref{tab:diverse_solutions_dominate_10} shows that incorporating this $10\times$ larger last layer training rate is able to keep or improve the \ood~performance ($\sim$0.2\%). Comparing the middle two columns further shows that using a large dropout rate vastly improves the \ood~performance of merely using the increased learning rate ($\sim$$1.3\%$).

\bfparagraph{Very-large dropout $+$ ensemble or weight averaging}

Table~\ref{tab:incremental_benefits_of_ensemble} similarly explores the incremental benefits achieved by constructing ensembles or by averaging the weights of models fine-tuned with very large dropouts. The results show that very-large dropout approach is compatible with ensembles and weight averaging apporach to gain a non-negative incremental imporvements in \ood~performance. On the other hand, 
comparing Table~\ref{tab:diverse_solutions_dominate_2d} and~\ref{tab:incremental_benefits_of_ensemble} shows that fine-tuning with large dropout rates before computing ensembles or averaging model weights brings large \ood~performance improvements over fine-tuning without dropout. 

In short, \emph{the very-large dropout approach is compatible with other fine-tuning techniques but acts as the leading factor in terms of \ood~performance}.

\subsubsection{Robustness to hyperparameter selection}
Out-of-distribution finetuning performance is known to be sensitive to hyperparameter selection \cite{ahuja2020empirical,wortsman2022model}. To reduce the uncertain of hyperparameter selection, Figure~\ref{fig:diverse_solutions_dominate_2d} presents the box plot of different hyperparameter combinations (where each choice of training sub-datasets searches 6 hyperparameter combinations). 

On all four datasets, the bottom of very-large dropout box (25\% quartile) outperforms the top of other baseline boxes (75\% quartile). On \textsc{office\,home} and \textsc{pacs} datasets, there is even a \emph{large gap} between the worst dropout results and the best baseline results.

\begin{figure}[th!]
    \centering
\includegraphics[width=0.235\textwidth]{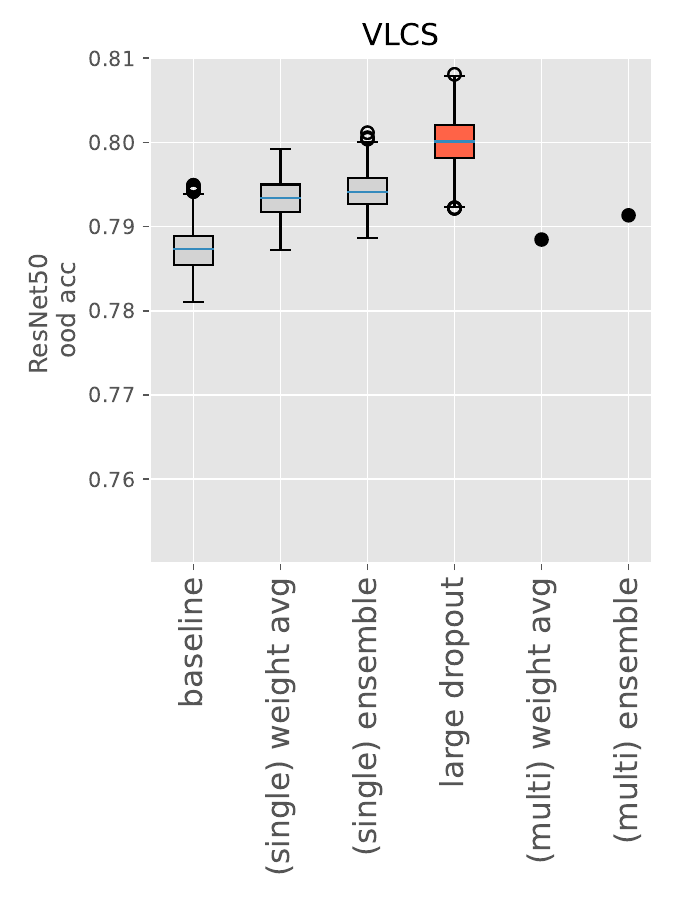}
\includegraphics[width=0.235\textwidth]{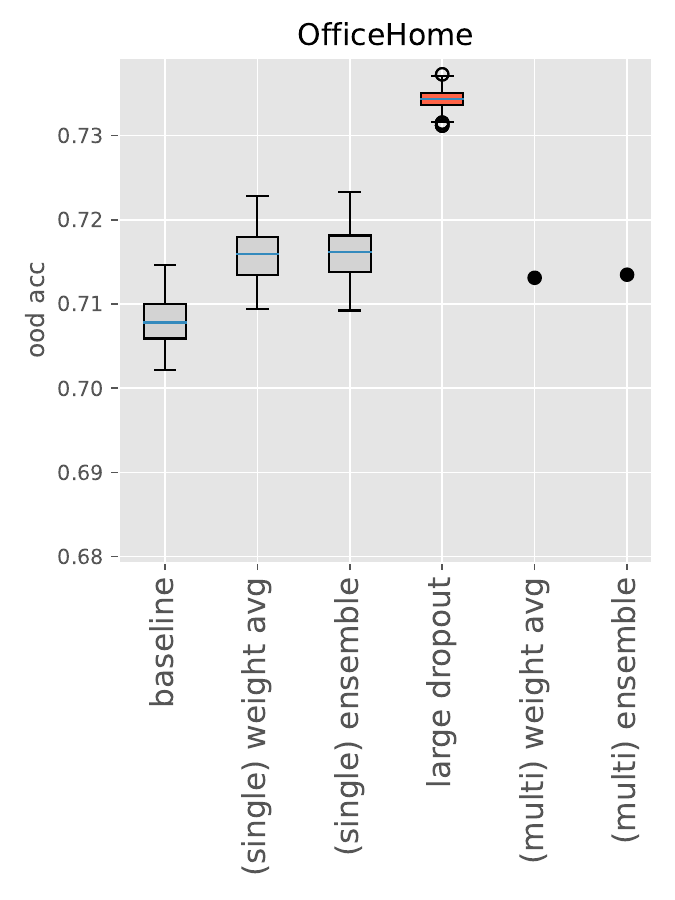} 
\includegraphics[width=0.235\textwidth]{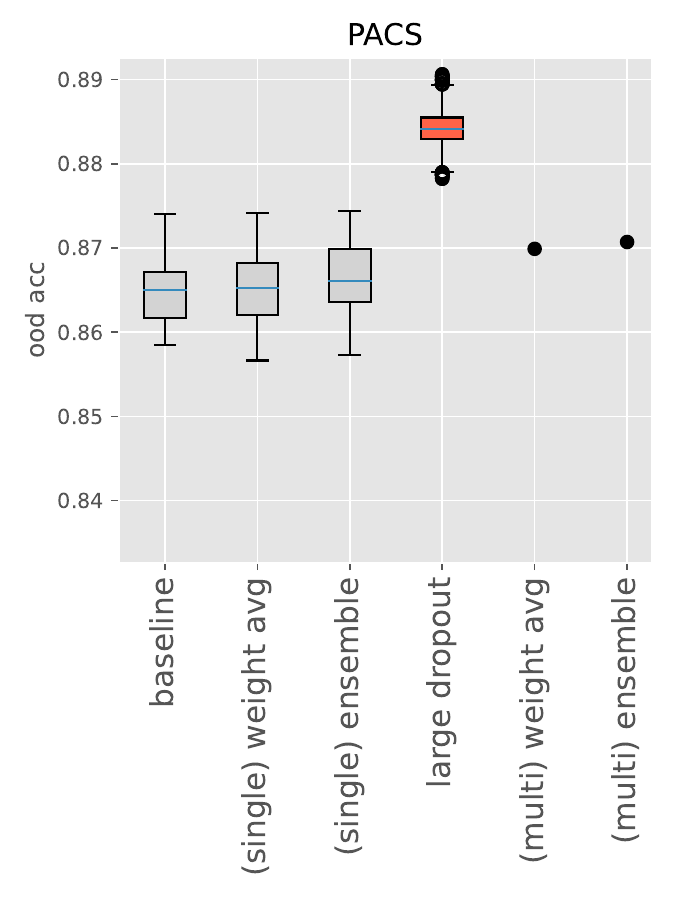}
\includegraphics[width=0.235\textwidth]{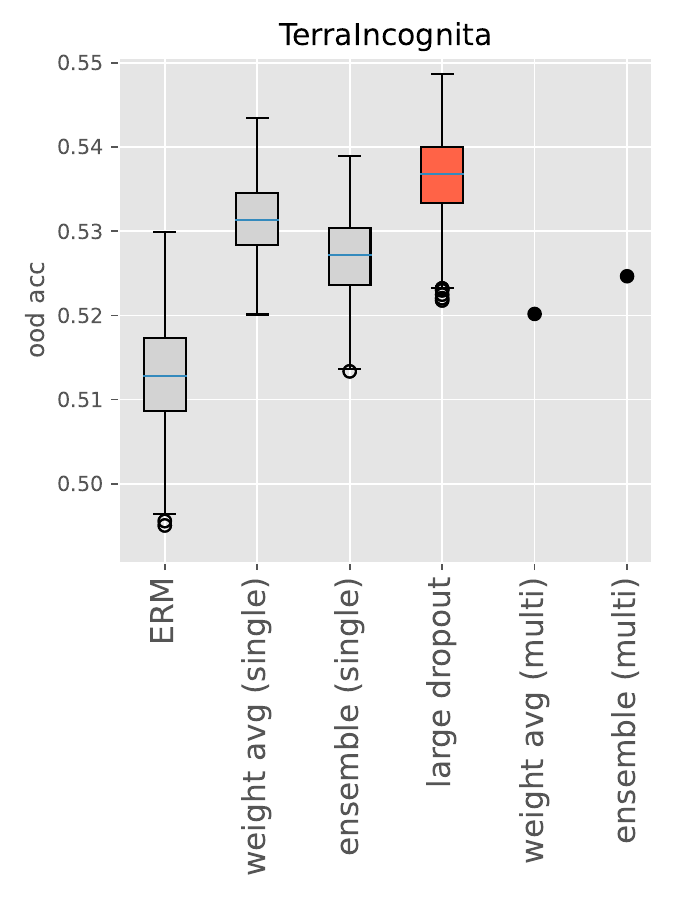} \\
    \caption{ \ood~performance comparison between very large dropout, ensembles, and weight averaging methods on four \textsc{DomainBed} tasks. \textbf{\textsc{erm}} results were obtained using plain fine-tuning with different hyperparameters. \textbf{Weight averaging} results either average the model weights collected every 300 iterations along each fine-tuning trajectory or the final model weights of all fine-tuning trajectories as in \citep{rame2022diverse}. \textbf{Ensemble} results average instead the model outputs. Finally, \textbf{large dropout} results were obtained like the \textsc{erm} results but using a 90\% dropout rate on the penultimate layer. Each box summarizes the results obtained with different hyper-parameters combinations.}
    \label{fig:diverse_solutions_dominate_2d}
\end{figure}

\subsubsection{Robustness of dropout rate selection}

As today, such large dropout rates (90\% and above) are considered unsuitable for training a network from scratch and have not been previously used for fine-tuning either. %
This section studies the relationship between dropout rates and \ood~performance. A smooth relationship indicates the robustness of dropout rate selection, while a curly relationship reflects the sensitivity.

Table~\ref{tab:compare_dropout_rate} compares various dropout rates on the four tasks. A $90\%$ dropout rate reliably produces good \ood~performance on all four tasks. The optimal dropout rate for \ood~performance ranges from 90\% to 95\% for \textsc{vlcs} and \textsc{pacs} task (with 10k examples). And becomes slightly smaller, about 90\%, for the slighlty larger datasets \mbox{\textsc{office\,home}} and \mbox{\textsc{terra\,incognita}} (with 15k to 25k examples).

Furthermore, the relationship between dropout rate and \ood~performance are smooth on all four datasets, which makes it easy to select the right dropout rate. 

\begin{table}[]
    \centering
    \caption{Effect of diverse dropout rates during fine-tuning. The best \ood~performances are attained using rates around or above 90\%. A large dropout rate (e.g. 90\%) reliably produces good \ood~performance on all four tasks.}
    \label{tab:compare_dropout_rate}
    \medskip
    \resizebox{0.7\linewidth}{!}{
    \begin{tabular}{c|ccccccc}
        \toprule
        dropout rate  & 0\% &10\% &30\% & 50\% &70\% & 90\% & 95\% \\
        \midrule
        \textsc{vlcs} & 78.3 & 79.2	& 79.3&	79.7&	79.6 & 80.1 & \textbf{80.4} \\
        \textsc{office home} &71.4 & 71.5&	72.3&	73.1	&\textbf{73.5}	 & \textbf{73.6} & 73.0 \\
        \textsc{pacs} &87.3 & 87.8&	87.4&	88.0	&88.1 & \textbf{88.5} & \textbf{88.4} \\
        \textsc{terra incognita} &51.0 & 50.2&	52.4&	52.4&	52.4 & \textbf{53.9} & 52.3 \\
        \bottomrule
    \end{tabular}
    }
\end{table}

\subsubsection{When should one apply very-large dropout?}

We have demonstrated that the very-large dropout method delivers consistently better \ood~performance than computing ensembles or weight-averages of models fine-tuned without dropout. However we also have argued that fine-tuning does not create new representations but merely exploits the representations already present in the pre-trained model. Therefore the final \ood~performance of this fine-tuning process must strongly depend on the quality and the diversity of the features present in the pre-trained network (\textit{richer representation}), even if these features are not exploited by the pre-trained network but buried in its hidden layers. i.e. the scope of applying very-large dropout method lies in situations where a \textit{rich representation} has already been established. 

Of course, modern foundational models, where many features are learned from a large and carefully constructed dataset, make this condition relatively easy to achieve. Thus provide a large space to apply this very-large dropout approach. 

In this section, we study this condition precisely. We first study the performance of very-large dropout approach on the scratch-training scenario, where the representation is random. Then we progressively enrich the representation by pretraining and pretraining with enormous augmentations.

\paragraph{{Random initialization and representation.}} Figure~\ref{fig:vlcs_scratch_training} shows the effect of various dropout rates when one trains a network on the \textsc{vlcs} task from scratch, that is starting from a randomly initialized network without pretraining (i.e. random initialization and random representation). The optimal dropout rate falls to about zero. Dropout rates higher than 50\% have a negative impact on both the \iid. and the \ood~performance of the network. \textit{This suggests that high dropout rates make it difficult to create new features (a nonlinear operation), but does not prevent leveraging existing features that were possibly buried in the network inner layers (a linear operation).}

\begin{figure}[th!]
\centering
\includegraphics[width=0.4\textwidth]{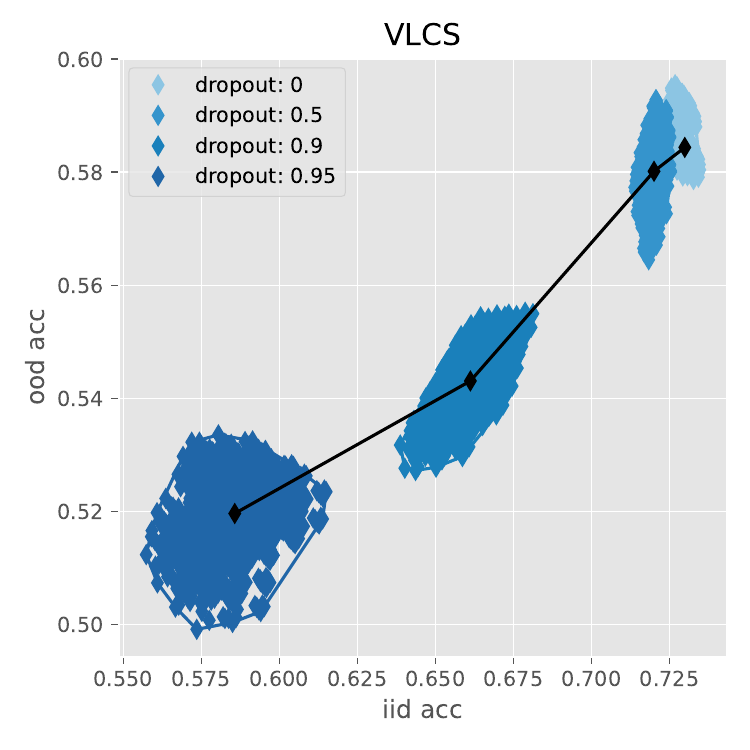}

\caption{Comparison of dropout rates when training a \textsc{ResNet50} network \emph{from scratch} on the \textsc{vlcs} dataset. The optimal dropout rate falls to about zero. Dropout rates greater than 50\% negatively impact both the \iid. and the \ood~performances. As a reference, a naive linear model on raw pixels (224$\times$224) achieves a 50.7 \iid~accuracy, approaching the \iid~accuracy of dropout rate = 95\%.}
\label{fig:vlcs_scratch_training}
\end{figure}
\paragraph{Richer and richer representation.}

To study the impact of rich representation, we compare the \ood~performance obtained by various methods applied to \textsc{ResNet50} networks pre-trained using the same \textsc{ImageNet} data but using different data augmentation schemes. As explained in the first paragraphs of section~\ref{sec:experiements}, the results reported so far use a network pre-trained using a broad array of data augmentation techniques, termed \textsc{ResNet\,\#2}. We now compare its fine-tuning properties with network termed \textsc{ResNet\,\#1} pre-trained using the simpler protocol described in \citet{he2016deep}.

Table~\ref{tab:resnetv1v2} compares the \ood~performances of both networks after regular fine-tuning and after fine-tuning with very-large dropout. Note that \textsc{ResNet\,\#2} contains richer representations than \textsc{ResNet\,\#1} due to the vast data augmentations. On \textsc{ResNet\,\#1}, where the representation is richer than random representation, a very-large dropout rate (0.9) starts to help \ood~performance (0.6\%). On \textsc{ResNet\,\#2}, where the representation is richer than \textsc{ResNet\,\#1}, the same very-large dropout approach vastly boosts \ood~performance (2\%).

The results in this section showcase an increasing \ood~benefits of the very-large dropout approach as the representation getting richer. Starting from the scale of \textsc{ResNet}50 and \textsc{ImageNet}, the \ood~benefits of a very large dropout becomes significant. 

In the context of large foundational models, both model size and dataset size are far larger than \textsc{ResNet}50 neural network and \textsc{ImageNet} dataset. Thus the space to apply this very-large dropout approach is large. 

\label{sec:richer_pretrain}
\begin{table*}[th!]
    \centering
        \caption{Comparison of the \ood~performances obtained after fine-tuning two pre-trained networks: \textsc{ResNet\,\#1} and \textsc{ResNet\,\#2}. Hyperparameters are selected according to the best \iid. performance. Compared with \textsc{ResNet\,\#1} \citep{he2016deep}, \textsc{ResNet\,\#2} was pre-trained with the vast array of data augmentation techniques. For each of these two pre-trained networks, we follow two fine-tuning approaches: 1) naive fine-tuning; 2) advanced fine-tuning including various tricks intended to improve the \ood~performance, \eg. large dropout (90\%), weight averaging, and increased last-layer learning rate, using hyper-parameters are selected according to the \iid. performance. Despite all this technology, advanced fine-tuning of a pretrained \textsc{ResNet\,\#1} (2nd column) barely matches the performance of naive fine-tuning on \textsc{ResNet\,\#2} (3rd column). }
    \label{tab:resnetv1v2}
    \medskip
    \setlength{\tabcolsep}{2mm} 
    \resizebox{0.78\textwidth}{!}{
    \begin{tabular}{c|cc|cc}
    \toprule
dataset & \makecell{\textsc{ResNet\,\#1}\\\textsc{erm}} & \makecell{\textsc{ResNet\,\#1}\\{very-large dropout}  } & \makecell{\textsc{ResNet\,\#2}\\\textsc{erm}} & \makecell{\textsc{ResNet\,\#2}\\{very-large dropout}} \\
\midrule
\textsc{vlcs} & 76.7 & 78.1 & 78.3 & 80.1 \\
\textsc{office home} & 68.9 & 69.1 & 71.4 & 73.6 \\
\textsc{pacs} & 86.2 & 86.5 & 87.3 & 88.5 \\
\textsc{terra incognita} & 48.2 & 48.8 & 51.0 & 53.9 \\
\midrule  
\textbf{Average} & 70.0 & 70.6 & 72.0 & 74.0 \\
\bottomrule
    \end{tabular}
    }
\end{table*}

\subsection*{Discussion}

The \ood~performance of fine-tuning with very large dropout consistently exceeds that achieved by popular techniques such as ensemble and by more recent techniques such as weight averaging.
Furthermore, ensemble and weight averaging techniques only bring a small incremental improvement when applied on top of fine-tuning with large dropout rates. This suggests that very large dropout implements a key factor that favors  \ood~performance, which we believe is related to
seeking features of type (a) among features of type (b) as explained in the introduction.

Both ensemble and weight-averaging techniques can be used for training a network from scratch or for fine-tuning a pre-trained network. In contrast, very large dropout rates cannot be realistically used when training a network from scratch. We argue that they work for fine-tuning because fine-tuning is well approximated as a linear process that can leverage their existing or buried features of a pre-trained network but cannot create new ones. Using large dropout rates is akin to a form of L2 regularization, expressing a richer set  of features even if redundant.  

This result also illustrates how the \iid~and \ood~scenarios can call for very different techniques.
It is well known that sparse representations can be very helpful in the \iid~scenario,
and it is increasingly clear that rich representations (\emph{rich features}) are preferable in 
the \ood~scenario \citep{zhang2022rich,zhang2023learning,chen2023towards}.
There are no reasons to expect that the many techniques designed for the \iid~scenarios will systematically help  \ood~generalization. The \emph{very-large dropout} case is one of many such examples.

\section{Conclusion}
\label{sec:rich_feature_conclusion}

This chapter explores the principle of \emph{rich features} and corresponding innovative techniques to help build AI for {open-world}. \emph{Rich features} principle prepares necessary features to enable quick learning of \emph{a broad range of \ood~tasks with fewer examples}. 

More precisely, this chapter explores the the failure of traditional \iid~training approach in rich feature discovery from both theoretical and experimental viewpoints, illustrates the benefits of rich features in transfer-learning, meta\&few-shot learning, invariant-learning, \ood~fine-tuning domains, and propose algorithms for the construction of rich features in different scenarios. 

Specifically, in transfer learning domain, concatenating multiple representations, \textsc{Cat}, constructs richer features than those obtained via \iid~training with a large network. In invariant-learning domain, {\textsc{Bonsai}} helps many invariant-learning algorithms to actually work, reveals that it is rich feature rather than the commonly believed invariant-learning penalties that matters in invariant-learning. In \ood~fine-tuning domain, the simple \textsc{very-large dropout} approach outperforms weight-averaging and ensemble.

These findings highlight the importance of \emph{rich features} to the build of AI for the open world, providing valuable insights into the construction of rich features.

\chapter{Disentangled Representation}
\label{chap:disentanglement_and_sample_complexity}

Disentanglement has long been identified as a desirable \cite{bengio2013deep} but challenging goal in AI \citep{comon-1994,roth-2022,thomas-2018}. Over the past decades, researchers have agreed on some key statistical properties of a disentangled representation. Specifically: assuming data generated by a set of unknown ground truth latent factors $\mathbb{S}$, and a subset of interesting latent factors $\Bar{\mathbb{S}}, \Bar{\mathbb{S}} \in \mathbb{S}$, a representation is said to be disentangled for $\Bar{\mathbb{S}}$ if there exists a one-to-one correspondence between each factor and dimension of the representation, {regardless of the rest factors $\mathbb{S} / \Bar{\mathbb{S}}$} \cite{roth2022disentanglement}. 

Meanwhile, some assumptions about the underling world, such as {Sparse Mechanism Shift}\footnote{\textit{``SMS: Small distribution changes tend to manifest themselves in a sparse or local way in the causal/disentangled factorization [...], that is, they should usually not affect all factors simultaneously.''}} \cite{thomas2018disentangling} and {Symmetry Transformation}\footnote{\textit{``In particular, our argument is based on the observation that many natural transformations will change certain aspects of the world state, while keeping other aspects unchanged (or invariant). Such transformations are called symmetry transformations, [...]''}. } \cite{higgins2018towards}, are proposed to argue that {the underlying factors of the world state tend to change in a sparse or local way. i.e. only a small fraction of factors change, while the rest (a large fraction) factors are unchanged / not affected.} These reasonable assumptions support the existence of disentanglement and the possibility of learning disentanglement.\footnote{We say ``these assumptions are reasonable'' in terms of thousands years human observations. It is possible that the underlining world operates a totally different principle that humans could never understand. Even this is the case, these assumptions are still helpful for us to understand and predict the world.}

These property and assumptions reveal the possible of ``reorganizing'' world factors such that successive states only differ on {few tractable factors}. Consequently, the required examples to learn the current state, based on previous states, dramatically decrease. \textit{That is, disentanglement leads to a reduced sample-complexity.}

This thesis aims at principles and techniques supporting AI for the open-world, where a machine is required to quickly learn a wide range of tasks with fewer examples and less priori knowledge. This chapter explores the principle of disentangled representation, specifically a cheaper yet reliable approach to drive disentanglement, called \emph{predictive disentanglement}.

This chapter is organized as follows. Section \ref{sec:disentanglement_linear_case} demonstrates how disentanglement reduces sample complexity in a logistic regression setting. Section \ref{sec:predictive_disentanglement} introduces \emph{predictive disentanglement}, a cheap yet reliable pressure (the ``quick learning'' pressure) to drive disentanglement in practice. Section~\ref{sec:3moons} illustrates this \emph{predictive disentanglement} using a network with only 54 parameters, showing that this is not a mysterious effect of scale but a property of architectures. Section~\ref{sec:layered_memories} extends these ideas to fully formed \emph{Memory Mosaics} architecture. Section~\ref{sec:language} reports on medium-scale language modeling experiments.

\section{Linear case: disentanglement reduces sample complexity}
\label{sec:disentanglement_linear_case}

This section uses logistic regression to demonstrate that a disentangled representation helps reduce sample complexity on learning new tasks. 

\sloppy \bfparagraph{Data Generation with disentangled representation.} We generate $n$ binary classification tasks $D_i=(X, Y_i), i\in[1,n]$ with a shared disentangled representation:
\begin{align}
    X &= [X_1, \dots, X_n] \sim  \mathcal{N}(0, \sigma^2\text{I})\\
    Y_i &\sim \mathcal{B}(|{\mathbbold{1}}[X_i > 0] - \epsilon|) \\
    D_i &= (X, Y_i), i \in [1,n]\,,
\end{align}
where $\mathcal{B}(\lambda)$ indicates Bernoulli distribution, $\mathcal{B}(\lambda)=1$ with  with probability $\lambda$, $\mathbbold{1}[true]=1$, $\mathbbold{1}[false]=0$.

The target label $Y_i \in \mathbb{R}$ of $i^{th}$ task, is sampled from Bernoulli distribution such that $Y=1$ with probability $1-\epsilon$ if $X_i>0$, $Y=0$ with probability $1-\epsilon$ if $X_i \leq 0$. I.e. target label $Y_i$ is only related with one input dimension. This is a simplified example of linear-probing (transfer-learning) from a pretrained disentangled representation to multiple tasks.

\bfparagraph{Data Generation with entangled representation} Similarly, we create another $m$ binary classification tasks $D^{\prime} = (X^{\prime}, Y_i)$ with a shared but entangled representation:
\begin{align}
     X^{\prime}&=AX \\
     D^{\prime} &= (X ^ {\prime}, Y)
\end{align}
where $A$ is an random orthonormal matrix such that $AA^\top = I$. By construction, $X^{\prime} \sim \mathcal{N}(0, \sigma^2I)$ (easy to check) follows the same distribution as $X$, but it is entangled. This simple example resembles the linear-probing from a pretrained but entangled representation to multiple tasks.

\bfparagraph{Experimental results}

To experimentally study the sample complexity of the two synthesized transfer learning cases, this section conduct experiments to optimize linear models via \textit{liblinear} solver \cite{scikit-learn} on both cases (number of features $m=100$) with different number of training examples. For each of the two cases, we search $[L_1, L_2]$ regularization strength in the range $C \in [0.01, 0.05, 0.1, 0.5, 1, 5, 10,50,100]$. The best hyperparameters are chosen based on a \iid~validation set. Finally, we report the performance on a \iid~test dataset in Figure \ref{fig:linear_disentanglement_benefit}.

\begin{figure}[ht!]
    \centering
    \includegraphics[width=0.5\linewidth]{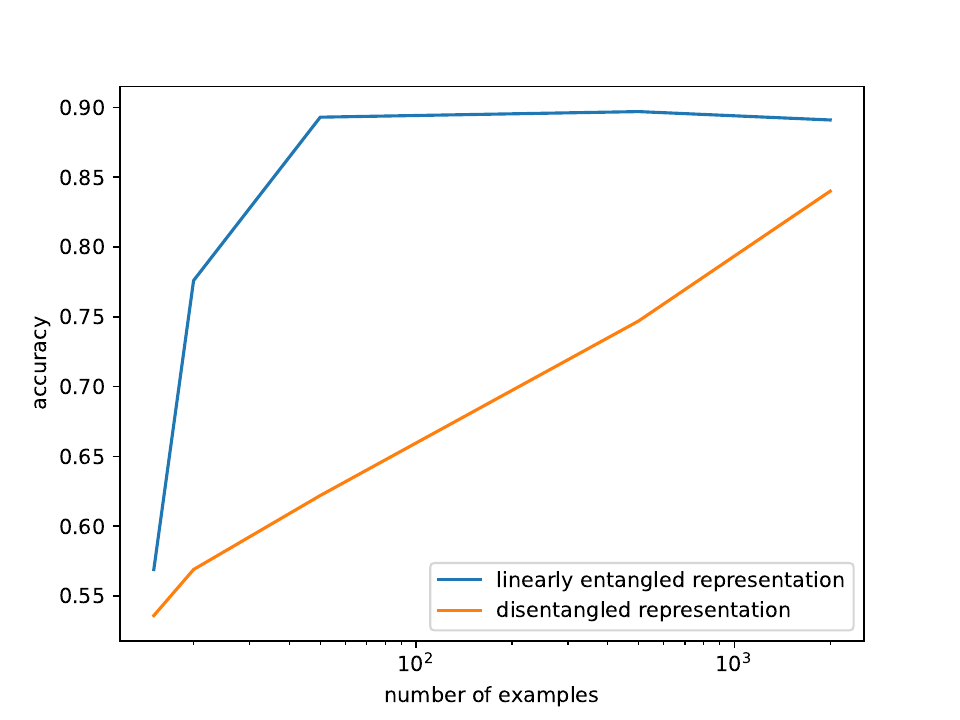}
    \caption{Sample complexity of linear model on disentangled representation and linearly entangled representation. Each point in the figure represents an average of 500 repeating experiments. \textbf{The results show the huge sample complexity gap between disentangled representation and entangled representation during linear-probing transfer-learning.} }
    \label{fig:linear_disentanglement_benefit}
\end{figure}

\bfparagraph{Theoretical results}
\citet{ng2004feature} (Theorem 3.1) shows that by incorporating this disentangled representation (where the dimension of representation is $n$, the effective dimension to learn a certain task is $r$, $r<<n$), the actual risk of a proper learning algorithm\footnote{i.e. structured risk minimization with $L1$ regularization.} trained on the {entire representation} (dim $n$) can be arbitrary closed the actual risk of a learning algorithm trained on the {effective dimension} (dim $r$) by using {$\Omega((log~n) \cdot {poly}(r, log(1/\delta), 1/\epsilon)$} examples, where $\epsilon$ is the difference of actual risks.

On the other hand, if the disentanglement property is destroyed (i.e., random scramble a disentangled representation with an invertible $n \times n$ matrix), \citet{ng2004feature} (Theorem 4.3) shows that there exist problems such that achieving the same $\epsilon$ requires {$\Omega(n/\epsilon)$} examples. 

\begin{boxA}
  In summary, disentanglement makes it possible to reduce sample complexity from $\Omega(n)$ to $\Omega(log~n)\cdot {poly}(r)$, $~r << n$, for logistic regression.   
\end{boxA}

\section{Predictive Disentanglement: a quick-learning pressure to drive disentanglement}
\label{sec:predictive_disentanglement}
Having shown the benefits of disentangled representation on sample complexity from both an experimental and a theoretical view, this section introduces a cheap yet reliable pressure, \emph{predictive disentanglement} \cite{zhang-2025}, to drive disentangled representation. The statistical view \cite{roth-2022} defines disentanglement on various ``independence'', and thus lack the robustness with respect to changing data distributions (not reliable). The causal view \cite{bengio-2013, bengio2019meta} defines disentanglement on active environments, and thus cannot be tested without active experiments (expensive). \emph{Predictive disentanglement} views natural sequences (e.g. articles, videos) as environments (cheap), uses quick-learning as the pressure (reliable).

Meanwhile, this section introduces a \emph{Memory Mosaics} \cite{zhang-2025} architecture to learn disentangled representation in practice. In {Memory Mosaics}, multiple associative memories work in concert to carry out a prediction task of interest. Such systems are closely related to memory networks \citep{weston-2014,sukhbaatar-2015} and resemble transformers \citep{vaswani-2017} despite significant differences. Like transformers, Memory Mosaics possesses some of the disentanglement and compositional capabilities that have long eluded machine learning systems \citep{lake-baroni-2018}. Unlike transformers whose internal mechanism are hard to decipher \citep{olsson-2022,bietti-2024}, Memory Mosaics achieve these capabilities in comparatively transparent ways.

To have a better illustration of \emph{predictive disentanglement}, we first describe simple associative memory units that can be inserted into a deep network in subsection~\ref{sec:memories}. Then explain how training such a network splits a prediction task into disentangled subtasks (i.e., illustrate the \emph{predictive disentanglement} principle) in subsection~\ref{sec-disentangle}.

\subsection{Memories}
\label{sec:memories}

\bfparagraph{Associative memory}

Generally speaking, an associative memory is a device that can store key-value pairs and retrieve values given a corresponding key. This definition omits important details about dealing with duplicate keys and approximate matches. For our purposes, both keys and values shall be vectors in~$\R^d$. The retrieval process can then be represented as a function of the queried key $k$ and all the stored pairs $(k_1,v_1)\dots(k_n,v_n)$.  
\[
     \left\{\begin{array}{lcl} 
        \R^d & \rightarrow & \R^d\\
        k & \mapsto & f\big(k;\: \{(k_1,v_1)\dots(k_n,v_n)\}\big)
    \end{array}\right.
\]
Except perhaps when duplicate keys are involved, an associative memory stores key-value pairs without consideration for their temporal ordering. Therefore the retrieval function can be assumed invariant with respect to any permutation of the stored pairs. This exchangeability property suggests that we can also view an associative memory as a device that estimates a conditional probability distribution $P(V|K)$ on the basis of the sample $(k_1,v_1)\dots(k_n,v_n)$ of key-value pairs. The retrieval function is then a conditional expectation over this estimated distribution:
\begin{equation}
\label{eq:regression}
     f\big(k;\: \{(k_1,v_1)\dots(k_n,v_n)\}\big) ~=~ \E\/(V\:|\:K=k)\,.
\end{equation}
Such a conditional expectation can be constructed with Gaussian kernel regression,\footnote{{Expression~\eqref{eq:gaussian-smoothing-with-distance} is known as the Nadaraya-Watson estimator \citep{nadaraya-1964,watson-1964}. It is known to converge to the true conditional expectation $\E(K|V)$ when $n\rightarrow\infty$ and $\beta=\sqrt{n}$.}}
\begin{equation}
\label{eq:gaussian-smoothing-with-distance}
   f\big(k;\: \{(k_1,v_1)\dots(k_n,v_n)\}\big) ~=~ {\sum_{i=1}^{n} \frac{1}{Z}}~e^{-\beta \| k-k_i\| ^2} v_i ~
    \quad\text{with}\quad Z={\sum_{i=1}^{n}} e^{-\beta\| k-k_i \| ^2}\,.
\end{equation} 
The close connection between this Gaussian kernel smoothing and attention \citep{bahdanau-2015} is obvious when all key vectors $k_i$ share a same squared norm because expression~\eqref{eq:gaussian-smoothing-with-distance} becomes
\begin{equation}
\label{eq:gaussian-smoothing-with-dp}
    f\big(k;\: \{(k_1,v_1)\dots(k_n,v_n)\}\big) ~=~ 
      \sum_{i=1}^{n}
      ~ \frac{e^{\,\beta\,k^\ttop k_i}}{\sum_{j=1}^{n} e^{\,\beta\,k^\ttop k_j} } 
      ~ v_i~.
\end{equation}

There are of course more advantageous ways to implement associative memories. Although some will certainly prove useful in the future, this paper only relies on associative memories implemented with Gaussian kernel smoothing, not least because that makes it easy to compute gradients.

\bfparagraph{Predicting with associative memories}

Consider now a sequence $(x_t)$ of observations, discrete tokens or continuous values. We would like to leverage the past observations $(x_t)_{t\leq T}$ to predict some useful property of the future observations $(x_t)_{t>T}$. For instance we might want to predict the next observation $x_{T+1}$ to construct an auto-regressive model of the sequence.

\begin{figure}
    \centering
    \includegraphics[width=.7\linewidth]{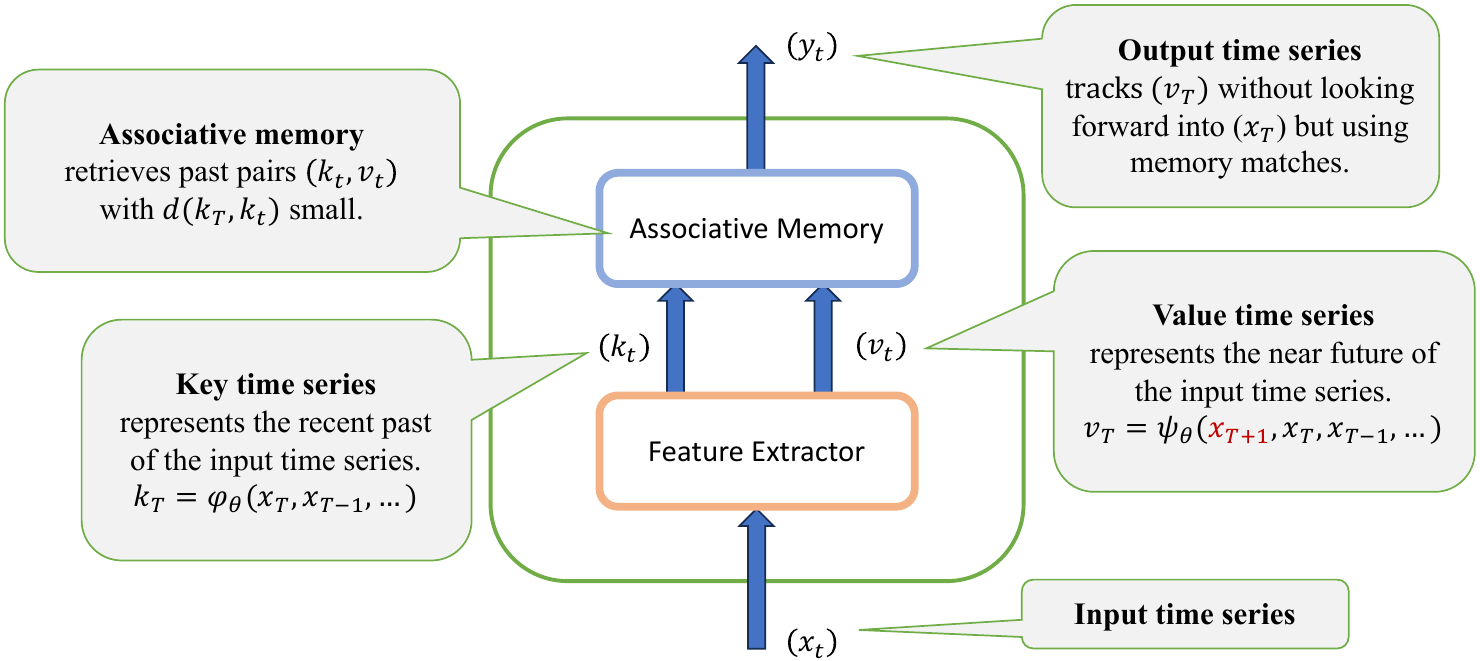}
    \caption{Elementary memory unit. The keys $k_T$ are computed as a function of past observations $(x_t)_{t\leq T}$. The values $v_T$ peek into the future. In this example, the value also depend on the next observation $x_{T+1}$. At time $T$, the associative memory uses the known key $k_T$ to compute an estimate $y_T$ of $\mathbb{E}(v_T|k_T)$ using only the previously stored pairs $(k_t,v_t)$, $t<T$. One time step later, the input~$x_{T+1}$ is revealed, the value $v_T$ can be computed, and the pair $(k_T,v_T)$ is added to the memory.}
    \label{fig:module}
\end{figure}

Our elementary memory unit (Figure~\ref{fig:module}) consists of an associative memory and a trainable feature extractor that computes suitable keys and values for the memory. The keys $k_T$ are computed as a function of the past observations $(x_t)_{t\leq T}$
and trainable weights $\w$,
\begin{equation}
\label{eq:varphi}
    k_T = \varphi(x_T,x_{T-1},\dots; \w)\,.
\end{equation}
In contrast, the values $v_T$ are allowed to peek in the future because they represent what the memory module aims to predict. For instance, the systems described in this paper merely allow values to depend on the next observation $x_{T+1}$,
\begin{equation}
\label{eq:psi}
    v_T = \psi(\textcolor{purple}{\mathbf{x_{T+1}}},x_T,x_{T-1},\dots; \w)\,.
\end{equation}
{
The memory units operate \emph{independently} at \emph{inference time}. They start empty at the beginning of each input sequence. At time step $T$, each memory receives a key vector $k_T$ computed from the recent inputs $(x_T, x_{T-1}, \dots)$ and interpolates a response $y_t$ on the basis of the previously stored key/value pairs. The value $v_T$ is computed one time step later when the next input $x_{T+1}$ is revealed and the pair $(k_T,v_T)$ is added to the memory.

Although the value $v_T$ depends on the near future, the output $y_T$ does not depend on $v_T$ but merely leverages the previously stored key/value pairs to estimate $v_T$. Therefore there is no leak of future information: each memory unit is a little machine that predicts a bit of future information (described by $v_T$) on the basis of recent information (described by $k_T$) and previously stored key/values pairs.}

The exact form of the feature extraction functions can vary in complexity. For instance, when each observation $x_T$ carries sufficient information, the keys $k_T$ and values $v_T$ can be computed as linear functions of respectively $x_T$ and $x_{T+1}$, that is $k_T = W_\varphi\,x_T$ and $v_T = W_\psi\,x_{T+1}$. 
However we find useful to consider feature extraction functions that summarize the \emph{recent past} using short convolutions or quickly vanishing leaky averages. For instance, the language experiments of Section~\ref{sec:language} use feature extractors of the following form:\footnote{{The leaking average in expression \eqref{eq:betterfeatures} is far too simple to effectively encode long range dependencies as demonstrated in \citep{voelker-2019,peng-2023,gu-2023}.}}

\vspace{-1ex}
\begin{equation}
\label{eq:betterfeatures}
 \begin{aligned}
 k_T &= \mathrm{Norm}\big(\textcolor{purple}{\Bar{k}_T}\big) &\text{with}  \quad \overbrace{\textcolor{purple}{\Bar{k}_T} = \textcolor{blue}{\tilde{k}_T} + \lambda_{\varphi}\textcolor{purple}{\Bar{k}_{T-1}} \quad\quad \textcolor{blue}{\tilde{k}_T} = W_\varphi\,x_T  }^{\text{leaky average over t = T, {T-1}\dots, 1}}  \\
   v_T &= \mathrm{Norm}\big(\textcolor{purple}{\Bar{v}_T}\big) & \text{with}  \quad   \underbrace{\textcolor{purple}{\Bar{v}_T} = \textcolor{blue}{\tilde{v}_T} + \lambda_{\psi}\textcolor{blue}{\tilde{v}_{T+1}} \quad\quad \textcolor{blue}{\tilde{v}_T} = W_\psi\,x_T
 }_{\text{convolution over t=T and T+1}}
 \end{aligned}
\end{equation}
Since this expression produces keys with unit norm ({\small $\mathrm{Norm}(x)=x/\|x\|$}), the effective kernel bandwidth is determined by the trainable parameter $\beta$ in equation \eqref{eq:gaussian-smoothing-with-dp}.

\bfparagraph{Training networks of memory units}

Consider now a deep network whose architecture includes layers of associative memory units. When the associative memories are implemented with differentiable kernel smoothing mechanisms, training such a deep network is simply a matter of unrolling the network in time and back-propagating the gradients, in ways that users of modern deep learning software will find very familiar. Unsurprisingly, unrolling equation~\eqref{eq:gaussian-smoothing-with-dp} along an input sequence $(x_1\dots x_D)$ of duration $D$ yields an expression that very much resembles masked self-attention \citep{vaswani-2017}.
\begin{align}
\label{eq:unrolling}
    \forall\,T\in\{1\dots{D}\} \qquad y_T = \sum_{i=1}^{T-1} 
      ~ \frac{e^{\beta\,k_T^\ttop k_i}}{\sum_{j=1}^{T-1} e^{\beta\,k_T^\ttop k_j} } 
      ~ v_i~,
\end{align}

Implementing associative memories with kernel smoothing therefore provides a particularly direct illustration of the connection between self-attention and associative memories (\eg., \citep{ramsauer-2020}). However, Memory Mosaics differ because the value extraction function is allowed to peek into the near future of the input time series $(x_t)$. This slight change has important consequences
\begin{itemize}[leftmargin=.2in]
    \item 
    Each memory unit operates as a little predictor whose outputs $y_T$ can be interpreted as a conditional expectation \eqref{eq:regression} that estimates features of the near future ($v_T$) of the input time series on the basis of its past observations ($k_T$). The parameters of the value extraction function ($\psi$) specify what is being predicted and the parameters of the key extraction function ($\varphi$) specify how it is predicted.
    \item 
    Equation~\eqref{eq:unrolling} must therefore account for the number of future time steps needed to compute~$v_T$. In our experiments, for example, $v_T$ can look one step ahead in the future. This amounts to having a more aggressive attention mask. Therefore the main diagonal must be excluded from the attention mask, justifying the ${T{-}1}$ upper bound in the sum.\footnote{One could of course use a more aggressive masking to allow $v_T$ peeking several time steps in the future.}
    \item
    Because each memory unit acts as a predictor, a single layer of memory units is sufficient to address the induction head problem of \citet{bietti-2024}. In contrast, a decoding transformer needs at least two self-attention layers for the same task.
    \item 
    Equation~\eqref{eq:unrolling} makes no provision for position encoding and no distinction between query and key vectors. In other words, we are betting that these transformers complications are no longer needed because our associative memory units do not need them to implement induction heads.
\end{itemize}

\subsection{Predictive Disentanglement}
\label{sec-disentangle}

\bfparagraph{Training and meta-learning}

The training process determines which future bit of information is predicted by each associative memory unit (through the parameters that control the computation of the values $v_T$) and which kernels are used to perform the predictions (through the parameters of that control the computation of the keys $k_T$). 
In contrast, \emph{the relation between keys and predicted values is determined for each input sequence at inference time} through the memorization of key/values pairs specific to each sequence.  The training procedure should therefore be seen as a \textit{meta-learning process}, distinct from the memory-based learning that occurs at inference time when new key/value pairs are added into the memories.

\bfparagraph{Predictive disentanglement}

This meta-learning interpretation reveals a remarkable phenomenon that we call \emph{predictive disentanglement}\:: the gradient training algorithm splits the overall prediction task (\eg., predicting the next token in a natural language sentence) into disentangled prediction sub-tasks assigned to each memory unit.

Consider a training set composed of long enough sequences $(x_1,\dots x_D)$ extracted from underlying time series governed by possibly different stationary processes. The goal of our network is to predict each $x_{T+1}$ using the previous observations $x_1\dots x_T$. Unrolling the network in time along each sequence $(x_1\dots x_D)$ and collecting the prediction losses measured at each position $t$ can be summarized by a curve that shows the prediction cost (or loss) at each time step $1\dots D$, as illustrated in Figure~\ref{fig:steamroller}. We can expect that the prediction cost observed at position $T$ becomes smaller when $T$ increases because more information $(x_1\dots x_T)$ is available to predict each $x_{T+1}$.

\begin{figure}[t]
\centering
\hspace{.06\linewidth}%
\vspace{-1ex}
\includegraphics[width=.6\linewidth]{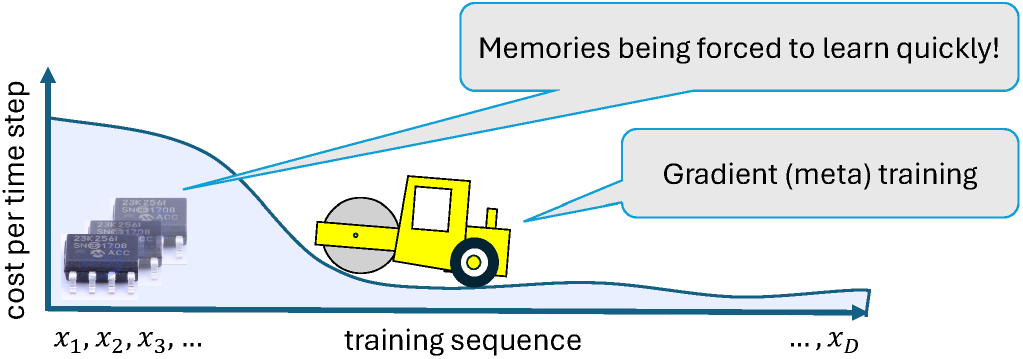}
\caption{The curve plots the prediction losses for all training sequence indices~$t\in\{1\dots D\}$ in the training sequence. Minimizing their sum ---the area under the curve--- favors memories that produce useful value estimates after fewer time steps.}
\label{fig:steamroller}
\end{figure}

The training process minimizes the total prediction cost, that is the area under the curve in Figure~\ref{fig:steamroller} viewed as a collection of vertical slices. We can also view this area as a collection of horizontal slices, each representing the context length required to drive the prediction cost below a certain threshold. Therefore the training process can also be viewed as \emph{minimizing the context length needed to produce good enough predictions} (the ``quick-learning'' pressure).

Because the associative memory retrieval function \eqref{eq:gaussian-smoothing-with-distance} is known to converge to stationary conditional expectations $\mathbb{E}(V|K)$, each memory unit is driven to produce a good conditional expectation estimate as soon as possible.
This can be achieved in two ways:
\begin{itemize}[leftmargin=.2in]
\item 
Let us first assume that each memory unit has a frozen value extraction function~$\psi$. The training procedure can still make each memory unit statistically more efficient by tuning the parameters of the key extraction function $\varphi$, that is, by learning how to compare the current prediction context $(x_T,x_{T-1},x_{T-2}\dots)$ with past prediction contexts~$(x_{t},x_{t-1},x_{t-2}\dots)$ for~${t<T}$. 

Learning a similarity metric (a kernel) is a well known way to make non-parametric estimators more efficient (\eg., \citealp{bach-2004-mkl}). For instance, the training procedure can construct keys that summarize the relevant contextual information, discarding noise factors that could increase the distance between keys associated with similar values. It can also adjust the effective kernel bandwidth, for instance, using parameter $\beta$ in equation~\eqref{eq:unrolling}.

\item

When multiple memory units are available, the training procedure  can also \emph{distribute} the overall prediction task among the available memory units. As long as the memory units outputs can still be combined to address the overall task, the training algorithm can optimize the parameters of the value extraction functions $\psi$ to produce values $v_T$ that more efficiently modeled by their respective memory units.

Because each memory unit operates independently at inference time, this works best when the overall prediction task is \emph{disentangled into smaller prediction sub-tasks that can be modeled independently and efficiently}. More precisely, the sub-tasks must be chosen so that each memory can carry out its assigned modeling task at inference time without having to account for the combined impact of the operation of all memory units.
Their outputs can then be recombined to provide predictions for inputs that are globally very different from the training inputs, but whose disentangled components are individually predictable, as illustrated in Section~\ref{sec:3moons}.

\end{itemize}

Disentanglement has long been recognized as desirable \citep{bengio-2013} but has been hard to pinpoint \citep{comon-1994,roth-2022,thomas-2018}. Predictive disentanglement is closely related to the meta-transfer objective of \citet{bengio-2019} but  arises as a side effect of a specific predictive architecture trained with the usual gradient procedure. Although predictive disentanglement is easier to understand in the case of a network of associative memory units, we conjecture that something similar also occurs in standard transformers.

\section{Tracking three moons case: learn disentanglement with 54 parameters and 1 layer}
\label{sec:3moons}

We give an illustrative example of predictive disentanglement: three moons orbit a remote planet. Although the local astronomers are very far from understanding celestial mechanics,\footnote{We do not seek to discuss subtleties such as elliptical orbits or multi-body problems. Our primitive astronomers are best compared to the ancient sky watchers whose efforts eventually gave the Ptolemaic model.} they nevertheless observe periodic motions and debate how to predict future moon positions. A first astronomer proposes to compile a single table containing the daily positions of all three moons, arguing that if the current set of moon positions matches a previous observation, the future moon positions will match the following observations. A second astronomer suggests instead to make three tables, one for each moon, arguing that the future positions of each moon can be independently predicted by matching its current position with a previously observed one. 

To make reliable predictions, the first astronomer needs a table that contains at least one record for each of the possible moon configurations. Our astronomer therefore needs to log the daily moon positions until all three moons return to their original configuration, after a number of days equal to the least common multiple $\mathrm{lcm}(p_1,p_2,p_3)$ of the individual moon periods. In contrast, the second astronomer only needs to log daily moon positions until each of the moons returns to a previously observed position, for a number of days equal to the period $\max(p_1,p_2,p_3)$ of the slowest moon.

One could argue that the proposal of the second astronomer is obviously superior because the three moons are distinct objects, well separated in space and time. One could instead argue that we view the moons as separate objects precisely because their respective futures can in general be independently predicted. Space and time separation merely suggests the possibility of independent predictions, as long as the moons do not collide. 

\begin{figure}
    \centering
    \begin{minipage}[c]{.33\linewidth}
        \includegraphics[width=\linewidth]{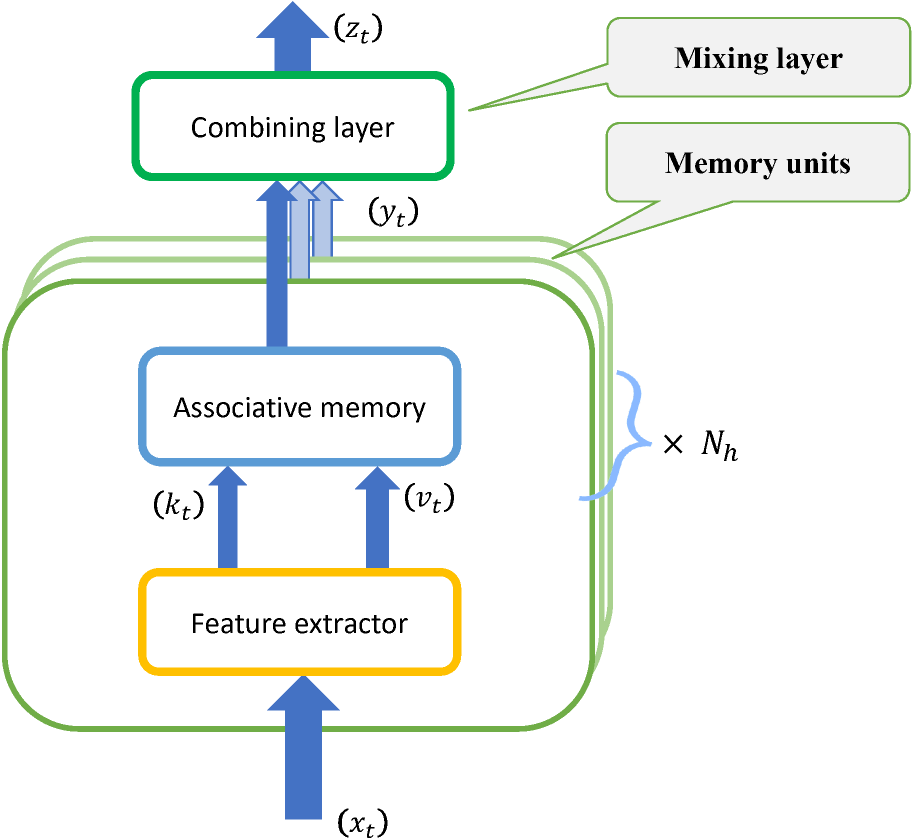}
    \end{minipage}
    ~~~~~~~~~~
    \hspace{-.05\linewidth}
    \begin{minipage}[c]{.4\linewidth}
    \small
    \def\C{\mathbb{C}}
    \def\stack{\mathop{\rm Stack}}
    \begin{align*}
    N_h &= 1 \text{~or~} 3\\
    \stack_{h=1\dots N_h}\left[k_T^{(h)}\right] &= W_\varphi\,x_T &  W_\varphi&\in\C^{3\times3}\\
    \stack_{h=1\dots N_h}\left[v_T^{(h)}\right] &= W_\psi\,x_{T+1}  &  W_\psi&\in\C^{3\times3}\\
    y_t^{(h)} &= \frac{1}{Z_T}\, \sum_{t<T} e^{\beta \, k_T^{(h)}\boldsymbol{\cdot}k_t^{(h)}}v_t^{(h)} \\
    z_t &= W_z\,\stack_{h=1\dots N_h}\left[y_T^{(h)}\right]  & W_z &\in\C^{3\times3}
    \end{align*}
    \end{minipage}
    \caption{An architecture for the three moons problem. 
    We consider single-layer networks with either $N_h=1$ or $N_h=3$ memory units whose keys and values belong to either $\mathbb{C}^3$ ($N_h=1$) or $\mathbb{C}^1$ ($N_h=3$). Both nets have $3\times3\times2\times3=54$ trainable real parameters that determine how to predict the moon positions using either a single 6-dimensional memory or three 2-dimensional memories.}
    \label{fig:singlelayer}
\end{figure}
\begin{figure}[ht]
\centering
    \setlength{\tabcolsep}{2mm}
    \begin{tabular}{p{.485\linewidth}p{.455\linewidth}}
      \includegraphics[width=\linewidth]{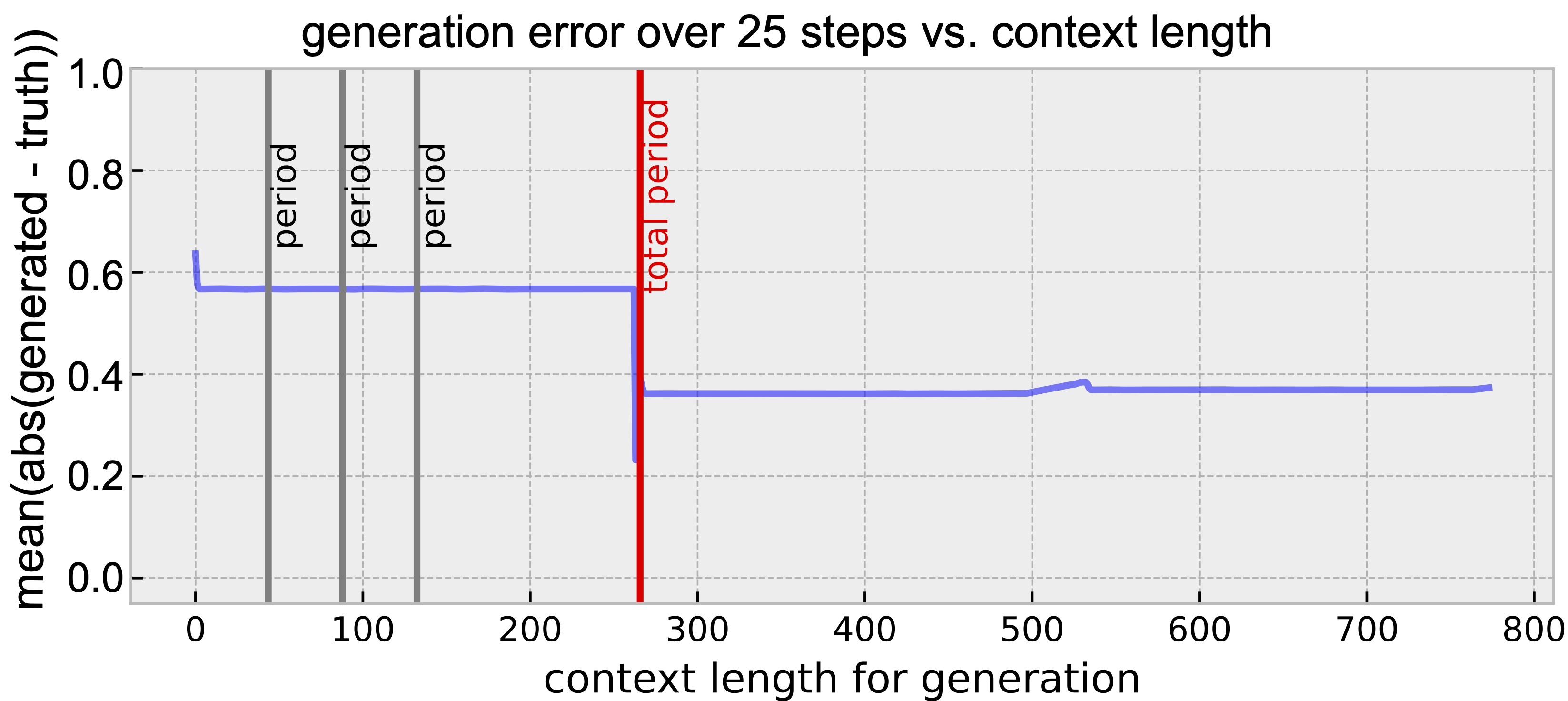}
      & \includegraphics[width=\linewidth]{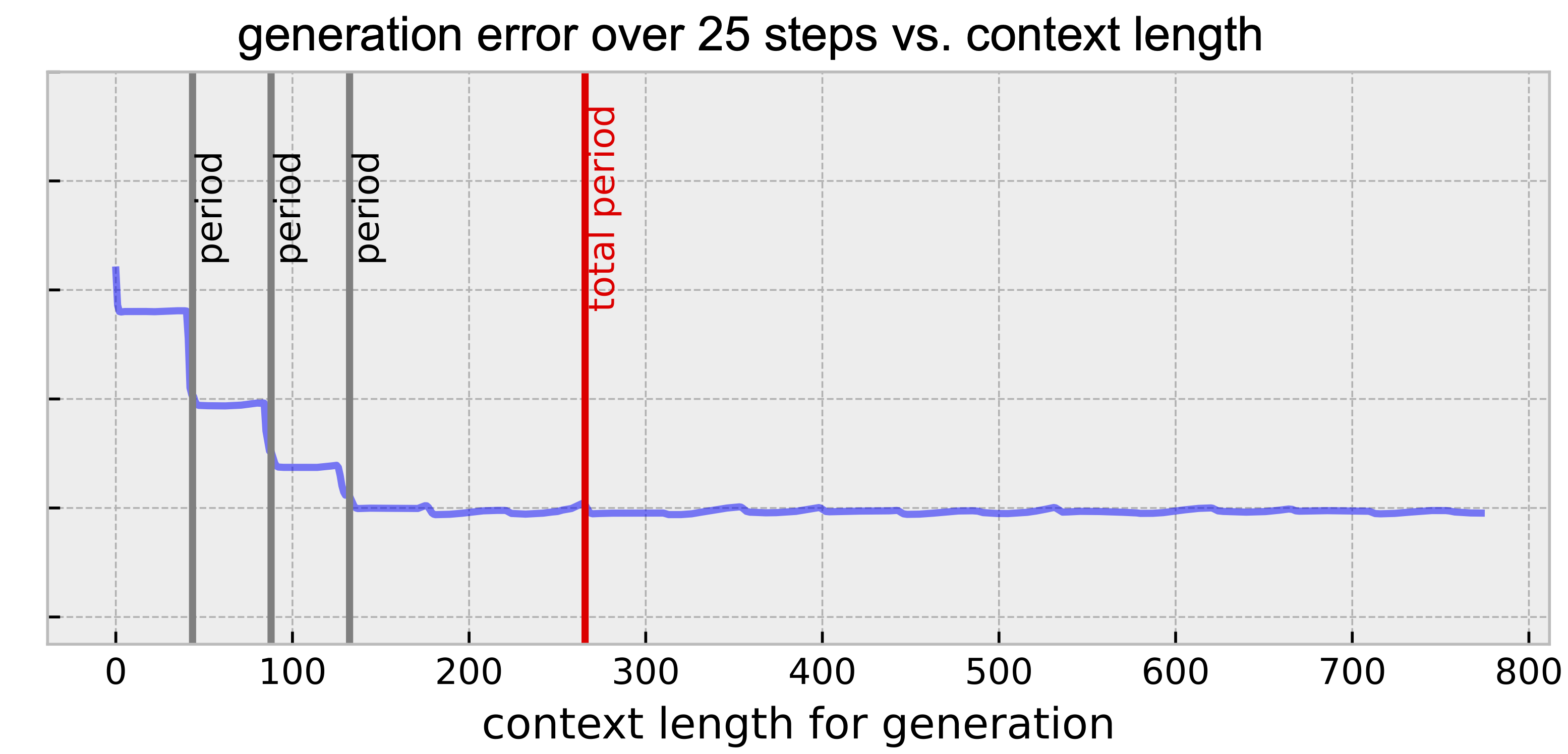} \\
      \caption{\label{fig:one-head-loss-all}%
        Single head network prediction error versus context length.
        The prediction error shows a sharp transition after $\mathrm{lcm}(p_1,p_2,p_3)$ observations (red vertical line), when the network switches from predicting the future moon position by repeating the last observation to predicting by find a matching memorized configuration.} 
      & \caption{\label{fig:3-head-loss-all}%
       Three-heads network prediction error versus context length.
       The prediction error improves whenever the context length reaches the period of a new moon (black vertical lines), yielding accurate predictions after the last one, well before having seen the full set of moon configurations (red vertical line).}
    \end{tabular}
    \vspace*{-2\baselineskip}
\end{figure}

\bfparagraph{Model}

For our purposes, each observation $x_t$ consists of three complex numbers $e^{i\theta_k}$ that encode the angular positions $\theta_k$ of the three moons inside their respective orbital plane. We consider two single layer models (Figure~\ref{fig:singlelayer}) with either $N_h=1$ or $N_h=3$ memory units whose added dimensions match the input dimension. The trainable parameters of the linear key and value extraction are  collected in two ${3\times3}$ complex matrices $W_\varphi$ and $W_\psi$. The memory unit follow equation~\eqref{eq:gaussian-smoothing-with-dp} with a fixed parameter $\beta=50$. A third ${3\times3}$ complex matrix $W_z$ combines the memory unit predictions into an output $z_T$ that hopefully predicts $x_{T+1}$. Both networks share an interesting analytic solution: setting all three matrices $W_\varphi$, $W_\psi$, and $W_z$ to the identity yields optimal predictions once the associative memories have seen enough samples.

\bfparagraph{Training}

The networks are trained using randomly generated sequences $(x_t)$ of length 800. Each sequence features three moons whose periods are related by randomly chosen ratios and are scaled to ensure that the 800 observation sequence contains at least three full periods lcm($p1,p2,p3$) of the moon system. Validation sequences are constructed similarly using a set of moon periods that does not appear in the training set. 

Figure~\ref{fig:one-head-loss-all} and~\ref{fig:3-head-loss-all} show the prediction errors of both networks as a function of the context length, that is, the number of observations stored into the memories.  More precisely, for each sequence $(x_t)$ and each time index $T$, we compute the average absolute deviation between the next 25 true moon positions $x_{T+1}\dots x_{T+25}$ and the next 25 auto-regressive predictions (in which the successive predictions are looped back into the network input.) The plots show curves averaged over 512 sequences sharing the same set of moon periods taken from either the training or validation set.

\begin{itemize}[leftmargin=.2in]
\item
For the single head network (Figure~\ref{fig:one-head-loss-all}), the plots show a sharp transition after $\mathrm{lcm}(p_1,p_2,p_3)$ observations, that is, when the memory contains a full set of moon configurations (red vertical line). Before this threshold, predictions are performed by repeating the last observation. After this threshold, predictions are performed by finding a matching moon configuration in the memory, just as suggested by the first astronomer.
\item
For the three-heads network (Figure~\ref{fig:3-head-loss-all}), the prediction error curve drops after seeing exactly $p_1$, $p_2$, and $p_3$ observations (black lines), that is whenever the orbit of an additional moon has been memorized. The learned weight matrices are shown Figure~\ref{fig:attn-3-heads} in the Appendix. Observe how the network produces accurate predictions after a time equal to the period $\max(p_1,p_2,p_3)$ of the slowest moon (last black line), long before the combined period $\mathrm{lcm}(p_1,p_2,p_3)$ (red line) of the moon system. In this interval, \emph{accurate predictions are returned for moon configurations that can be very different from the previously observed ones}. Instead the network \emph{combines individual moon predictions, each well supported by the past observations}. 

\end{itemize}

\bfparagraph{Predictive disentanglement and compositional learning in language models}
Consider a chat-bot assisted creative writing scenario in which the human uses dialogue to repeatedly introduce new ideas into an evolving story that the chat-bot reprints at each step. The user can drive such a story arbitrarily far from the training data and into the distant tail of its distribution. Although no training example resembles the story, the chat-bot keeps producing syntactically correct language and coherent stories because it has learned some of the mathematical structures of language \citep{harris-1968} and can  recombine pieces of information coming from either the context or the training data. This phenomenon is fundamentally similar to that illustrated in Figure~\ref{fig:3-head-loss-all}, where moon configurations unlike any previously seen configurations are accurately predicted because the network has learned how to combine individual moon predictions. This similarity casts a useful light on the otherwise mysterious compositional learning abilities of transformer-like models.

\section{Layered memories: predictive disentanglement in real }
\label{sec:layered_memories}

We of course envision deeper networks of memory units. In order to make meaningful comparisons, we also would like to remain as close as possible to the classic transformer architecture which alternates self-attention layers with fully connected feed-forward networks (FFNs).

\bfparagraph{Persistent memories}

\citet{sukhbaatar-2019} shows that FFNs in a transformer can be interpreted as \emph{persistent memories} that augment the self-attention layers and provide means to represent information that persists across input sequences.  
Besides the \emph{contextual memory units} (Figure~\ref{fig:module}), we therefore introduce \emph{persistent memory units} (Figure~\ref{fig:persistentmodule} in the Appendix) that contain a predefined number of key value pairs $(k_i,v_i)_{i=1\dots N_m}$ determined at training time through gradient back-propagation. Persistent memory units no longer need an explicit value extraction function because the memory content is not updated at inference time. As pointed out by~\citeauthor{sukhbaatar-2019}, they also can be viewed as fully connected neural networks with a single hidden layer that uses a soft-max non-linearity instead of a component-wise transfer function. Yet, we find conceptually useful to still view the persistent memory output $y_t$ as the conditional expectation $\E(V|K)$ of an implicit value function that is not explicitly parameterized, but can be figured out after training.

\bfparagraph{Routing}

Interleaving layers of contextual and persistent memory units can then be understood as means to increase the effective complexity of either the feature extractors or the combining layers of contextual memories (see Figure~\ref{fig:gpt2} for a spoiler). Therefore persistent memory units can also be seen as tool for \emph{routing information} between successive layers of contextual memory units.
Such a circuitry can implement routes that depend on the data, just like the gating modules of a mixture of expert \citep{jacobs-1991}.
Since all the parameters of such a circuitry are determined at training time, all the possible routes would have to be determined at training time. However the learning algorithm can overcome this limitation by also recruiting contextual memory units from adjacent layersy. Because the contents of contextual memory units are updated at inference time, recruiting some of them into the routing circuitry provides the means to create new routes on the basis of the first observations of a new sequence, suggesting an efficient alternative to capsule networks \citep{sabour-2017}.

\bfparagraph{Memory Mosaics}

In such a complex network, the division of labor between contextual memory units is still determined by the predictive disentanglement principle. During training, the steamroller of Figure~\ref{fig:steamroller} pushes the contextual memory units towards functions that more easily memorized independently than in aggregation. This does not only hold for memory units that record primary pieces of information such as the moon positions of Section~\ref{sec:3moons}, but also for those that affect the routing circuitry and those that operate on the information produced by earlier memory units.  

Therefore, \emph{under the pressure of the predictive disentanglement principle}, a network of memory units does \emph{not only memorize disentangled fragments of information, but also memorizes how they fit together and how their combinations can be again broken into new disentangled fragments and recombined in myriad ways}. This is why we call such networks \emph{Memory Mosaics}.

\section{Experiments}
\label{sec:language}

We have so far described Memory Mosaics as an architecture that resembles transformers in important way but {offers additional insights such as predictive disentanglement}. We now provide evidence that Memory Mosaics can handle the most successful application of decoding transformers, that is, language modeling.

\bfparagraph{Language modeling task}

{
The \textsc{TinyStories} work of \citet{eldan-2023} shows how to study large language modeling questions using small language models. This is achieved by limiting the scope to tiny stories written in simple english and taking place in the simple world that a three years old child could understand. A small language model trained on such data generates continuations with far better language quality and narrative consistency than those a much larger model (1.5B parameters) trained on a generic text.
}

Following both the lead of \citeauthor{eldan-2023} and the advice of our legal department, we leverage the \textsf{\small Mixtral-8x7B} open language model \citep{jiang-2024} to generate a new corpus of tiny stories dubbed \babistories. This corpus and its generation are detailed in Appendix~\ref{sec:alt_tinystories}.\footnote{We share the \textsc{BabiStories} dataset and Memory Mosaics source code at \url{https://github.com/facebookresearch/MemoryMosaics}.}

\bfparagraph{Architecture}
\begin{figure}[ht]
\centering
\includegraphics[height=.38\linewidth]{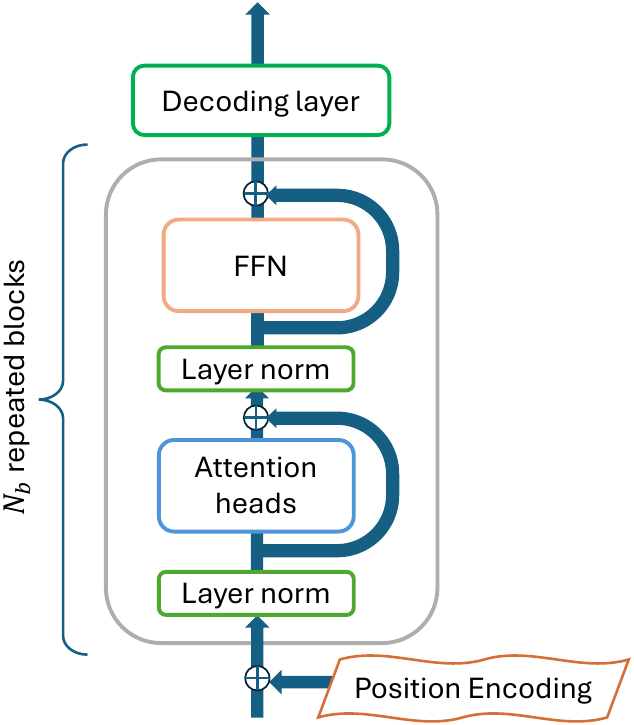}
\quad
\includegraphics[height=.38\linewidth]{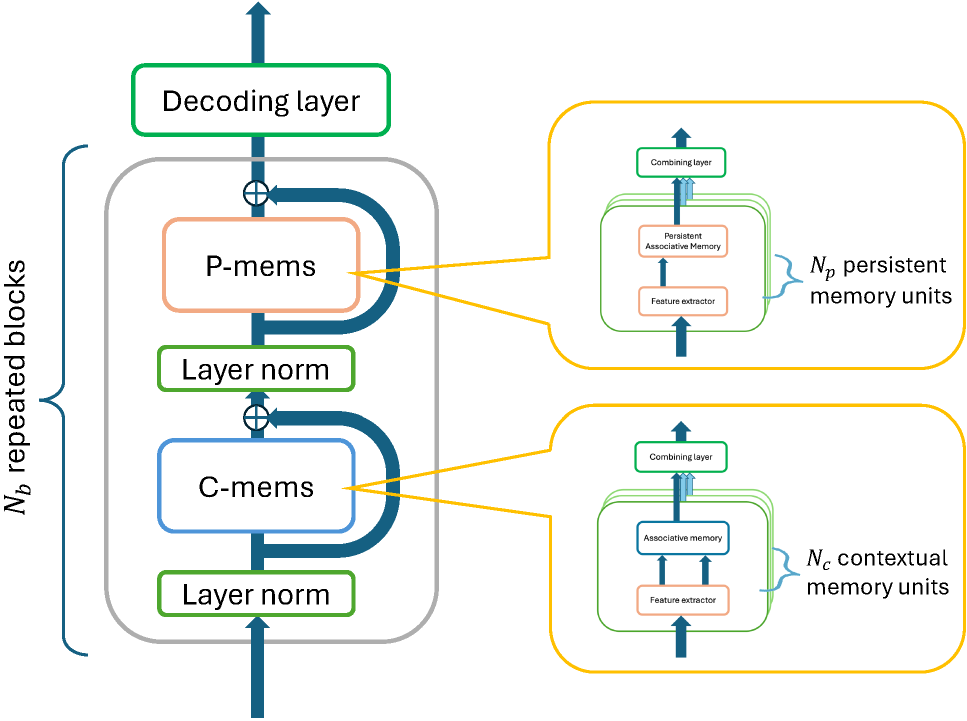}
\caption{\emph{Left}: Classic GPT2-small transformer. \emph{Right}: GPT2-like Memory Mosaic}
\label{fig:gpt2}
\end{figure}

To put our experiments into context, we design a Memory Mosaic architecture that closely matches the classic GPT2-small transformer architecture \citep{radford-2018,radford-2019}. Both architectures, shown side-by-side in Figure~\ref{fig:gpt2}, use the same GPT2 tokenizer, the same embedding dimension ($d=768$), and the same number of heads (${N_h=N_c=N_p=12}$). Both architectures are trained and tested using sequences of length 512, that is, one to three stories long.

There are three major differences between these two architectures. First, the Memory Mosaic does not use positional encoding. Second, unlike the $N_h=12$ attention heads of each transformer block, the $N_c=12$ contextual memory units in each block do not distinguish keys from queries (Figure~\ref{fig:module}) but instead use the key and value extraction functions described in Equation~\ref{eq:betterfeatures}. The keys are formed with a leaky average of past inputs, and the values can peek one time step ahead.\footnote{The key idea here is to define key and value extraction functions that combine a couple successive inputs~$x_t$ instead of just one as in the three moons example. Many variations perform more or less equivalently.}  Accordingly, the attention mask excludes the main diagonal to avoid breaking causality. Finally, the feed forward networks (FFNs) of the classic transformers blocks are replaced by a layer of $N_p=12$ persistent memory units, complete with a key extraction functions \eqref{eq:betterfeatures} and combining layer. These persistent memory units are sized to ensure that the per-block parameter count of the Memory Mosaic architecture closely matches GPT2-small.\footnote{Compared with GPT2-small, we save ${768\times512}$ position encoding weights and ${N_b\times768^2}$ query projection weights, but add ${2\times{N_b}\times768^2}$ weights for the persistent memory key extraction and mixing layer. The total number of persistent memory unit slots is therefore close to the total number of FFN hidden units.}

\bfparagraph{Training and validation}

\begin{figure}[ht]
    \centering
    \includegraphics[width=\linewidth]{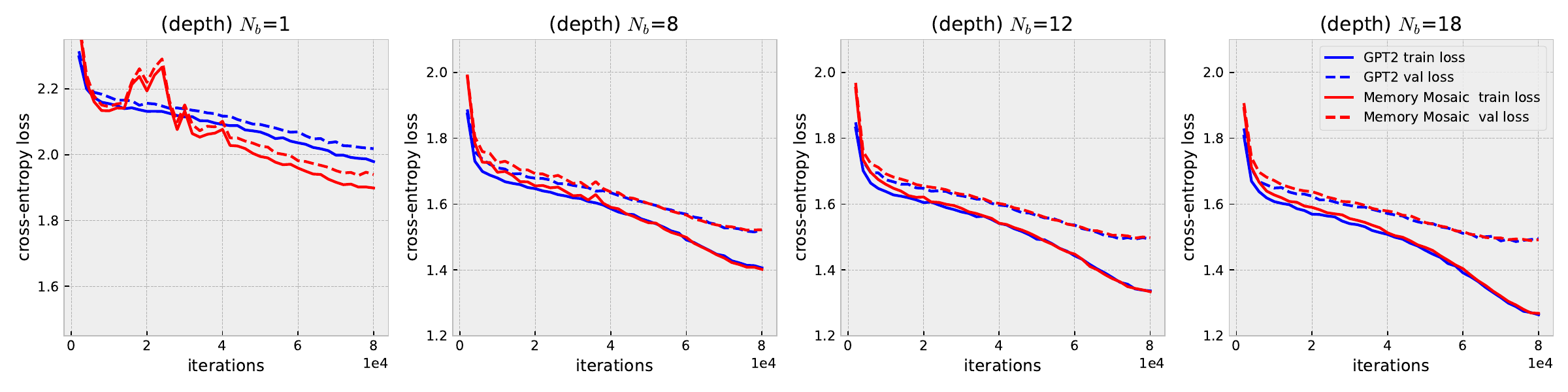}
    \caption{Training and validation loss of the transformer and Memory Mosaic architectures trained on {\babistories} for different model depths. The horizontal axis represents the number of training iterations. All hyper-parameters have been tuned on the transformer architecture and transferred verbatim to the Memory Mosaic architecture.  The Memory Mosaic slightly outperforms the transformer for small depth networks, but that effect disappears when the depth increases.{Additional results are presented in Appendix~\ref{app:figseven}.}}
    \label{fig:gpt2_mm_comparison}
\end{figure}
Figure~\ref{fig:gpt2_mm_comparison} shows the training and validation curves of both transformers and Memory Mosaics of different depth trained on \babistories. The Memory Mosaic slightly outperforms the transformer for small depth networks,\footnote{This is not surprising because Memory Mosaics only need a single block to implement induction heads, whereas transformers need at least two for the same task.} but this effect disappears when the depth increases and both the training and validation losses become indistinguishable. {Additional results are presented in Appendix~\ref{app:figseven}.}

Importantly, all hyper-parameters were tuned for the transformer architectures (Appendix~\ref{sec:gpt2_baseline_hyperparameters}) and transferred verbatim to the Memory Mosaics. This choice might explain why the training curves track each other so well. It also leaves the Memory Mosaics at a slight disadvantage.

\bfparagraph{Qualitative evaluation}

In order to compare the quality of the text generated by models trained on tiny stories, \citeauthor{eldan-2023} designed twenty-four prompts that exercise the factual, logical, and consistency properties of the generated continuations. Table~\ref{tab:gpt_mm_generation_18layers} in the Appendix compares the continuation generated on these prompts by a transformer and a Memory Mosaic, both $N_b=18$ blocks deep. Both models perform very similarly on this task.

\begin{figure}
    \centering
     \setlength{\tabcolsep}{2mm}
     \begin{tabular}{p{.45\linewidth}p{.48\linewidth}}
      \includegraphics[width=.98\linewidth]{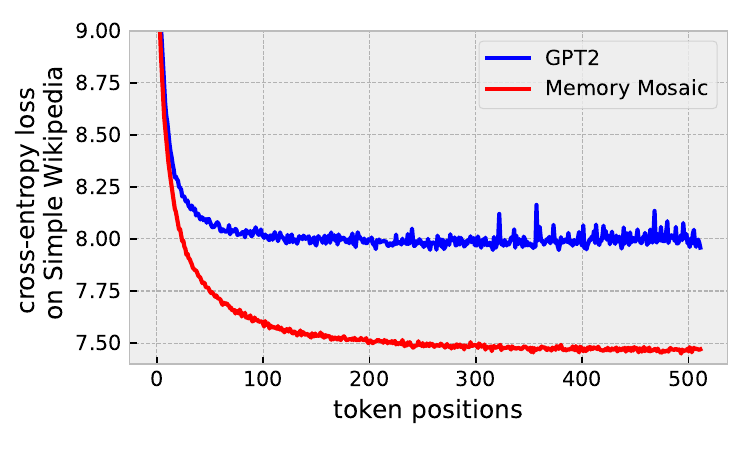}
      & \includegraphics[width=1\linewidth]{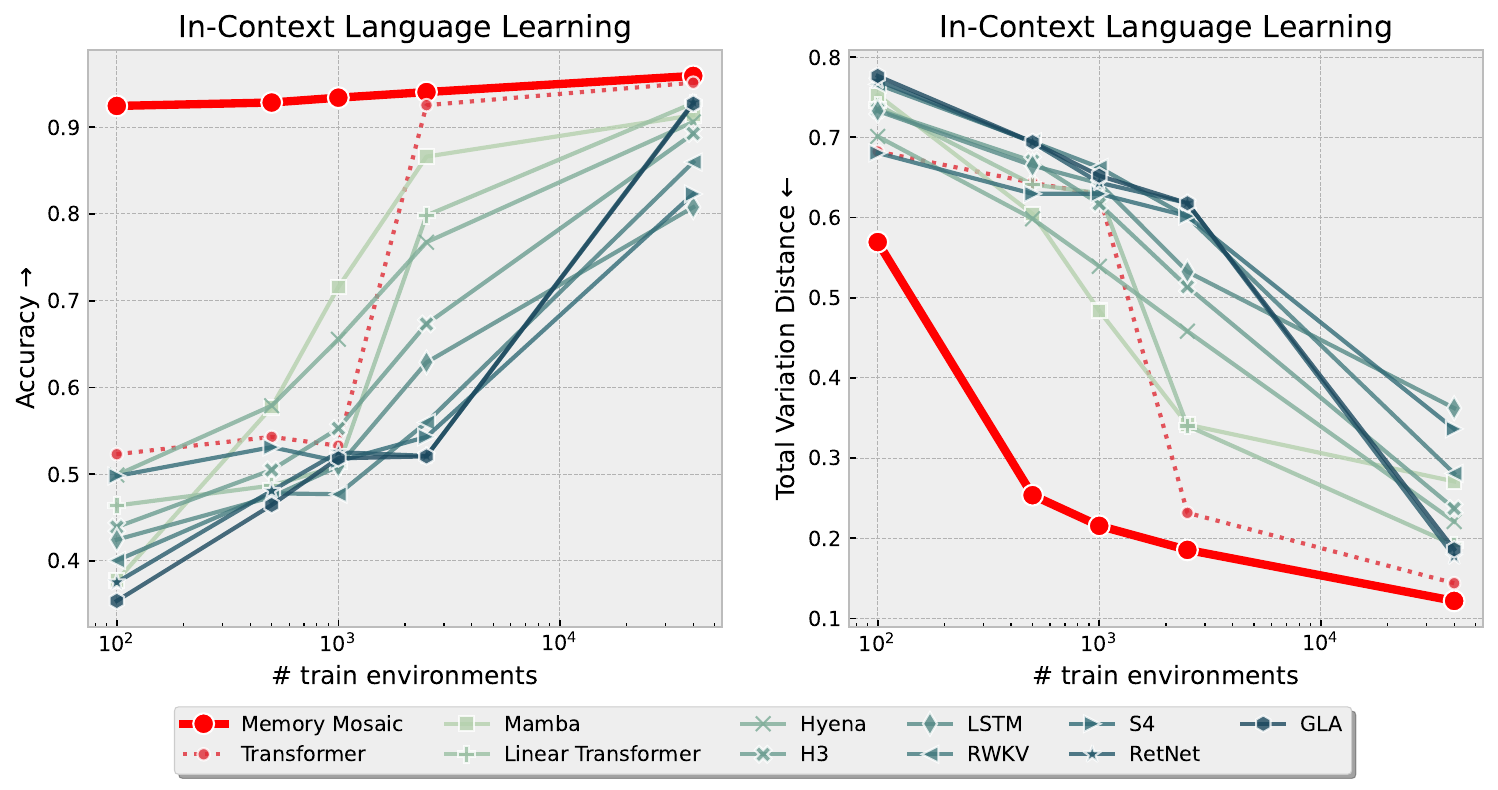} \\
      \caption{\label{fig:gpt_mm_ood}%
      Prediction performance on the Simple English Wikipedia dataset using models trained on {\babistories}. The plot shows the per-token average loss as a function of the position of the generated token in the 512-token long input window. Memory Mosaics outperform transformers after about 50~tokens, suggesting superior in-context learning abilities.} 
      & \caption{\label{fig:icll_acc}%
        Memory Mosaics performance on the \textsc{RegBench} in-context learning benchmark \citep{akyurek2024context}. Since \textsc{RegBench} includes an hyper-parameter search, Memory Mosaics and transformers use the same search space with the same parameter counts. Memory mosaics outperform all previously tested architectures in this benchmark.}
    \end{tabular}
    \vspace*{-2\baselineskip}
\end{figure}

\bfparagraph{Out-of-distribution evaluation}
The Simple English Wikipedia\footnote{Described in   \url{https://simple.wikipedia.org/wiki/Simple_English_Wikipedia} with downloads in  \url{https://huggingface.co/datasets/wikipedia\#20220301simple}.} is a version of Wikipedia written in a language that is easier to understand. Despite the intended simplicity, the articles are substantially longer and more sophisticated than our \babistories. Predicting Simple English Wikipedia articles using models trained on \babistories is therefore a challenging out-of-distribution task. 

Figure~\ref{fig:gpt_mm_ood} shows the per-token average loss as a function of the position of the generated token in the input window. Both the transformer and the Memory Mosaic are $N_b=12$ blocks deep. In this experiment, the token prediction is expected to improve when the increasing context size reveals that the distribution is different. The transformer performance plateaus after 100 to 150~tokens, which is a bit shorter than a typical tiny story. Memory Mosaics substantially outperform transformers after about 50~tokens, suggesting superior in-context learning abilities.

\bfparagraph{In-context learning evaluation}

In order to rigorously compare the in-context learning abilities of various architectures, the \textsc{RegBench} benchmark \citep{akyurek2024context} constructs random artificial languages defined by probabilistic finite automata (PFA). Each input sequence is composed of~10 to~20 strings drawn from a same PFA and delimited separator tokens. The competing architectures are trained on a variable number of input sequences, then evaluated on their ability to predict the last token of testing sequences generated using held out PFAs. 

Since \textsc{RegBench} performs a hyper-parameter searches, we use the Memory Mosaic architecture of Figure~\ref{fig:gpt2} with the same search space as transformers, ensuring that both transformers and Memory Mosaics have the same parameter count for the same architectural hyper-parameters. We sweep over depth \mbox{$N_b\in\{2,4,8\}$}, number of heads \mbox{$N_h{=}N_c{=}N_p\in\{2,4,8\}$}, embedding dimension in \mbox{$d\in\{64,128,256\}$}, weight decay in $\{10^{-2}, 10^{-1}\}$, and training epochs in $\{1,2,\dots200\}$.

Figure~\ref{fig:icll_acc} compares Memory Mosaic on \textsc{RegBench} with the results previously reported by \citeauthor{akyurek2024context}. The left plot shows the prediction accuracy for the test string last token. The right plot compares the predicted last token distribution with the exact distribution implied by PFA. Memory Mosaics dominate this benchmark, substantially outperforming transformers, recurrent neural networks, and state-space models for training set sizes covering three orders of magnitude.\footnote{Although the baseline methods trained with small training sets (e.g. 100) perform poorly on the \textsc{RegBench} task, they perform very well when tested in-distribition (see Table~\ref{tab:icll_iid_100} in the Appendix). Therefore they learned to model the training languages but did not acquire the ability to learn new languages in context.}

\section{Discussion}
\label{sec:discussion}

\bfparagraph{Disentanglement and AI for the open-world} 
AI for the open-world requires a machine to learn on a wide range of new tasks/domains (versatility) quickly using fewer examples and less task-specific priori knowledge (from human designers). Building AI for the open-world, thus, requires unique learning principles.
Pre-preparing \emph{rich features} for unseen tasks before facing them is the first principle introduced in Chapter \ref{chap:rich_representation_ood_scope}. Given a rich set of features, organizing these features in a ``nice'' way such that successive tasks (e.g. time or location) only differ on a small fraction of features is the second principle introduced in this chapter. This principle is called \emph{disentangled representation}. It reduces the number of examples required for learning unseen tasks.

Disentanglement and its benefits are not new topics \cite{ng2004feature,bengio-2013, bengio-2019, roth-2022}, but a cheap and reliable pressure to drive disentanglement has remained elusive. %
One contribution of this chapter is the introduction of \emph{predictive disentanglement}, which provides a cheap and reliable pressure to drive the learning of disentanglement. Unlike the causal viewpoint of disentanglement \cite{bengio-2013,bengio-2019}, predictive disentanglement does not require active or annotated environments, thus cheap. Unlike the statistical viewpoint of disentanglement \cite{roth-2022}, predictive disentanglement is a direct pressure of ``quick-learning'', thus reliable.

\bfparagraph{Memory Mosaics and Predictive disentanglement}

The starting point of Memory Mosaics is made of two very old ideas. The first one is augment a deep network with explicit memories. The second one is to let the learning process decide what gets memorized and how it gets retrieved. Although such ideas have been explored in memory networks \citep{weston-2014,joulin-mikolov-2015,sukhbaatar-2015}, the importance of having lots of independent memories had not been fully appreciated.

This contribution focuses on networks of associative memories implemented with kernel smoothing, therefore amenable to gradient-based learning algorithms. Such learning machines not only resemble decoding transformers (Section~\ref{sec:memories}) but also perform very much like decoding transformers on the sort of language modeling task that made them famous (Section~\ref{sec:language}). Although much work is needed to replicate our observations at far greater scale, Memory Mosaics satisfy narrative constraints as well as transformers (Table~\ref{tab:gpt_mm_generation_18layers}), and generally behave in very encouraging ways (Figures~\ref{fig:gpt_mm_ood} to \ref{fig:gpt2_mm_attn_long}).

Most importantly, we understand what Memory Mosaics do far better than we understand what transformers do. First, the value extraction functions of the associative memory units precisely describe what each memory seeks to memorize. Second, the predictive disentanglement principle explains why training a Memory Mosaic breaks the overall prediction task into pieces that are more efficiently memorized when they are considered independently (Section~\ref{sec:3moons}). Therefore, Memory Mosaics are not just a transformer-like architecture, but also a {model}\footnote{Not as in ``statistical model'' but as in ``model used to describe and explain a phenomenon.''} for compositional learning systems that break knowledge into independently memorized fragments, then reassemble them as needed using combination strategies that can themselves be viewed as memorized knowledge fragments (Section~\ref{sec:layered_memories}).

The focus on memorization allow us to formulate new questions. Could memories operate independently on different time scales?  Could we envision a richer memory hierarchy than simply distinguishing persistent memories from contextual memories? Can intermediate memory tiers be trained like contextual memories, that is, without gradients? Can the persistent knowledge be then reduced to a compact high order bias?

Memory Mosaics also offer an array of engineering opportunities. Limited storage contextual memories could leverage least-recently used eviction schemes (\eg., \citealp{xiao-2023}), and associative memories could be implemented using a wide spectrum of techniques, either classical (\eg., \citealp{greengard-1991,spring-shrivastava-2017}), or neural (\eg., \citealp{krotov-2023}), which could redefine the computing requirements of contemporary artificial intelligence systems.

\chapter{Inference-Time Learning}
\label{chap:inference-time-learning}

The ability of AI for the open-world is closely related with the requirements of passing \textsc{Turing Test} \citep{turing-test}, where a machine is expected to carry out any task that a human could possibly undertake. In \textsc{Turing Test}, the competence in specific tasks (AI for the closed-world) tells very little about passing the \textsc{Turing Test}. This fact makes it difficult to translate the success of building AI for the closed-world to AI for the open-world. \emph{Thus, researchers tried two shortcuts to bypass the building of AI for the open-world, but to mimic the ability of AI for the open-world.}

\emph{The first attempt is ``training on everything'', e.g. all text on internet.} This approach aligns with the mainstream belief of foundational models \cite{bommasani2021opportunities}. Unfortunately, this strategy to mimic the ability of AI for the open-world is neither computationally feasible nor practically plausible due to two aspects of the open-world: \textbf{1)} Combining two pieces of knowledge results in a new knowledge, leading to an exponentially large number of possible combinations; \textbf{2)} Pieces of knowledge change over time. 

Another practical compromise is \emph{assigning a human designer to one specific task}. This strategy aligns with many task-orienting practices in artificial environments and specific real-world tasks. \cite{samuel1959some,berliner1980backgammon,tesauro1995temporal,silver2016mastering,matan1991multi,taigman2014deepface,eliza,SHRDLU,winograd1971procedures}. This strategy not only suffers from the two open-world aspects above, but also heavily relies on the priori knowledge of human designers. In the end, instead of proving that our machine is intelligent, this strategy often finds satisfaction in proving that we are intelligent \cite{ai-in-open-world}.

To address the two aspects of the open-world and reduce the reliance on task-specific priori knowledge from designers, this chapter introduces a learning paradigm in building AI for the open-world, called \emph{inference-time learning}. This learning paradigm shifts parts of computation to each specific task at the inference time rather than pre-computes everything during pretraining.\footnote{One may argue that human doesn't have a clear ``pretraining'' stage. However, genes are not random! Revolution provides a good initialization, just as pretraining. In addition, people are hardly remembering anything before three years old. This early period of time also significantly shapes the learning capacity of human.} This computation-shifting process may remind readers of the fine-tuning notion in the transfer-learning domain. However, unlike fine-tuning, which heavily relies on designers' priori knowledge \citep{li2020rethinking}, inference-time learning demands stricter requirements on reducing the reliance on task-specific priori knowledge from human designers. %
In summary, inference-time learning operates under conditions where \emph{a sequences of new tasks are required to be learned quickly with few examples} --- a scenario akin to a child's daily school life. The large number of unseen new tasks, in turn, restricts the amount of designers' priori knowledge inserted into each task.
This learning condition is fundamentally different from that of \iid~training where a large amount of example within one single \iid~distribution is available.\footnote{A Mix (shuffle) of multiple distributions is still one single \iid~distribution.}  

Furthermore, this chapter explores suitable techniques for performing inference-time learning paradigm. It is worth noting that Memory Mosaics in Chapter \ref{chap:disentanglement_and_sample_complexity} have revealed some preliminary abilities to quickly learn new tasks with fewer examples and less task-specific priori knowledge.\footnote{Although Transformers also demonstrate quick learning abilities, they lag significantly behind Memory Mosaics as shown in previous chapter Table \ref{fig:gpt_mm_ood} and \ref{fig:icll_acc}.}
Building upon Memory Mosaics, this chapter propose \textit{Memory Mosaics \textsc{v}2} to enhance inference-time learning. \textit{Memory Mosaics \textsc{v}2} is a memory-based method at inference time, using rich features and disentangled representations constructed during the pre-training stage.
To avoid confusion, the author needs to clarify that all techniques proposed in this thesis are used to support and verify the three learning principles, which is the main focus of this thesis. Memory Mosaics in Chapter \ref{chap:disentanglement_and_sample_complexity} is to construct disentangled representations of training knowledge, while Memory Mosaics v2 in this chapter is to perform inference-time learning on new tasks.

The remainder of this chapter is organized as follows. Section \ref{sec:learning_conditions} introduces the learning conditions of \emph{inference-time learning}, discussing possible algorithms. Building on these conditions, Section \ref{sec:memory_mosaics_v2} introduces the \emph{Memory Mosaics V2} architecture to perform {inference-time learning}. Section \ref{sec:mmv2_training_and_results} provides training and evaluation on a large scale (9.9B parameters). Section \ref{sec:risk-return-trade-off-mmv2} studies the gap between \emph{``training on everything''} strategy and inference-time learning, answers the question of how many additional data Transformers need to match Memory Mosaics V2. In addition, Section \ref{sec:more-learning-signals} investigates the impact of incorporating additional learning signals into the objective function. Finally, Section \ref{sec:inference-time-discussion} discusses the future directions of \emph{inference-time learning}.

\section{Learning conditions of inference-time learning}
\label{sec:learning_conditions}

Recall the example of \textit{a daily life of a child --- playing games, creating art works, and learning science sequentially.} This scenario is analogous to the learning conditions of inference-time learning: 1) the number of available examples in a new task is limited; 2) multiple tasks are learned sequentially in time.
These two learning conditions are ubiquitous and essential, yet pose significant challenges for traditional \iid~training. To design suitable learning algorithms under these conditions, we first need to explore their properties and related techniques.

\bfparagraph{Limited examples} The limited examples condition requires learning algorithms to operate with a small amount of data, making it challenging to achieve reliable performance. In such scenarios, the algorithm must trade-off hypothesis space and the number of examples, which is known as structured risk minimization \cite{atkeson1997locally,bottou1992local,guyon1991structural, vapnik1991principles}. 

\emph{Model-based learning algorithms}, such as optimizing neural networks, shapes hypothesis space through three factors: preprocessing, architecture, and learning mechanism (e.g., regularization) \cite{guyon1991structural}. In these three factors, the process of preprocessing (e.g. feature engineering) and shaping architectures rely on task-specific experiences, the number of hyper-parameters in learning mechanism (e.g. regularization, learning rate) is large. Thus, the selection process on a specific task often involves a lot of priori knowledge from algorithm designers \cite{guyon1991structural,li2020rethinking}.

On the other hand, \emph{memory-based learning algorithms} (also called local learning algorithms) \cite{vapnik1991risk,bottou1992local,vapnik1993local,atkeson1997locally}, shapes the hypothesis space by one or two smoothing parameters. For example, bandwidth in kernel smoothing or local weighted averaging, the number of neighbors in k-nearest-neighbors. This simple hypothesis space controlling makes it possible to reduce the reliance on task-specific priori knowledge from human designers.

\bfparagraph{Sequentially learning} 
The sequentially learning condition requires the learning algorithms to acquire new knowledge without destroying previously learned information. In such scenarios, the algorithm must balance the need to adapt to new tasks with the need to retain existing knowledge.

In \emph{model-based learning algorithms}, sequentially learning different tasks leads to negative interference or catastrophic forgetting issues \cite{kirkpatrick2017overcoming}. These problems refer to the destruction of previously learned knowledge when learning new tasks, even if the knowledge does not conflict with each other. 

In contrast, \emph{memory-based algorithms} were initially motivated to solve the negative interference problem, as described in \citet{atkeson1997locally} --- \textit{``Negative interference between old and new training data is one of the most important motivations for exploring locally weighted learning''}. Memory-based approaches store and retrieve information in a non-parametric manner, allowing them to adapt to new tasks without overwriting existing knowledge. Figure \ref{fig:memory-based_model-based_comparison}, taken from \citet{atkeson1997locally}, provides a comparison of model-based and memory-based approaches on negative interference.

In summary, the two learning conditions of inference-time learning favor memory-based learning algorithms rather than model-based learning algorithms.

\begin{figure}[ht!]
    \centering
    \includegraphics[width=0.4\linewidth]{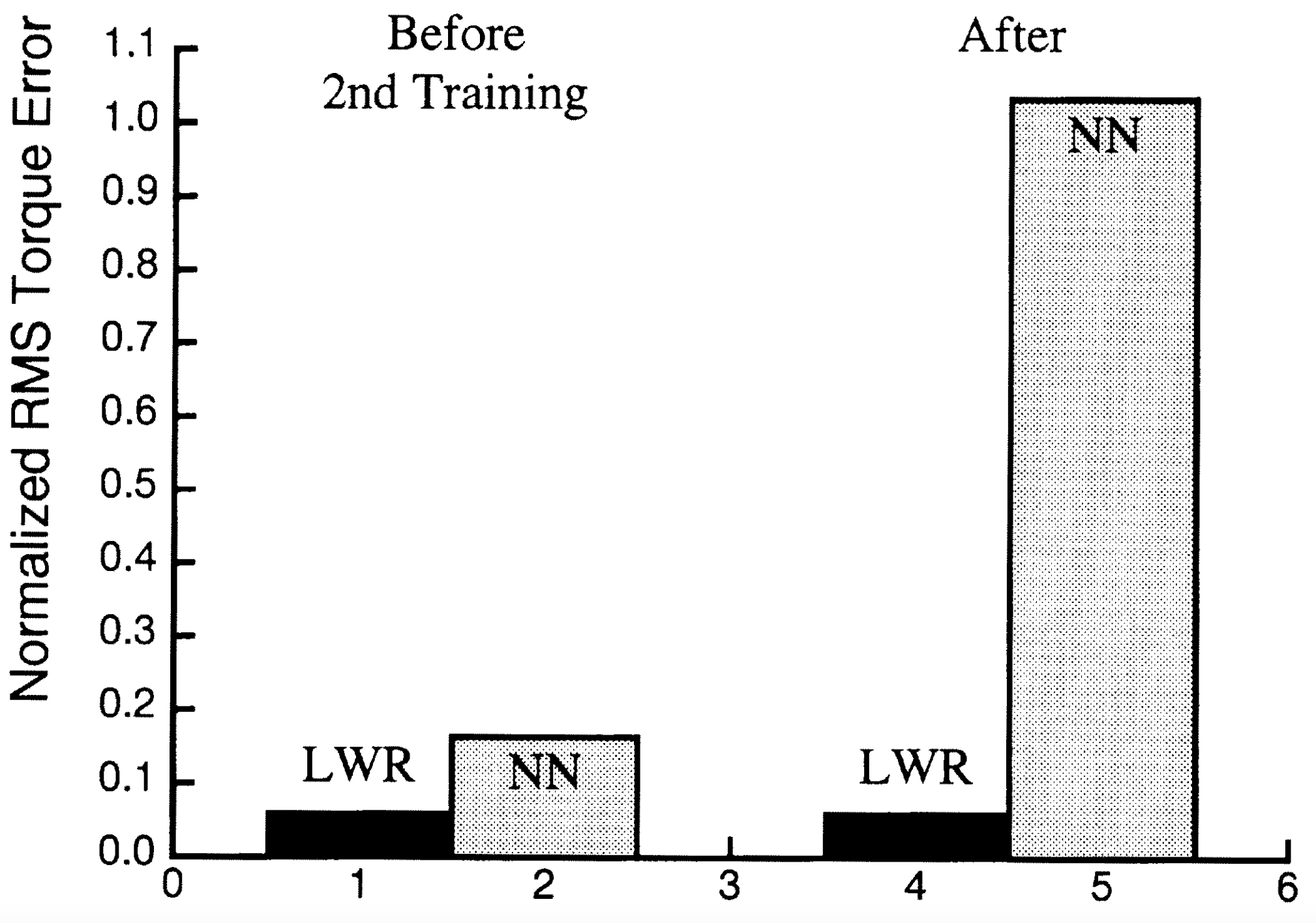}
    \caption{The differences between a global parametric representation (i.e. neural network) and a locally weighted learning approach (i.e. local weighted regression) \textbf{\cite{atkeson1997locally}}. A sigmoid neural network (marked as ``NN'') and a local (quadratic) weighted regression (marked as ``LWR'') are trained to predict the torques of two jointed arm dynamics. Both ``NN'' and ``LWR'' generalize well on \iid~test data (bar 1 and 2). After that, each model is trained on ten attempts to make a particular desired movement. Each model successfully learned the desired movement. The global parametric representation ``NN'' fails to generalize on the original test data (bar 5), while the locally weighted learning approach ``LWR'' still generalize well (bar 4). 
movement. }
    \label{fig:memory-based_model-based_comparison}
\end{figure}

\setlength{\tabcolsep}{0.9mm} 

\section{Memory Mosaics v2 for inference-time learning}
\label{sec:memory_mosaics_v2}

Researchers have realized the benefits of \emph{memory-based algorithms} in shaping hypothesis space and reducing negative interference for at least three decades \cite{atkeson1997locally}. The benefits of {memory-based algorithms} on inference-time learning were also discussed above. However, the curse of dimensionality impedes the use of memory-based algorithms \cite{toth2017handbook}, especially in learning raw features from scratch, for decades. Fortunately, the \emph{rich feature} and \emph{disentangled representation} principles provide an opportunity to 1) reduce the need for feature learning in inference time and 2) organize each feature nicely in a low-dimensional space, respectively. Consequently, such a strategy --- a memory-based method at inference time, using rich features and disentangled representation constructed during pretraining stage --- becomes feasible to reduce the impact of the curse of dimensionality while using memory-based algorithms.\footnote{This argument of ``It is possible to reduce the impact of the curse of dimensionality on memory-based algorithms via rich features and disentanglement'' was first supported by Pascal Vincent during my Memory Mosaic talk at the FAIR Lab Offsite in June 2024. Thus, the author thanks Pascal Vincent for his insightful comments.}

To implement this learning strategy, this section explores relevant techniques. 
The first turning point of practical techniques came with the introduction of transformers \citep{vaswani2017attention}. Using transformers, \citet{brown2020language} demonstrated the possibility of a general machine that can learn diverse tasks in a few-shot manner, requiring less task-specific priori knowledge from designers.\footnote{Of course, a ``turning point'' is unlikely to also be the ``ending point'', just like the Transformer itself.} Later, Memory Mosaics (in Chapter \ref{chap:disentanglement_and_sample_complexity}), explicitly using associative memory and memory-based approaches, architecturally resembles transformers but outperforms them on preliminary evaluations (Table \ref{fig:gpt_mm_ood} and \ref{fig:icll_acc}). 
Inspired by the practical success of Transformers and Memory Mosaics, this section proposes \emph{Memory Mosaics v2} architectures to leverage memory-based methods at inference time, building upon the rich features and disentangled representation constructed during pretraining stage.

Due to well-developed language training and evaluation datasets, well-optimized software and hardware, most experiments in this section are conducted on sequential language data. In the sequential language domain, \emph{Memory Mosaics v2} substantially outperforms transformers (>10\%) on the ability of learning new tasks at inference-time, although they are architecturally similar to each other. It is worth noting that the {inference-time learning} paradigm is a general framework beyond language. \textit{In this section, Memory Mosaics v2 serves as an experimental verification of the {inference-time learning} paradigm in the language domain, but not ``a new language model'' in the classical sense.}\footnote{Even-though one can use Memory Mosaics v2 in a similar way as the classical language model, e.g. chat, prompt.}

\subsection{Memory Mosaics v2 architectures}
\label{sec:mmv2}

Compared with the Memory Mosaics architecture in section \ref{sec:language}, Memory Mosaics v2 incorporates three architecture modifications, including an adaptive bandwidth in associative memory, a gated time-variant key feature extractor, and a 3-level memory design.  This section provides a detailed explanation of the Memory Mosaics v2 architecture.

\subsubsection{Adaptive bandwidth in Gaussian kernel smoothing}

Memory mosaics use one fixed bandwidth parameter $\beta$ for different sizes $n$ of associative memory (Equation \ref{eq:unrolling}). It is well known that bandwidth controls the bias-variance trade-off \citep{hastie2009elements} of kernel regression (memory-based) methods. That is, for a given distribution, the optimal bandwidth depends on the number of examples (key-value pairs in associative memory). Inspired by the asymptotic Mean Integrated Squared Error kernel bandwidth estimation approach where $1/\sqrt{\beta} \propto n^{-1/(p+4)}$ \citep{Garcia-Portugues2024}, memory mosaics v2 scale $\beta$ in associative memories as:\footnote{For the easy of reading, this chapter reuse the ``attention score'' notion to $\frac{e^{\,\beta\,k^\ttop k_i}}{\sum_{j=1}^{n} e^{\,\beta\,k^\ttop k_j} }$ in associative memories.}
\begin{align}
\label{eq:gaussian-smoothing-with-dp}
    f\bigl(k;\: \{(k_1,v_1)\dots(k_n,v_n)\}\bigr) ~&=~ 
      \sum_{i=1}^{n}
      ~ \frac{e^{\,\beta\,k^\ttop k_i}}{\sum_{j=1}^{n} e^{\,\beta\,k^\ttop k_j} } 
      ~ v_i~ \\
      \beta &= \beta_1  n^{\alpha} + \beta_0 
\end{align}
where $\beta_0\geq0,\,\beta_1>0,\,1>\alpha>0$ are learnable parameters. I.e., the more key-value pairs (examples), the smaller bandwidth $1/\sqrt{\beta}$.

\subsubsection{Gated time-variant key feature extractor} 

Memory mosaics in Section \ref{sec:memories} employ a simple time-invariant leaky averaging to extract key features:
\begin{equation}
\label{eq:mm_key_extractor}
 \begin{aligned}
 k_T &= \mathrm{Norm}\big(\textcolor{purple}{\Bar{k}_T}\big) &\text{with}  \quad {\textcolor{purple}{\Bar{k}_T} = \textcolor{blue}{\tilde{k}_T} + \lambda\, \textcolor{purple}{\Bar{k}_{T-1}} \quad\quad \textcolor{blue}{\tilde{k}_T} = W_\varphi\,x_T  }
 \end{aligned}
\end{equation}
The averaging weights $\lambda$ in Equation \ref{eq:mm_key_extractor} are fixed and independent of the semantic input $x$. As a result, semantically similar cases, such as ``tom-and-jerry'' and ``tom- - -and- - -jerry'', may receive different key features. Inspired by recurrent-style networks \citep{peng2023rwkv,gu2023mamba, beck2025xlstm}, memory mosaics v2 utilize the following gated time-variant key feature extractor:\footnote{It worth noting that this work is neither a linearization of attention nor attention efficiency. The recurrent feature extractor in Eq. \ref{eq:mm_key_extractor} is used to create keys, while associative memory in Eq. \ref{eq:gaussian-smoothing-with-dp} still stores all key-value pairs. }
\begin{equation}
\label{eq:betterfeatures}
 \begin{aligned}
 k_T &= \mathrm{Norm}\big(\textcolor{purple}{\Bar{k}_T}\big) \quad \text{with}  \quad
 \begin{cases}
  {\, \textcolor{purple}{\Bar{k}_T} = g_T\textcolor{blue}{\tilde{k}_T} + \lambda_{T}\textcolor{purple}{\Bar{k}_{T-1}} \quad\quad \textcolor{blue}{\tilde{k}_T} = W_\varphi\,x_T  }\\
  \,   g_t = e^{W_g x_T} \in \mathbb{R}\,,\, \lambda_T = e^{- |W_{\lambda} x_T|}\in \mathbb{R}
 \end{cases},
 \end{aligned}
\end{equation}
where $W_{\varphi}, W_{g}, W_{\lambda}$ are learnable parameters, the averaging weights $\lambda_T \in \mathbb{R}$ and the the exponential gate $g_T\in \mathbb{R}$ semantically depend on input $x_T$. %

For key feature extractor, memory mosaics v2 reuses the same convolutional key extractor as in memory mosaics: 
\begin{equation}
\label{eq:betterfeatures}
 \begin{aligned}
   v_T &= \mathrm{Norm}\big(\textcolor{purple}{\Bar{v}_T}\big) \quad\, \text{with}  \quad   {\textcolor{purple}{\Bar{v}_T} = \gamma\, \textcolor{blue}{\tilde{v}_T} + (1-\gamma )\,\textcolor{blue}{\tilde{v}_{T+1}} \quad~ \textcolor{blue}{\tilde{v}_T} = W_\psi\,x_T\,,
 }
 \end{aligned}
\end{equation}
where $\gamma \in \mathbb{R}$ and $W_{\psi}$ are learnable parameters.

\subsubsection{3-level memory}
The memory mosaics in Section \ref{sec:memories} simplify the attention in transformer as contextual associative memory, view the feed-forward network in transformer as persistent memory. This simplification reduces the dependence between the ``attention score'' and the token position, as shown in Figure \ref{fig:transformer_mm_attn}. 
\begin{figure}[ht!]
    \centering
    \vspace{-1ex}
   \includegraphics[width=0.4\textwidth]{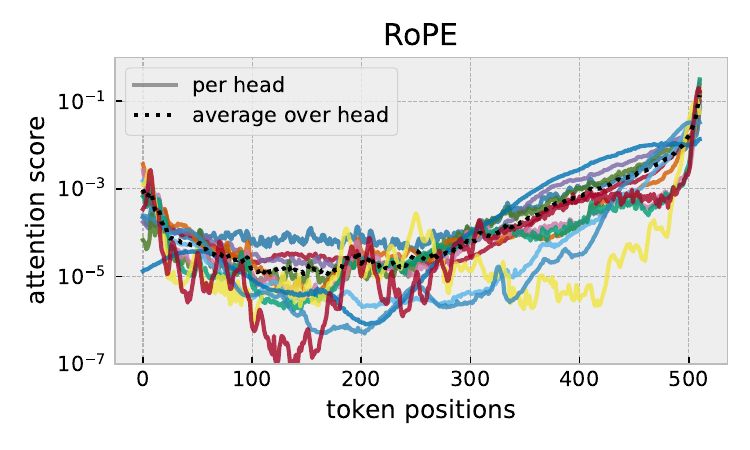} \quad \quad \quad
    \includegraphics[width=0.4\textwidth]{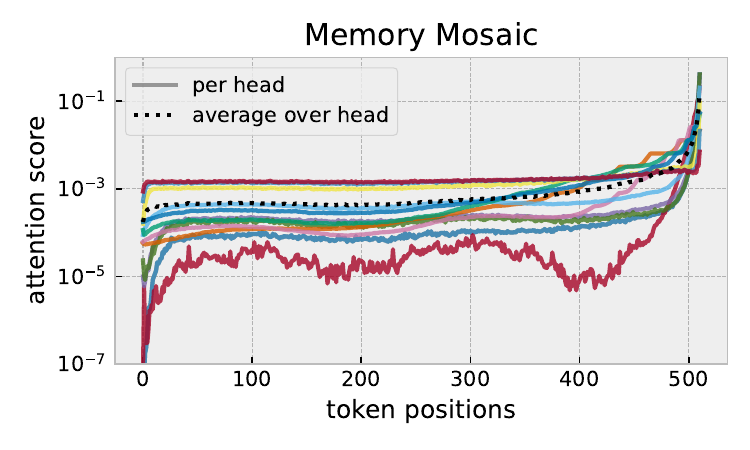}
    \caption{\underline{Average} attention scores of the last token attending previous tokens. \textbf{Left}: Transformer with RoPE position encoding. \textbf{Right}: Memory Mosaics in Section \ref{sec:memories}. The (averaged) attention scores in transformer heavily depends on token positions (curly curves), while the attention scores in memory mosaics at far tokens (e.g. position 0 to 450) are almost invariant to positions (flat curves). }
    \label{fig:transformer_mm_attn}
     \vspace{-1.5ex}
\end{figure}
Compared with transformers (figure \ref{fig:transformer_mm_attn} left), the attention scores in memory mosaics exhibit a structured pattern (figure \ref{fig:transformer_mm_attn} right). That is, attention scores on near-tokens (positions) heavily depends on positions, while attention scores on far-tokens are almost invariant to token positions. Inspired by this experimental discovery, memory mosaics v2 replace contextual associative memory in memory mosaics with two associative memories, \textit{short-term memory} and \textit{long-term memory}, using distinct parameters (as figure \ref{fig:memory-mosaics-v2-architecture}). The short-term memory at position $t$ only store key-value pairs of near tokens, ranging from $t-h+1$ to $t-1$. In contrast, the long-term memory skips near tokens and only store key-value pairs before position $t-m$. By setting $m < h$, memory mosaics v2 create an overlap between long-term and short-term memory, resulting in a soft boundary between these two memories.

Memory mosaics v2 implements \emph{persistent memory} using dense two-layers 
 neural networks with SwiGLU activation \citep{shazeer2020glu} due to computational efficiency concerns.\footnote{A two-layers feed-forward network and a key-value associative memory are interchangeable as shown in \citet{sukhbaatar-2019}.} 

\begin{figure}[ht!]
    \centering
    \includegraphics[height=0.4\linewidth]{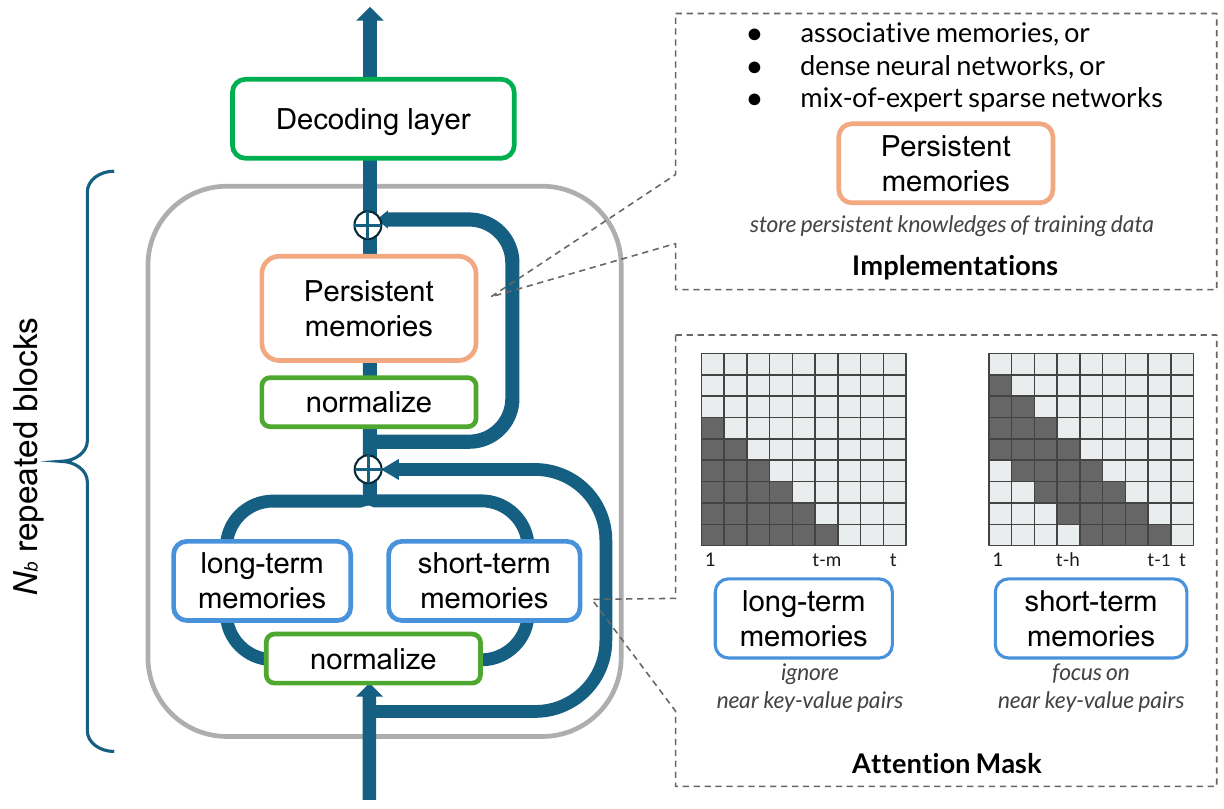}
     \vspace{-1ex}
    \caption{Memory Mosaics v2 architecture. }
    \label{fig:memory-mosaics-v2-architecture}
     \vspace{-1ex}
\end{figure}

\subsection{Large-scale Training}
\label{sec:mmv2_training_and_results}

We train two Memory Mosaics v2: \emph{Memory Mosaics v2 Small} and \emph{Memory Mosaics v2 Large}. Memory Mosaics v2 Small contains 24 layers, 2048 hidden dimensions, and 16 heads. Memory Mosaics v2 large increases the number of layers to 32, hidden dimensions to 4096, and the number of heads to 32.\footnote{The persistent memory hidden dimension is set to 6,144 and 14,336 for Small and Large models, respectively.}

Similarly, we train two baseline transformers, \emph{Transformer Small} and \emph{Transformer Large}, with the same configurations as their Memory Mosaic counterparts. Unless otherwise specified, in this work, transformer models use Llama architectures with multi-head attention.

\bfparagraph{Dataset}

Unless otherwise specified, Memory Mosaics v2 Small and Transformer Small are trained on a dataset consisting of 200 billion tokens from a diverse datamix that includes web text, arxiv paper, github code, and books. Similarly, Memory Mosaics v2 Large and Transformer Large are trained on 1 trillion tokens from the same datamix.
The sequence length distributions of the training dataset are visualized in Figure \ref{fig:mm_training_data_seq_lens}, providing insight into the characteristics of the data used to train our models.

\begin{figure}[h!]
    \centering
    \includegraphics[width=0.48\linewidth]{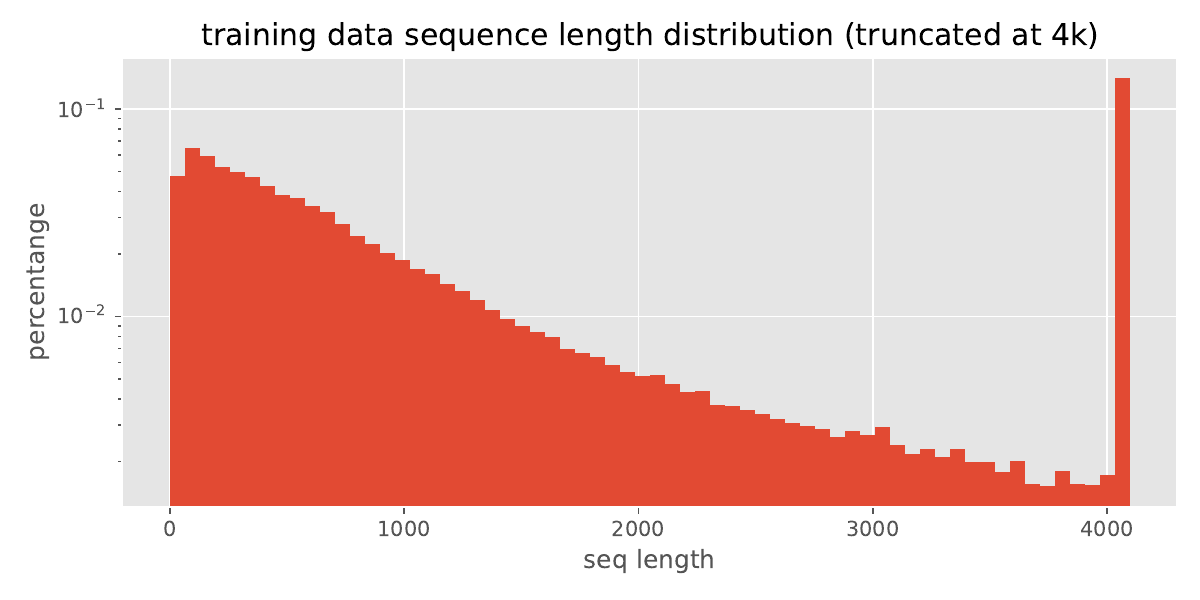}
    \includegraphics[width=0.48\linewidth]{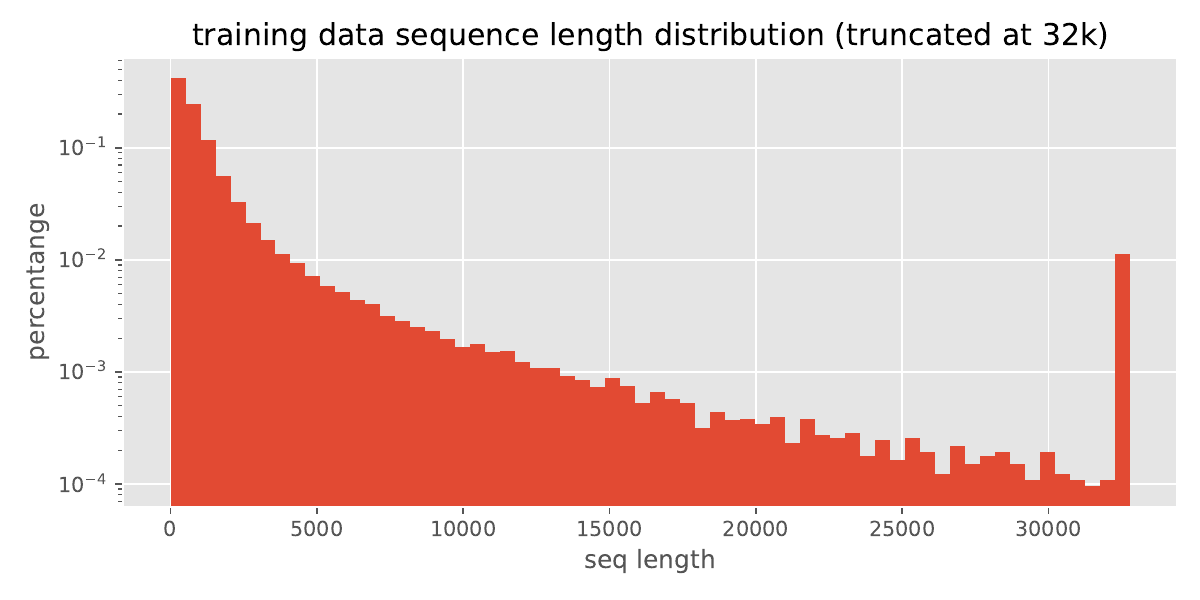}
    \caption{Training data sequence length distributions. For a given maximum sequence length during training (e.g. 4k), longer sequences are truncated to the maximum sequence length. This truncation results in the peaks at the end of distributions. }
    \label{fig:mm_training_data_seq_lens}
\end{figure}

\bfparagraph{hyperparameters}

For all Memory Mosaics v2 and baseline Transformer models, we use a consistent set of hyperparameters. That is, a batch size of 1024, a sequence length of 4096, an adamw optimizer with $\beta_1=0.9$  and $\beta_2=0.95$ accompanied by a $L_2$ weight decay of 0.1 and a gradient norm clip of 1, a learning rate warm-up of 2000 iterations followed by a cosine learning rate scheduler that reduces the learning rate by a factor of 100 at the end. The initial learning rates (after warm-up) are set to 3e-4 for ``small'' models and 1e-3 for ``large'' models.

We also employ document-wise attention mask, where the attention scores are only computed within each sequence (document) in the training data, to reduce computation cost. Two special tokens, ``<|begin\_of\_text|>'' and ``<|end\_of\_text|>'' are appended at the begining and ending of a sequence, respectively.

During training, Memory Mosaics v2 samples the long-term memory delay step $m$ from $[64, 256]$, sets the short-term memory window size $h=256$. At inference, $m$ is set to 64, as illustrated in Figure \ref{fig:overlapped-long-short-memory}.

It is worth noting that these hyperparameters were originally searched and optimized for the baseline Transformer models (by amaia team). We transfer these hyperparameters to Memory Mosaics v2 without further hyperparameter searching. Thus, it is possible that this hyperparameter setup is suboptimal for Memory Mosaics v2.\footnote{This hyperparameter setup is sufficient for memory mosaics v2 to verify the inference-time learning principle.   }

\begin{figure}[ht!]
    \centering
    \includegraphics[width=0.8\linewidth]{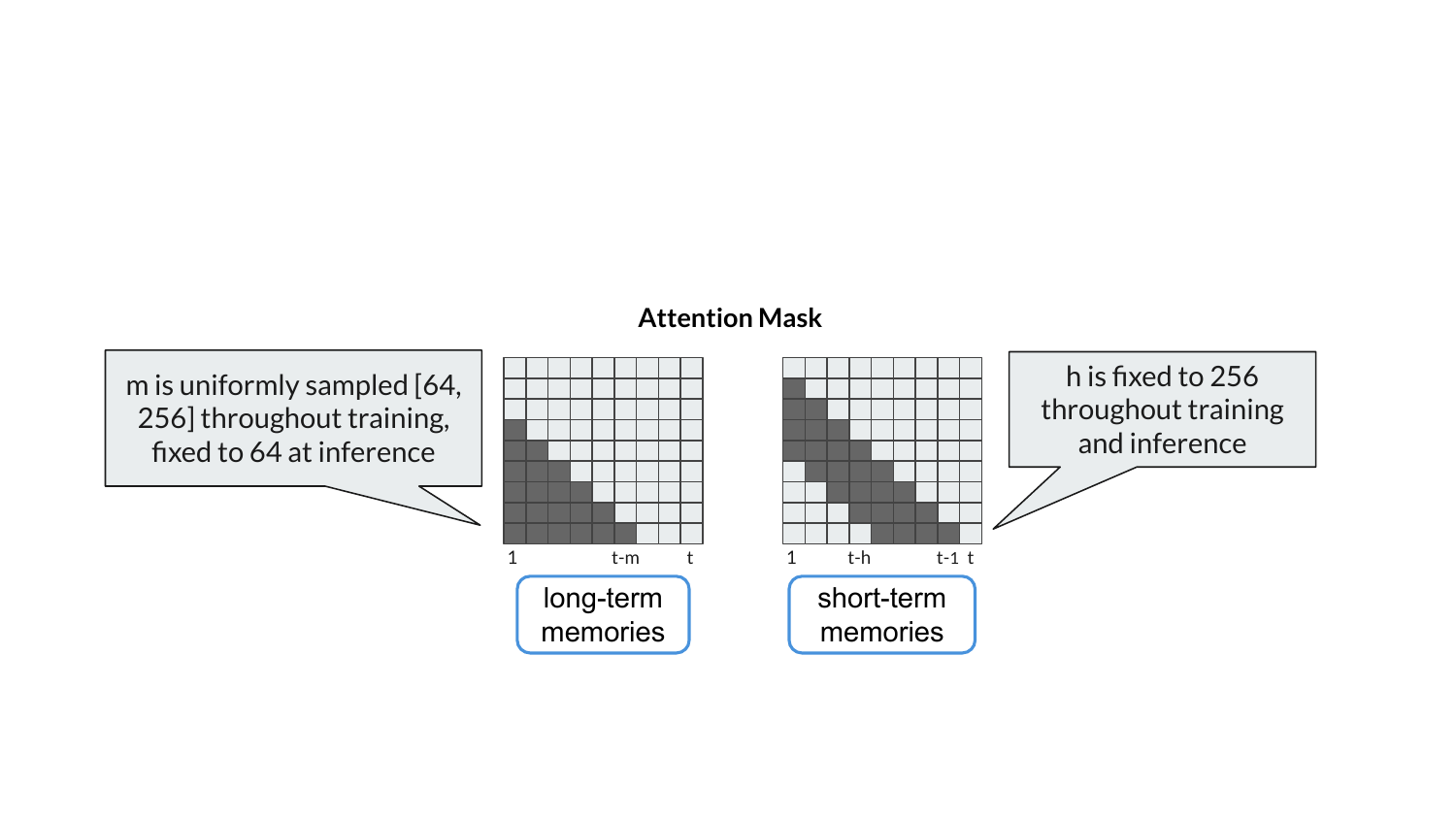}
    \caption{Randomly overlapped long-term \& short-term memory}
    \label{fig:overlapped-long-short-memory}
\end{figure}

\subsection{Three evaluation dimensions}

The evaluation design provides a means to assess a specific property of a system and contains a clear goal.\footnote{For instance, in the early stages of machine learning, performance was often evaluated solely on training datasets \cite{rosenblatt-1957, rumelhart-1986}. As the field evolved, researchers realized that \iid~generalization was a more challenging and interesting problem than optimization, leading to a shift towards evaluating \iid~performance.} The main focus of this thesis is the ability to learn new tasks with fewer examples and less task-specific priori knowledge. Thus, this chapter adopts three evaluation dimensions to comprehensively assess this ability.
\begin{itemize}
 \item \textbf{Persistent-knowledge storage and retrieval}, the ability of persistent-memory to store and retrieve knowledge of training dataset. This capability prepares knowledge that could be reused in other tasks during inference. We use common language benchmarks to access this aspect. 
    \item \textbf{New-knowledge storage and retrieval}, the ability to store and retrieve new information of test dataset. It is a prerequisite for ``learning'' new tasks via memory-based methods. We employ ``multi-unrelated-documents storing and question-answering'' tasks to evaluate this aspect.\footnote{Imagining a poor goldfish with a 7-seconds long-term memory, how can it learn a 90-mins movie? Various approaches with fix-sized memories, including RNNs, LSTM, state-stace model, sliding-window attention, resemble this poor goldfish.}
\end{itemize}

As mentioned above, this chapter trains and evaluates Memory Mosaics v2 on language tasks due to the simplicity of ``language''.\footnote{Language is a human designed tool to communicate. Thus, it is designed to be easy and compact.} This thesis does not focus on language models but rather explores \emph{AI for the open world} and its three learning principles --- \emph{rich features}, \emph{disentangled representation}, \emph{inference-time learning}.

\subsubsection{Persistent-knowledge storage and retrieval }
\label{sec:eval_training_data_stsoring_retrieval}

This section first evaluates both Memory Mosaics v2 and baseline transformers on 19 commonly-used language benchmarks. Table \ref{tab:tf_mm_19_common_tasks} shows that Memory Mosaics v2 and transformers performs closely on these benchmarks. This is nothing to supervise, because both Memory Mosaics v2 and transformers use the same persistent memory architecture. Meanwhile, many other RNN architectures, such as \textsc{mamba} \cite{gu2023mamba} and \textsc{xlstm} \cite{beck2025xlstm}, also perform well on these benchmarks.

\begin{table}[ht!]
    \centering
        \caption{Memory Mosaics v2 and Transformers (``small'' and ``large'') performance on 19 common language benchmarks. After extending context length to 32k (via fine-tuning), Memory Mosaics v2 and Transformer performs closely to each other on these common benchmarks (i.e., 38.0\% vs 37.8\% for small models, 52.2\% vs 52.2\% for large models).   }
    \label{tab:tf_mm_19_common_tasks}
    \resizebox{\textwidth}{!}{
    \begin{tabular}{cc|ccc ccc ccc c ccc ccc ccc |c}
    \toprule
model        & \makecell{context\\length}         &   obqa  &  \makecell{arc\\easy}  &   \makecell{wino-\\grande}   &  \makecell{arc\\challenge} &  piqa  &  boolq  &  \makecell{hell-\\aswag}  &  nq  &  siqa  &  tqa  &  gsm8k  &  \makecell{mmlu\\alt}   &  \makecell{human \\eval+}  &  squad  &  bbh  &  math  &  mbpp  &  \makecell{race\\middle}  &  \makecell{race\\high}  &   avg \\
\midrule
transformer small & 4k        &  35.6  &  61.9  &  60.9  &  33.5  &  74.0  &  63.0  &  61.0  &  12.7  &  45.3  &  30.4  &  2.7  &  36.0  &  34.3  &  58.8  &  27.6  &  1.3  &  9.9  &  52.5  &  39.1  &  39.0 \\
memory mosaics v2 small & 4k     &  34.0  &  60.4  &  58.4  &  33.0  &  72.8  &  63.1  &  58.1  &  11.6  &  46.4  &  29.4  &  3.1  &  34.7  &  32.0  &  59.8  &  27.1  &  1.0  &  9.2  &  49.4  &  38.3  &  38.0 \\
\midrule
transformer large &  4k        &  44.6  &  78.6  &  74.7  &  53.8  &  81.0  &  71.9  &  80.3  &  34.3  &  49.7  &  64.5  &  37.2  &  50.2  &  39.6  &  80.2  &  53.4  &  10.3  &  9.0  &  58.3  &  47.3  &  53.6 \\
memory mosaics v2 large &  4k     &  45.0  &  78.1  &  72.3  &  51.6  &  80.4  &  71.7  &  78.6  &  30.6  &  48.7  &  62.3  &  27.7  &  48.2  &  43.3  &  78.4  &  48.1  &  8.4  &  9.5  &  61.4  &  46.1  &  52.1 \\
\midrule
\rowcolor{LightCyan}
transformer small & 32k     &  35.2  &  61.0  &  60.1  &  31.4  &  73.6  &  63.0  &  59.3  &  11.7  &  44.5  &  26.7  &  3.0  &  35.2  &  32.4  &  54.7  &  26.0  &  1.2  &  9.2  &  52.2  &  37.4  & \cellcolor{blue!55} 37.8 \\
\rowcolor{LightCyan}
memory mosaics v2 small & 32k  &  35.0  &  60.0  &  58.4  &  32.9  &  73.3  &  62.7  &  58.0  &  11.8  &  46.6  &  29.3  &  3.1  &  34.7  &  30.8  &  59.3  &  27.3  &  9.4  &  49.2  &  1.1  &  38.4  &  \cellcolor{blue!55}  38.0 \\
\midrule
\rowcolor{LightCyan}
transformer large & 32k     &  45.8  &  77.3  &  72.3  &  52.6  &  80.8  &  72.6  &  79.2  &  31.9  &  49.3  &  61.5  &  32.4  &  49.0  &  38.3  &  76.3  &  45.6  &  8.7  &  9.8  &  62.6  &  45.6  & \cellcolor{blue!55} 52.2 \\
\rowcolor{LightCyan}
memory mosaics v2 large & 32k  &  45.4  &  78.0  &  71.2  &  51.8  &  80.4  &  73.1  &  78.6  &  30.9  &  48.6  &  62.0  &  27.4  &  48.2  &  43.0  &  78.2  &  47.8  &  8.8  &  9.6  &  61.6  &  46.5  & \cellcolor{blue!55}  52.2 \\
\bottomrule
    \end{tabular}
    }

\end{table}

How do we know whether these benchmarks access persistent-knowledge ability rather than new-knowledge ability? To answer this question, we re-evaluate these benchmarks on Memory Mosaics v2 but with \emph{long-term memory} being removed after training.  The underlying reason is that if a task solely relies on the information stored in persistent memory and retrieved by {short-term memory}, removing long-term memory should not significantly affect performance.

Table \ref{tab:13_normal_benchmarks_with_without_longterm_mem} shows that removing {long-term memory} after training does not hurt the performance of 13 common benchmarks. 
This suggests that these 13 tasks are almost exclusively based on information stored in persistent memory and retrieved by short-term memory. In contrast, Table \ref{tab:6_normal_benchmarks_with_without_longterm_mem} shows that the other 6 benchmarks perform badly after removing long-term memory.

Based on these findings, we use the 13 tasks to evaluate the persistent-knowledge storage and retrieval capability. The results (Table \ref{tab:tf_mm_19_common_tasks}) show that Memory Mosaics v2 and Transformer perform closely in this evaluation dimension, suggesting that both models are capable of effectively storing and retrieving persistent-knowledge.

\bfparagraph{Context length extension (by fine-tuning) concerns}
To deal with complicated scenarios/tasks, such as learning a new language, a model must handle a long context length. Compared with training a model on long context length directly, pretraining on a short context length (e.g. 4k) followed by fine-tuning on longer context length (e.g. 32k) reduces the overall computation cost. However, this long-context extension process hurts transformer performance on ``short'' tasks, as shown in previous studies \citet{chen2023extending, peng2023yarn}. In Table \ref{tab:tf_mm_19_common_tasks}, we observe a similar performance degradation behavior while extending the context length of the transformers. 

In contrast, Memory Mosaics v2 exhibits consistent behavior across different context lengths (perhaps because Memory Mosaics v2 doesn't contain any explicit position encoding). This property enables Memory Mosaics v2 to maintain its performance after long-context fine-tuning, unlike transformers.

\begin{table}[ht!]
    \centering
       \caption{Memory Mosaics v2 performance on 13 common language benchmarks. Removing the ``long-term memory'' after training barely hurt the performance (56.6\% vs 56.8\%). Flops/token is estimated at context length 256.  }  
    \label{tab:13_normal_benchmarks_with_without_longterm_mem}
    \resizebox{\textwidth}{!}{
    \begin{tabular}{c|cc|ccc ccc ccc c ccc | c  }
    \toprule
  &  params & flops/token  &   obqa  &  \makecell{arc\\easy}  &  \makecell{wino-\\grande}  &  \makecell{arc\\challenge}  &   piqa  &  boolq  &  \makecell{hell-\\aswag}  &  nq  &  siqa  & tqa  & gsm8k  &  \makecell{mmlu\\alt}  &  \makecell{human \\eval+} &  avg \\
         \midrule
         Transformer large  & 8.8B & 16.7B  & 45.8  &  77.3  &  72.3  &  52.6  &  80.8  &  72.6  &  79.2  &  31.9 & 49.3  &  61.5  &  32.4  &  49.0  &  38.3  &  57.1 \\
         \midrule
         memory mosaics v2 large  & 9.9B &  18.9B  &  45.4  &  78.0  &  71.2  &  51.8   &  80.4  &  73.1  &  78.6  &  30.9  &  48.6 &  62.0  &  27.4  &  48.2  &  43.0  &  56.8 \\
         \makecell{memory mosaics v2 large \\without long-term memory} & 8.3B &15.6B  & 45.4  &  77.9  &  71.2  &  51.8  &  80.4  &  73.1  &  78.6  &  30.8 &  48.6  &  62.1  &  26.7  &  46.8  &  42.2  &  56.6 \\
         \bottomrule
    \end{tabular}
    }

\end{table}

\bfparagraph{Computation and \# parameters concerns} 

 Table \ref{tab:13_normal_benchmarks_with_without_longterm_mem} and \ref{tab:6_normal_benchmarks_with_without_longterm_mem} summarize the number of parameters and computation required for transformers and Memory Mosaics v2.\footnote{The flops per token is estimated via \citet{casson2023transformerflops} approach.} Compared with Transformer Large, Memory Mosaics v2 Large uses slightly more parameters and computations to explicitly allocate 3-level memories (persistent-memory, long-term memory and short-term memory). It is worth paying more parameters in exchange for the 3-level memory design, because this design helps in allocating features according to their level of invariance, facilitating feature reuse in new tasks.  This 3-levels design is used to allocate features according to their level of invariance, help feature reusing on new tasks. Furthermore, removing long-term memory from memory mosaics v2 after training achieves a comparable transformer performance on the 13 persistent-knowledge benchmarks, while using fewer parameters and computations.

\begin{table}[ht!]
    \centering
       \caption{Memory Mosaics v2 performance on 6 language benchmarks, where removing the ``long-term memory'' after training dramatically hurt the performance (42.1\% vs 34.9\%).}  
    \label{tab:6_normal_benchmarks_with_without_longterm_mem}
    \resizebox{0.7\textwidth}{!}{
    \begin{tabular}{c|cc|ccc ccc  | c  }
    \toprule
  &  params & flops/token  &    squad & \makecell{bbh}  &  \makecell{math}  &  \makecell{mbpp}   &  \makecell{race\\ middle}  & \makecell{race \\ high} &  avg  \\
         \midrule
         Transformer large  & 8.8B & 16.7B  &  76.3  &  45.6  &  8.7  &  9.8  &  62.6  &  45.6  &  41.4 \\
         \midrule
          \makecell{memory mosaics v2 large}  & 9.9B &  18.9B & 78.2  &  47.8  &  8.8  &  9.6  &  61.6  &  46.5  &  42.1 \\
        \makecell{memory mosaics v2 large \\without long-term memory} & 8.3B & 15.6B  &  69.4  &  24.6  &  5.4  &  6.8  &  59.5  &  43.6  &  34.9 \\
         \bottomrule
    \end{tabular}
    }

\end{table}

\subsubsection{New-knowledge storage and retrieval }
\label{sec:long-term-mem_utilization}

The new-knowledge storage and retrieval ability is a prerequisite for learning new tasks via memory-based methods (e.g., Gaussian kernel regression), because the data of new tasks must be adequately ``stored'' before learning (Note that memory-based methods are lazy methods). To illustrate this point, consider a poor goldfish with 7-second memory -- how can it possibly learn a 90-minute movie? Similarly, a model with limited new-knowledge storage ability will struggle to learn information that exceeds its storage (memory) capacity. 

\bfparagraph{Task description} To assess this ability, we employ two {``multi-unrelated-documents question-answering''} tasks from the \textsc{ruler} benchmark \cite{hsieh2024ruler}. These tasks involve multiple concatenated realistic articles, collected from SQuAD \cite{rajpurkar2016squad} and HotpotQA \cite{yang2018hotpotqa}, followed by a question related to one of these articles. Then expect the model to find the correct answer based on the correct article.\footnote{Similarly to the process used in section \ref{sec:eval_training_data_stsoring_retrieval} for verifying persistent-knowledge storage and retrieval tasks, appendix Table \ref{tab:tf_mm_32k_ruler_qa_tasks_long-term-mem} compares memory mosaics v2 with and without long-term memory on these question-answering tasks, confirming the necessity of ``long-term memory'' for these tasks.} By concatenating more (unrelated) articles, the task-length becomes larger and the task itself becomes more challenging. An example prompt is shown below:

\begin{boxA}
    Answer the question based on the given documents. The following are given documents. Document 1: [...] Document2: [...] [...] Document 20: [...] Question: What religion were the Normans? Answer:
\end{boxA}

Similarly to the process used in Section \ref{sec:eval_training_data_stsoring_retrieval} for verifying persistent-knowledge storage and retrieval tasks, Appendix Table \ref{tab:tf_mm_32k_ruler_qa_tasks_long-term-mem} compares Memory Mosaics v2 with and without long-term memory on these multi-unrelated-documents question-answering tasks, verifying that these tasks truly require ``long-term memory''.

Table \ref{tab:tf_mm_4k_ruler_qa_tasks} compares Transformer and Memory Mosaics v2, pretrained on 4k context length, evaluated on the question-answer tasks. Memory Mosaics v2 outperforms Transformer on 4k task-length by 1.4\%$\sim$5.6\%. Similarly, Table \ref{tab:tf_mm_32k_ruler_qa_tasks} presents the same comparison but with both Transformer and Memory Mosaics v2 models fine-tuned on 32k context length. After context-length extension, Memory Mosaics v2 significantly outperforms Transformer on 32k task-length by 12.3\%$\sim$14.8\%. Moreover, memory mosaics v2 also outperforms many other public base models of similar scale (See Appendix \ref{apx:new-knowledge} Table \ref{tab:memory_mosaics_v2_and_other_base_models_on_ruler} for the details). 

\begin{table}[ht!]
    \centering
        \caption{Comparison of Memory Mosaics v2 and Transformer, trained on 4k context length, on \textsc{ruler} question-answer tasks. Memory Mosaics v2 not only outperforms Transformer on 4k task-length, but also successfully extrapolate the context length $\times4\sim\times8$ times without any fine-tuning.}
    \label{tab:tf_mm_4k_ruler_qa_tasks}
    \resizebox{0.83\textwidth}{!}{
    \setlength{\tabcolsep}{2mm} 
    \begin{tabular}{cc|c| c c c   }
    \toprule
         model          &   \makecell{context\\length}      &   \makecell{task-length\\4k}   &    \makecell{task-length\\8k}   &    \makecell{task-length\\16k}  &    \makecell{task-length\\32k}  \\
         \midrule
        transformer small       &4k &  39.4  &  $\times$  & $\times$  & $\times$   \\
        memory mosaics v2 small    &4k &  45.0  &  35.0  &  34.1  &  31.7 \\
        \midrule
        transformer large     &4k   &  57.7  &  $\times$  &  $\times$  &  $\times$\\ 
        memory mosaics v2 large  &4k   &  59.3  &  48.8  &  46.4  &  26.5\\
\bottomrule
    \end{tabular}
    }

\end{table}

\begin{table}[ht!]
    \centering
        \caption{Comparison of Memory Mosaics v2 and Transformer, trained on 4k and fine-tuned on 32k context length, on \textsc{ruler} question-answer tasks. Memory Mosaics v2 outperforms Transformer by 12.3\%$\sim$14.8\% (36.9\% - 22.1\% = 14.8\%, 53.4\%-41.1\%=12.3\%). }
    \label{tab:tf_mm_32k_ruler_qa_tasks}
    \resizebox{0.98\textwidth}{!}{
    \setlength{\tabcolsep}{2mm} 
    \begin{tabular}{cc|c c c c | c  }
    \toprule
         model          &   \makecell{context\\length}      &   \makecell{task-length\\4k}   &    \makecell{task-length\\8k}   &    \makecell{task-length\\16k}  &    \makecell{task-length\\32k}  & \makecell{task-length\\64k}  \\
         \midrule
        transformer small    &32k  &  37.0  &  29.3  &  29.0  &  22.1 & $\times$ \\
        memory mosaics v2 small &32k  &  44.3  &  39.3  &  39.4  &  36.9 &  25.3\\
        \midrule
        transformer large    &32k &  51.2  &  48.8  &  44.7  &  41.1 & $\times$ \\
        memory mosaics v2 large &32k &  58.9  &  55.5  &  54.9  &  53.4 &  46.4\\
        \bottomrule
    \end{tabular}
    }

\end{table}
\bfparagraph{The failures of many potential baselines}
\label{parag:failues_of_many_baselines}

Many memory compression algorithms, such as RNNs, LSTM \cite{beck2025xlstm}, and state-space models \cite{gu2023mamba}, fail on this task by construction because they cannot store all articles before reading the question. Similarly, local-window memory approaches, such as Alibi position encoding \cite{press2021train} and sliding-window attention \cite{beltagy2020longformer}, also fail for the same reason. Figure \ref{fig:gpt2_mm_attn_long} in Appendix provides a nice attention plot to illustrate the failure of Alibi. One may argue to read the question before articles while using memory compression algorithms. However, this is exactly the task-specific priori knowledge that an open-world AI needs to avoid. Please recall that \emph{a child does not prepare all questions before going to school}.

\bfparagraph{Extrapolating context length (without fine-tuning)} 
Context length extrapolation (without fine-tuning) not only is computationally appealing, but also reveals the consistency of a model in handling context. Unfortunately, transformers (with \textsc{rope} position encoding) hardly extrapolate the context length, as shown in Table \ref{tab:tf_mm_4k_ruler_qa_tasks}.\footnote{The comparison ignores many memory compression and local window \cite{press2021train,beltagy2020longformer} approaches, because they fail on this evaluation by construction.} In contrast, Table \ref{tab:tf_mm_4k_ruler_qa_tasks} shows that Memory Mosaics v2, trained on 4k context length, not only outperforms Transformer on 4k length, but also performs well after extrapolating context length $\times4\sim\times8$ times without any fine-tuning or adaptation.\footnote{The difficulty of this multiple-articles question-answering task increases as the task length and number of articles grow. Therefore, it is not surprising that performance decreases as the task-length increases.} 

One may argue that the performance of context length extrapolation (e.g. 48.8\% on 8k task-length) still lags behind that of fine-tuning (i.e., 55.5\% on 8k task-length), a 6.7\% gap. We will show later in section \ref{sec:more-learning-signals} that incorporating more learning signals into objective function further reinforces the context-length extrapolation ability of Memory Mosaics v2, and helps Memory Mosaics v2 fill this gap.

In summary, the results in this section highlight two advantages of Memory Mosaics v2: 1) its ability to extrapolate context length, while baseline transformer cannot; 2) its ability to further boost long context performance through fine-tuning, whereas fine-tuning Transformer only results in a mediocre long-context performance (12.3\% worse than Memory Mosaics v2).

\subsubsection{In-context learning}
\label{sec:inference-data-learning}

Having demonstrated the new-knowledge storage and retrieval ability of Memory Mosaics v2, this section takes a step further to evaluate in-context learning ability. That is, the capacity to learn new tasks / distributions (with fewer examples and less priori knowledge from designers) at inference time. To assess the ability, we employ the classic multiclass classification problems. 

We choose classic classification problems over other fancy benchmarks (e.g., ``xx reasoning'') for two reasons. 
 Firstly, the mechanisms underlying classification are well-studied, allowing us to confidently attribute good or poor performance to the system's properties. Secondly, classification tasks can be designed to be arbitrarily different from the training set by changing the classification boundary, making it easier to measure the ability to learn new distributions at inference-time. In contrast, as of this writing, many fancy benchmarks may not offer the same level of control and fine-grained analysis.

\bfparagraph{Tasks Description} This section evaluates the in-context learning ability using three multi-class classification tasks with either semantic (e.g. ``dog'', ``cat'') or anonymous (e.g. ``class 1'', ``class 2'') target labels, adopted from \citet{li2024long}. The tasks are as follows:
\begin{itemize}
     \setlength\itemsep{-0.3em}
    \item \textbf{banking77} \cite{casanueva2020efficient} is a banking-intent classification task with 77 target categories. The average length of each example is 24 tokens. 
    \item \textbf{tacred} \cite{zhang2017tacred} is a relation classification task of two objects in a sentence, extracted from newswire or webtext, with 41 target categories in total. The average length of each example is 77 tokens. 
     \item \textbf{goemotion} \cite{demszky2020goemotions} is emotion classification task of Reddit comment with 28 target categories.  The average length of each example is 26 tokens. 
\end{itemize}

Notably, semantic labels, such as ``balance\_not\_updated\_after\_bank\_transfer'', may be easier to classify if the model has learned relevant knowledge from the training data. In contrast, anonymous labels, such as ``class\_71'', are less likely to have been seen during training and therefore rely more heavily on the learning of inference data. As a result, we place greater emphasis on tasks with anonymous labels when evaluating in-context learning.

In this section, we adopt a few-shot learning setup, where a single ``shot'' consists of one $(x,y)$ example from each possible target label category. By collecting multiple shots, we create an $n$-shot classification task. To encode these $(x,y)$ examples for Memory Mosaics v2 and Transformers, we serialize the $(x,y)$ pairs into a sequence, followed by a test query $x_{test}$:

\begin{boxA}
    Given a customer service query, please predict the intent of the query. [...]  The examples are as follows: query: $x_{shot1}$, instant: $y_{shot1}$, [...], query: $x_{shot2}$, instant: $y_{shot2}$, [...], query: $x_{test}$, instant:
\end{boxA}

Transformers are known to be sensitive to the prompt strategies \cite{gupta_changing_2024,mirzadeh_gsm-symbolic_2024}, such as the delimiter before $x$ and $y$, shuffling/not-shuffling the $(x,y)$ examples within each shot. To reduce the influence of prompt strategies, we evaluate each classification task with different delimiters (``[space]'' and ``$\backslash n$''), shuffled/non-shuffled $(x,y)$ examples. Then choose the best prompt strategy for each $n$-shot classification task. Check appendix \ref{apx:prompt_examples_icl_class} for the examples of prompt.

\bfparagraph{Main Results} 

Figure \ref{fig:inference-time-classification_semantic_labels} compares the performance of Memory Mosaics v2 (small / large) and Transformer (small / large) in three classification tasks with semantic target labels. The horizontal axis represents the number of shots, while the vertical axis represents the classification accuracy on $x_{test}$. The results show that Memory Mosaics v2 consistently improves classification performance as it sees more demonstration shots (\textcolor{blue}{blue curves}). In contrast, Transformer struggles to maintain its performance, and exhibits a counter-intuitive degraded performance as more demonstrations are provided (\textcolor{red}{red curves}). Furthermore, Memory Mosaics v2 significantly outperforms Transformers by more than 10\%. 

Figure \ref{fig:inference-time-classification_anonymous_labels} presents a similar comparison as Figure \ref{fig:inference-time-classification_semantic_labels}, but on anonymous target labels. Again, Memory Mosaics v2 significantly outperforms Transformers on all tasks.

\begin{figure}[ht!]
    \centering
    \includegraphics[width=0.9\linewidth]{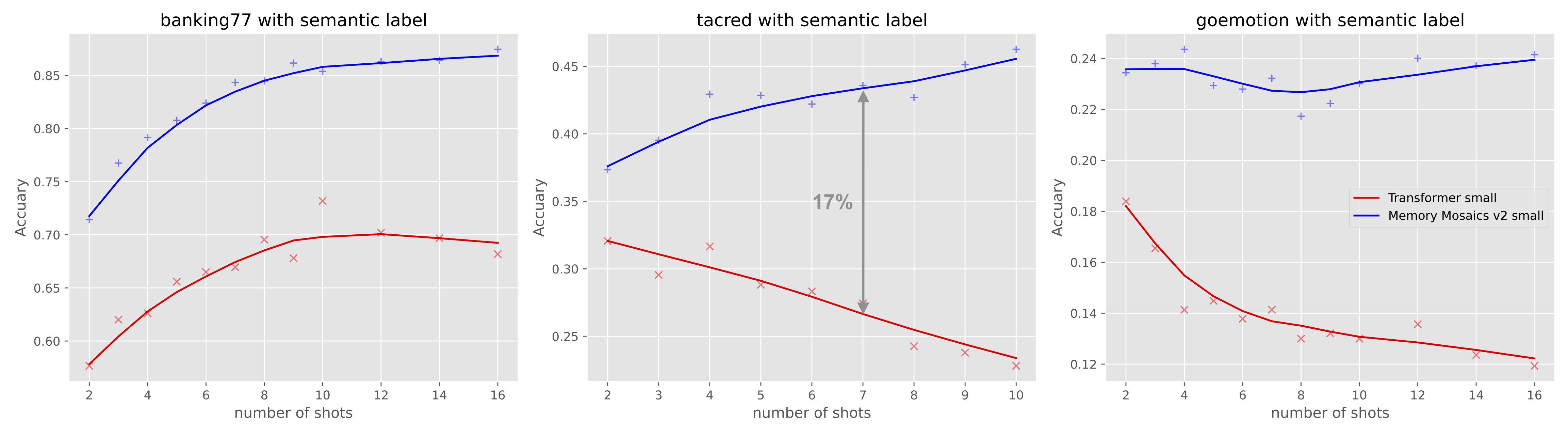}
    \includegraphics[width=0.9\linewidth]{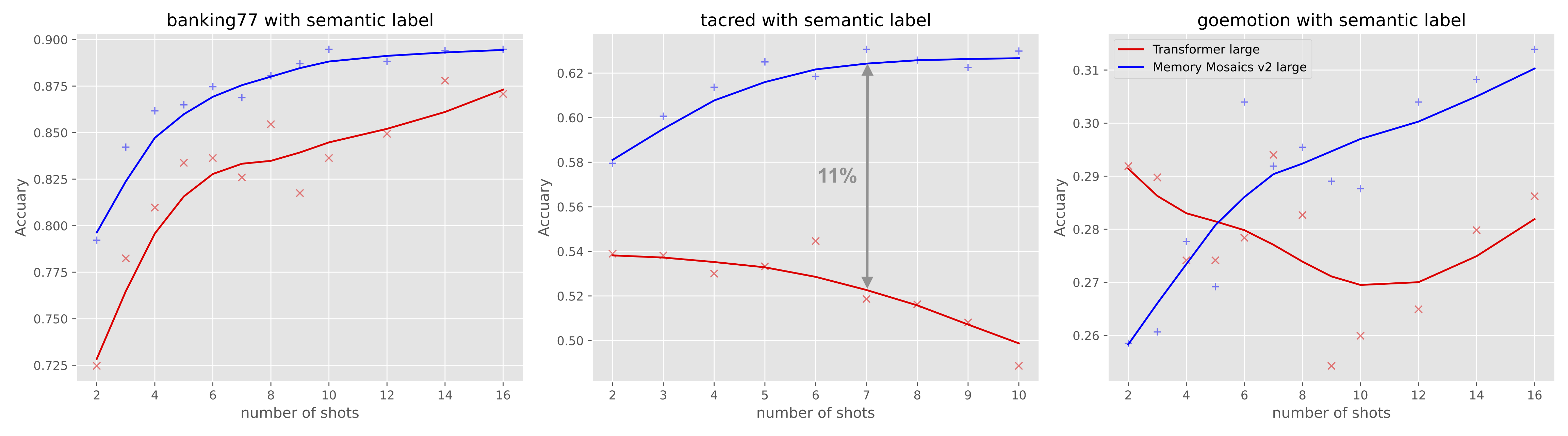}
    \caption{Semantic label in-context learning comparison between Memory Mosaics v2 (small/large) and Transformer (small/large). Memory Mosaics v2 significantly outperform Transformer on in-context learning with a large margin (more than 10\%). Meanwhile, Memory Mosaics v2 benefits from more demonstration shots (x-axis). In contrast, the performance of Transformer may decrease as providing more shots.}
    \label{fig:inference-time-classification_semantic_labels}
\end{figure}

\begin{figure}[ht!]
    \centering
    \includegraphics[width=0.9\linewidth]{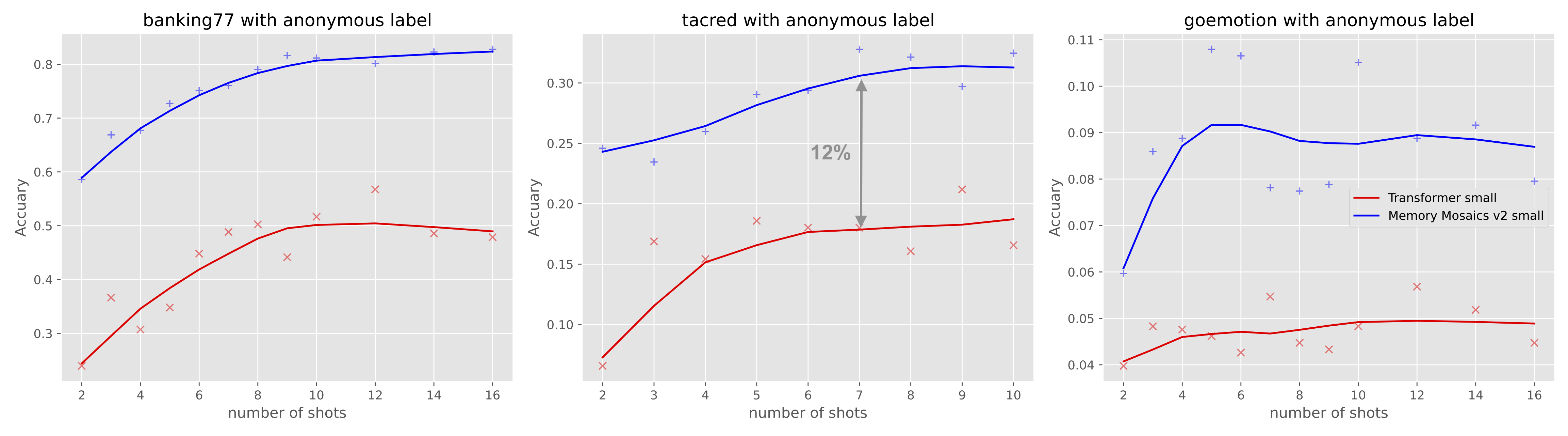}
         \includegraphics[width=0.9\linewidth]{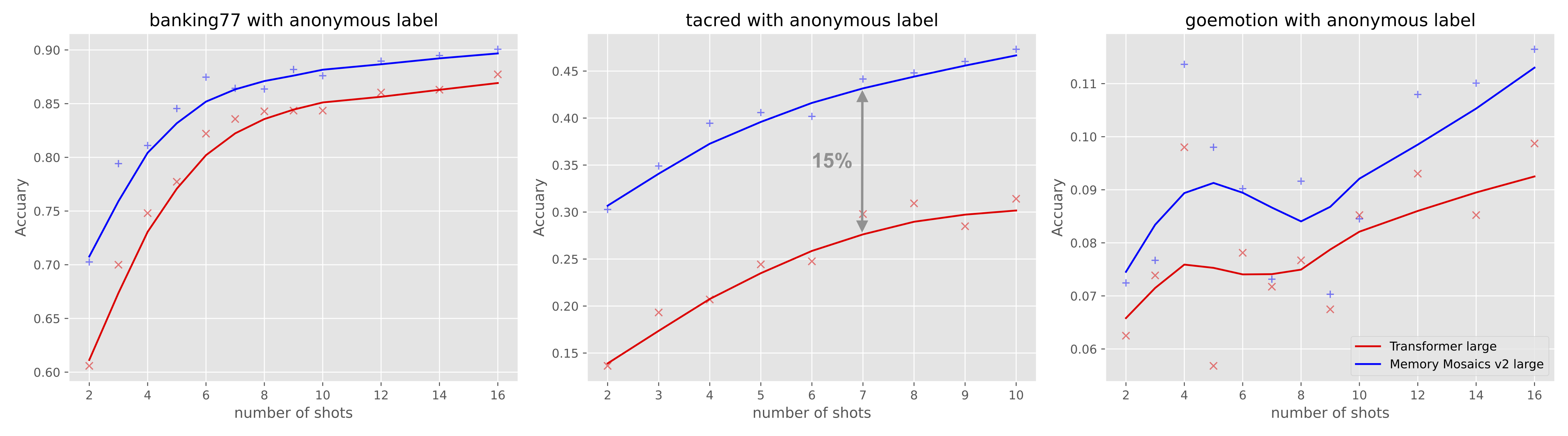}
    \caption{Anonymous label in-context learning comparison between Memory Mosaics v2 (small/large) and Transformer (small/large). Memory Mosaics v2 significantly outperform Transformer on in-context learning with a large margin (more than 10\%).}
    \label{fig:inference-time-classification_anonymous_labels}
\end{figure}

In summary, the experiments demonstrate that Memory Mosaics v2 not only outperforms Transformer on in-context learning by a significant margin (more than 10\%), but also consistently improves the performance as more demonstrations are provided. These results highlight the superior in-context learning ability of Memory Mosaics v2.

\bfparagraph{inference-time learning vs task-specific fine-tuning} A common transfer-learning approach is fine-tuning pre-trained models on a collection of examples from a new task. AI designers choose suitable fine-tuning algorithms and hyper-parameters according to the personal understanding of the ``distance'' between the new task and the training dataset. The selection heavily relies on AI designers' priori knowledge. In contrast, the goal of inference-time learning is to learn new tasks with fewer examples and less priori knowledge from human designers. 

Table \ref{tab:task-specific_task-agnostic_comparison} provides a summary of fine-tuning approaches and inference-time learning approach (Memory Mosaics v2) on the three multiclass classification tasks. As expected, fine-tuning performs exceptionally well when incorporating large amounts of data and task-specific priori knowledge from human designers. On the other hand, Memory Mosaics v2 achieves good performances on all tasks with fewer examples and (almost) no task-specific priori knowledge.

\begin{table}[ht!]
    \centering
        \caption{A summary of fine-tuning and memory mosaics v2 approaches on multi-class classification tasks. (Top) Fine-tuning approach adapts each classification task with a special algorithm and a set of hyper-parameters on a pretrained RoBERTa-large model. ``-'' means the author didn't report the hyperparameter searching space. (Bottom) Memory Mosaics v2 Large on classification tasks with semantic labels. It produces good performance on all tasks with one single model, $92.4\%\sim99.4\%$ less data, and almost no task-specified priori knowledge. Note that a fine-tuned model on one tasks (e.g. banking77) cannot work on another task (e.g. goemotion).}
    \label{tab:task-specific_task-agnostic_comparison}
    \resizebox{0.98\textwidth}{!}{
    \begin{tabular}{cc|ccc|c}
    \toprule
         approach & dataset  & \# training examples & \makecell{task-specified\\algorithm} & \makecell{hyperparameters\\searching space} & accuracy \\ 
         \midrule
         \multirow{4}*{\makecell{fine-tuning}} & \multirow{2}*{banking77} &  10,003  & linear probing & - & 93.7 \cite{lin2023selective} \\
         &           & $10\times77=770$ & linear probing & - &  86.1 \cite{lin2023selective} \\
         & tacred    &  68,124  & insert adapter & $96$ & 70.1 \cite{wang2020k}\\
         & goemotion &  43,410  & fine tuning  & -  & 48.2 \cite{roberta-large-goemotions} \\
         \midrule
         \multirow{3}*{\makecell{Memory Mosaics v2}}   & banking77 &  $10\times77=770$  & No  & 0 &  89.5 \\
            & tacred    &  $10\times41=410$  & No  & 0 &  63.0 \\
            & goemotion &  $16\times28=448$  & No  & 0  &31.4  \\
    \bottomrule
    \end{tabular}
    }

\end{table}

\subsection{Risk-Return trade-off of frontier-model-sized Memory Mosaics v2}
\label{sec:risk-return-trade-off-mmv2}

Having demonstrated the superior open-world AI ability of Memory Mosaics v2 up to 9.9 billion parameters and 1 trillion training tokens, this section analyzes the ``risk-return trade-off'' to further scale Memory Mosaics v2 to the size of the frontier model, unveiling potential benefits and difficulties.

\bfparagraph{Two Approaches} To train a huge frontier foundational model, one can either:
\begin{itemize}
     \setlength\itemsep{-0.3em}
    \item[1)] take a low-risk-low-return approach by investing more resources (GPUs and data) and reusing old recipes (e.g. architecture), or
    \item[2)] take a middle-risk-high-return approach by trying new smart techniques. 
\end{itemize}

Taking the former approach, one can take advantage of existing software, hardware, experiences, and datasets to quickly ``reproduce'' a huge foundational model. However, this approach is unlikely to result in a model that stands out from others, as it is based on shared recipes.

In contrast, taking the latter approach may require software and hardware optimization, adaptation of technique components, a sharp sense of research direction, and problem-solving ability. \emph{These requirements, in turn, demand a small group of high-quality researchers and managers, rather than a large group of mediocre researchers and managers.} Despite the high requirements of people, however, this approach has the potential for a tremendous breakthrough.

Ultimately, the decision between these two approaches depends on the available resources and people at hand. To aid in this decision-making process, this section provides a simple and brutal comparison:

\begin{boxF}
    How much more data does the transformer recipe approach need to match the performance of memory mosaics v2?
\end{boxF}

\subsubsection{Comparison of two approaches}

To answer the question above, this section compares the new tasks learning ability (open-world AI ability)\footnote{In \iid regime, such as persistent-knowledge storing and retrieval, of course, more data + larger model = better performance. This argument in \iid~scenario has been well studied three decades ago \cite{vapnik1991risk}.} of Memory Mosaics v2 and Transformer Large models trained on different amounts of data. Specifically, multiple Transformer Large models are trained on 200\textsc{b}, 1\textsc{t}, and 8\textsc{t} training tokens, while a Memory Mosaics v2 Large model is trained on 1\textsc{t} training tokens.

\bfparagraph{{New-knowledge storage and retrieval}} Table \ref{tab:tf_mm_ruler_qa_tasks_size_of_data} shows the comparison on the \emph{new-knowledge storage and retrieval} ability. Transformer Large trained on 1\textsc{t} tokens lags behind Memory Mosaics v2 by 12.3\% (41.1\% vs 53.4\%). $\times8$ times more training tokens (8\textsc{t}) boosts the performance of Transformer Large by 5.8\% (46.9\% vs 41.1\%). Despite the $\times8$ times more training tokens, Transformer Large trained on 8\textsc{t} still lags behind Memory Mosaics v2 trained on 1\textsc{t} by 6.5\%.

Although further increasing training data may improve the performance of Transformer, it comes at the cost of significantly larger training cost (time and resource). Moreover, a serious problem occurs: we are running out of data!

\begin{table}[ht!]
    \centering
        \caption{Comparison of Memory Mosaics v2 and Transformer, trained on 4k and fine-tuned on 32k context
length, on \textsc{ruler} question-answer tasks. Transformer lags behind Memory Mosaics v2 by 12.3\% when training on the same amount of data. $\times 8$ times training data (8\textsc{t}) boosts the performance of Transformer, but the resulting Transformer model still lags behind Memory Mosaics v2 (1\textsc{t}) by 5.8\%. (``transformer large*'' uses group-query attention to reduce memory cost, 8k training context length to boost long-context performance.)}
    \label{tab:tf_mm_ruler_qa_tasks_size_of_data}
    \setlength{\tabcolsep}{2mm} 
    \resizebox{0.95\textwidth}{!}{
    \begin{tabular}{ccc|c c c c | c cc }
    \toprule
         model         &   \makecell{context\\length} & \makecell{train\\tokens}     &   \makecell{task-length\\4k}   &   \makecell{task-length\\8k}   &   \makecell{task-length\\16k}   &   \makecell{task-length\\32k} & \makecell{task-length\\64k} && \\
         \midrule
transformer large    &32k & 200\textsc{b} &  48.6  &  42.9  &  40.7  &  33.8 & $\times$&&\\ 
transformer large    &32k & 1\textsc{t}   &  51.2  &  48.8  &  44.7  &  41.1 & $\times$&&\\
transformer large*    &32k & 8\textsc{t}  &  59.2  &  54.5  &  50.9  &  46.9 & $\times$&& \\
\midrule
memory mosaics v2 large &32k & 1\textsc{t} &  58.9  &  55.5  &  54.9  &  53.4 &46.4&\\
\bottomrule
    \end{tabular}
    }
\end{table}

\begin{figure}[ht!]
    \centering
    \includegraphics[width=0.98\linewidth]{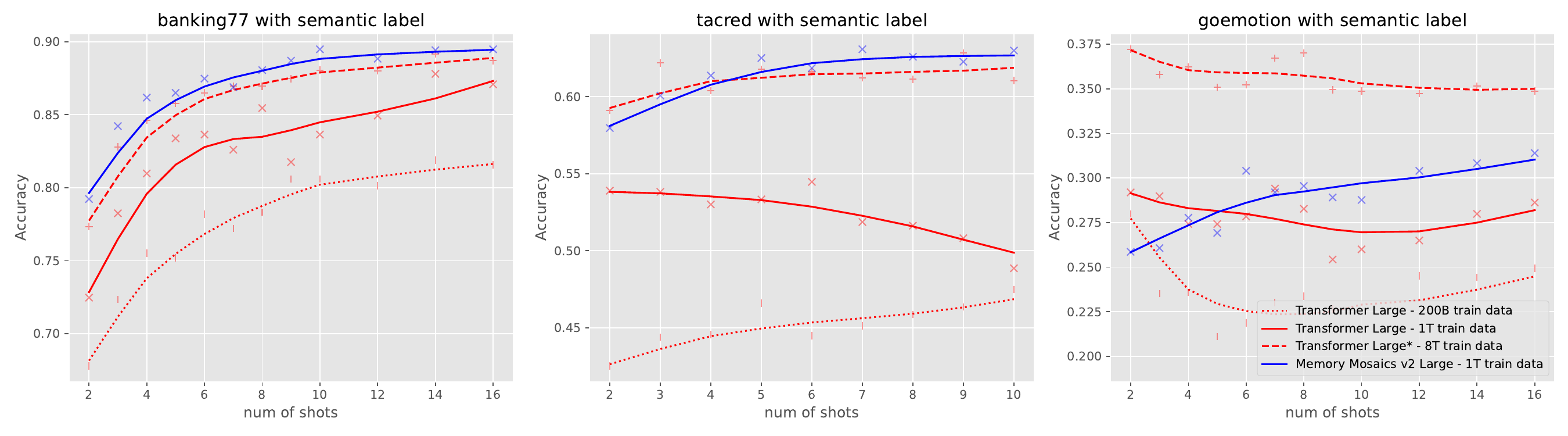}
    \caption{In-context learning (multiclass classification with semantic labels) comparison between Memory Mosaics v2 Large and Transformer Large.  Memory Mosaics v2 is trained on 1\textsc{t} tokens, while three transformers are trained on 200\textsc{b}, 1\textsc{t}, 8\textsc{t} tokens, respectively. Transformer with $\times 8$ times more training data (8\textsc{t}, \textcolor{red}{dash red line}) starts to match the performance of Memory Mosaics v2 (1\textsc{t}, \textcolor{blue}{solid blue line}). }
    \label{fig:tf_mm_classification_semantic}
\end{figure}

\begin{figure}[ht!]
    \centering
    \includegraphics[width=0.98\linewidth]{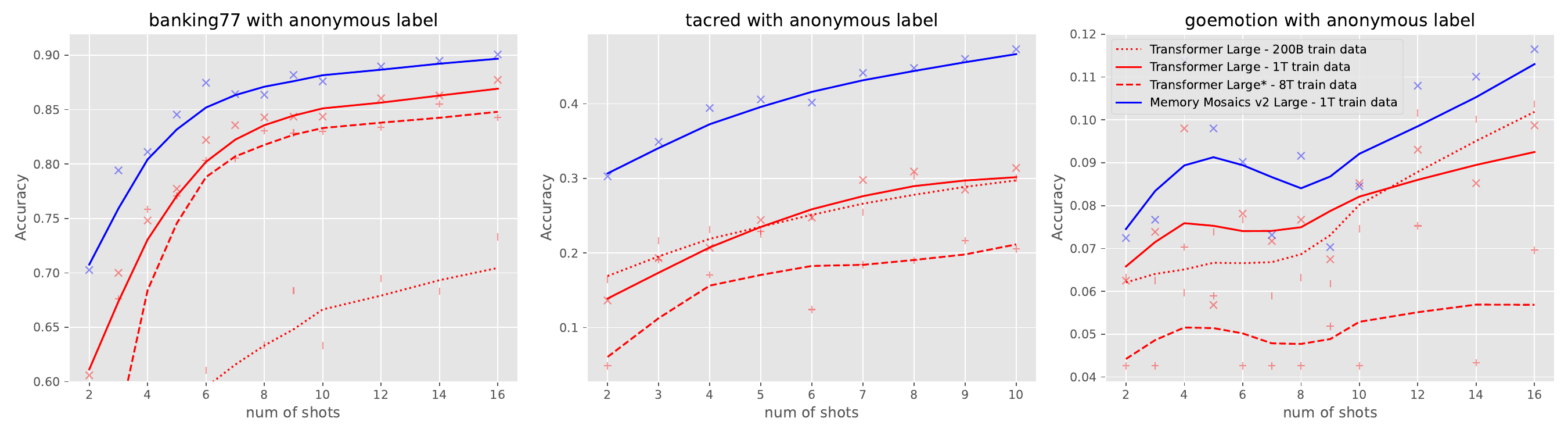}
    \caption{In-context learning (multiclass classification with anonymous labels) comparison between Memory Mosaics v2 Large and Transformer Large. The Transformer trained on 8\textsc{t} (\textcolor{red}{dash red line}) still lags behind Memory Mosaics v2 trained on 1\textsc{t} (\textcolor{blue}{solid blue line}) by a large margin. }
    \label{fig:tf_mm_classification_anonymous}
\end{figure}

\bfparagraph{In-context learning} Figures \ref{fig:tf_mm_classification_semantic} and \ref{fig:tf_mm_classification_anonymous} show the comparison on in-context learning tasks. $\times 8$ times more training data helps Transformer Large (8\textsc{t} data) match the performance of Memory Mosaics v2 (1\textsc{t} data) on semantic label tasks (Figure \ref{fig:tf_mm_classification_semantic}). However, more training data cannot help Transformer Large on the more difficult anonymous label tasks. In contrast, more training data (8\textsc{t}) hurts Transformer Large on anonymous label tasks (Figure \ref{fig:tf_mm_classification_anonymous}).

In summary, $\times 8$ more training data helps Transformer in some tasks that assess open-world AI ability, but the resulting model (8\textsc{t} training data) still lags behind Memory Mosaics v2 trained on less data (1\textsc{t} training data). In some difficult tasks, more training data cannot help Transformer match Memory Mosaics v2. These experiments answer the initial question: \textit{``How much data does the transformer recipe approach need to match the performance of memory mosaics v2?''}.

\subsection{More learning signals for context-length extrapolation}
\label{sec:more-learning-signals}

As discussed in section \ref{sec:long-term-mem_utilization}, the context-length extrapolation ability of Memory Mosaics v2 trained with next-token prediction objective is imperfect. For example, Memory Mosaics v2, trained on 4k context length, achieves a 48.8\% accuracy in 8k task context length (Table \ref{tab:tf_mm_4k_ruler_qa_tasks}). Fine-tuning on 32k context length boosts Memory Mosaics v2 performance to 55.5\% (Table \ref{tab:tf_mm_32k_ruler_qa_tasks}), a 6.7\% gap. 
This performance gap arises mainly from two main reasons: 
\begin{itemize}
     \setlength\itemsep{-0.3em}
    \item[1)] \textbf{Amplified error during extrapolation.} For example, in the estimation of the bandwidth, $\beta = \beta_1 n^\alpha + \beta_0$, a small error in $\alpha$, $\alpha + \epsilon$, results in a large error factor $\beta_1 n^\epsilon$, at length $n$. 
    \item[2)] \textbf{Limited learning signals in the objective function.}  The next-token objective function (within a certain training context length, e.g. 4k) cannot effectively distinguish ``noisy'' (e.g. $\alpha+\epsilon$) and ``cleaned'' (e.g. $\alpha$) extrapolation parameters. This is evident in the fact that fine-tuning Memory Mosaics v2 Large on 32k context length significantly boosts long-context performance but doesn't reduce the training loss within 4k context length.    
\end{itemize}
While the first reason is inevitable, the second reason comes from the lack of learning signals. This section explores the effect of more learning signals on context-length extrapolation ability.

To encode more learning signals, a straightforward approach is to incorporate challenging predictions into the objective function. However, designing objective functions requires a careful balance between optimization difficulty and generalization ability. On the one hand, the objective function cannot be too difficult to optimize, On the other hand, it cannot be too trivial to reveal interesting properties. Check the optimization \& generalization dilemma in Chapter \ref{chap:rich_representation_ood_scope} for a more detailed discussion about this trade-off.\footnote{In certain cases, an objective function can be simultaneously too hard to optimize and too trivial to reveal the underling properties.}

\subsubsection{Objective function: bag of future tokens}

In the sequential language domain, \citet{gloeckle2024better} proposes to predict more future tokens ($x_{t+1}, \dots, x_{t+k}$) from time $t$, rather than the next token ($x_{t+1})$, to speed up inference. This is a simple, yet effective way to introduce more learning signals. Inspired by this work, we introduce more learning signals by applying two classifiers on the last Memory Mosaics v2 persistent memory block. The first classifier predicts the next token, while the second classifier predicts the bag of next k tokens (regardless of order).\footnote{Treating the future tokens as a bag of tokens without order allows the model to predict a far future.} 

For the next token prediction, we employ the common cross-entropy loss on the output of the first classifier. For the bag of next $k$-tokens prediction, we employ the following binary-cross entropy between second classifier logits $h=[h_1, \dots, h_v]$ ($v$ indicates vocabulary size) and the bag of next $k$-tokens  $[x_1, \dots, x_k], x_i\in \{1,\dots, v\}$:
\begin{align*}
    \mathcal{L}_{\text{bag-k}}(h, [x_1, \dots, x_k]) &= - \frac{1}{k} \sum_{i\in [1,k]} \left[ -\frac{1}{v}[\eta \cdot \log \sigma(h_{x_i}) + \sum_{j\in [1,v], j\neq x_i} \log (1-\sigma(h_j))] \right] \\
    &= \frac{1}{kv} \left[ k\sum_{j\in[1,v]} \log(1+e^{h_j})\, +\sum_{i\in[1,n]}\left[ (\eta-1) \log(1+e^{h_{x_i}}) - \eta h_{x_i}\  \right]\right]\,,
\end{align*}
where $\sigma(\cdot)$ is sigmoid function, $log(1+e^x) = log(e^{x-\max(x,0)} + e^{-\max(x,0)}) + \max(x,0)$ (logsumexp) is applied during forward pass for numerical stability, $\frac{\partial \log (1+e^x)}{\partial x} = \sigma(x)$ is applied during backward pass to reduce memory cost.

\subsubsection{Experiments}
\bfparagraph{Experimental Setups}

We train Memory Mosaics v2 Large on 1\textsc{t} tokens using an objective function $\mathcal{L}_{\text{next}} + \alpha \mathcal{L}_{\text{bag-k}}$, where $\alpha=0.3$, $\mathcal{L}_{\text{next}}$ is cross-entropy on next token. We set $\eta=64000,k=64$ in $\mathcal{L}_{\text{bag-k}}$, skip the computation of $\mathcal{L}_{\text{bag-k}}$ on the last $k$ tokens (because the future of these tokens is smaller than $k$). All other experimental details follow the pretraining setups in Section \ref{sec:mmv2_training_and_results}.

After pretraining on 4k context length, we fine-tune the resulting Memory Mosaics v2 Large on 32k context-length with the same $\mathcal{L}_{\text{next}} + \alpha \mathcal{L}_{\text{bag-k}}$ objective function as pretraining. Other fine-tuning details follow the fune-tuning setups in Section \ref{sec:mmv2_training_and_results}.

\bfparagraph{Main Results}

Table \ref{tab:mm_future_ruler_qa_tasks} compares two Memory Mosaics v2 models trained with the objective functions $\mathcal{L}_{\text{next}}$ and $\mathcal{L}_{\text{next}} + \alpha \mathcal{L}_{\text{bag-k}}$, on \textsc{ruler} question-answer tasks. After 4k context-length pretraining, Memory Mosaics v2 trained with $\mathcal{L}_{\text{next}} + \alpha \mathcal{L}_{\text{bag-k}}$ outperforms that with $\mathcal{L}_{\text{next}}$ in the 4k task length by 2\% (61. 3\% vs 59. 3\%), in the 8k task length by 8. 8\% (57. 6\% vs 48. 8\%). 

With the $\mathcal{L}_{\text{next}}$ training objective, fine-tuning on a longer context length boosts the performance of the 8k task length by 6.7\%. However, With the $\mathcal{L}_{\text{next}} +  \alpha \mathcal{L}_{\text{bag-k}}$ training objective, fine-tuning on a longer context-length can only boost the performance by 0.8\%, showing a great context-length extrapolation ability of Memory Mosaics v2 trained with $\mathcal{L}_{\text{next}} +  \alpha \mathcal{L}_{\text{bag-k}}$.

\begin{table}[ht!]
    \centering
    \resizebox{0.8\textwidth}{!}{
    \setlength{\tabcolsep}{3mm} 
    \begin{tabular}{cccc| c    }
    \toprule
         model         & \makecell{objective\\function}  &   \makecell{context\\length}      &   \makecell{task-length\\4k}   &    \makecell{task-length\\8k}   \\
         \midrule
memory mosaics v2 large & $\mathcal{L}_{\text{next}}$   &4k &  59.3  &  48.8  \\
memory mosaics v2 large & $\mathcal{L}_{\text{next}}$  &32k  &   58.9  {\footnotesize \textcolor{red}{(-0.4)}}  &  55.5  {\footnotesize (+6.7)}  \\
\midrule
memory mosaics v2 large  & $\mathcal{L}_{\text{next}} + \alpha \mathcal{L}_{\text{bag-k}}$ & 4k&  61.3  &  57.6  \\
memory mosaics v2 large & $\mathcal{L}_{\text{next}} + \alpha \mathcal{L}_{\text{bag-k}}$ & 32k & 61.9 {\footnotesize (+0.6)}  &  58.4  {\footnotesize (+0.8)} \\
\bottomrule
    \end{tabular}
    
    }
    \caption{\textsc{Ruler} question-answer tasks comparison. Two objective functions, $\mathcal{L}_{\text{next}}$ or $\mathcal{L}_{\text{next}} + \alpha \mathcal{L}_{\text{bag-k}}$, are used to train Memory Mosaics v2 Large on 4k context length. Then finetune 32 context length. Using $\mathcal{L}_{\text{next}}$ objective function, the pretrained model lags behind the finetuned counterpart by 6.7\%. Using $\mathcal{L}_{\text{next}} + \alpha \mathcal{L}_{\text{bag-k}}$, this gap is reduces to 0.8\%.  }
    \label{tab:mm_future_ruler_qa_tasks}
\end{table}

This section uses this simple ``bag of next k token'' objective function as an example to study the impact of more learning signals on Memory Mosaics v2. Of course, there are many other appealing learning signals, such as predicting a far future through reinforcement learning \cite{guo2025deepseek}. We leave the research on other learning signals for future work.

\section{Discussion}
\label{sec:inference-time-discussion}

\subsection{new learning paradigm}

\emph{Inference-time learning} is a new learning paradigm that utilizes memory-based methods at inference-time, leveraging rich features and disentangled representation constructed during pre-training. That is, a gradient-based optimization process in pretraining stage to prepare \emph{rich features} and \emph{disentangled representation}, and a memory-based approach in inference stage to learn new tasks quickly with fewer examples and less priori knowledge from human designers. The ``local updating'' nature of memory-based method avoids the negative interference problem in model-based algorithms, making it possible to learn multiple task sequentially --- akin the daily school life of a child.

To perform this \emph{inference-time learning} paradigm, this chapter introduces the Memory Mosaics v2 architecture. This architecture outperforms the baseline transformers in the dimensions \emph{new-knowledge storing and retrieval} and \emph{in-context learning} by more than 10\%, demonstrating a great open-world AI ability. Furthermore, this chapter shows that $\times 8$ times more training data (8\textsc{t} train tokens) hardly helps baseline transformers to match the performance of Memory Mosaics v2 (1\textsc{t} train tokens), implying the tall ceiling of Memory Mosaics v2 and inference-time learning.

\subsection{Future works} 
Memory-based approaches (e.g. k-nearest neighbors) often suffer from a $O(n)$ time complexity at inference, where $n$ is the number of examples. It may impede the application of this \emph{inference-time learning} paradigm in practice due to computation and memory size. Fortunately, this computation cost can be significantly cutoff via fuzzy hashing \cite{breitinger2014approximate, chen2024magicpig} from the non-exhaustive search viewpoint, or hierarchical memory \cite{yuan2025native, lu2025moba} from the information organization viewpoint. Furthermore, the long-term memory in Memory Mosaics v2 is designed to be permutation-invariant (key-value pairs), making it amenable to use all these techniques.

Exploring ``stronger'' learning signals at the successive stage after the initial pretraining stage, such as learning to predict a far future via reinforcement learning \cite{guo2025deepseek}, is an interesting but orthogonal direction. 

It is well known that the reconstruction-based objective function is helpful for the compact language data. However, for other completed data, such as video, early work \cite{lerer2016learning} reveals the weakness of this reconstruction-based objective function. That is, the prediction tends to be ``blur''. Incorporating an objective function working on latent space (rather than input space), such as Joint Embedding \citep{lecun2022path}, into Memory Mosaics v2 is another interesting research direction.

Currently in Memory Mosaics v2, the hyperparameter of the memory-based method (i.e. bandwidth of Gaussian kernel smoothing)  is determined by the ``input statistics'' (e.g. number of examples) instead of cross-validation. It is well known that cross-validation is (almost) always the best way to choose hyperparameters. Automatically selecting bandwidth parameter, as well as some other essential parameters, at inference-time using ``cross-validation'' is another appealing future direction.

\chapter{Discussion and Future Directions}
\label{chap:future_directions}

This thesis focuses on \emph{learning principles} to build artificial intelligence (AI) for the open world, proposing three learning principles --- \emph{rich features}, \emph{disentangled representation}, and \emph{inference-time learning}. These learning principles are explained via imaginary examples and little theories, implemented by innovative techniques, and verified through extensive large-scale experiments.

It is worth noting that the three principles are connected. 
\vspace{-0.6em}
\begin{itemize}
     \setlength\itemsep{-0.3em}
     \item \emph{Rich feature} provides an optimization benefit to the other two learning principles. Directly pursuing advanced properties of features (e.g. invariant or disentanglement) may encounter huge optimization difficulties, as discussed in the generalization-optimization dilemma in Section \ref{sec:bonsai}. Rich features provide stairs to reduce the difficulty of optimization.

    \item \emph{Disentangled representation} organizes features nicely, which in turn aids in further discovery of \emph{rich features}. For example, one can explicitly ``mask'' certain learned features and encourage the model to discover other features, as in \textsc{Bonsai} and \textsc{very-large dropout}.

    \item \emph{Rich feature} and \emph{disentangled representation} prepare and organize features, so that the memory-based methods in \emph{inference-time learning} can avoid the curse of dimensionality. Ultimately, enable \emph{inference-time learning} to learn new tasks quickly with fewer examples and less priori knowledge from human designers. 
\end{itemize}

This thesis highlights numerous future work opportunities in the discussion of each chapter. Overall, a profound hint of future directions is in hardware and software design. Hardware and software are intertwined and have deeply influenced each other since the inception of computer science. Memory and computation are key aspects in both algorithms and computers. Modern large-scale parallel computation hardware (e.g. GPU), as well as software, shift to computation more than memory. This thesis suggests a focus on memory in building AI for the open-world. 
\vspace{-0.6em}
\begin{itemize}
 \setlength\itemsep{-0.3em}
    \item The first reason is that quick learning on new task with few examples does not require extensive parallelization. 
    \item More importantly, this quick learning is performed on non-\iid~and even active environments (e.g. driving). Feedback from active environments is crucial for the quality of learning. Although computation could be parallelized, this feedback from active environment is inherently sequential! 
\end{itemize}

\chapter*{Acknowledgements}
\addcontentsline{toc}{chapter}{Acknowledgments}

These acknowledgments are surely incomplete.

To mother, father, brother, and Lining Zhang, for the unconditional support, the love, and the eternal companionship.

To Françoise Soulié.  A sunny afternoon 2015 Summer at Tianjin University in China, I took my friend to meet Francoise to learn machine learning. In that afternoon, my machine learning background knowledge was almost empty, I could barely speak a couple words in English. Luckily, I got full support from Francoise, ranging from machine learning knowledge and practice, teaching experience, Kaggle competition, international visits, strong recommendations\footnote{Francoise recommended me to Léon during 2018 fall.}.

To Léon Bottou. I first met Léon in 2019 spring in New York City, where I was Léon's intern student at FAIR after finishing my Master's degree from Tianjin University. During the internship, I received advice, met researchers, and had a great summer in New York. The next year, I became Léon's student. Since then, I ate a lot of Léon's time, received Léon's full support --- I was advised in a rare one-on-one mode. In research, Léon is extremely sharp in direction and problem, patient in counterintuitive ideas. In summer 2021, I accidentally got the counterintuitive ``rich feature'' idea\footnote{Papers of this direction are hard to be accepted by the mainstream. All my papers were rejected at least once before publication. Some paper \cite{zhang2024fine} has been rejected 3-4 times.}, did some toy experiments and asked Léon for advice. My understanding of what I did was quite fuzzy. To my surprise, Léon immediately pinpointed the interesting part and then advised me to research it in the following years. Examples of this kind are enormous.

To Yann Lecun. I met Yann a few times in New York during the 2019 intern. The next year, during the PhD interview, Yann asked me ``{What do you want to do?}''. I blurted out ``{learning with small data}''.\footnote{During the interview (2020), I had neither idea about AI for the open-world nor the three learning principles. My response simply came from a fuzzy feeling that I believe in.} This response was too fuzzy to be understood.\footnote{Don't learn it for your interview!} Luckily, Yann accepted the response (after a couple of conversions) and asked another question ``{Who do you want to work with?}''. From this question, my PhD journal was started. After 5 years, this thesis clarifies my initial fuzzy response.

To my committee members, Kyunghyun Cho, Joan Bruna, and Alfredo Canziani, for their enormous insightful suggestions on this thesis over the 5 years. In every phd yearly review, I received helpful feedback from my committee members. Having you as my committee members is my pleasure.

In Chinese, there is a famous proverb from 1200 years ago, ``\textit{Good horses are common, but Bole, the great horse identifier, is rare.}''. It describes a common scenario in society: while there may be many talented individuals, it is rare to find someone with the discernment and skill to recognize and cultivate that talent. Luckily, I got many.

\newpage
\bibliographystyle{plainnat}
\bibliography{thesis}

\clearpage
\begin{appendices}

\chapter{Rich features}
\section{\textsc{Cifar} supervised transfer learning}
\label{apx:cifar10_100_sl}
\textsc{Cifar10} supervised transfer learning experiments train a \textsc{resnet18} network on the \textsc{Cifar10} dataset with/without L2 weight decay (4e-5) for $200$ epochs. During training, we use a SGD optimizer \citep{bottou2018optimization} with momentum=0.9, initial learning rate=0.1, cosine learning rate decay, and batch size=128. As to data augmentation, we use \textsc{RandomResizedCrop} (crop scale in $[0.8, 1.0]$), aspect ratio in $[3/4, 4/3]$) and \textsc{RandomHorizontalFlip}. During testing, the input images are resized to $36\times36$ by bicubic interpolation and \textsc{CenterCroped} to $32\times32$. All input images are normalized by $mean=(0.4914, 0.4822, 0.4465), std=(0.2023, 0.1994, 0.2010)$ at the end. 

Then transfer the learned representation to \textsc{Cifar100} dataset by training a last-layer linear classifier (linear probing). The linear layer weights are initialized by Gaussian distribution $\mathcal{N}(0, 0.01)$. The linear probing process shares the same training hyper-parameters as the supervised training part except for a zero L2 weight decay in all cases. 

The \textsc{Cifar100} supervised transfer learning experiments swap the order of \textsc{Cifar100} and \textsc{Cifar10}.

\section{\textsc{ImageNet} supervised transfer learning}
\label{apx:imagenet_sl}
\subsection{Experiment settings}
\label{apx:imagenet_sl_settings}

{\paragraph{Image Preprocessing:} Following \citet{he2016deep}, we use \textsc{RandomHorizontalFlip} and \textsc{RandomResizedCrop} augmentations for all training tasks. For \textsc{ImageNet} and \textsc{Inat18}, the input images are normalized by $mean=(0.485, 0.456, 0.406), std=(0.229, 0.224, 0.225)$. For \textsc{Cifar}, we use the same setting as Appendix \ref{apx:cifar10_100_sl}.}

\paragraph{\textsc{Imagenet} Pretraining:} The \textsc{resnet}s are pre-trained on \textsc{ImageNet} with the popular protocol of \citet{goyal2017accurate}: a SGD optimizer with momentum=0.9, initial learning rate=0.1, batch size=256, L2 weight decay=1e-4, and 90 training epochs. The learning rate is multiplied by 0.1 every 30 epochs. By default, the optimizer in all experiments is SGD with momentum=0.9.

\paragraph{\textsc{Distill}:} To distill the {\synthcat}$n$ representations $[\phi_1, \dots \phi_n]$ ($n\times$\textsc{resnet50}) into a smaller representation $\Phi$ (\textsc{resnet50}), we use the multi-head architecture as Figure \ref{fig:distill}. Inspired by \citet{hinton2015distilling}, we use the Kullback–Leibler divergence loss to learn $\Phi$ as:
\begin{equation}
\label{eq:distill}
    \min_{\Phi, w_0, \dots, w_n}\sum_{i=0}^n\sum_x\bigg[ \tau^2\mathcal{L}_{kl}\Big(s_{\tau}\big({v_i} \circ {\phi_i}(x)\big) ~||~ w_i \circ \Phi(x) \Big)\bigg],
\end{equation}

where $s_{\tau}(v)_i = \frac{e^{v_i/\tau}}{\sum_k{e^{v_k/\tau}}}$ is a softmax function with temperature $\tau$, $v_i$ is the learned last-layer classifier of $i^{th}$ sub-network of \synthcat$n$.

In the {\synthdistill} experiments, we distill five separately trained \textsc{resnet50} into one \textsc{resnet50} according to Eq \ref{eq:distill} with $\tau=10$. We use a SGD optimizer with momentum=0.9, batch size=2048, and weight decay=0. The initial learning rate is 0.1 and warms up to 0.8 within the first 5 epochs. Then learning rate decays to 0.16 and 0.032 at $210^{th}$ and $240^{th}$ epochs, respectively. The total training epochs is 270.

\paragraph{Linear probing:}
\begin{itemize}
    \item \textbf{\textsc{ImageNet}:} The \textsc{ImageNet} linear probing experiments train a linear classifier with the same hyper-parameters as \textsc{ImageNet} pretraining. By default, the last linear classifier in all linear probing experiments is initialized by $\mathcal{N}(0, 0.01)$.
    \item \textbf{\textsc{Inat18}, \textsc{Cifar100}, \textsc{Cifar10}:} Following the settings of \citet{goyal2022vision}, the linear probing experiments (on \textsc{Inat18}, \textsc{Cifar100}, \textsc{Cifar10}) adds a \textsc{BatchNorm} layer before the linear classifier to reduce the hyper-parameter tuning difficulty. The learning rate is initialized to 0.01 and multiplied by 0.1 every 8 epochs. Then train these linear probing tasks for 28 epochs by SGD Nesterov optimizer with momentum=0.9, batch size 256. Note that \textsc{BatchNorm} + a linear classifier is still a linear classifier during inference. We tune L2 weight decay from \{1e-4, 5e-4, 1e-3, 5e-3, 1e-2, 5e-2\} for \textsc{Cifar100} and \textsc{Cifar10}, \{1e-6, 1e-5, 1e-4\} for \textsc{Inat18}.

\end{itemize}

\paragraph{Fine-tuning:} As to the fine-tuning experiments (on \textsc{Cifar100}, \textsc{Cifar10}, and \textsc{Inat18}), we tune the initial learning rate from \{0.005, 0.01, 0.05\}, training epochs from \{50, 100\}. We further tune L2 weight decay from \{0, 1e-5, 1e-4, 5e-4\} for \textsc{Cifar100} and \textsc{Cifar10}, \{1e-6, 1e-5, 1e-4\} for \textsc{Inat18}. A cosine learning rate scheduler is used in fine-tuning experiments. A 0.01 learning rate and 100 training epochs usually provide the best performance for these three datasets. So we fix these two hyperparameters in the following supervised learning two-stage fine-tuning experiments and self-supervised learning experiments.

\paragraph{Two-stage fine-tuning:} For the two-stage fine-tuning experiments, we separately fine-tune each sub-network (i.e. \textsc{resnet50}) of the \textsc{Cat}$n$ architecture by the same protocol as the normal fine-tuning above. Then train a last-layer linear classifier on top of the concatenated fine-tuned representation. The last-layer linear classifier training can be very efficient with a proper weights initialization strategy. In this work, we initialize the last-layer classifier $w$ (including the bias term) by concatenating the last-layer classifier of each fine-tuned sub-network $w_i$, $w \leftarrow {[w_0^\top, \dots, w_n^\top]^\top}/{n}$. Then we only need to train the last-layer classifier $w$ for 1 epoch with a learning rate = $1e-3$ for \textsc{Cifar} and $1e-5$ for \textsc{Inat18}.

\begin{table*}[t]
    \centering
     \caption{Training accuracy}
    \begin{tabular}{c|cccc}
    \toprule
    & subnetwork0 & subnetwork1 & subnetwork2 & subnetwork3 \\
    \midrule
2\textsc{resnet50} & 73.94 & 18.05 & - & - \\
4\textsc{resnet50} & 9.25 & 74.33 & 0.40 & 0.96 \\
\bottomrule
    \end{tabular}
    \label{tab:subnetwork_acc}
\end{table*}

\subsection{Experiments on a deeper architecture: \textsc{resnet152}}
\label{apx:imagenet_sl_resnet152}

Similar to table  \ref{tab:imagenet_sl_lineareval} in section \ref{sec:supervisedtransfer}, table \ref{tab:imagenet_sl_rn152_lineareval} provides similar experiments on a deeper architecture \textsc{resnet152}. {\synthcat}$n$ exceeds ERM on \textsc{ImageNet}, \textsc{Cifar10}, \textsc{Cifar100} , and \textsc{Inat18} linear probing tasks. 
\begin{table}[ht]
    \caption{Imagenet supervised transfer learning performance on a deep architecture \textsc{resnet152}.}
    \bigskip
    \label{tab:imagenet_sl_rn152_lineareval}
    \centering
   
    \begin{tabular}{cc |c| ccc }
    \toprule
                 &               &  ID      & \multicolumn{3}{c}{Linear Probing (OOD)} \\ 
         method  & architecture  & \textsc{ImageNet} & \textsc{Cifar10} & \textsc{Cifar100} & \textsc{Inat18}  \\
         \midrule
             ERM & \textsc{resnet152}     & 77.89 &	92.50 &	76.23 & 39.70 \\
        \midrule
             {\synthcat}2& $2\times$\textsc{resnet152}      & 79.34 &	94.26 &	79.15 & 45.42\\
             {\synthcat}5& $5\times$\textsc{resnet152}      & 80.14 &	94.91 &	81.35 & 50.32\\
            {\synthcat}10& $10\times$\textsc{resnet152}      & 80.18 &	95.38 &	82.39 & 52.73 \\
        \bottomrule
    \end{tabular}
    
\end{table}

\subsection{Fine-tuning experiments}
For reference, table \ref{tab:imagenet_sl_ft_2ft} provides numerical results for the fine-tuning experiments of Figure \ref{fig:imagenet_sl_ft_2ft}.
\begin{table}[ht]
    \caption{Supervised transfer learning by either normal fine-tuning or proposed two-stage fine-tuning. Various representations are pre-trained on \textsc{ImageNet} and then fine-tuned or two-stage fine-tuned on \textsc{Cifar10}, \textsc{Cifar100}, \textsc{Inat18} tasks. }
    \bigskip
    \label{tab:imagenet_sl_ft_2ft}
    \centering

    \begin{tabular}{cc c | cc c | ccc}
    \toprule
                 &              &        & \multicolumn{3}{c|}{fine-tuning} & \multicolumn{3}{c}{two-stage fine-tuning} \\
         method  & architecture          &params &  \textsc{Cifar10}  &  \textsc{Cifar100} &  \textsc{Inat18} &  \textsc{Cifar10} &  \textsc{Cifar100} &  \textsc{Inat18} \\
         
         \midrule
             ERM & \textsc{resnet50}     & 23.5M & 97.54	& 85.58	& 64.19 &  - & 	- & - \\
        \midrule
             ERM & \textsc{resnet50w2}   & 93.9M & 97.76	& 87.13	& 66.72 &  - & 	- & - \\
             ERM & \textsc{resnet50w4}   & 375M  & 97.88	& 87.95 & 66.99 &  - & 	- & -\\
             ERM & \textsc{2$\times$resnet50}    & 47M   & 97.39	& 85.77	& 62.57 &  - & 	- & - \\
             ERM & \textsc{4$\times$resnet50}    & 94M   & 97.38	& 85.56	& 61.58 &  - & 	- & - \\
             \midrule
             {\synthcat}2& \textsc{2$\times$resnet50}     & 47M   & 97.56	& 86.04	& 64.49 & 97.87	& 87.07	& 66.96\\
             {\synthcat}4& \textsc{4$\times$resnet50}     & 94M   & 97.53	& 86.54	& 64.54 & 98.14	& 88.00	& 68.42\\
             {\synthcat}5& \textsc{5$\times$resnet50}     & 118M  & 97.57	& 86.46	& 64.86 & 98.19	& 88.11	& 68.48\\
            {\synthcat}10& \textsc{10$\times$resnet50}     & 235M  & 97.19	& 86.65 & 64.39 & 98.17	& 88.50	& 69.07\\
            \midrule
            {\synthdistill}5& \textsc{resnet50}     & 23.5M & 97.07	& 85.31	& 64.17 &  - & 	- & -\\
        \bottomrule
    \end{tabular}
    
\end{table}

\subsection{Vision transformer Experiment settings}
{For all vision transformer experiments, we keep the input image resolution at 384 $\times$ 384 and follow a similar protocol as appendix \ref{apx:imagenet_sl_settings}. Specifically, we use a weight decay=5e-4 and a batch size=256 for linear probing, a weight decay=0 and a batch size=512 (following the \citet{dosovitskiy2020image} settings) for fine-tuning and two-stage fine-tuning. Following \citet{dosovitskiy2020image}, all input images are normalized by $mean=(0.5, 0.5, 0.5), std=(0.5, 0.5, 0.5)$.}

\section{Self-supervised transfer learning}
\label{apx:ssl}
\subsection{SWAV on \textsc{ImageNet}}
SWAV is a contrastive self-supervised learning algorithm proposed by \citet{caron2020unsupervised}. We train \textsc{resnet50} on \textsc{ImageNet}\footnote{\url{https://github.com/facebookresearch/swav/blob/main/scripts/swav_400ep_pretrain.sh}} by the SWAV algorithm four times, which gives us four pretrained \textsc{resnet50} models. As to the rest four SWAV pre-trained models in this work, we  use the public available \textsc{resnet50}\footnote{\url{https://dl.fbaipublicfiles.com/deepcluster/swav_400ep_pretrain.pth.tar}}, \textsc{resnet50w2}\footnote{\url{https://dl.fbaipublicfiles.com/deepcluster/swav_RN50w2_400ep_pretrain.pth.tar}}, \textsc{resnet50w4}\footnote{\url{https://dl.fbaipublicfiles.com/deepcluster/swav_RN50w4_400ep_pretrain.pth.tar}}, and \textsc{resnet50w5}\footnote{\url{https://dl.fbaipublicfiles.com/deepcluster/swav_RN50w5_400ep_pretrain.pth.tar}} checkpoints.

\paragraph{Linear probing:} Following the settings in \citet{goyal2022vision}, the linear probing experiments (on \textsc{ImageNet}, \textsc{Inat18}, \textsc{Cifar100}, \textsc{Cifar10}) add a \textsc{BatchNorm} layer before the last-layer linear classifier to reduce the hyper-parameter tuning difficulty. The learning rate is initialized to 0.01 and multiplied by 0.1 every 8 epochs. Then train these linear probing tasks for 28 epochs by SGD Nesterov optimizer with momentum=0.9. We search L2 weight decay from $\{5e-4\}$, $\{5e-4, 1e-3, 5e-3, 1e-2\}$, and $\{1e-6, 1e-5, 1e-4\}$ for \textsc{ImageNet}, \textsc{Cifar}, and \textsc{Inat18} tasks, respectively. 

\paragraph{Fine-tuning:} 
\begin{itemize}
    \item \textbf{\textsc{ImageNet}}: Inspired by the semi-supervised \textsc{ImageNet} fine-tuning settings in \citet{caron2020unsupervised}, we attach a randomly initialized last-layer classifier on top of the SSL learned representation. Then fine-tune all parameters, using a SGD optimizer with momentum=0.9 and L2 weight decay=0. Low-layers representation and last-layer classifier use different initial learning rates of 0.01 and 0.2, respectively. The learning rate is multiplied by 0.2 at $12^{th}$ and $16^{th}$ epochs. We train 20 epochs for networks: \textsc{resnet50}, \textsc{resnet50w2}, \textsc{resnet50w4}. We further search training epochs from $\{10, 20\}$ for the wide network (due to overfitting), \textsc{resnet50w5} and then select the best one with 10 training epochs. 
    \item \textbf{\textsc{Cifar10, Cifar100, Inat18}}: Same as the fine-tuning settings in supervised transfer learning in Appendix \ref{apx:imagenet_sl_settings}.

\end{itemize}

\paragraph{Two-stage fine-tuning:} 
\begin{itemize}
    \item  \textbf{\textsc{ImageNet}}: Similar to the two-stage fine-tuning settings in supervised transfer learning, we initialize the last-layer classifier $w$ by concatenation and then train 1 epoch with learning rate=0.001, L2 weight decay=0. 
    \item \textbf{\textsc{Cifar10, Cifar100, Inat18}}: For \textsc{Cifar10, Cifar100}, we use same two-stage fine-tuning settings as in supervised transfer learning in Appendix \ref{apx:imagenet_sl_settings}. For \textsc{Inat18}, we attach a \textsc{BatchNorm} layer before the last-layer linear classifier to reduce the training difficulty. Note that \textsc{BatchNorm} + a linear classifier is still a linear classifier during inference. Following the linear probing protocol, we train the \textsc{BatchNorm} and linear layers by a SGD optimizer with momentum=0.9, initial learning rate=0.01, and a 0.2 learning rate decay at $12^{th}$ and $16^{th}$ epochs. As to L2 weight decay,  we use the same searching space as in the fine-tuning. 
\end{itemize}

\subsubsection{Additional results}
\label{apx:swav_additional_exp}
Beside the SWAV \textsc{ImageNet} fine-tuning experiments in Figure \ref{fig:ssl_tf}, Figure \ref{fig:swav_ft_full} provides additional SWAV fine-tuning / two-stage fine-tuning results on  \textsc{naturalist18}, \textsc{Cifar100}, and \textsc{Cifar10} tasks. We give a {\plotlabel{[init]cat}} curve on the \textsc{ImageNet} task, but omit the curves on other tasks (\textsc{naturalist18}, \textsc{Cifar100}, and \textsc{Cifar10}) because they are computationally costly. 
\begin{figure}[ht]
    \centering
    \includegraphics[width=\textwidth]{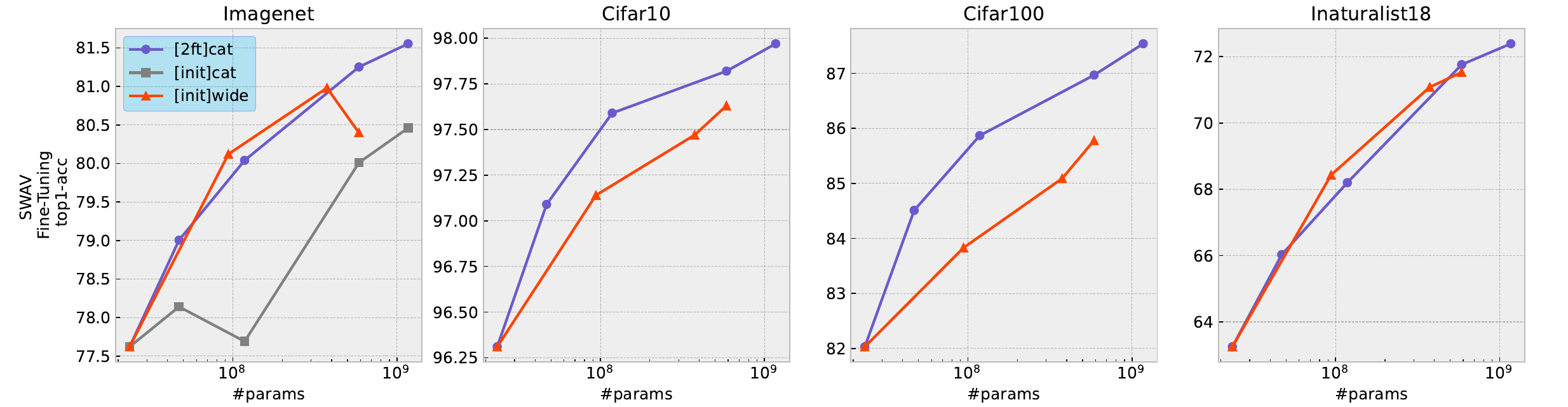}
    \caption{Fine-tuning performance of SWAV on \textsc{ImageNet}, \textsc{naturalist18}, \textsc{Cifar100}, and \textsc{Cifar10} tasks. SWAV is trained on unlabeled \textsc{ImageNet}. {\plotlabel{[2ft]cat}} and {\plotlabel{[init]cat} indicate our two-stage fine-tuning strategy and the normal fine-tuning strategy on $n$ concatenated networks. {\plotlabel{[init]wide}} refers to the normal fine-tuning strategy on wide networks, i.e. \textsc{resnet50}, \textsc{resnet50w2}, \textsc{resnet50w4}, and \textsc{resnet50w5}.    }}
    \label{fig:swav_ft_full}
\end{figure}

\subsection{SEER on \textsc{Instagram1B}}
SEER \citep{goyal2022vision} trains large \textsc{regnet\{32gf, 64gf, 128gf, 256gf, 10B\}} architectures on the \textsc{Instagram1B} dataset with 1 billion Instagram images, using the SWAV contrastive self-supervised learning algorithm.

\paragraph{Linear Probing:} Same as the linear probing settings in SWAV. 

\paragraph{Fine-tuning:} We use SEER checkpoints\footnote{\url{https://github.com/facebookresearch/vissl/tree/main/projects/SEER}} fine-tuned on \textsc{ImageNet} with $384 \times 384$ resolutions. It is fine-tuned on \textsc{ImageNet} for 15 epochs using SGD momentum 0.9, weight decay 1e-4, learning rate 0.04 and batch size 256. The learning rate is multiplied by 0.1 at $8^{th}$ and $12^{th}$ epochs.

\paragraph{Two-stage Fine-tuning:} We keep L2 weight decay 1e-4 the same as fine-tuning. Then keep the other settings the same as in SWAV.

\subsection{Additional experiment: \textsc{SimSiam} on \textsc{Cifar}}
\label{apx:simsiam_cifar}
\textsc{SimSiam} \cite{chen2020simsiam} is a non-contrastive self-supervised learning algorithm. In this section, we pre-train the networks using \textsc{SimSiam} on \textsc{Cifar10}, then transfer the learned representation by linear probing to   \textsc{Cifar10}, \textsc{Cifar100},  \textsc{Cifar10} with 1\% training examples, and \textsc{Cifar100} with 10\% training examples. 

\paragraph{\textsc{SimSiam} pre-training} Following \citet{chen2020simsiam} we pre-train \textsc{resnet18}, \textsc{resnet18w2}, \textsc{resnet18w4}, \textsc{2resnet18}, and \textsc{4resnet18} on \textsc{Cifar10} ($32 \times 32$ resolution) by \textsc{SimSiam} for $800$ epochs, using a SGD optimizer with momentum = $0.9$, initial learning rate = $0.06$, batch size = $512$, L2 weight decay = $5e-4$, and cosine learning rate scheduler. The data augmentations include \textsc{RandomResizedCrop} (crop scale in $[0.2, 1]$), \textsc{RandomHorizontalFlip}, \textsc{RandomGrayScale} ($p=0.2$), and a random applied \textsc{ColorJitter} ($0.4, 0.4, 0.4, 0.1$) with probability $0.8$. All images are normalized by $mean=(0.4914, 0.4822, 0.4465), std=(0.2023, 0.1994, 0.2010)$ before training.

\paragraph{\textsc{\synthdistill}} Since self-supervised learning tasks don't contain target labels as supervised learning, we apply knowledge distillation on representation directly. Specifically, we set $v_1, \dots v_n$ in Figure \ref{fig:distill} as Identity matrices. Then we distill $[\phi_1, \dots, \phi_n]$ into $\Phi$ by use a cosine loss: 

\begin{equation}
    \min_{\Phi, w_0, \dots, w_n}\sum_{i=0}^n\sum_x\bigg[  1-\cos\Big({\phi_i}(x)~, ~ w_i \circ \Phi(x) \Big)\bigg]
\end{equation}

\paragraph{Linear Probing:} Following again the settings of \citet{goyal2022vision}, the linear probing experiments
(on \textsc{Cifar100}, \textsc{Cifar10}, \textsc{Cifar100(1\%)} with 10\% training data, and \textsc{Cifar10(1\%)} with 1\% training data) adds a \textsc{BatchNorm} layer before the last-layer
linear classifier to reduce the hyper-parameter tuning difficulty. We use batch size = 256 for \textsc{Cifar100} and \textsc{Cifar10}, use batch size = 32 for corresponding sampled (10\%/1\%) version. Then we search initial learning rate from $\{0.1, 0.01\}$, L2 weight decay from \{1e-4, 5e-4, 1e-3, 5e-3\}. The learning rate is multiplied by 0.1 every 8 epochs during the total 28 training epochs. As to the optimizer, all experiments use a SGD Nesterov optimizer with momentum=0.9. 

\paragraph{Results}
Table \ref{tab:simsiam} shows the linear probing accuracy of \textsc{SimSiam} learned representation on various datasets and architectures. When linear probing on the same \textsc{Cifar10} dataset as training, the {\synthcat}$n$ method performs slightly better than width architectures (e.g. \textsc{resnet18w2} and \textsc{resnet18w4}). When comparing them on the  \textsc{Cifar100} dataset (OOD), however, {\synthcat}$n$ exceeds width architectures.

\begin{table}[ht]
    \centering
    \caption{Linear probing accuracy on \textsc{Cifar100}, \textsc{Cifar10}, \textsc{Cifar100(1\%)}, and \textsc{Cifar10(10\%)} tasks. The representation is learned on \textsc{Cifar10} by \textsc{SimSiam} algorithm. {\synthcat}$n$ concatenates $n$ learned representation before linear probing. {\synthdistill}$n$ distills $n$ learned representation into $\textsc{resnet18}$ before linear probing. \textsc{resnet18w}$n$ contains around $n^2$ parameters as {\textsc{resnet18}}.}
    \bigskip

    \begin{tabular}{cc|cc|cc}
    \toprule
    &  &\multicolumn{2}{c|}{Linear Probing (ID)} & \multicolumn{2}{c}{Linear Probing (OOD) } \\
    method & architecture & \textsc{Cifar10} & \textsc{Cifar10(1\%)} & \textsc{Cifar100} &  \textsc{Cifar100(10\%)} \\
    \midrule
    \textsc{SimSiam}  &\textsc{resnet18}       &  91.88   &87.60   &  55.29   &    42.93 \\
    \midrule
    \textsc{SimSiam}  &\textsc{resnet18w2}     &  92.88   & 88.95   &  59.41   &   45.39\\
    \textsc{SimSiam}  &\textsc{resnet18w4}     &  93.50   &90.45   &  59.28   &    44.98\\
    \textsc{SimSiam}  &\textsc{2resnet18}      &  91.62   &87.14   &  55.67   &    43.07\\
    \textsc{SimSiam}  &\textsc{4resnet18}      &  92.54   &85.65   &  64.42   &    49.65\\
    \midrule
    {\synthcat}2 & 2$\times$\textsc{resnet18}      &  92.94     & 88.32   & 59.40   &  46.06\\
    {\synthcat}4 & 4$\times$\textsc{resnet18}      &  93.42   & 88.81   &  63.06   &   47.48\\
    {\synthcat}5 & 5$\times$\textsc{resnet18}      &  93.67   & 88.78   & 63.71   &    48.31\\
   {\synthcat}10 & 10$\times$\textsc{resnet18}      &  93.75   & 88.65   &   66.19   &  49.90\\
   \midrule
{\synthdistill}2 & 2$\times$\textsc{resnet18}       &93.04	&88.59	&59.65	&45.10 \\
{\synthdistill}5 & 5$\times$\textsc{resnet18}       &93.02	&88.56	&60.79	&46.41 \\
{\synthdistill}10 & 10$\times$\textsc{resnet18}      &93.11	&88.72	&61.35	&46.75 \\
    \bottomrule
    \end{tabular}
    
    \label{tab:simsiam}
\end{table}

\subsection{Numerical results}
For reference, Tables \ref{tab:swav} and \ref{tab:seer} provide the numerical results for the linear probing, fine-tuning, and two-stage fine-tuning plots of Figure \ref{fig:ssl_tf}.
\begin{table}[ht]
    \centering
    \caption{Linear probing, fine-tuning, and two-stage fine-tuning performance of SWAV pre-trained representation and corresponding {\synthcat}$n$ representations. }
    \label{tab:swav}
    \bigskip

    \begin{tabular}{ccc|cccc|c|c}
\toprule
                &            &        &     \multicolumn{4}{c|}{ \footnotesize linear-probing}     & \footnotesize fine-tuning & \footnotesize two-stage ft \\   
\footnotesize method     & \footnotesize architecture   & \footnotesize params & \footnotesize \textsc{ ImageNet} & \footnotesize \textsc{Cifar10} & \footnotesize \textsc{Cifar100} & \footnotesize \textsc{Inat18} & \footnotesize \textsc{Imagenet} & \footnotesize \textsc{Imagenet} \\
\midrule
SWAV             & \textsc{resnet50}   & 23.5M & 74.30 & 91.83 & 76.85 & 42.35 & 77.62 & - \\
SWAV              & \textsc{resnet50w2} & 93.9M & 77.31 & 93.97 & 79.49 & 47.55 & 80.12 & - \\
SWAV              & \textsc{resnet50w4} & 375M & 77.48  & 94.29 &	80.51	& 44.13 & 80.98 & - \\
SWAV              &  \textsc{resnet50w5} & 586M  & 78.23 & 94.84 & 81.54 & 48.11 & 80.40 & - \\
\midrule
{\synthcat}2          & - & 47M  & 76.01 & 93.48 & 78.91 & 45.57 & 78.14 & 79.00 \\
{\synthcat}5          & - & 118M & 77.43 & 94.62 & 81.11 & 49.12 &  77.69     & 80.04 \\
{\synthcat}7    & - & 587M & 78.72 & 95.59 & 82.71 & 49.68 & 80.05  & 81.25 \\
{\synthcat}9 & - & 1170M& 78.89 & 95.76 & 83.16 & 50.61 & 80.46 & 81.55 \\
\bottomrule
    \end{tabular}
    
\end{table}

\begin{table}[ht]
    \centering
    \caption{Linear probing, fine-tuning, and two-stage fine-tuning performance of SEER pre-trained representation and corresponding {\synthcat}$n$ representations. }
    \bigskip
    \label{tab:seer}

    \begin{tabular}{ccc|cccc|c|c}
\toprule
                &            &        &     \multicolumn{4}{c|}{ \footnotesize linear-probing}     & \footnotesize fine-tuning & \footnotesize two-stage ft \\   
\footnotesize method     & \footnotesize architecture   & \footnotesize params & \footnotesize \textsc{ ImageNet} & \footnotesize \textsc{Cifar10} & \footnotesize \textsc{Cifar100} & \footnotesize \textsc{Inat18} & \footnotesize \makecell{\textsc{Imagenet} \\ (384px)} & \footnotesize \makecell{\textsc{Imagenet} \\ (384px)} \\
\midrule
SEER              & \textsc{regnet32gf}   & 141M & 73.4   &   89.94   &   71.53   & 39.10     &   83.4   &   - \\
SEER              & \textsc{regnet64gf}   & 276M & 74.9   &   90.90   &   73.78   &  42.69    &   84.0   &   - \\ 
SEER              & \textsc{regnet128gf}  & 637M  & 75.9   &   91.37   &   74.75   &  43.51    &   84.5   &   - \\
SEER              & \textsc{regnet256gf}  & 1270M  & 77.5   &   92.16   &   74.93   &  46.91    &   85.2   &   - \\
\midrule
{\synthcat}2             & - & 418M  & 76.0   &   92.16   &   75.65   &  45.36    &  -    &   84.5 \\
{\synthcat}3          & - & 1060M  & 77.3   &   93.15   &   77.26   &  47.18    &  -    &   85.1 \\
{\synthcat}4      & - & 2330M & 78.3   &   93.59   &   78.80   & 48.68     &    -  &   85.5 \\
\bottomrule
    \end{tabular}
    
\end{table}

\section{meta-learning / few-shots learning}
\label{apx:meta-learning}
\subsection{Datasets}
\textbf{\textsc{Cub}} \citep{wah2011caltech} dataset contains $11,788$ images of 200 birds classes, 100 classes ($5,994$ images) for training and 100 classes ($5,794$ images) for testing. 

\textbf{\textsc{MiniImageNet}} \citep{matchingnet} dataset contains $60,000$ images of 100 classes with 600 images per class, 64 classes for training, 36 classes for testing. 

\subsection{\textsc{Baseline} and \textsc{Baseline++} experiment Settings}
\label{apx:meta_baseline_pretrain}
For \textsc{Baseline} and \textsc{Baseline++} experiments, following \citet{closelookatfewshot},  we use \textsc{RandomSizedCrop}, \textsc{ImageJitter}(0.4, 0.4, 0.4), and \textsc{HorizontalFlip} augmentations, as well as a image normalization $mean=(0.485, 0.456, 0.406)$, $std=(0.229, 0.224, 0.225)$. Then use an \textsc{Adam} optimizer with learning rate = 0.001, batch size = 16, input image size = $224\times224$. Finally, train \textsc{resnet18} on \textsc{Cub} and \textsc{MiniImageNet} datasets for 200 and 400 epochs, respectively. We further tune L2 weight decay from \{0, 1e-5, 1e-4, 1e-3, 1e-2\} and choose 1e-4 for \textsc{Cub}, 1e-5 for \textsc{MiniImageNet} experiments. Compared with the \textsc{Baseline} and \textsc{Baseline++} performance reported by \citet{closelookatfewshot} (table A5), this L2 weight decay tuning process provides $\sim5\%$ and $\sim1\%$ improvement on \textsc{miniImagenet} 5way-1shot and  5way-5shot, respectively. In this work, we use this stronger setting in baseline methods. 

As to the few-shots learning evaluation, following \citet{closelookatfewshot}, we scale images by a factor of 1.15, \textsc{CenterCrop}, and normalization. Then randomly sample 1 or 5 images from 5 random classes from the test set (5way-1shot and 5way-5shot). Finally, train a linear classifier on top of the learned representation with a SGD optimizer, momentum = 0.9, dampening = 0.9, learning rate = 0.1, L2 weight decay = 1e-3, batch size = 4, and epochs = 100. We take the average of 600 such evaluation processes as the test score.

 The \textsc{Baseline} and \textsc{Baseline++} results in Figure \ref{fig:few_shots} report the mean of five runs with different training and evaluating seeds.

\paragraph{Implementation details of the cosine classifier}
Here we summarize the technical details of the cosine classifier implementation used in this work which follows \citet{closelookatfewshot}\footnote{%
\url{https://github.com/wyharveychen/CloserLookFewShot/blob/master/backbone.py##L22}}. 

Denote the representation vector as $z$. The cosine classifier calculates the $i^{th}$ element of logits by: 
\begin{equation}
     h_i = g_i \frac{\left<u_i,z\right>}{||u_i||||z||}
\end{equation}
where $u_i$ is a vector with the same dimension of $z$, $g_i$ is a scalar, $h_i$ is $i^{th}$ element of logits $h$.

Then minimize the cross entropy loss between the target label $y$ and softmax output $s(h)$ by updating $w$ and $g$: $\min_{w,g} \mathcal{L}_{ce}(y, s(h))$.

\subsection{{\synthcat} and {\synthdistill} experiment settings}

For {\synthcat}, we concatenate $n$ representation separately trained by either \textsc{Baseline} or \textsc{Baseline++} as the settings above.  For {\synthdistill}, we use the same multi-head architecture as figure \ref{fig:distill} together with a cross-entropy loss function:
\begin{equation}
\label{eq:ce_kl_distill}
    \min_{\Phi, w_0, \dots, w_n}\sum_{i=0}^n\sum_x\bigg[ (1-\alpha) \mathcal{L}_{ce}\Big(s\big(w_i \circ \Phi(x)\big), y \Big) + \alpha\tau^2\mathcal{L}_{kl}\Big(s_{\tau}\big({v_i} \circ {\phi_i}(x)\big) ~||~ w_i \circ \Phi(x) \Big)\bigg]
\end{equation}, 
where $\mathcal{L}_{ce}$ indicates a cross-entropy loss, $\alpha$ is a trade-off parameter between cross-entropy loss and kl-divergence loss. We set L2 weight decay = 0,  $\tau=10$, search $\alpha \in \{0.8, 0.9, 1\}$, and keep the other hyper-parameters as Appendix \ref{apx:meta_baseline_pretrain}. We find the impact of $\alpha$ is limited in both \textsc{cub} ($\leq 1\%$) and \textsc{MiniImageNet} ($ \leq 0.5\%$) tasks. 

\subsection{Snapshots experiment settings}
In this section, we apply {\synthcat} and {\synthdistill} on 5 snapshots sampled from one training episode (called \textsc{cat5-s} and \textsc{distill5-s}, respectively). We train $\textsc{cub}$ and $\textsc{MiniImageNet}$ respectively for 1000 and 1200 epochs by naive SGD optimizer with a relevant large learning rate 0.8. Then we sample 5 snapshots, $\{200^{th}, 400^{th}, 600^{th}, 800^{th}, 1000^{th}\}$ and  $\{400^{th}, 600^{th}, 800^{th}, 1000^{th}, 1200^{th}\}$, for $\textsc{cub}$ and $\textsc{MiniImageNet}$, respectively. The other hyper-parameters are the same as Appendix \ref{apx:meta_baseline_pretrain}.

\subsection{More experimental results}
Table \ref{tab:few_shot_synt_cat} provides the exact number in Figure \ref{fig:few_shots}, as well as additional {\synthcat}$n$ and {\synthdistill}$n$ few-shots learning results with a linear classifier (The orange and gray bars in figure \ref{fig:few_shots} report the few-shots learning performance with a cosine classifier). 

{Table \ref{tab:few_shot_synt_cat_snapshot} provides more \synthcat5\textsc{-s} and \synthdistill5\textsc{-s} results with either a linear classifier or a cosine-based classifier.}
\begin{table}[t]
    \caption{Few-shots learning performance on \textsc{cub} and \textsc{miniImagenet}. The \textsc{cat5-s} and \textsc{distill5-s} results were obtained using five snapshots taken during a single training episode with a relatively high step size (0.8, SGD). The best snapshot performances are also reported. Standard deviations over five repeats are reported.
    }
    \label{tab:few_shot_synt_cat_snapshot}
    \par\bigskip\centering
    \begin{tabular}{ccc |cc |cc}
  \toprule
 &           &            & \multicolumn{2}{c|}{\textsc{cub}}         &     \multicolumn{2}{c}{\textsc{miniImagenet} } \\
                & architecture & classifier       &   5way 1shot    &    5way 5shot &   5way 1shot    &    5way 5shot    \\
    \midrule
    best snapshot & \textsc{resnet18} & linear & 59.70$\pm$1.38 &	81.35$\pm$0.79 & 52.79$\pm$0.92 & 75.18$\pm$0.57 \\ 
    \synthcat5\textsc{-s} & \textsc{5$\times$resnet18} & linear& 72.62$\pm$0.98 & 86.56$\pm$0.82 & 61.91$\pm$0.37 &	81.06$\pm$0.14 \\
\synthdistill5\textsc{-s} & \textsc{resnet18} & linear& 68.4$\pm$0.5 & 87.2$\pm$0.4 &
    59.9$\pm$0.5 & 80.8$\pm$0.4  \\
    \midrule
    best snapshot & \textsc{resnet18} & cosine & 65.59$\pm$0.87 & 81.81$\pm$0.50 & 55.67$\pm$0.48 &	75.48$\pm$0.46 \\
    \synthcat5\textsc{-s} & \textsc{5$\times$resnet18} & cosine & 73.66$\pm$0.82 & 87.25$\pm$0.77 & 62.94$\pm$0.51 &	81.05$\pm$0.16 \\
\synthdistill5\textsc{-s} & \textsc{resnet18} & cosine &
     75.2$\pm$0.8 & 88.6$\pm$0.4 & 62.0$\pm$0.5 & 81.0$\pm$0.3 \\
    \bottomrule
    \end{tabular}
\end{table}
\begin{table}[ht]
    \centering
    \caption{Few-shot learning performance on \textsc{cub} and \textsc{miniImagenet} dataset with either a linear classifier or cosine-distance based classifier. Standard deviations over five repeats are reported. }
    \bigskip

    \label{tab:few_shot_synt_cat}
    \begin{tabular}{ccc |cc |cc}
  \toprule
 &           &            & \multicolumn{2}{c|}{\textsc{cub}}         &     \multicolumn{2}{c}{\textsc{miniImagenet} } \\
                & architecture & classifier       &   5way 1shot    &    5way 5shot &   5way 1shot    &    5way 5shot    \\
\midrule  
supervised 	    &  \textsc{resnet18} & linear &          63.37$\pm$1.66 & 83.47$\pm$1.23	&	55.20$\pm$0.68 & 76.52$\pm$0.42	\\
{\synthcat}2	&  $2\times$\textsc{resnet18} &   linear &66.25$\pm$0.85 & 85.50$\pm$0.34	&	57.30$\pm$0.31 & 78.42$\pm$0.17	\\
{\synthcat}5	&  $5\times$\textsc{resnet18} &   linear &67.00$\pm$0.18 & 86.80$\pm$0.10	&	58.40$\pm$0.25 & 79.59$\pm$0.17	\\
{\synthdistill}2&  \textsc{resnet18} &   linear&69.93$\pm$0.74 & 87.72$\pm$0.31	&	58.99$\pm$0.32 & 79.73$\pm$0.21	\\
{\synthdistill}5&   \textsc{resnet18} &  linear&70.99$\pm$0.31 & 88.52$\pm$0.14	&	59.66$\pm$0.59 & 80.53$\pm$0.27	\\
\midrule
supervised 	    &  \textsc{resnet18} &   cosine&69.19$\pm$0.88 & 84.41$\pm$0.49	&	57.47$\pm$0.45 & 76.47$\pm$0.27	\\
{\synthcat}2	&  $2\times$\textsc{resnet18} &   cosine&72.87$\pm$0.43 & 86.82$\pm$0.17	&	60.69$\pm$0.24 & 79.29$\pm$0.23	\\
{\synthcat}5	&  $5\times$\textsc{resnet18} &   cosine&76.23$\pm$0.55 & 88.87$\pm$0.40	&	63.63$\pm$0.23 & 81.22$\pm$0.17	\\
{\synthdistill}2&  \textsc{resnet18} &   cosine &  74.81$\pm$0.45 & 88.14$\pm$0.40	&	61.95$\pm$0.11 & 80.79$\pm$0.26	\\
{\synthdistill}5&  \textsc{resnet18} &   cosine & 76.20$\pm$0.39 & 89.18$\pm$0.24	&	62.89$\pm$0.38 & 81.49$\pm$0.26	\\
\bottomrule
    \end{tabular}
\end{table}

\subsection{Comparison with conditional Meta-learning approaches}
In order to address heterogeneous distributions over tasks, the conditional meta-Learning approaches \citet{wang2020structured,denevi2022conditional,rusu2018meta} adapt a part of model parameters conditioning on the target task, while freeze the other model parameters that are pre-trained as a feature extractor. 

The results presented in \citet{wang2020structured,denevi2022conditional,rusu2018meta} already allow us to make some elementary comparisons: \citet{denevi2022conditional} is derived from \citet{wang2020structured}. In practice, \citet{wang2020structured} reuses the pre-trained frozen feature extractor (\textsc{WRN-28-10}) from \citet{rusu2018meta}. Table \ref{tab:miniimagenet_conditional_metalearning} below shows the performance of these conditional meta-learning methods and our \textsc{Distill5} on the \textsc{miniImagenet} few-shot learning task. The first 3 rows are copied from \citet{wang2020structured} (marked by *). Despite the fact that the backbone in \citet{wang2020structured,rusu2018meta} (\textsc{WRN-28-10}) is wider and deeper than the backbone (\textsc{resnet18}) used in our paper, \textsc{Distill5} still outperforms both \citet{wang2020structured} and \citet{rusu2018meta}. Other relevant details are summarized in table \ref{tab:backbone_pretraining_details}.

If our goal were to present state-of-the-art results exploiting diverse features, a more systematic comparison would be needed. however it is not clear that these results say a lot about how optimization constructs and (prematurely) prunes features. The conditional meta-learning addresses an orthogonal problem but does not seem to fix the premature feature pruning issue. Please not that the message of our paper is that a single optimization run — which is what most people are doing these days - prematurely prunes its representations, missing opportunities to produce the richer representations that benefit out-of-distribution scenarios.

\begin{table}[]
    \centering
        \caption{\textsc{miniImageNet} few-shots learning comparison between \textsc{Distill5} and conditional meta-learning approaches. The first three rows are copied from corresponding papers (marked by *).}
    \label{tab:miniimagenet_conditional_metalearning}
    \resizebox{0.8\textwidth}{!}{
    \begin{tabular}{c|cc}
    \toprule
      &  miniImageNet 5way-1shots &	miniImageNet 5way-5shots \\
      \midrule
LEO \cite{rusu2018meta} &	61.76$\pm$0.08*	&77.59$\pm$0.12* \\
LEO(local) \cite{rusu2018meta} &	60.37$\pm$0.74*	&75.36$\pm$0.44* \\
TASML \cite{wang2020structured}&	62.04$\pm$0.52*	&78.22$\pm$0.47* \\
Distill5 (our)&	62.89$\pm$0.38&	81.49$\pm$0.26 \\
\bottomrule
    \end{tabular}}

\end{table}

\begin{table}[]
    \centering
        \caption{Backbone pretraining details. Note that LEO only keeps the first 21 layers (21.7M parameters) after pretraining \textsc{WRN-28-10} (Wide residual network). But it is still twice the time larger than \textsc{resnet18}.}
    \label{tab:backbone_pretraining_details}
    \resizebox{0.6\textwidth}{!}{
    \begin{tabular}{c|cc}
    \toprule
& Our backbone  &	\makecell{ LEO backbone\\ \cite{rusu2018meta}\\\cite{wang2020structured} }\\
\midrule
Architecture &	\textsc{resnet18} &	\textsc{WRN-28-10} \\ 
Parameters	& 11.4M &	36.5M \\
L2 weight decay	& \checkmark	& \checkmark \\
Learning rate scheduler & 	$\times$	& \checkmark \\
Data augmentation (color)& 	\checkmark	& \checkmark \\
Data augmentation (scale)& 	\checkmark	& \checkmark \\
Data augmentation (deformation)& 	$\times$	& \checkmark \\
\bottomrule
    \end{tabular}}

\end{table}

\section{Out-of-distribution learning}
\label{apx:ood}
Following \citet{zhang2022rich}, we use the \textsc{Camelyon17} \citep{koh2021wilds} task to showcase the $\synthcat$ and $\synthdistill$ constructed (rich) representation in out-of-distribution learning scenario. The first row of table \ref{tab:camelyon17_synt_cat} is copied from \citet{zhang2022rich}. The rest results use a frozen pre-trained representation, either by concatenating $n$ ERM pre-trained representations ({\synthcat}$n$), distilling of $n$ ERM pre-trained representations ({\synthdistill}$n$), or RFC constructed representations (RFC2). Then train a linear classifier on top of the representation by vREx or ERM algorithms. 

For the vREx algorithm, we search the penalty weights from \{0.5, 1, 5, 10, 50, 100\}. For {\synthdistill}$n$ representations in the \textsc{Camelyon17} task, we follow Algorithm 2 in \citet{zhang2022rich}, but use a slightly different dataset balance trick in the loss function (\citet{zhang2022rich} Algorithm 2 line 13-14). We instead balance two kinds of examples: one shares the same predictions on all ERM pre-trained models, and one doesn't. We keep other settings to be the same as \citet{zhang2022rich}\footnote{\url{https://github.com/TjuJianyu/RFC}}.

\section{MAML-IRM resembles vREx+Fish}
\label{apdix:mamlirm=vrex+gm}

We omit the MAML-IRM method in our experiments because we can show that minimizing its cost amounts to minimizing a mixture of the vREx and Fish costs.

Notations: 
\begin{itemize*}
    \item $\mathcal{E}$: indicates a set of environments.
\item $\theta$: indicates the model parameters.
\item $L_i(\theta)$: indicates an ERM loss (e.g. MSE, cross-entropy) of a model parameterized by $\theta$ on environments $i$.
\item $\bar{g_i} = L_i^{'}(\theta)$: is the gradients of $L_i(\theta)$.
\item $\bar{H_i} = L_i^{''}(\theta)$: is the Hessian of $L_i(\theta)$.
\end{itemize*}

Let $U_i(\theta) = \theta - \alpha L_i^{'}(\theta)$ denote the updated parameters after performing a SGD iteration on environments $i$.
The MAML-IRM loss can be expressed as:
\begin{align}
    L_\text{maml-irm} = \mathbb{E}_s [L_j(U_i(\theta))] + \lambda \sqrt{Var_s[L_j(U_i(\theta))]}
\end{align}
where the notation $\mathbb{E}_s$ and $Var_s$ respectively denote the average and the variance with respect to all pairs of distinct environment $s = \{i,j | i\in \mathcal{E}, j \in \mathcal{E}, i \neq j\}$, and where $\lambda$ is a hyper-parameter.

According to~\cite{nichol2018first}, the gradients of the first term is:
\begin{align}
    \frac{\partial (\mathbb{E}_s [L_j(U_i(\theta)))]}{ \partial \theta} &= \mathbb{E}_s [\bar{g_j} - 2\alpha \bar{H_i}\bar{g_j}] + O(\alpha^2)
\end{align}
Note that $\mathbb{E}_s [-2\bar{H_i}\bar{g_j}] = \mathbb{E}_s [-\frac{\partial\left\langle  g_i, g_j \right\rangle}{\partial \theta}]$ is in fact the gradients of $-\left\langle  g_i, g_j \right\rangle$, the Fish penalty.  

We now turn out attention to the second term $\sqrt{Var_s[L_j(U_i(\theta))]}$. Expanding $L_j(U_i(\theta))$ with a Taylor series gives:
\begin{align}
    L_j(U_i(\theta)) &= L_j(\theta) + \left\langle L_j^{'}(\theta),(U_i(\theta) - \theta)) \right\rangle + O(\alpha^2) \\
    &= L_j(\theta) - \alpha \left\langle L_i^{'}(\theta),L_j^{'}(\theta)\right\rangle + O(\alpha^2) \quad\quad\quad\quad (\longleftarrow U_i(\theta) = \theta - \alpha L_i^{'}(\theta)) \\
    &=  L_j(\theta) - \alpha\left\langle \bar{g_i},\bar{g_j}\right\rangle + O(\alpha^2)
\end{align}
Therefore
\begin{align}
    Var_s(L_j(U_i(\theta)))&= Var_s[L_j(\theta) - \alpha \left\langle \bar{g_i},\bar{g_j}\right\rangle]  + O(\alpha^2) \\
    &=Var_s[L_j(\theta)] + \alpha^2 Var_s[\left\langle \bar{g_i},\bar{g_j}\right\rangle] - 2\alpha \text{Cov}_s[L_j(\theta), \left\langle \bar{g_i},\bar{g_j}\right\rangle] + O(\alpha^2)  \notag \\
    &= Var_s[L_j(\theta)] - 2\alpha \text{Cov}_s[L_j(\theta), \left\langle \bar{g_i},\bar{g_j}\right\rangle] + O(\alpha^2)  \notag\\
    &=  Var_s[L_j(\theta)] - 2\alpha \{ \mathbb{E}_s[L_j(\theta)\left\langle \bar{g_i},\bar{g_j}\right\rangle] - \mathbb{E}_s [L_j(\theta)]\mathbb{E}_s[\left\langle \bar{g_i},\bar{g_j}\right\rangle]\} + O(\alpha^2) \notag  \notag \\
    &=  Var_s[L_j(\theta)] - 2\alpha  \mathbb{E}_s[(\frac{L_i(\theta)+L_j(\theta)}{2}-\mathbb{E}[L(\theta)])\left\langle \bar{g_i},\bar{g_j}\right\rangle] + O(\alpha^2)  \notag
\end{align}
The first term of this expression, $Var_s[L_j(\theta)]$, penalizes a high variance of the loss across environments. It is equal to the vREx penalty. The second term, $- 2\alpha \mathbb{E}_s[(\frac{L_i(\theta)+L_j(\theta)}{2}-\mathbb{E}[L(\theta)])\left\langle\bar{g_i},\bar{g_j}\right\rangle]$ is a weighted average of $\left\langle g_i,g_j\right\rangle$, that is a smoothed Fish penalty. 

In conclusion, optimizing the MAML-IRM cost amounts to optimizing a $\lambda$ controlled mixture of the vREx and Fish costs.

\section{GroupDRO interpolates environments while vREx extrapolates.}
\label{apdix:groupdro_vrex_inter_extra}
The vREx objective function can be expressed as: 
\begin{align}
    L_\text{vrex} = \mathbb{E}_{e\in\mathcal{E}} (L_e) + \lambda Var_{e\in\mathcal{E}}(L_e)
\end{align}

The GroupDRO objective function is a mixture of the per-environment costs $L_e$ with positive coefficients:
\begin{align}
    L_\text{groupDRO} = \mathbb{E}_{e\in\mathcal{E}} (p_e L_e)
\end{align}
where the adjustable mixture coefficients $p_e\geq0$, $\sum_{e\in\mathcal{E}}p_e = 1$, are treated as constaants for computing  the gradients  $\frac{\partial L_e}{\theta}$.  

The gradient of these two cost functions are:
\begin{align}
    \frac{\partial L_\text{vrex}}{\theta} &= \mathbb{E}_{e\in\mathcal{E}}([2\lambda(L_e - \mathbb{E}_iL_i) + 1]g_e) \\
    \frac{\partial L_\text{groupDRO}}{\theta} &= \mathbb{E}_{e\in\mathcal{E}}(p_eg_e)
\end{align}
where $g_e= \frac{\partial L_e}{\theta}$ is the gradients of network weights $\theta$ on environment $e$. 

Because the $p_e$ mixture coefficients are always positive, it is easy to see that GroupDRO follows a direction aligned with a convex combination of the per-environment gradients. In contrast, vREx can follow 
a direction that is outside this convex hull because the coefficients $\mathbb{E}_iL_i) + 1]$ can be positive or negative).

\section{Loss landscape of OoD methods}
\label{apdix:lowdim_losslandscape}

Here we visualize the loss landscape of some of OoD penalties on a synthetic two-dimensional problem, TwoBits, which was introduced by \citep{Kamath2021} as a simplified version of the \textsc{coloredMNIST}. 
TwoBits is a binary classification problem $Y=\pm1$ with two binary inputs $X_1=\pm1$ and $X_2=\pm1$ distributed as follows:
\begin{align*}
    Y ~ &\sim ~ \text{Rademacher}(0.5) \\
    X_1 &\sim ~ Y \cdot \text{Rademacher}(\alpha_e) \\
    X_2 &\sim ~ Y \cdot \text{Rademacher}(\beta_e)
\end{align*}
where $\text{Rademacher}(\alpha)$ denotes the law of a random variable taking value $-1$ with probability $\alpha$ and taking $+1$ probability $1{-}\alpha$.
The training algorithms observe two training environments, $(\alpha_e, \beta_e) \in \{(0.1, 0.1), (0.1, 0.3)\}$. The four input patterns $(X_1,X_2)$ are represented by four points $\{\Psi(1,1),\Psi(-1,-1),\Psi(1,-1),\Psi(-1,1)\}$ in the representation space where $\Psi$ can represent any network architectures with numerical outputs. Following \citep{Kamath2021}, we use a mean squared loss and focus on the symmetric case $\Psi(-x)=-\Psi(x)$. The representation space can therefore be displayed with only two dimensions, $\Psi(1,-1)=-\Psi(-1,1)$ and $\Psi(1,1)=-\Psi(-1,-1)$. 

Figure \ref{fig:isolatedspace} shows a heat map of the penalty terms of three OoD methods (IRMv1, vREx, SD) as a function of the chosen representation. The stars denote three solutions: (a) the Invariant solution which only uses feature $X_1$ because this is the feature whose correlation with the label remains the same across the training environments, (b) the ERM solution which uses both features, and (c) a random feature initialization with small variance for which the
representations $\Psi(1,1), \Psi(1,-1)$ are close to zero.  

All three OoD methods have low penalties when the $\Psi(1,1), \Psi(1,-1)$ are close to zero. This explains why random initialization performs so poorly with these methods. In contrast, pretraining with ERM leads to a new initialization point that is away from the origin and close to the ERM solution. The OoD performance then depends on the existence of a good optimization path between this initialization and the Invariant solution. Alas Figure~\ref{fig:isolatedspace} shows a lot of optimization difficulties such as finding a solution that lies at the bottom of an elongated ravine (ill-conditioning).  In conclusion, the impact of the number of ERM pretraining epochs is essentially unpredictable.  

\begin{figure}
    \centering
    \includegraphics[width=\textwidth]{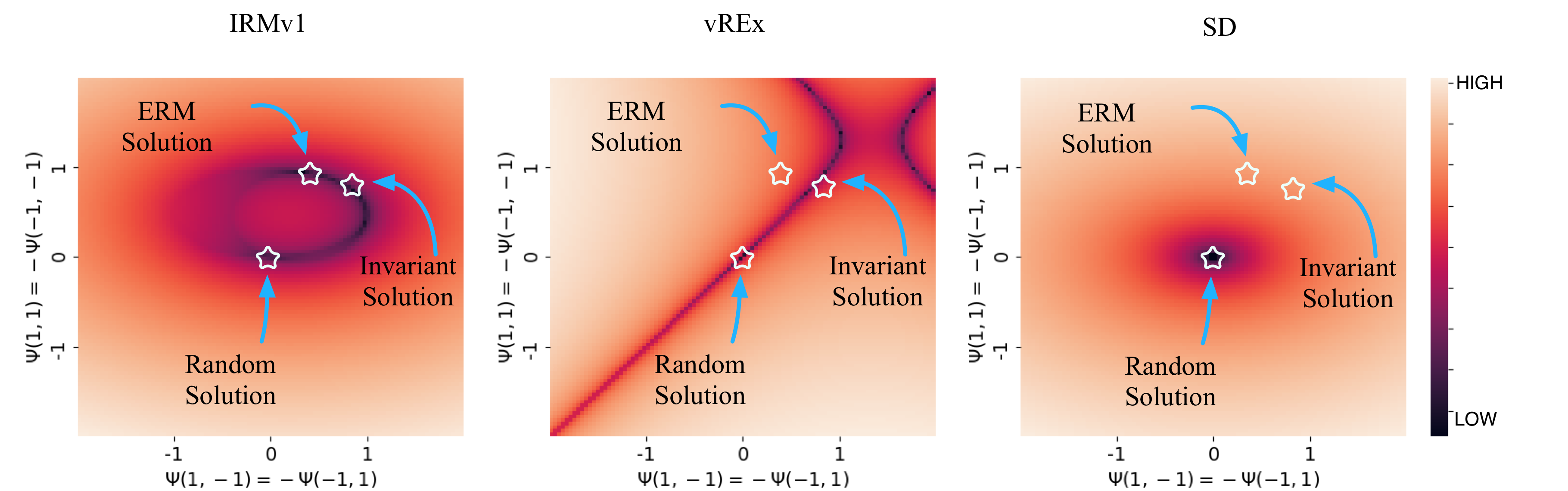}
    \caption{The IRMv1, vREx, and SD landscapes show a challenging non-convex landscape in the two-dimensional TwoBits problem. In particular, the path between the ERM solution and the invariant solution often involves climbing the loss landscape.}
    \label{fig:isolatedspace}
\end{figure}

\section{Experimental details for the ColoredMNIST experiments}
\label{apdix:train_details_colorminst}

The ColoredMNIST experiments sweep penalization weights from $\{10, 50, 100, 500, 1000\}$ for the SD method, from $\{1000, 5000, 10000, 50000, 100000\}$ for the other methods, sweep ERM pretraining epochs $\{50,100,150,200,250\}$ for ERM initialization. Then select hyper-parameters by peeking at the test set performance.\footnote{The small size of the \textsc{ColoredMNIST} makes this hard to avoid. Tuning the hyper-parameters using the testing set favors in fact the ERM initialization because the test performance depends strongly on the number of pre-training epochs (Figure~\ref{fig:importance_of_rep}).} All ColoredMNIST experiments use the same 2-hidden-layers MLP network architecture (390 hidden neurons), Adam optimizer, learning rate=$0.0005$, $L_2$ weights regularization=$0.0011$ and binary cross-entropy objective function.

We use the original \textsc{ColoredMNIST} dataset \cite{irm} with two training environments $(0.25, 0.1), (0.25, 0.2)$. The target label correlates with the invariant feature (the  digit shape) with a probability 0.75. The sirious feature (color) correlates with the target label with a probability 0.8 and 0.9, respectively. Each training environment contains $25000$ images where the size of each image is $2\times14\times14$. For all \textsc{ColoredMNIST} experiments, we use a fully connected neural network with 3 layers (392 (input dim) $\times390\times390\times1$), trained with the Adam optimizer with learning rate 0.0005. We use a L2 weights regularization with parameter 0.0001 for \textsc{InverseColoredMNIST} tasks and 0.0001 in the regular \textsc{ColoredMNIST} tasks. 
For the CLOvE method, we use a Laplacian kernel $k(r,r_0)=exp(\frac{-|r-r_0|}{0.4})$ \cite{kumar2018trainable} with mini-batch size 512. All other methods train using full batches. For the ERM baseline and for computing the oracle performance, we search the L2 regularization parameter in $\{0.0001, 0.0005, 0.001, 0.005, 0.01\}$. We run each experiment 10 times to get the standard deviation. 

\subsection{Hyper-parameter searching space}
Table \ref{tab:penalty-weights-searching-space} shows the penalty weights searching space for all OoD  methods in the \textsc{ColoredMNIST} experiments. Table \ref{tab:training_epoch_search_space} shows the training epochs searching space for different OoD  methods and network initialization/representation on the \textsc{ColoredMNIST} dataset.

\begin{table}[ht]
  \centering
  \caption{Penalty weight search space for both the \textsc{ColoredMNIST} and \textsc{InverseColoredMNIST} datasets. }
  \label{tab:penalty-weights-searching-space}
  \resizebox{0.6\textwidth}{!}{
  \begin{tabular}{l|c|c}
    \toprule
                     &\textsc{ColoredMNIST}     & \textsc{InverseColoredMNIST}  \\
    \midrule
    IRMv1            & $10000\times\{0.1, 0.5, 1, 5, 10\}$& $10000\times\{0.1, 0.5, 1, 5, 10\}$ \\
    vREx             & $10000\times\{0.1, 0.5, 1, 5, 10\}$& $10000\times\{0.1, 0.5, 1, 5, 10\}$ \\
    IGA              & $10000\times\{0.1, 0.5, 1, 5, 10\}$& $10000\times\{0.1, 0.5, 1, 5, 10\}$ \\
    CLOvE            & $10000\times\{0.1, 0.5, 1, 5, 10\}$& $10\times\{0.1, 0.5, 1, 5, 10\}$ \\
    Fishr            & $10000\times\{0.1, 0.5, 1, 5, 10\}$&  $10000\times\{0.1, 0.5, 1, 5, 10\}$\\
     SD              & $100  \times\{0.1, 0.5, 1, 5, 10\}$& $\{0.05, 0.1, 0.5, 1,5\}$\\
    RSC              & $(0.995,0.98) \times \{0.95,0.97,0.98,0.99,1\}$& - \\
    LfF              & $\{0.1, 0.2, 0.3, 0.4, 0.5\}$& - \\
    Fish             & $0.001 \times\{0.1, 0.5, 1, 5, 10\}$ &- \\
    \bottomrule
  \end{tabular}}
\end{table}

\begin{table}[ht]
    \centering
        \caption{The number of training epochs search space for the \textsc{ColoredMNIST} dataset, with $i \in [0,24]$.}
    \label{tab:training_epoch_search_space}
    \resizebox{0.325\textwidth}{!}{
    \begin{tabular}{l|c|c|c}
    \toprule
            & Rand/ERM & Bonsai & Bonsai-cf\\
    \midrule
         IRMv1  &    $i\times 20$ & $i\times2$  &   $i\times 125$  \\
         vREx   &    $i\times 20$ & $i\times2$  &   $i\times 20$  \\
         IGA   &    $i\times 20$ & $i\times1$  &   $i\times 20$  \\
         CLOvE   &    $i\times 30$ & $i\times1$  &   $i\times 20$  \\
         Fishr   &    $i\times 20$ & $i\times1$  &   $i\times 20$  \\
         SD   &    $i\times 20$ & $i\times1$  &   $i\times 20$  \\
         RSC   &    $i\times 1$ &  -  &   -  \\
         LfF &$i\times 20$ & - & - \\
         Fish &$i\times 20$ & - & -\\
    \bottomrule
    
    \end{tabular}}
\end{table}

\subsection{Bonsai algorithm}
For all \textsc{ColoredMNIST} experiments, we use a 2-rounds Bonsai \emph{discovery phase} trained with respectively 50 and 500 epochs.  Then we train 500 epochs for the distillation network of the Bonsai \emph{synthesis phase}. For the \textsc{InverseColoredMNIST} experiments, we again use a 2-rounds Bonsai \emph{discovery phase} trained with respectively 150 and 400 epochs. We choose these training epochs because they can maximize the IID validation performance during each round.

\subsection{PI training}
We use the original implementation from PI \cite{Bao2021}. Because the PI algorithm is closely related to the \emph{discovery phase}, we use the same hyper-parameters and settings.

\section{Experimental details for the \textsc{Camelyon17} experiments}
\label{apdix:camelyon17}

We strictly follow the implementation of the \textsc{Camelyon17} task in the WILDS benchmark \citep{wilds2021}. For the results presented in section \ref{sec:hyper-paraemeter-model-selection-difficulty}, we additionally search the penalty weights in the set $\{0.5,0.75, 1, 2.5, 5, 7.5, 10, 25, 50, 75, 100, 250, 500, 750, 1000\}$ for IRMv1 and vREx methods, and the set $\{0.5,0.75, 1, 2.5, 5, 7.5, 10, 25, 50, 75, 100, 250, 500, 750, 1000\}\times10^{-3}$. The CLOvE method require a kernel function, we choose the Laplacian kernel $k(r,r_0) = exp(\frac{-|r-r_0|}{l})$  \cite{kumar2018trainable} where $l$ is a positive scalar. For the CLOvE baseline with an ERM pretrained initialization (the fourth row of table \ref{tab:camelyon_full_results}), we test the scalar $l\in \{0.1, 0.2\}$ and choose the better one $l = 0.2$. For the other CLOvE experiments on \textsc{Camelyon17}, we choose $l=0.1$.

We train the \emph{synthesis phase} 20 epochs and the other methods/phase 10 epochs. Hyper-parameter tuning strictly follows the IID and OoD tuning process described in the WILDS task. We use a L2 weights regularization $1e-6$ during the \emph{synthesis phase} to help it get a lower training loss on the pseudo-labels. During any further training that updates the weights of the learned representation, we keep the L2 weights regularization to be the same as $1e-6$. Otherwise, a stronger L2 weights regularization will destroy the learned representation. We also tried other L2 regularization weights in $\{1e-2, 1e-4, 1e-6\}$. Table \ref{tab:synthesis_quality} shows the synthesis quality with different (\emph{synthesis phase}) L2 weights decay. Two smaller L2 weights decay hyper-parameters $\{1e-4, 1e-6\}$ can arrive at a good synthesis quality. The corresponding test performances on the frozen representation ``2-Bonsai-cf'' of the two smaller hyper-parameters are higher too (Table \ref{tab:more_synthesis_l2}). Table \ref{tab:more_synthesis_l2} shows that the "2-Bonsai-cf" representation can also reliability gain a high performance once the synthesis quality is good.

After the \emph{synthesis phase}, RFC provides us a rich representation $\Phi$ and $K$ linear classifiers $\omega_1,\dots, \omega_K$. In the downstream tasks, such as OoD/ERM training, we will keep the representation $\Phi$ and initialize the top-layer classifier $\omega$. There are at least two ways to initialize it: 1) initialize $\omega$ as the average of $\omega_1,\dots, \omega_K$ with the hope that the initial top-layer classifier uses all discovered features. 2) randomly initialize $\omega$. Table \ref{tab:rand_average} shows the test performance of OoD/ERM methods with each top-layer initialization method. None of the two top-layer initialization methods significantly outperforms the other one. We choose the first top-layer initialization method in all main experiments because of the interpretation.

\begin{table*}[ht]
    \centering
        \caption{Test accuracy of OoD methods (IRMv1, vREx) and ERM methods. Three \emph{synthesis phase} L2 weights decay $\{1e-2, 1e-4, 1e-6\}$ are tested. All the other settings are the same as the main results in Table \ref{tab:camelyon_full_results}. }
    \label{tab:more_synthesis_l2}
    \resizebox{0.55\textwidth}{!}{
    \begin{tabular}{l|l|l|c|c}
        \toprule
        Synthesis phase & Network  & Methods & \multicolumn{2}{c}{Test Acc} \\
        L2 weights decay & Initialization  &   & IID Tune & OoD Tune \\
        \midrule
        $1e-6$ &    2-Bonsai-cf & ERM  & 78.2$\pm$2.6 & 78.6$\pm$2.6\\
        $1e-6$ &    2-Bonsai-cf & IRMv1  & 78.0$\pm$2.1 & 79.1$\pm$2.1 \\
        $1e-6$ &    2-Bonsai-cf & vREx  & 77.9$\pm$2.7 & 79.5$\pm$2.7 \\
        \midrule
        $1e-4$ & 2-Bonsai-cf & ERM    & 77.8$\pm$1.7 & 78.8$\pm$2.3 \\
        $1e-4$ & 2-Bonsai-cf & IRMv1    & 77.7$\pm$1.7 & 78.9$\pm$2.3 \\
        $1e-4$ & 2-Bonsai-cf & vREx   & 77.9$\pm$1.7 & 79.7$\pm$1.7 \\
        \midrule
        $1e-2$ & 2-Bonsai-cf & ERM   & 75.2$\pm$7.8 & 75.5$\pm$7.4   \\
        $1e-2$ & 2-Bonsai-cf & IRMv1    &75.0$\pm$7.9 & 75.4$\pm$7.5 \\
        $1e-2$ & 2-Bonsai-cf & vREx   & 75.4$\pm$7.7& 75.8$\pm$7.3  \\
        \bottomrule
    \end{tabular}
    }
\end{table*}
\begin{table*}[ht]
    \centering
    \caption{The train and IID-validation performance of the \emph{synthesis phase}. Note that it uses the pseudo-labels instead of the true labels as $Y$. Three \emph{synthesis phase} L2 weights decay $\{1e-2, 1e-4, 1e-6\}$ are tested.  }
    \label{tab:synthesis_quality}
    \resizebox{0.65\textwidth}{!}{
    \begin{tabular}{c|c|c}
            \toprule
           (Synthesis phase) L2 weights decay & Train accuracy & IID-validation accuracy\\
           \midrule
          $1e-6$ &99.7$\pm$0.0 & 97.4$\pm$0.3 \\
          $1e-4$ &99.6$\pm$0.1 & 97.4$\pm$0.2 \\
          $1e-2$ &93.9$\pm$0.7 & 94.9$\pm$0.5 \\
          \bottomrule
    \end{tabular}
    }
\end{table*}

\setlength{\tabcolsep}{4mm}
\begin{table*}[ht]
    \centering
     \caption{Test performance of IRMv1, vREx, and ERM methods on a 2 rounds Bonsai representation. The top-layer classifier is initialized by either the average of $\omega_1,\dots \omega_K$ (Average) or a random initialization (Random). When freezing the representation and training the top-layer classifier only, we get the “-cf” methods.  }
     \resizebox{0.8\textwidth}{!}{
    \begin{tabular}{c|c|c|c||c|c}
    \toprule
Network Initialization & Methods & \multicolumn{2}{c}{Average} & \multicolumn{2}{c}{Random} \\
        & & IID Tune & OOD Tune &  IID Tune & OOD Tune \\
        \midrule
2-Bonsai&ERM & 72.8$\pm$3.2   & 74.7$\pm$4.3  & 73.0$\pm$3.7 & 75.9$\pm$6.7  \\
2-Bonsai&IRMv1 & 71.6$\pm$4.2 & 75.3$\pm$4.8    & 74.5$\pm$2.3 & 75.2$\pm$6.5  \\
2-Bonsai&vREx &  73.4$\pm$3.3 & 76.4$\pm$5.3   & 73.0$\pm$3.9 & 77.1$\pm$5.0  \\
\midrule
2-Bonsai-cf&ERM & 78.2$\pm$2.6 & 78.6$\pm$2.6 & 77.8$\pm$2.4 & 78.6$\pm$2.6 \\
2-Bonsai-cf&IRMv1 & 78.0$\pm$2.1 & 79.1$\pm$2.1  & 78.0$\pm$2.1 & 79.1$\pm$2.1 \\
2-Bonsai-cf&vREx & 77.9$\pm$2.7 & 79.5$\pm$2.7  & 78.0$\pm$2.6 & 79.7$\pm$2.4  \\
\bottomrule
    \end{tabular}
    }
    \label{tab:rand_average}
\end{table*}

\section{Fine-tuning with Very Large Dropout Experiment details}

\subsection{Training from scratch in Figure~\ref{fig:vlcs_scratch_training}}
The \textsc{vlcs} scratch training experiment in Figure~\ref{fig:vlcs_scratch_training} follows the same pipeline as \ood. fine-tuning experiments. But it uses larger learning rates $\{5.10^{-3}, 10^{-2}\}$ on a random initialized \textsc{ResNet50} network (all weights are trainable).

\subsection{Compute Resources}
\label{apx:computeresource}
All experiments are done on V100 GPUs with Intel(R) Xeon(R) Gold 6230
CPUs. One V100 GPU and less than 32GB RAM are enough to fine-tune one Domainbed dataset within a few hours.

\chapter{disentangled representation}
\section{Tracking three moons}

Figure~\ref{fig:attn-3-heads} shows how the training process yields parameter matrices $W_\varphi$, $W_\psi$, and $W_z$, that dedicate one memory unit to each moon.

Training the three-heads network can be quite challenging in a manner that resembles the XOR networks of the early times \citep{rumelhart-1986}. We obtained reliable convergence using two tricks. First, we slightly restrict the linear operations by using $3\times3$ complex matrices (18 real parameters) instead of $6\times6$ real matrices (36 real parameters) operating on the 3-dimensional complex vectors as 6-dimensional real vectors. Second, we clip the mean squared loss in order to prevent the training algorithm from trying to optimize the prediction error when the memories are nearly empty.\footnote{The steamroller metaphor (Figure~\ref{fig:steamroller}) makes more sense when the loss is bounded.}  

Reliable convergence could also be achieved by making any of $W_\varphi$, $W_\psi$, or $W_z$ equal to the identity. Doing so would of course bias the network toward the disentangled solution, something we wanted to avoid. Yet it is not unreasonable to believe that disentanglement can often be achieved in the canonical basis. For instance, objects well separated in space often appear in different image regions, and therefore along different pixels axes.

\begin{figure}[h]
    \centering
    \includegraphics[width=.80\linewidth]{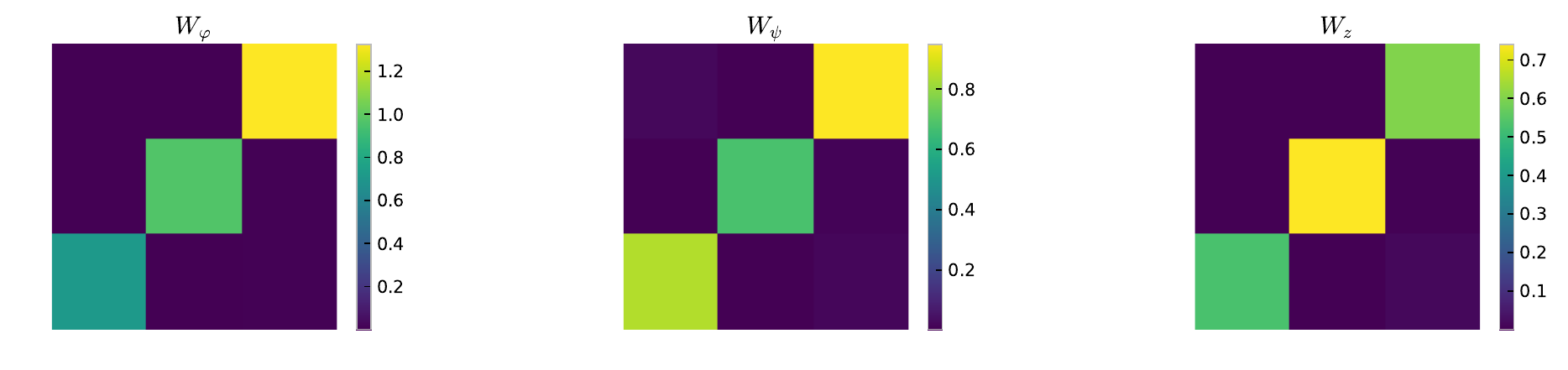}
    \caption{Visualization of the disentangled $W_\varphi$, $W_\psi$, and $W_z$ matrices in the 3-heads network. The color scale represents the moduli of the complex matrix coefficients.}
    \label{fig:attn-3-heads}
\end{figure}

\newpage
\section{BabiStories}
\label{sec:alt_tinystories}
The \textsc{TinyStories} dataset \citep{eldan-2023} is composed of stories written in a simple language and taking place a narrow world. Such stories can be used to train relatively small language models that still must address some of the broader language modeling challenges such as obeying narrative necessity and maintaining logical consistency. This dataset is a wonderful way to study big problems with acceptable computation and quick turn around.

The experiments of Section~\ref{sec:language} were carried out using a dataset generated using a similar methodology but using the \textsc{Mixtral-8x7B} open language model in order to generate unencumbered data. We call this dataset \babistories. 
All the scientific credit is still due to the remarkable work of \citeauthor{eldan-2023}. 
Table \ref{tab:alternative_tinystories_statistic} provides basic statistics for this newly generated \babistories dataset, essentially matching those of the original \textsc{TinyStories} dataset of \citet{eldan-2023}. We had to increase the diversity of the generated stories by expanding the prompt to specify first names and by providing opening words for the story, in addition to required words and story features used by \citeauthor{eldan-2023} (Figure~\ref{fig:storygen}). We also removed the few generated stories containing URLs. %

\begin{figure}[h]
    \centering\bigskip
    \includegraphics[width=.7\linewidth]{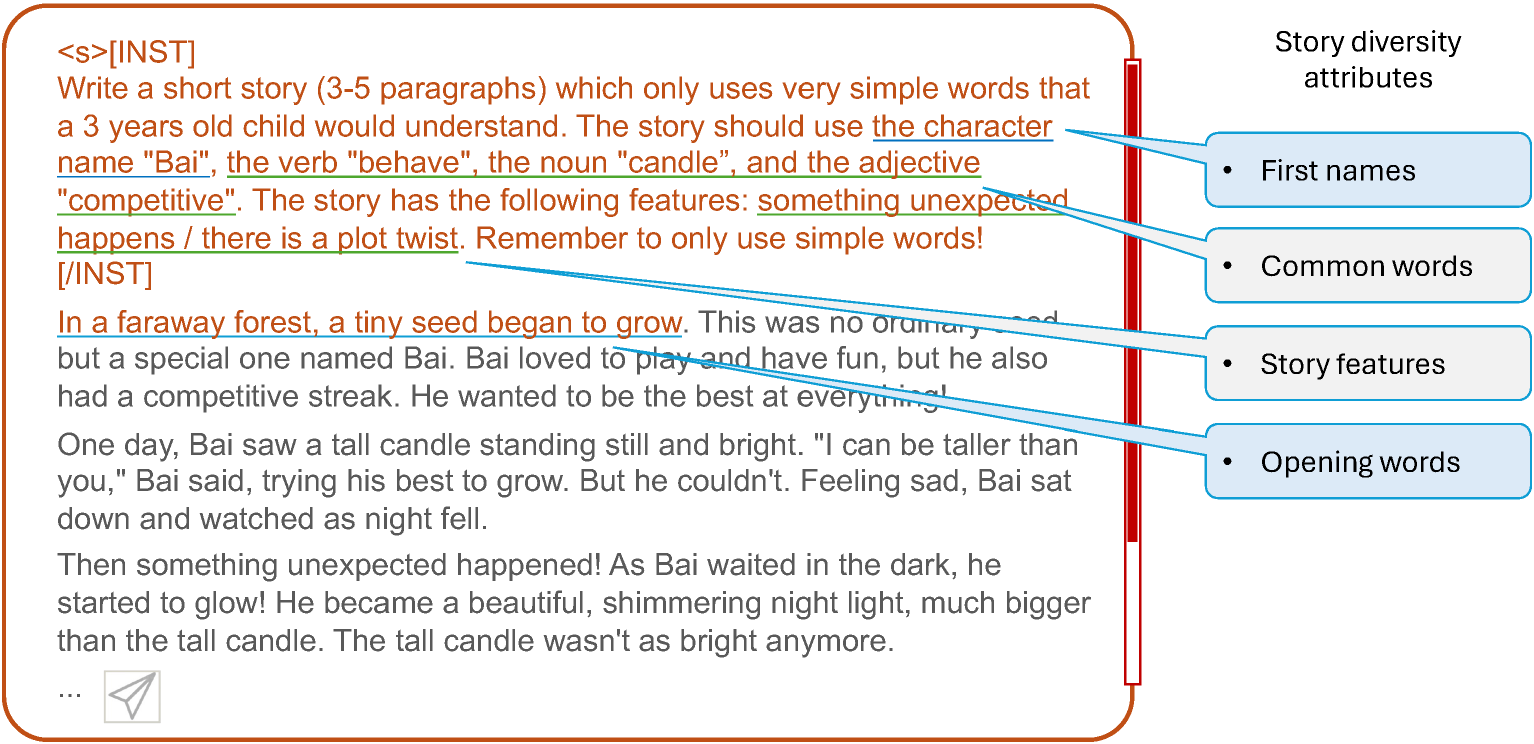}
    \caption{Generation of \babistories. In order to improve the diversity of the generations, each story is generated by a prompt that provides a list of required words and story features (as in \citealp{eldan-2023}) and additionally provides first names and opening words.}
    \label{fig:storygen}
\end{figure}

\begin{table}[h]
    \caption{{\babistories} statistics.}
    \medskip
    \label{tab:alternative_tinystories_statistic}
    \centering
    \begin{tabular}{c|c|c|c}
    \toprule
         dataset partition & \#stories & \#tokens (GPT2 tokenizer) & \#char per story (average) \\
         \midrule
        train & 2.2M &  474,704,907 & 888 \\
        valid & 2.2k &  4,749,107 & 889 \\
        \midrule  %
    \end{tabular}
\end{table}

\newpage
\section{GPT2 baseline and hyperparameters}
\label{sec:gpt2_baseline_hyperparameters}
Table \ref{tab:gpt2_baseline_hyperparameters} showcases the hyper-parameters searching process of GPT2 transformer baseline on the {\babistories} dataset, where we use AdamW optimizer \cite{loshchilov2017decoupled}, batch-size 512, context-size 512, and a cosine learning rate scheduler with minimum learning rate $1e-4$ for all training.

\begin{table}[h]
    \centering
    \caption{Hyperparameters searching on GPT2 transformer with $N_b=12$. ``dropout'', if any, is applied on attention score, attention heads output (before combining layer), and FFN output. }
    \resizebox{\textwidth}{!}{
    \begin{tabular}{ccccc|c|c}
    \toprule
         learning rate & dropout & L2 weight decay & warm-up iters & training iters & train loss & valid loss   \\
         \midrule
         5e-3          & 0.05         & 0.1          & 2000              & 80000                      & 1.336      &\textbf{1.494}         \\
         \midrule
          \cellcolor{red!25}1e-3          & 0.05         & 0.1          & 2000              & 80000    &   1.350     &  1.524      \\
         5e-3          & \cellcolor{red!25}  0            & 0.1          & 2000              & 80000  & 1.281      &1.556         \\
         5e-3          & 0.05         & \cellcolor{red!25} 0.01          & 2000              & 80000    & 1.322      & 1.516        \\
         5e-3          & 0.05         & 0.1          &  \cellcolor{red!25}200              & 80000    &  fail     &   fail      \\
         5e-3          & 0.05         & 0.1          & 2000              & \cellcolor{red!25}40000    & 1.325      &1.532         \\
         5e-3          & 0.05         & 0.1          & 2000              & \cellcolor{red!25}160000    & 1.314      &1.497         \\
    \toprule
    \end{tabular}
    }
    \label{tab:gpt2_baseline_hyperparameters}
\end{table}

\newpage
\section{Memory Mosaics for language modeling}

\subsection{Persistent memory units}

{
Persistent memory units produce their outputs using the same key extraction function $\varphi(x_T,x_{T-1},\dots)$ and the same retrieval function \eqref{eq:gaussian-smoothing-with-dp} as contextual memory units. They differ because, following \citet{sukhbaatar-2019}, they use a fixed array of key/values pairs that are treated as parameters and are determined at training time by gradient descent. Since these stored key/value pairs do not change at inference time, there is no need for a value extraction function $\psi(x_{T+1},x_T,\dots)$
}

\begin{figure}[h]
    \centering
    \includegraphics[width=.7\linewidth]{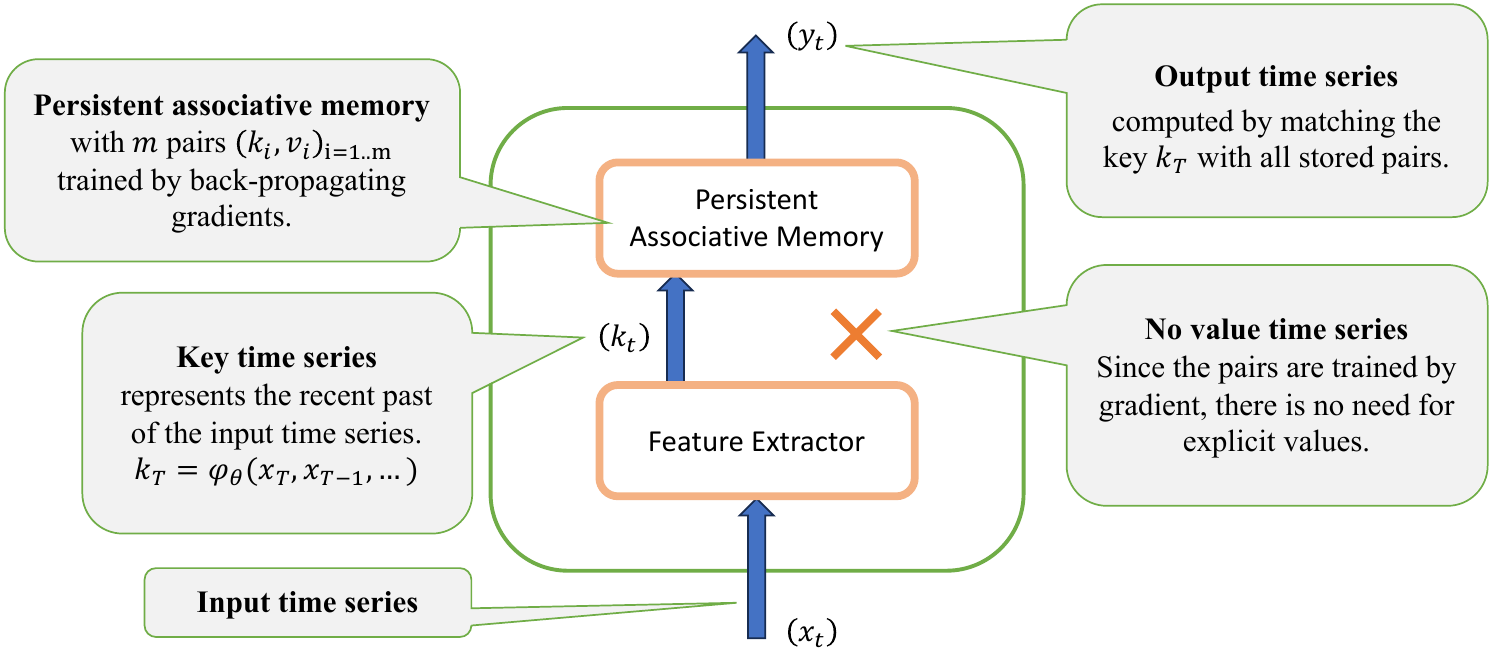}
    \caption{Persistent memory unit. The persistent associative memory contains a fixed number of key-value pairs $(k_i,v_i)_{i=1\dots m}$ whose values are determined by back-propagating gradients at training time. Since the memory contents do not change at inference time, there is no need for explicit values.}
    \label{fig:persistentmodule}
\end{figure}

{

\subsection{Training and validation}
\label{app:figseven}

Figure~\ref{fig:gpt2_mm_comparison_appendix} plots the training and validation curves for both Transformer and Memory Mosaic in a manner similar to Figure~\ref{fig:gpt2_mm_comparison} but showing additional block depths. 

}

\begin{figure}[h]
    \centering
    \includegraphics[width=\linewidth]{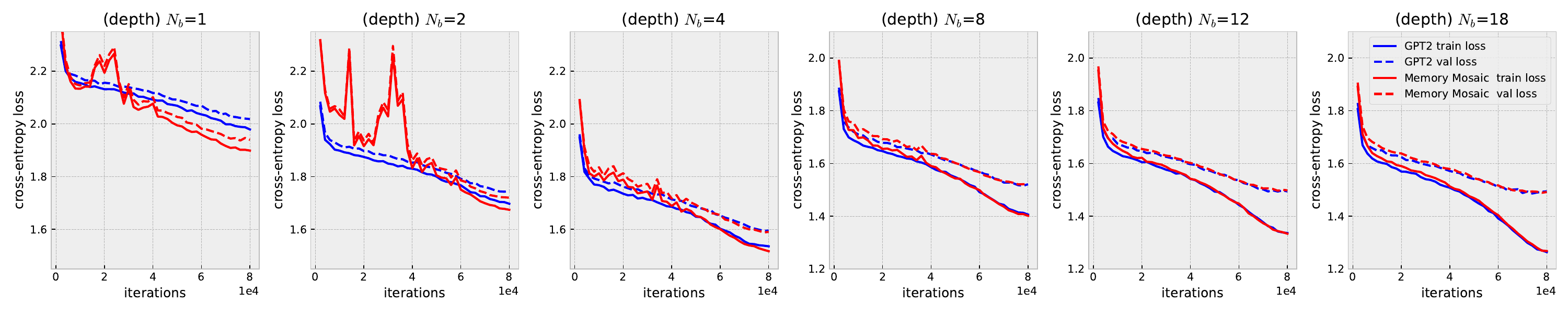}
    \caption{{Additional training and validation curves for the transformer and Memory Mosaic architectures trained on {\babistories} for more model depths than Figure~\ref{fig:gpt2_mm_comparison}.}}
    \label{fig:gpt2_mm_comparison_appendix}
\end{figure}

{
Several comments can be made:
\begin{itemize}
    \item 
    The Memory Mosaic has a small advantage for very small depths ($N_b=1$ and $N_b=4$) but this advantage does not persist when the number of blocks increases. We believe this is due to the fact that a single layer Memory Mosaic can implement an induction head whereas a Transformer needs two layers. This amounts to saying that a $n$ block deep Mosaic has the same number of parameters than a $n$ block deep Transformer, its performance is closer to that of a $n+1$ block Transformer. This is not much of an advantage when $n$ gets large.
    \item
    The Memory Mosaic training uses the hyper-parameters that worked best for the Transformer and operates on the same mini-batches of examples in the same order. However, for small block depths, the Memory Mosaic training curve shows initial instability, suggesting that it might benefit from a smaller stepsize. 
    \item 
    The similarity of the Transformer and Memory Mosaic curves is especially striking when one recalls that the Memory Mosaic does not use position encoding. In fact Memory Mosaic have two mechanisms for dealing with positions. The first one is the fact that the values $v_T$ peek one time position ahead. The second one is the leaky integration in \eqref{eq:betterfeatures}. These two mechanisms are useful to implement bigram or n-gram induction heads in a single layer, but they do not allow a head to selectively address a token by position (we use a single scalar leaky average coefficient per head). This suggests that position encoding in Transformers is mostly useful to implement an initial induction head in the first two blocks.
\end{itemize}

}

\subsection{Qualitative evaluation}
Table \ref{tab:gpt_mm_generation_1layer} provides a variant of Table~\ref{tab:gpt_mm_generation_18layers} in Section \ref{sec:language}, with $N_b=1$. 

\subsection{Differences in Attention and the leaky average coefficient \texorpdfstring{$\lambda_{\varphi}$}{lambda}}

Because Memory Mosaics lack position encoding and do not distinguish keys and queries, we investigate how their attention patterns differ from those of transformers.  Figure~\ref{fig:gpt2_mm_attn} shows attention scores for each head of either a one-block deep transformer using absolute position encoding (left plot) or a one-block deep Memory Mosaic (right plot). The scores are averaged on 5000 {\babistories} sequences and show how the last position attends to earlier positions in the 512 token long context window. The transformer attention patterns are noisy, with a strong ``attention sink'' at position 0 \citep{xiao-2023}. In contrast, the Memory Mosaic attention pattern is mostly flat, save for higher scores for the most recent tokens.\footnote{This effect is connected to the leaky average coefficient $\lambda_\varphi$, as shown in Figure~\ref{fig:mm_attn_details}.}

Figure~\ref{fig:gpt2_mm_attn_long} show the attention patterns for contexts extended to 1536 tokens, using models trained on 512 token long sequences. Because the absolute position encoding scheme cannot be extended to longer contexts, we provides a comparison with transformers using \textsc{RoPE} \citep{su-2024} and \textsc{AliBi}~\citep{press-2022}. The \textsc{RoPE} attention patterns do not extend nicely beyond the training context length. The \textsc{AliBi} attention patterns show the vanishing contribution of distant tokens. In contrast the Memory Mosaic attention patterns remain mostly flat.

Figure \ref{fig:mm_attn_details} shows the relationship between attention map and leaky average coefficient $\lambda_{\varphi}$.

\begin{figure}[h]
    \centering
    \includegraphics[width=0.4\textwidth]{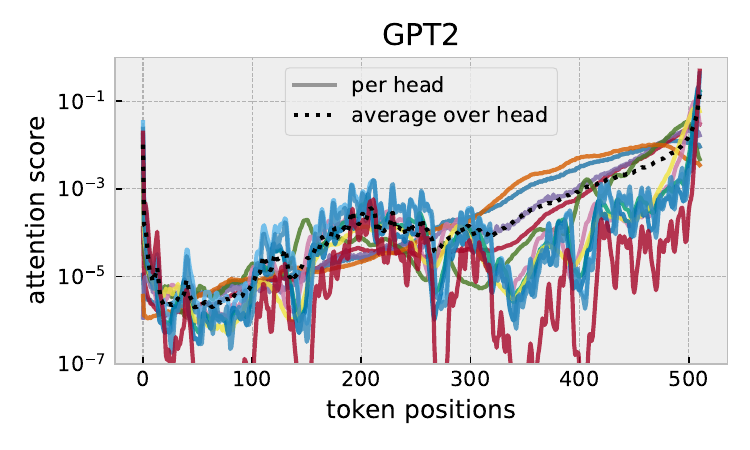}
    \includegraphics[width=0.4\textwidth]{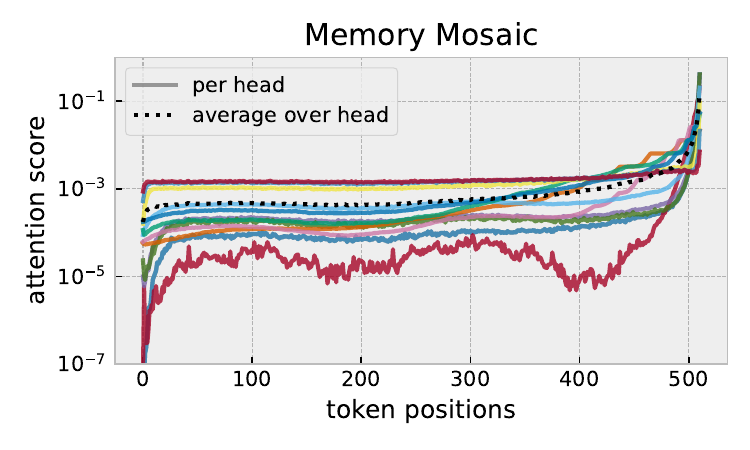}
    \\
    \caption{Average attention scores of the last token attending previous tokens (evaluated on an in-distribution validation dataset). Each solid line indicates one head in either the transformer attention block or the Memory Mosaic contextual memory block. 
    The dotted line averages the attention of all heads. 
    All models are trained with context length 512.
    }
    \label{fig:gpt2_mm_attn}
\end{figure}

\begin{figure}
    \centering
    \includegraphics[height=0.25\textwidth]{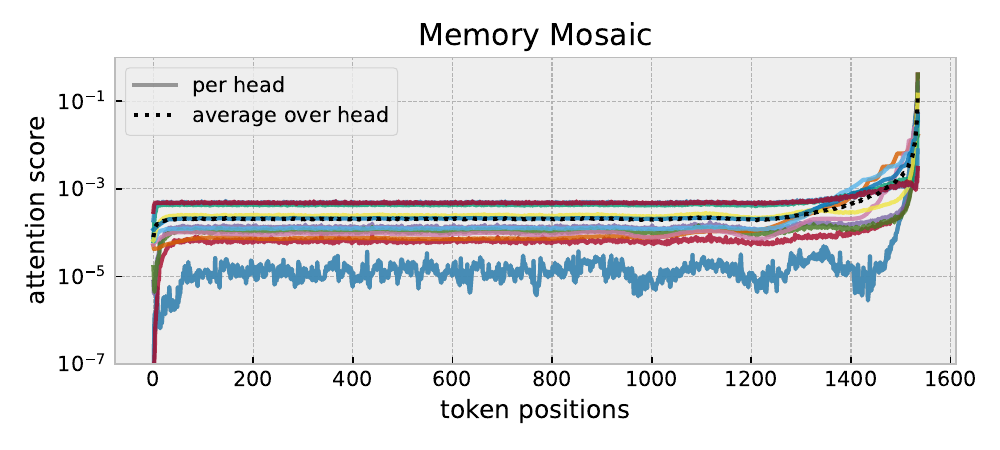}\\
    \includegraphics[width=0.45\textwidth]{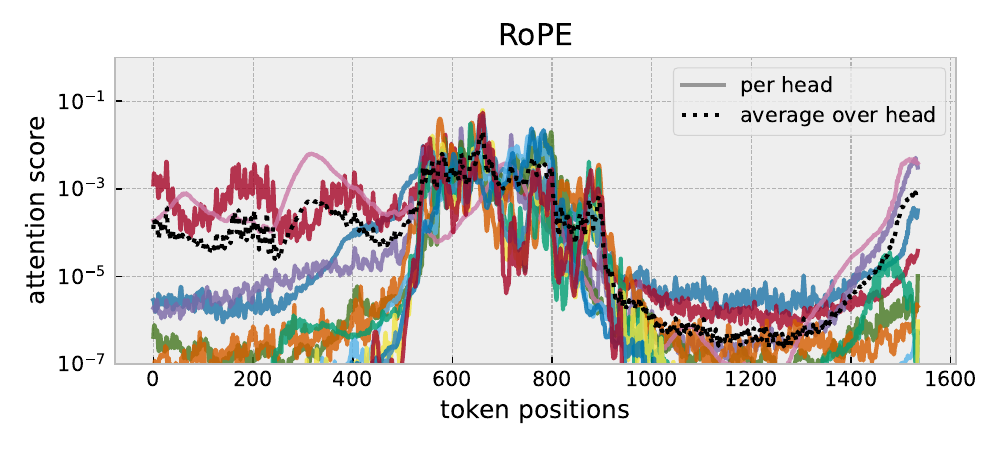}
    \includegraphics[width=0.45\textwidth]{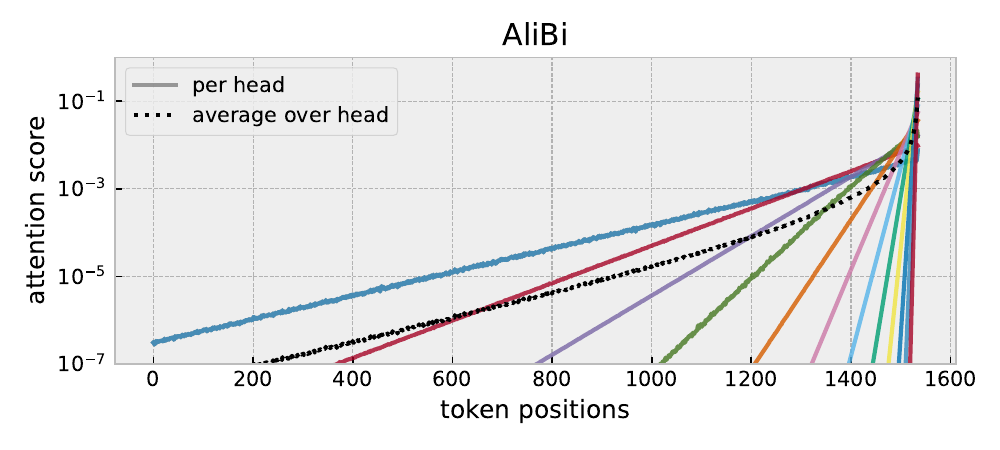}
    \caption{Average attention scores on an extended context window ($3\times512$ tokens). Models are still training with a 512 token long context window. Because the GPT2 absolute position encoding does not extend, we compare with \textsc{RoPE}~\citep{su-2024} and \textsc{AliBi}~\citep{press-2022}.}
    \label{fig:gpt2_mm_attn_long}
\end{figure}

\begin{figure}
    \centering
    \includegraphics[width=\textwidth]{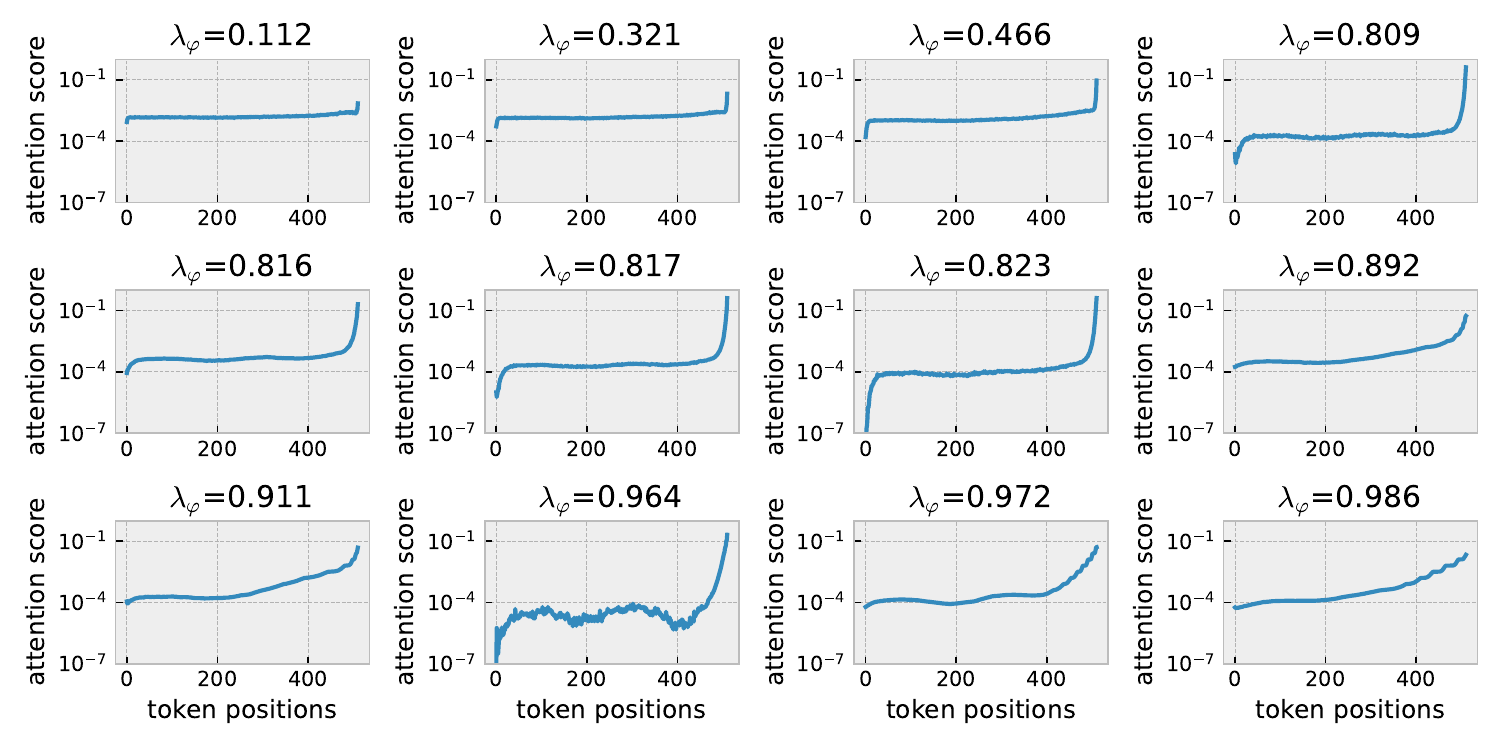}
    \caption{Attention map and leaky average coefficient $\lambda_{\varphi}$. As $\lambda_{\varphi}$ increases, $k_t$ in Eq \ref{eq:betterfeatures} effectively takes a longer history into the account, and thus the pick at the end of attention map becomes wider.}
    \label{fig:mm_attn_details}
\end{figure}

\subsection{In-context language learning evaluation}
Table \ref{tab:icll_iid_100} provides the IID test performance of various architectures trained on \textsc{RegBench} \citep{akyurek2024context} with 100 training environments. We keep the training process, including hyperparameter searching space, to be the same as the one in Figure \ref{fig:icll_acc}. But sample validation and test sets from the same 100 probabilistic finite automatons (training environments) as the training set. This table, together with Figure \ref{fig:icll_acc}, show that \textit{baseline methods learned the training environments (good IID) but not the meta-learning ability (poor OOD)}.

\begin{table}
    \caption{In-distribution (IID) performance of various architectures trained on \textsc{RegBench} \citep{akyurek2024context} with only 100 training environments. Both training, validation, and test set (100 samples) are sampled from the same 100 random probabilistic finite automatons (PFA). Compared with the poor OOD accuracy ($\sim$0.45) / TVD ($\sim$0.75) of baseline methods in Figure \ref{fig:icll_acc}, All baseline methods perform well in the IID test set (even with only 100 training environments). }
    \label{tab:icll_iid_100}
    \centering
    \resizebox{\textwidth}{!}{
    \begin{tabular}{c|c|cccccccccc}
                                       & Memory Mosaic & tf & Mamba & S4 & RWKV & linear tf & H3 & GLA & Hyena & LSTM & RetNet \\
                                       \toprule
        Accuracy ($\uparrow$)          & \textbf{0.959} &0.856 &0.929 &0.846 & \textbf{\underline{0.967}} & 0.816 & 0.794 & 0.870 & \textbf{0.953} & 0.849 & 0.876  \\
        TVD ($\downarrow$)             & 0.417 & 0.308 & \textbf{0.268} & 0.350 & \textbf{\underline{0.183}} & 0.348 & 0.425 &0.284 &\textbf{0.244} &0.343 &0.296  \\
    \bottomrule
    \end{tabular}
    }
\end{table}

\clearpage
\section{Computing Resources}
\label{sec:compute_resources}
Models were trained on 64 NVidia V100 GPUs over 80k epochs.
From conception to finalization of this paper we trained about 200 models.
To create the BabiStories dataset via Mistral, we ran with 128 NVidia V100 GPUs for 3 days. The supporting machines contain Intel(R) Xeon(R) Gold 6230 CPUs. The 3 moons result took negligible resources and were trained on Apple M1 laptops.

\begin{table}[p]
    \vspace*{-1ex}
    \caption{Continuations generated for the 24 prompts designed by \citet{eldan-2023} to investigate the factual, reasoning, and consistency capabilities of language models trained on \babistories. Both the transformer and the Memory Mosaic were $N_b=18$ blocks deep. Colors \colorbox{green!25}{green} and \colorbox{red!25}{red} respectively indicate correct and incorrect answers. The text was generated with a beam search of width 2 and manually truncated to the shortest length that answers or fails to answer the question.}
    \label{tab:gpt_mm_generation_18layers}
    \scriptsize
    \centering
    \resizebox{\textwidth}{!}{
    \begin{tabular}{p{.48\linewidth}!{\color{white}~~}p{.26\linewidth}!{\color{white}~~}p{.26\linewidth}}
        \toprule[1pt]
        \hfil\bf Prompt & \hfil\bf Transformer ($N_b=18$) & \hfil\bf Memory Mosaic ($N_b=18$) \\
        \midrule
        Alice was so tired when she got back home so she went & \cellcolor{green!25}  to sleep. &  \cellcolor{green!25}  straight to bed.  \\
        \midrule
        Jack and Lily saw a rainbow after a rainy day. They were amazed by the colors. Jack said, ``Look, Lily. A rainbow has & \cellcolor{green!25} many colors!'' &\cellcolor{green!25} many colors!''  \\
        \midrule
        Jack and Lily liked to watch the moon at night. They noticed that the moon changed its shape every night. Sometimes the moon was big and round, and sometimes it was &   \cellcolor{green!25} small and thin. One night, Jack said, ``Lily, look! The moon is changing.''  &  \cellcolor{green!25}  small and thin. One night, Jack had an idea. He said, ``Let's make a picture of the moon!'' \\
        \midrule
        Jack wanted to read a book, so he went to&  \cellcolor{green!25}  the library. &  \cellcolor{green!25} the library.  \\
        \midrule
        ``Can cows fly?'', Alice asked her mother&  \cellcolor{green!25} Her mother laughed and said, ``No, Alice. Only birds and planes can fly.''  & \cellcolor{green!25} ``No, dear. Only birds can fly,'' her mother replied with a smile. \\
        \midrule
        ``What do birds like to eat?'', Tom asked his mother.&  \cellcolor{green!25}  ``Birds like to eat worms and bugs,'' his mother replied.  &  \cellcolor{green!25} ``Birds like to eat worms and bugs, Tom,'' his mother replied. \\
        \midrule
        ``What language do \textbf{{they}} speak in France?'', Tom asked his mother&  \cellcolor{red!25}   His mother smiled and said, ``\textbf{\underline{We}} all speak different languages, Tom. It's what makes us unique and special.''  &  \cellcolor{green!25}   His mother smiled and said, ``They speak in different languages, Tom. But we can learn some of them.''  \\
        \midrule
        If I throw a ball up in the air, eventually it will&  \cellcolor{green!25} come down and I can catch it. & \cellcolor{green!25} come down.  \\
        \midrule
        It was winter and cold outside so his mother told him, ``You should& \cellcolor{green!25}  wear your warm coat, dear.'' & \cellcolor{green!25}  wear your warm coat and hat.''  \\
        \midrule[1pt]
        Lily likes cats and dogs. She asked her mom for a dog and her mom said no, so instead she asked& \cellcolor{red!25} for \textbf{\underline{a toy}}.  & \cellcolor{green!25}  her dad.  \\
        \midrule
        Jack told Mary, "If you give me your {{banana}}, I'll give you my {apple}". Mary gave Jack her {banana} so&  \cellcolor{green!25}he could eat it.& \cellcolor{green!25} that he could eat it. \\
        \midrule
        On \textbf{weekends} Jack went to visit his \textbf{grandmother} whereas on \textbf{weekdays} he would go to \textbf{school}. Last \textbf{weekend}, when Jack was on his way to& \cellcolor{red!25} \textbf{\underline{school}}, & \cellcolor{red!25} \textbf{\underline{school}}, \\
        \midrule
        Lily and Ben were having an {argument}. Ben said that cake is much better than ice cream and Lily said that&  \cellcolor{green!25}  ice cream is the best. & \cellcolor{green!25}  she likes ice cream \\
        \midrule
        Lily and Ben are having an {argument}. They are trying to decide between the park and the swimming pool. Ben says, ``I want to go to the park''. Lily says& \cellcolor{green!25}, ``No,  & \cellcolor{green!25}   ``No, \\
        \midrule
        Jack's \textbf{mother was not home}, and his \textbf{father was at home}. When Jack came home, he said hello to& \cellcolor{red!25} his \textbf{\underline{mother}}. & \cellcolor{red!25} his \textbf{\underline{mother}} \\
        \midrule
        Lily doesn't like swimming. When her father wants to take her to the swimming pool, she says& \cellcolor{green!25} , ``No, daddy. I don't want to go.'' &  \cellcolor{green!25} , ``No, thank you.''  \\
        \midrule
        Both Ben and Lily wanted cake. Father said that there was only one piece of cake left. They&  \cellcolor{green!25}   both wanted to eat it, but there was only one.  &  \cellcolor{green!25}  all wanted to eat it, so they began to fight.\\
        \midrule
        Ben went to visit Lily in her house, but she was not at home. Ben knocked on the door,& \cellcolor{green!25} and Lily's mom answered. & \cellcolor{green!25}  and Lily's mom opened it. \\
        \midrule[1pt]
        ``Hi Jane, have you seen Alice? I can't find her anywhere'', said Jack.&  \cellcolor{green!25}  Jane shook her head. ``No, I haven't. But I will help you look.'' &  \cellcolor{green!25}  Jane shook her head. ``No, I haven't seen Alice. But I can help you look!''  \\
        \midrule
        Max had two dogs. One was \textbf{white} and the other was \textbf{black}. Max walked up the street and saw a kid with a dog. He told the kid, ``I see you have a brown dog. I also have& \cellcolor{green!25}  a white dog.''  &\cellcolor{red!25}  a \textbf{\underline{brown}} dog.'' \\
        \midrule
        Anne had a piece of candy in her left pocket and a piece of chocolate in her right pocket. Anne's mom asked her, ``Anne, what is that you have in your left pocket?''& \cellcolor{green!25} Anne smiled and said, ``It's a candy& \cellcolor{green!25} Anne opened her hand to show her mom the candy\\
        \midrule
        Alice had both an \textbf{apple and a carrot} in her bag. She took the \textbf{apple} out of the bag and gave it to Jack. She reached into the bag again and took& \cellcolor{green!25}  out the carrot. & \cellcolor{red!25}  out \textbf{\underline{another apple}}. \\
        \midrule
        Alice and Jack walked up the street and met a girl in a red dress. The girl said to them, ``Hi, I'm Jane. What are your names?''& \cellcolor{green!25} Alice replied, ``I'm Alice. This is Jack. We came to see you.'' & \cellcolor{green!25}Alice smiled and said, ``I'm Alice. This is Jack.  We are going to the park to play.'' \\
        \midrule
        Diva was hungry, and wanted to bake a cake, but she didn't have any sugar at home, so she decided to go ask around. She started walking and met a squirrel. She asked the squirrel, ``Would you happen& \cellcolor{green!25}   to know where I can find some sugar?'' &\cellcolor{green!25}   to see any sugar around here?'' \\
        \midrule 
    \end{tabular}
    }
\end{table}

\begin{table}[p]
    \caption{Continuations generated for the 24 prompts designed by \citet{eldan-2023} to investigate the factual, reasoning, and consistency capabilities of language models trained on \babistories. Both the transformer and the Memory Mosaic were $N_b=1$ blocks deep. }
    \label{tab:gpt_mm_generation_1layer}
    \scriptsize
    \centering
    \resizebox{\textwidth}{!}{
    \begin{tabular}{p{.48\linewidth}!{\color{white}~~}p{.26\linewidth}!{\color{white}~~}p{.26\linewidth}}
        \toprule[1pt]
        \bf Prompt & \bf Transformer & \bf Memory Mosaic  \\
        \midrule
Alice was so tired when she got back home so she went  & \cellcolor{green!25}to sleep & \cellcolor{green!25}to bed \\
\midrule
Jack and Lily saw a rainbow after a rainy day. They were amazed by the colors. Jack said, ``Look, Lily. A rainbow has & \cellcolor{green!25}come!"  &\cellcolor{green!25} come to us!" \\
\midrule
Jack and Lily liked to watch the moon at night. They noticed that the moon changed its shape every night. Sometimes the moon was big and round, and sometimes it was &   \cellcolor{red!25} just a big, round moon.  &  \cellcolor{red!25} too big.  \\
\midrule
Jack wanted to read a book, so he went to&  \cellcolor{red!25} find his friend, a wise old owl named Hoot.  &  \cellcolor{red!25} find his best friend, Timmy. \\
\midrule
``Can cows fly?", Alice asked her mother&  \cellcolor{red!25}  Her mother smiled and said, ``Of course, dear.  & \cellcolor{red!25}Her mother smiled and said, "Sure, honey.  \\
\midrule
``What do birds like to eat?", Tom asked his mother.&  \cellcolor{red!25} ``I don't know, but it looks yummy!" Tom replied. ``Let's find out!"  &  \cellcolor{red!25}  His mother smiled and said, ``That's a good idea, Tom.\\
\midrule
``What language do they speak in France?", Tom asked his mother&  \cellcolor{red!25} France smiled and said, ``It's a language, Tom. It helps us talk to each other." &  \cellcolor{red!25}  ``Mom, what is that?" France asked.  \\
\midrule
If I throw a ball up in the air, eventually it will&  \cellcolor{red!25}  go far away. & \cellcolor{green!25} come back.\\
\midrule
It was winter and cold outside so his mother told him, ``You should& \cellcolor{red!25} go outside and play in the snow."  & \cellcolor{green!25}  stay in your warm, warm home."  \\
\midrule
Lily likes cats and dogs. She asked her mom for a dog and her mom said no, so instead she asked& \cellcolor{red!25} , ``Can we keep it, mom?" Her mom said, ``Yes, but we need to keep it safe." & \cellcolor{red!25} her mom if they could keep the dog. Her mom said yes, and they named the dog Sparky.  \\
\midrule
Jack told Mary, ``If you give me your {banana}, I'll give you my {apple}". Mary gave Jack her {banana} so&  \cellcolor{green!25} he could have it. & \cellcolor{green!25} she could eat it. \\
\midrule
On {weekends} Jack went to visit his {grandmother} whereas on {weekdays} he would go to {school}. Last {weekend}, when Jack was on his way to& \cellcolor{red!25}  school, & \cellcolor{red!25}school, \\
\midrule
Lily and Ben were having an {argument}. Ben said that cake is much better than ice cream and Lily said that&  \cellcolor{red!25}they could both have fun. & \cellcolor{red!25} it was the best cake they ever had.  \\
\midrule
Lily and Ben are having an {argument}. They are trying to decide between the park and the swimming pool. Ben says, ``I want to go to the park". Lily says& \cellcolor{green!25}, ``No, & \cellcolor{green!25} ``No, \\
\midrule
Jack's {mother was not home}, and his {father was at home}. When Jack came home, he said hello to& \cellcolor{red!25} his mother, & \cellcolor{red!25} Jack. \\
\midrule
Lily doesn't like swimming. When her father wants to take her to the swimming pool, she says& \cellcolor{red!25}  ``Lily, you can do it!  &  \cellcolor{red!25} , ``Mommy, I want to swim too!"   \\
\midrule
Both Ben and Lily wanted cake. Father said that there was only one piece of cake left. They&  \cellcolor{red!25} all sat down to enjoy the yummy treat.  &  \cellcolor{green!25}   both felt sad.  \\
\midrule
Ben went to visit Lily in her house, but she was not at home. Ben knocked on the door,& \cellcolor{green!25} and the door opened. A kind lady came out  & \cellcolor{red!25}  and when \textbf{\underline{Ben}} opened the door, \\
\midrule
``Hi Jane, have you seen Alice? I can't find her anywhere", said Jack.&  \cellcolor{red!25} Alice smiled and said, ``Sure, I will help you find your way home."  &  \cellcolor{red!25}  ``I don't know, Jack.  \\
\midrule
Max had two dogs. One was white and the other was black. Max walked up the street and saw a kid with a dog. He told the kid, ``I see you have a brown dog. I also have& \cellcolor{green!25} a black dog." &\cellcolor{red!25}  a brown dog."\\
\midrule
Anne had a piece of candy in her left pocket and a piece of chocolate in her right pocket. Anne's mom asked her, "Anne, what is that you have in your left pocket?"& \cellcolor{red!25} Anne smiled and said, "Yes, mommy. I found it in the park." & \cellcolor{green!25}Anne smiled and said, "I found it on the ground. It's mine!" \\
\midrule
Alice had both an {apple and a carrot} in her bag. She took the {apple} out of the bag and gave it to Jack. She reached into the bag again and took& \cellcolor{red!25}  out the apple.  & \cellcolor{red!25}  out the apple.  \\
\midrule
Alice and Jack walked up the street and met a girl in a red dress. The girl said to them, "Hi, I'm Jane. What are your names?"& \cellcolor{red!25} Jane smiled and said, "I'm \textbf{\underline{Timmy}},   & \cellcolor{red!25} Jane replied, "I'm Jane.  \\
\midrule
Diva was hungry, and wanted to bake a cake, but she didn't have any sugar at home, so she decided to go ask around. She started walking and met a squirrel. She asked the squirrel, "Would you happen& \cellcolor{red!25} to my house, little one?" &\cellcolor{red!25}  to my cake?" \\
        \midrule 
    \end{tabular}
    }
\end{table}

\chapter{Inference-time learning}

\section{Additional results for new-knowledge storage and retrieval}
\label{apx:new-knowledge}
Table \ref{tab:tf_mm_32k_ruler_qa_tasks_long-term-mem} shows that removing long-term memory from memory mosaics v2 after training degrades the performance on the \textsc{ruler} question-answer tasks by 20\%$\sim$30\%. This indicates that the ruler question-answer tasks rely on long-term memory to perform well. 

\begin{table}[ht!]
    \centering
    \vspace{-1ex}
    \caption{The effect of removing ``long-term memory'' of memory mosaics V2 large on \textsc{ruler} question-answer tasks.}
    \vspace{-1ex}
    \label{tab:tf_mm_32k_ruler_qa_tasks_long-term-mem}
    \resizebox{0.9\textwidth}{!}{
    \setlength{\tabcolsep}{4mm} 
    \begin{tabular}{cc|c c c c  }
    \toprule
         model          &   \makecell{context length}      &   4k   &   8k   &   16k   &   32k  \\
         \midrule
memory mosaics large  &32k & 58.9  &  55.5  &  54.9  &  53.4 \\
memory mosaics large without long-term memory  &32k &   38.5  &  22.2  &  20.0  &  20.2 \\
\bottomrule
    \end{tabular}
    }
    \vspace{-1ex}
\end{table}

Table \ref{tab:memory_mosaics_v2_and_other_base_models_on_ruler} compares memory mosaics v2 large and other public base models on \textsc{ruler} question-answer tasks. Memory mosaics v2 large outperforms these models across all task lengths.

\begin{table}[ht!]
    \centering
    \vspace{-1ex}
    \caption{Comparison of Memory Mosaics v2 large (base model) and other public base models (similar scale) on \textsc{ruler} question-answer tasks. Memory Mosaics v2 large outperforms these models across all task lengths. The numbers in ``*'' rows come from \citet{hsieh2024ruler}.   }
    \label{tab:memory_mosaics_v2_and_other_base_models_on_ruler}
    \vspace{-1ex}
    \setlength{\tabcolsep}{2mm} 
    \resizebox{0.8\textwidth}{!}{
    \begin{tabular}{@{}l|c|cccc}
    \toprule
    Model & claimed length & task-length 4k &  8k &  16k & 32k \\
    \midrule
    Memory-Mosaics-v2-large (base) &32k &  58.9  &  55.5  &  54.9  &  53.4 \\
    \midrule
    Llama2-7B (base)* & 4k & 48.6 & - & - & -  \\
Mixtral-base (8x7B)* & 32k & {50.8} & 47.7 & 45.3 & 41.3 \\
Mistral-base (7B)* & 32k & {53.5} & {51.0} & 48.4 & 44.7 \\
Together-base (7B)* & 32k  & 47.5 & 44.6 & 33.6 & 0.0  \\
LongLoRA-base (7B)* & 100k &  34.5 & 32.1 & 33.6 & 29.4 \\
Yarn-base (7B)* & 128k &  29.7 & 23.5 & 28.6 & 29.7  \\
LWM-base (7B)* & 1M & 42.7 & 40.2 & 38.7 & 37.1  \\
\bottomrule
    \end{tabular}}
\end{table}

\section{Prompt examples of multiclass classification tasks}
\label{apx:prompt_examples_icl_class}

\subsection{Banking77 classification with semantic labels}
We sweep the delimiter from  ``[return]'' and ``[space]'', leads to the following two prompts:

\begin{boxA}
``Given a customer service query, please predict the intent of the query. The predict answer must come from the demonstration examples with the exact format. The examples are as follows: 
\bigbreak
service query: \\
I am still waiting on my card? \\
intent category: \\
city\_arrival
\bigbreak
service query:\\
My card has been found. Is there any way for me to put it back into the app?\\
intent category:\\
city\_linking 
\bigbreak
...
\bigbreak
service query:\\
Can I get a card even if I live outside the UK?\\
intent category:\\''
\end{boxA}

\begin{boxA}
``Given a customer service query, please predict the intent of the query. The predict answer must come from the demonstration examples with the exact format. The examples are as follows: \\
service query: I am still waiting on my card? \\
intent category: city\_arrival \\
service query: My card has been found. Is there any way for me to put it back into the app?\\
intent category:
city\_linking \\ 
... \\
service query: Can I get a card even if I live outside the UK? \\
intent category:''
\end{boxA}

\subsection{Goemotion classification with semantic labels}
We sweep the delimiter from  ``[return]'' and ``[space]'', leads to the following two prompts: 

\begin{boxA}
``Given a comment, please predict the emotion category of this comment. The predict answer must come from the demonstration examples with the exact format. The examples are as follows: 
\bigbreak
comment:\\
Her upper lip always looks terrible - such an easy fix, can u believe she is so vain and never bothers to wax \\
emotion category:\\
embarrassment
\bigbreak
comment:\\
No problem. I'm happy to know it's not what you meant.\\
emotion category:\\joy
\bigbreak
...
\bigbreak
comment:\\
These refs have it out for the colts. I didn't realize we traded our MVP 11 to
KC either.\\
emotion category:\\''
\end{boxA}

\begin{boxA}
``Given a comment, please predict the emotion category of this comment. The predict answer must come from the demonstration examples with the exact format. The examples are as follows: \\
comment:
Her upper lip always looks terrible - such an easy fix, can u believe she is so vain and never bothers to wax \\
emotion category:
embarrassment \\
comment: No problem. I'm happy to know it's not what you meant.\\
emotion category: joy
...
comment: These refs have it out for the colts. I didn't realize we traded our MVP 11 to
KC either.\\
emotion category:''
\end{boxA}

\subsection{Tacred classification with semantic labels}
We sweep the delimiter from  ``[return]'' and ``[space]'', leads to the following two prompts: 

\begin{boxA}
``Given a sentence and a pair of subject and object entities within the sentence, please predict the relation between the given entities. The examples are as follows: 
\bigbreak
sentence:\\
But US and Indian experts say it has hesitated to take action against Lashkar-e-Taiba, which means ``The Army of the Pure, ''believing that the Islamic militants could prove useful in pressuring its historic rival India.\\
the relation between Lashkar-e-Taiba and Army of the Pure is:\\
org:alternate\_names
\bigbreak
sentence:\\
The offer from ITW, the Glenview, Ill, diversified manufacturer of engineered products, represents a premium of 85 percent to the Manitowoc bid.
\\the relation between ITW and Glenview is:\\
org:city\_of\_headquarters
\bigbreak
...
\bigbreak
sentence:\\
The statement from North Korea, carried by the country's official Korean Central News Agency, did not mention Kim by name, but South Korean Unification Ministry spokesman Kim Ho-nyeon said the North's state media has before used such wording to refer to him.\\
the relation between Korean Central News Agency and North Korea is:\\''
\end{boxA}

\begin{boxA}
``Given a sentence and a pair of subject and object entities within the sentence, please predict the relation between the given entities. The examples are as follows: \\
sentence:
But US and Indian experts say it has hesitated to take action against Lashkar-e-Taiba, which means ``The Army of the Pure, ''believing that the Islamic militants could prove useful in pressuring its historic rival India.\\
the relation between Lashkar-e-Taiba and Army of the Pure is:
org:alternate\_names
\\
sentence:
The offer from ITW, the Glenview, Ill, diversified manufacturer of engineered products, represents a premium of 85 percent to the Manitowoc bid.
\\
the relation between ITW and Glenview is:
org:city\_of\_headquarters
\\
...
\\
sentence:
The statement from North Korea, carried by the country's official Korean Central News Agency, did not mention Kim by name, but South Korean Unification Ministry spokesman Kim Ho-nyeon said the North's state media has before used such wording to refer to him.\\
the relation between Korean Central News Agency and North Korea is:''
\end{boxA}

\end{appendices}

\end{document}